\documentclass[conference,compsoc]{IEEEtran}
\usepackage{graphicx}
\graphicspath{{_img/}}
\usepackage[skip=6pt]{caption} 
\usepackage{placeins}                  % \FloatBarrier
\usepackage{needspace}                 % keep headings off page bottoms
\usepackage{stfloats}
\usepackage{float}
\usepackage{indentfirst}
\makeatletter
\renewcommand\subsubsection{\@startsection{subsubsection}{3}{\z@}%
  {1.8ex plus 1ex minus .2ex}%
  {1.2ex plus .2ex}%
  {\normalfont\normalsize\itshape}}
\makeatother

\usepackage[skins,breakable,theorems,listings]{tcolorbox}
\usepackage{changepage}
\usepackage{enumitem}
\usepackage{amssymb}
\usepackage{newfloat}

\usepackage{booktabs, multirow, makecell, siunitx}
\DeclareFloatingEnvironment{boxes} %\box and \Box is taken
\newcommand{\boxref}[1]{\hyperref[{#1}]{Box~\ref*{#1}}}

\ifCLASSOPTIONcompsoc
  \usepackage[nocompress]{cite}
\else
  \usepackage{cite}
\fi
\ifCLASSINFOpdf
\else
\fi

\usepackage[hidelinks]{hyperref}
\usepackage{url}
\usepackage{booktabs} 
\usepackage{adjustbox}
\usepackage{amsmath}
\usepackage{colortbl} 
\usepackage{xcolor} 
\usepackage[skins]{tcolorbox}
\usepackage[utf8]{inputenc}
\usepackage{dirtytalk}

\usepackage{mathtools}
\usepackage{amsfonts}

\begin{document}
%

% \tableofcontents

\title{Image Generation Models: A Technical History}

\author{
\IEEEauthorblockN{Rouzbeh Shirvani}
\IEEEauthorblockA{rouzbeh.asghari@gmail.com}
}

\maketitle

\begin{abstract}

Image generation has advanced rapidly over the past decade, yet the literature seems fragmented across different models and application domains. This paper aims to offer a comprehensive survey of breakthrough image generation models, including variational autoencoders (VAEs), generative adversarial networks (GANs), normalizing flows, autoregressive and transformer-based generators, and diffusion-based methods. We provide a detailed technical walkthrough of each model type, including their underlying objectives, architectural building blocks, and algorithmic training steps. For each model type, we present the optimization techniques as well as common failure modes and limitations. We also go over recent developments in video generation and present the research works that made it possible to go from still frames to high quality videos. Lastly, we cover the growing importance of robustness and responsible deployment of these models, including deepfake risks, detection, artifacts, and watermarking.

\end{abstract}

\IEEEpeerreviewmaketitle

\section{Introduction} 

Generative image modeling is a problem in computer vision and machine learning that is concerned with learning the underlying structure of the input image in order to generate new image samples that look realistic and diverse while preserving the high level structure. It wasn't until 2014 that the field began to evolve from a niche research topic into a widely used technology leading to applications such as image editing and manipulation, content creation, and multimodal generation models. Progress has mainly been driven by algorithmic innovation, data at scale, and compute. On the algorithmic side, the field has steadily improved model architectures, training objectives, and optimization techniques. In parallel, the availability of large-scale social media and internet training data (image–text pairs) has enabled models to learn stronger priors and better align outputs with the intent of the user.

% Finally, scaling compute has not only made training feasible, but has also set new standards: larger models can be more compute-efficient~\cite{peebles2023scalable} and achieve better sample quality as measured by standard metrics such as FID (Fréchet Inception Distance).

Despite the volume of work in this area, the literature remains dispersed across model types, training objectives, and applications. This fragmentation makes it difficult, especially for newcomers and researchers to develop a coherent understanding of why different approaches work, how they are trained in practice, and where their limitations originate. Motivated by the lack of a single, comprehensive technical review that spans the major breakthrough models, this paper surveys the major body of image generation models. In this work our main emphasis is on (i) the detailed technical formulation behind each approach, (ii) algorithmic and training overviews that clarify how these models are optimized in practice, (iii) key applications and variants of each model, and (iv) recurring shortcomings and failure modes of each model family.

We present the material in a roughly chronological order, following how the field has evolved. We begin with VAEs in Section 2 and show how VAEs provided a probabilistic framework for learning latent-variable models with tractable training objectives, but introduced practical challenges such as posterior collapse and difficulties in producing sharp reconstructions. Nevertheless, VQ-VAEs~\cite{van2017neural} are a key ingredient in the success of diffusion models, which we study later.

In Section 3, we review Generative Adversarial Networks (GANs) that demonstrated a compelling alternative by learning through an adversarial game, producing high quality image samples. Later on different styles of GANs emerged that gave users more flexibility for generating, manipulating, and editing the generated images. Like VAEs, GANs also had optimization and stability issues that we will study.

We dedicate Section 4 to Normalizing Flows, which offer a likelihood-based alternative for generative modeling by learning an invertible transformation from a simple distribution such as Gaussian to the image distribution. A key advantage of normalizing flows is that they enable exact log-likelihood computation and straightforward inference. At the same time, the invertibility constraint and the need for efficient Jacobian calculations restrict model design and can make training and sampling expensive at high resolutions. This in turn can limit their use in large-scale image generation compared to GANs and diffusion models.

In Section 5, we focus on transformer/autoregressive approaches that generate images sequentially by modeling a series of conditional distributions over pixels or image tokens. These methods are appealing due to stable likelihood-based training and strong flexibility for conditional generation. At the same time, sequential sampling can be slow and difficult to scale to high resolutions. This led to the discrete latent representations of pixel and hybrid designs that improve efficiency and visual fidelity.

In Section 6, we show how Diffusion-based models have further improved the state of the art results for high-quality generation by combining powerful image based models with text based models. This turned them into a foundation for many modern text-to-image systems. Diffusion models have evolved from pixel-space denoisers to a more efficient latent-space generators.

Section 7 is dedicated to recent developments in image generation, more specifically, Flow Matching (FM) and Rectified Flows (RF). These methods rely on continuous-time normalizing flows that try to learn an ordinary differential equation (ODE) vector field that can transport samples from a simple distribution, such as a Gaussian, to the data distribution. Overall, these methods provide a stable training regime and are able to generate high quality images with fewer steps.

In Section 8, we turn to video generation and discuss how image generation methods are adapted to model temporal structure, motion, and consistency across frames. We survey major design choices and conditioning mechanisms used in practice. We outline the main bottlenecks, such as long-range coherence, controllability of motion, and substantially higher computational demands.

Finally, we dedicate a separate Section (9) to the societal and security implications of synthetic image/video generation, including detection, deepfake benchmarks, artifacts, and watermarking.

% \subsubsection{Bayes}

% \[
% P(\theta \mid x) \;=\; \frac{P(x \mid \theta)\, P(\theta)}{P(x)}.
% \]

% \begin{itemize}
%     \item \(\boldsymbol{\theta}\): Parameters of the model.
%     \item \textbf{x}: Input to the model.
%     \item \textbf{Posterior \(\boldsymbol{P(\theta \mid x)}\)}: 
%     How likely the parameter \(\theta\) is given the observed data \(x\). 
%     \item \textbf{Likelihood \(\boldsymbol{P(x \mid \theta)}\)}:  
%     The probability of the observed data \(x\) given the parameter \(\theta\).
%     \item \textbf{Prior \(\boldsymbol{P(\theta)}\)}:  
%     Our belief about the parameter \(\theta\) before observing any data.
%     \item \textbf{Evidence \(\boldsymbol{P(x)}\)}:
%     The probability of observing the data \(x\), regardless of \(\theta\).  
%     Acts as a normalizing constant:
%     \[
%     P(x) \;=\; \int P(x \mid \theta)\, P(\theta)\, d\theta.
%     \]
% \end{itemize}

% \subsection{Metrics}

\section{Variational Autoencoders}\label{sec:vae}
Autoencoders (AEs) work by compressing the input data into a representation called latent space (code). Using the latent space, we can later reconstruct the input. Given $x \in \mathbb{R}^D$, one can write the encoder and decoder as

\begin{equation}
\begin{aligned}
z &= f_{\phi}(x) \ , \\
\hat{x} &= g_{\theta}(z) \\
\end{aligned}
\end{equation}

where $\phi$ and $\theta$ are encoder and decoder networks and $z\in\mathbb{R}^d$ where $d<D$. The two networks are trained by minimizing the reconstruction error between $x$ and $\hat{x}$ through the following MSE (Mean Squared Error) loss term:

\begin{equation}
\min_{\phi,\theta}\ \mathbb{E}_{x}\!\left[\|x-\hat{x}\|_2^2\right]
\end{equation}

Rumelhart, Hinton, and Williams~\cite{Rumelhart1986LearningRB} introduced the idea of compressing the input into a bottleneck representation and reconstructing it back. Vincent et~al.~\cite{Vincent2008DAE} showed that corrupting inputs and training the model to reconstruct them produces robust feature representations. Yoshua Bengio and his coauthors~\cite{bengio2013generalized} generalized this framework, proving that denoising autoencoders can be seen as consistent estimators of the data distribution and proposing the Markov-based training procedure to improve the quality of generative sampling. Variational Autoencoders (VAEs) improve the autoencoders in two ways. First, the autoencoders did not enforce structure on the latent space. On the other hand, VAEs use the Kullback-Leibler (KL) term to force the latent space $z$ to follow a particular distribution such as Gaussian as shown in Eq.~\ref{eq:vae-kl}.

\vspace{-0.75em}
\begin{equation}
\mathrm{KL}\!\big(\mathcal{N}(\mu,\sigma^2)\,\|\,\mathcal{N}(0,I)\big)
= \tfrac{1}{2}\sum_{i}\Big(\mu_i^2 + \sigma_i^2 - \log\sigma_i^2 - 1\Big)
\label{eq:vae-kl}
\end{equation}
\vspace{-0.75em}

\begin{figure}[h]
  \centering
  \includegraphics[width=0.6\linewidth]{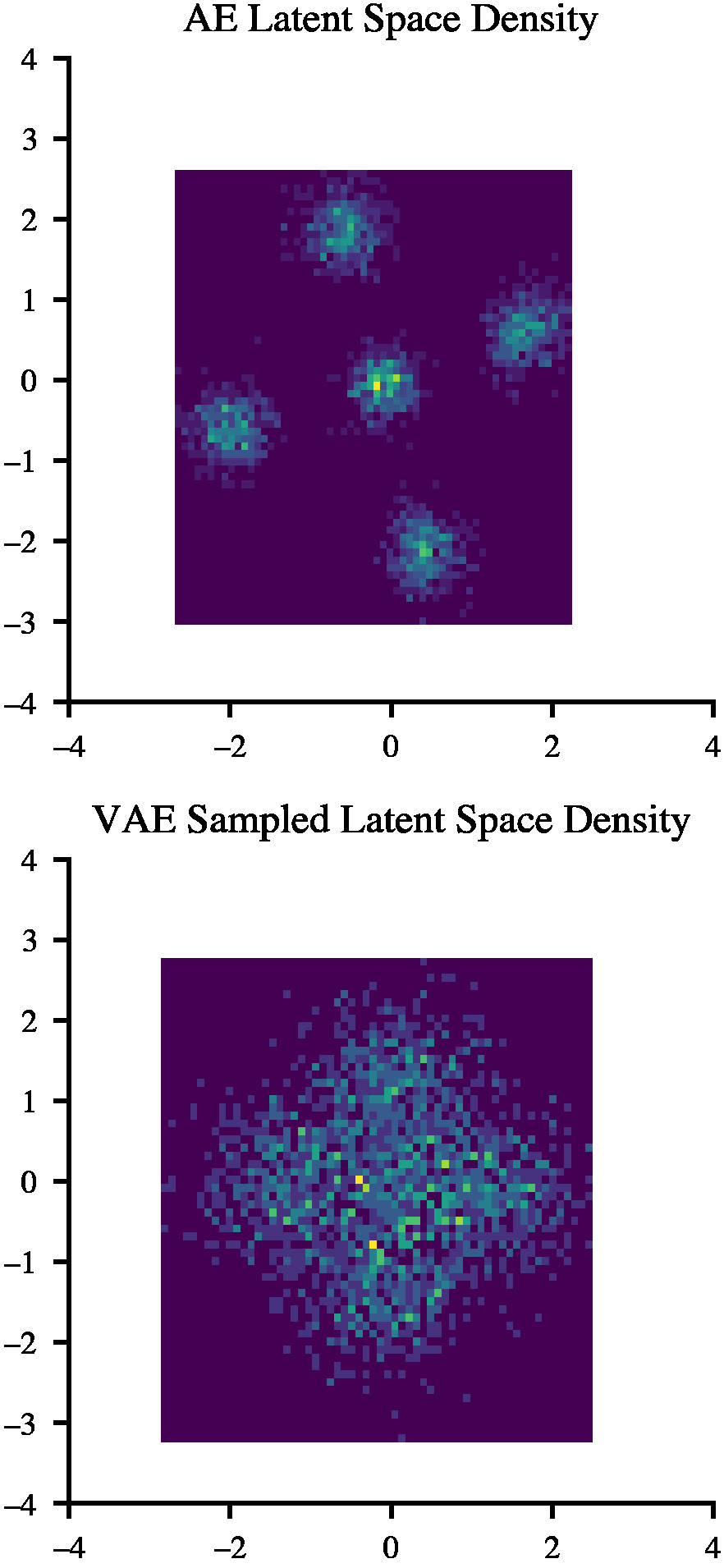}
  \caption{Comparison of AE latent density and VAE sampled latent $z$. VAE has a smooth and favorable latent space compared to the latent space in AE. Image source:~\cite{kingma2013auto}
  }
  \label{fig:vae-vs-ae-density}
\end{figure}

As shown in Fig.~\ref{fig:vae-vs-ae-density} (top), in early generations of autoencoders a randomly sampled latent code $z$ often decoded to meaningless outputs, VAEs regularize the latent space so that random samples can be reliably mapped back into the data space. VAEs force $z$ to follow a particular probability distribution via prior distribution $p(z)$. As a result, $z$ is no longer irregular with holes and rather a smooth distribution as shown in Fig.~\ref{fig:vae-vs-ae-density} (bottom).

Second, VAEs introduce a structured likelihood-based training structure that provides a clean mathematical framework for training both the encoder and decoder simultaneously.

\subsection{How VAEs work?}
Given the model parameters $\theta$, we are interested in knowing how likely it is to see  the real data $x$, $p_\theta(x)$. In order to maximize the likelihood of seeing $x$, we would like to fit it such that the probability assigned to real data points $x$ is as high as possible, Eq.~\ref{eq:mle}.

\begin{equation}
\theta^* = \arg\max_{\theta} \sum_{x \in \text{dataset}} \log p_{\theta}(x)
\label{eq:mle}
\end{equation}

Maximizing $p_\theta(x)$ allows the model to learn the data distribution which results in a more realistic generation. If we bring the latent code, $z$, into the equation, the marginal likelihood would become

\begin{equation}
p_\theta(x) = \int p_\theta(x \mid z)\, p(z)\,dz
\label{eq:marginal}
\end{equation}
This integral sums over all possible latent codes $z$ that could have generated $x$. In high-dimensional space, calculating this integral is intractable.  Instead, we use a neural network $q_\phi(z \mid x)$ (the encoder) that gives a tractable approximation of the distribution of $z$ given $x$. We can rewrite:

\begin{equation}
\log p_\theta(x) 
= \log \int q_\phi(z \mid x)\, \frac{p_\theta(x,z)}{q_\phi(z \mid x)}\, dz
\label{eq:log-marginal}
\end{equation}
this can be expressed as an expectation:

\begin{equation}
\log p_\theta(x) 
= \log \, \mathbb{E}_{q_\phi(z \mid x)}\!\left[\frac{p_\theta(x,z)}{q_\phi(z \mid x)}\right]
\label{eq:log-marginal-expectation}
\end{equation}
Applying Jensen's inequality, $\log \mathbb{E}[X] \geq \mathbb{E}[\log X]$, results in:

\begin{figure}[t]
  \centering
  \includegraphics[width=\columnwidth]{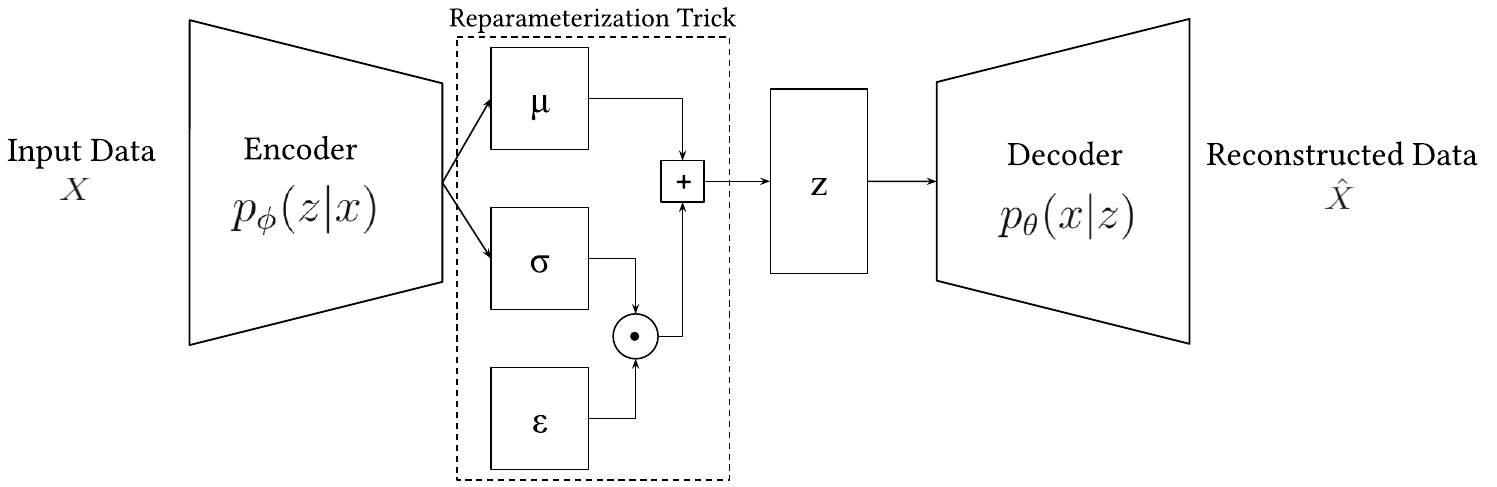}
  \caption{Diagram of the end to end VAE model architecture with reparameterization trick}
  \label{fig:vae-diagram}
\end{figure}

\begin{equation}
\log p_\theta(x) 
\;\geq\; \mathbb{E}_{q_\phi(z \mid x)} 
\Bigg[ \log \frac{p_\theta(x,z)}{q_\phi(z \mid x)} \Bigg]
\label{eq:jensen-log}
\end{equation}
The right-hand side is the Evidence Lower Bound (ELBO)

\begin{equation}
\mathcal{L}(\theta, \phi; x) 
= \mathbb{E}_{q_\phi(z \mid x)} \!\left[ \log p_\theta(x,z) - \log q_\phi(z \mid x) \right]
\label{eq:elbo-loss}
\end{equation}
The ELBO is always a lower bound, as a result maximizing the ELBO pushes it closer to $\log p_\theta(x)$

\begin{equation}
\log p_\theta(x) \;\geq\; \mathcal{L}(\theta, \phi; x).
\label{eq:elbo-inequality}
\end{equation}
Since $p_\theta(x,z) = p_\theta(x \mid z)\, p(z)$, ELBO can be rewritten as:

\begin{align}
\mathcal{L}(\theta, \phi; x) 
&= \underbrace{\mathbb{E}_{q_\phi(z \mid x)}\!\left[ \log p_\theta(x \mid z) \right]}_{\text{reconstruction term}} \nonumber \\
&\quad - \underbrace{\mathrm{KL}\!\left(q_\phi(z \mid x) \,\|\, p(z)\right)}_{\text{regularizer}}
\label{eq:total-loss}
\end{align}

The reconstruction term forces the decoder $p_\theta(x \mid z)$ to explain $x$ well when $z$ is sampled from the encoder $q_\phi(z \mid x)$. The regularizer $\mathrm{KL}(q_\phi(z \mid x)\,\|\,p(z))$ pushes the approximate posterior (encoder output) closer to the prior $p(z)$. In practice, the negative ELBO is minimized as the loss function

\begin{align}
\mathcal{J}(\theta, \phi; x) 
&= - \mathbb{E}_{q_\phi(z \mid x)}\!\left[\log p_\theta(x \mid z)\right] \nonumber \\
&\quad + \mathrm{KL}\!\left(q_\phi(z \mid x) \,\|\, p(z)\right)
\label{eq:loss-kl}
\end{align}

In implementations, $q_\phi(z \mid x)$ and $p_\theta(x \mid z)$ are estimated by encoder and decoder neural networks as depicted in Fig.~\ref{fig:vae-diagram}. $z$ is commonly drawn from a Gaussian prior, i.e., \(z \sim \mathcal{N}(0, I)\)

\subsection{Reparameterization Trick}

As shown in Fig.~\ref{fig:vae-diagram}, in order to move the randomness out of the neural network parameters, 
Kingma et~al.~\cite{kingma2013auto}  redefine the latent space as

\begin{equation}
z = \mu_\phi(x) + \sigma_\phi(x) \odot \varepsilon, 
\quad \varepsilon \sim \mathcal{N}(0, I).
\label{eq:reparam-trick}
\end{equation}
This makes the gradients a standard chain-rule derivative and $\varepsilon$ becomes the only random input

\begin{equation}
\mathbb{E}_{q_\phi(z|x)}[f(z)]
= \mathbb{E}_{\varepsilon \sim \mathcal{N}(0, I)}
   \big[f(g_\phi(\varepsilon, x))\big]
\label{eq:chain-gradient}
\end{equation}
This approach leads to a much lower variance, and end-to-end backpropagation becomes possible. Intuitively, this means that they are sampling from fixed noise and pushing it through a differentiable function that depends on $\phi$. Almost simultaneously to~\cite{kingma2013auto}, Rezende et~al.~\cite{rezende2014stochastic} published a similar work where they presented stochastic backpropagation as a general rule for backpropagating through stochastic nodes and derived gradient identities for Gaussians. Both of these papers introduced reparameterization and also the idea of backpropagation through stochastic nodes.

\subsection{KL Collapse: when the model ignores \texorpdfstring{$z$}{z}}

KL collapse, also known as posterior collapse, happens when a VAE stops encoding useful information about $x$. In such a case, the approximate posterior $q_\phi(z \mid x)$ becomes nearly identical to the prior $p(z)$ for almost all $x$:
\begin{equation}
    q_\phi(z \mid x) \approx p(z), \quad \text{for almost all } x.
\end{equation}
As a result, the decoder learns to reconstruct or predict without using $z$. The KL term vanishes, and the VAE effectively acts as a simple autoencoder. If the decoder $p_\theta(x \mid z)$ is powerful enough to model $p_\theta(x)$ without depending on $z$, then the easiest way to increase the objective is to let $\mathrm{KL} \to 0$ which is undesirable. 

To alleviate this issue, Higgins and coauthors~\cite{higgins2017beta} introduced the $\beta$-VAE, a variant of VAEs that introduces a hyperparameter $\beta$:
\begin{align}
\mathcal{L}(\theta, \phi; x) 
&= \mathbb{E}_{q_\phi(z \mid x)} 
   \big[ \log p_\theta(x \mid z) \big] \nonumber \\
&\quad - \beta \, \mathrm{KL}\!\big(q_\phi(z \mid x) \;\|\; p(z)\big)
\label{eq:beta-vae-loss}
\end{align}
The adjustable hyperparameter $\beta$ controls the balance between the latent space capacity (through KL) and the reconstruction loss (log-likelihood term).

Burgess et~al.~\cite{burgess2018understanding} proposed a modified $\beta$-VAE where the information capacity of the latent space is progressively increased during training. Instead of a fixed $\beta$, they schedule the KL capacity $C_t$ from zero to some maximum $C_{\max}$:

\begin{align}
\mathcal{L}(\theta, \phi; x) 
&= \mathbb{E}_{q_\phi(z \mid x)} 
   \big[ \log p_\theta(x \mid z) \big] \nonumber \\
&\quad - \gamma \Big| \mathrm{KL}\!\big(q_\phi(z \mid x) \;\|\; p(z)\big) 
   - C_t \Big|
\label{eq:capacity-loss}
\end{align}
where $C_t$ gradually increases from $0$ to $C_{\max}$ over the course of training. Early in training when the encoder has very limited capacity, it must focus on transmitting only the most useful coarse factors. As $C_t$ increases, the encoder gains more capacity to capture fine-grained details.

\subsection{Blurry Reconstructions in VAEs}

Most vanilla VAEs rely on a Gaussian decoder:
\begin{equation}
p_\theta(x \mid z) = \mathcal{N}\!\bigl(x;\,\mu_\theta(z),\,\sigma^2 I\bigr)
\end{equation}
The log-likelihood will look like:

\begin{equation}
\log p_\theta(x \mid z)
= \text{Constant} - \frac{1}{2\sigma^2}\,\left\lVert x - \mu_\theta(z) \right\rVert_2^2
\end{equation}
Maximizing $\log p_\theta(x \mid z)$ is the same as minimizing $\left\lVert x - \mu_\theta(z) \right\rVert_2^2$. This means that $x$ can be pushed towards the mean. For the output image, this is undesirable since the image can get blurry by converging towards the mean of the pixels.

\begin{figure}
  \centering
  \includegraphics[width=\columnwidth]{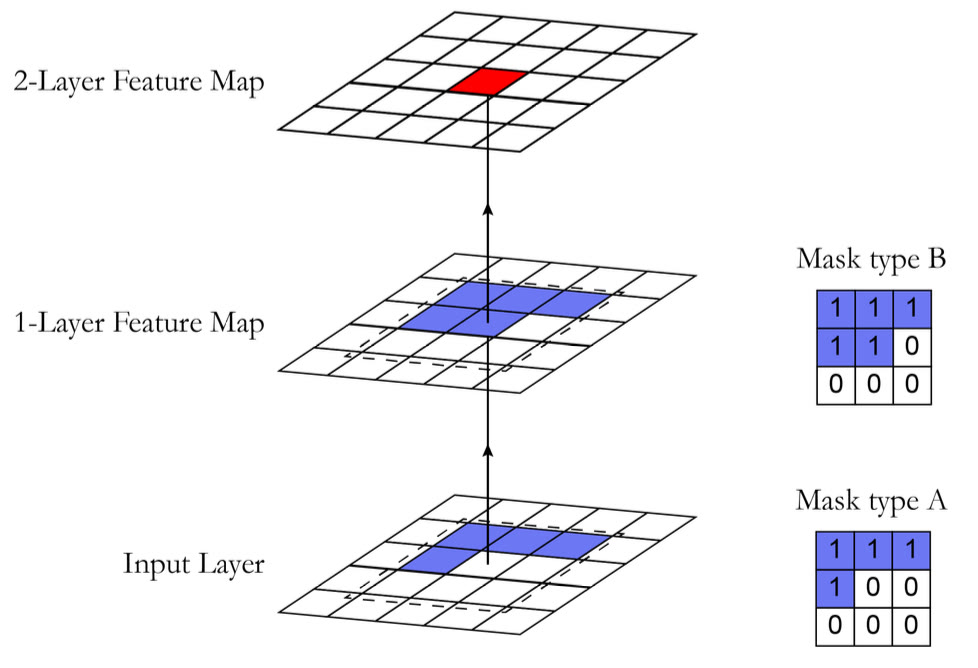}
  \caption{PixelCNN and conditioning on prior pixels. Image adapted from~\cite{daCosta_autoencoder2021}}
  \label{fig:autoencoder-costa}
\end{figure}

In order to alleviate this problem, Oord et~al.~\cite{van2016conditional} introduced PixelCNN for conditional image generation where they predict each pixel from the surrounding pixels in a raster order:

\begin{equation}
p(x) = \prod_{i=1}^{HW} p(x_i \mid x_{<i}),
\label{eq:p-pixel}
\end{equation}
where $x_{<i}$ are earlier pixels in raster order. As shown in Fig.~\ref{fig:autoencoder-costa} PixelCNNs are capable of producing sharper images since they model a joint distribution.
Later, Gulrajani et~al.~\cite{gulrajani2016pixelvae} introduced PixelVAE where they use PixelCNN to model an autoregressive decoder in the decoder section of the VAE. This can help VAEs generate sharper output images. 

\begin{figure}
  \centering
  \includegraphics[width=\columnwidth]{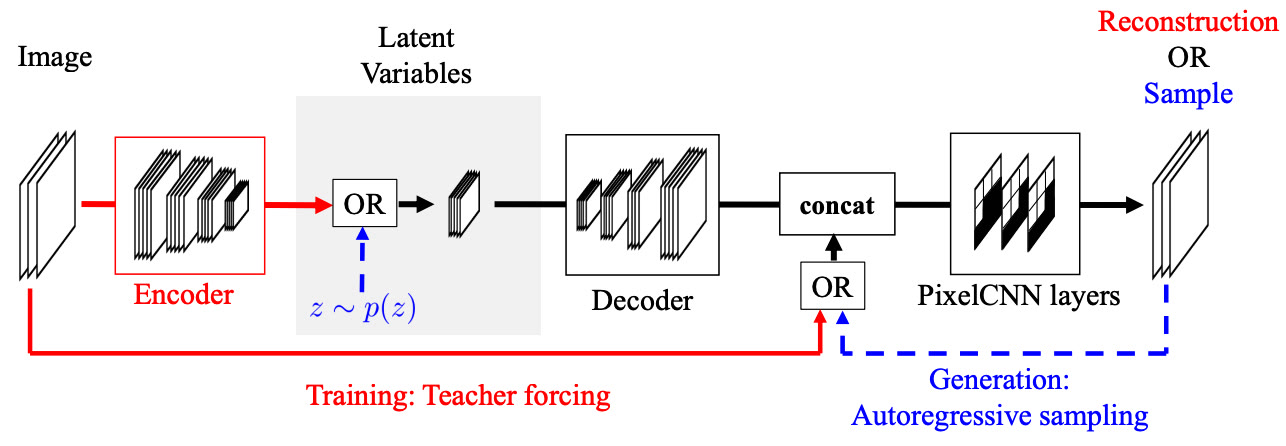}
  \caption{PixelVAE utilizing PixelCNN to generate sharper images. Image adapted from~\cite{gulrajani2016pixelvae}}
  \label{fig:pixel-vae}
\end{figure}

% The model can be conditioned on different types of vectors such as labels, tags, or embeddings from other models:

% \begin{equation}
% p(x \mid h) = \prod_{i=1}^{HW} p(x_i \mid x_{1}, \ldots, x_{i-1}, h),
% \label{eq:x-given-h}
% \end{equation}

% where $h$ is a high-level description represented as a latent vector $h$.
\begin{equation}
p_{\theta}(x \mid z) = \prod_{i=1}^{HW} p_{\theta}(x_i \mid x_{<i}, z), 
\quad z \sim p(z).
\label{eq:z-given-x}
\end{equation}
As in Fig.~\ref{fig:pixel-vae}, the encoder part $q_\phi(z \mid x)$ is like a vanilla VAE, while the decoder $p_\theta(x \mid z)$ is a masked convolution. PixelVAE divides the job, letting $z$ handle global semantics, while PixelCNN in the decoder handles local details.

\subsection{Conditional VAEs}

\begin{figure}[t]
  \centering
  \includegraphics[width=\columnwidth]{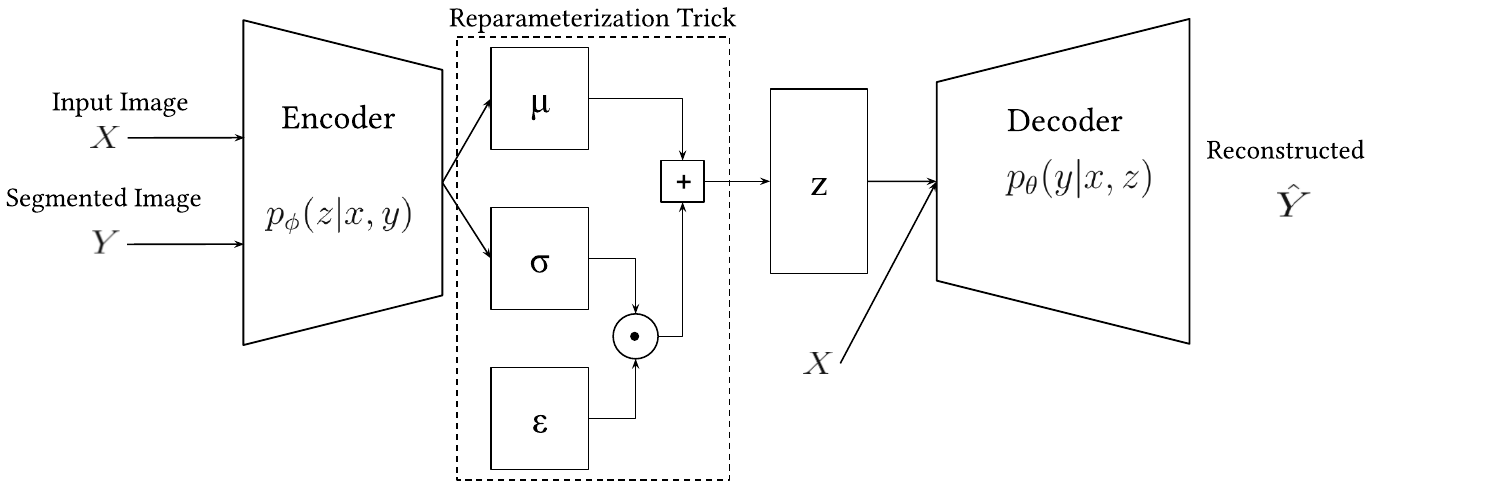}
  \caption{Conditional VAE: One can think of $X$ as the condition upon which the output is being generated.}
  \label{fig:CVAE-diagram}
\end{figure}

So far we have talked about the generation process in VAEs without much emphasis on controlling the generated output. Sohn et~al.~\cite{sohn2015learning} introduced Conditional VAE (CVAE), where the generated output is conditioned on the input. Fig.~\ref{fig:CVAE-diagram} shows a simplified diagram of conditional VAE. Here are two examples where the condition is represented as input \(x\) and a target represented by \(y\) in the \((x,y)\) pairs: (input image, segmented image) or (painted input, realistic photo).
With this setup, the conditional ELBO loss will look like:

\begin{align}
\mathcal{L}_{\text{CVAE}}(x,y;\theta,\phi)
&= - \mathrm{KL}\!\big(q_\phi(z \mid x,y)\,\|\,p_\theta(z \mid x)\big) \nonumber \\
&\quad + \frac{1}{L} \sum_{\ell=1}^{L} 
   \log p_\theta\!\big(y \mid x, z^{(\ell)}\big)
\label{eq:cvae-loss}
\end{align}

where \(L\) is the number of samples and 

\begin{align}
z^{(\ell)} &= g_\phi(x, y, \varepsilon^{(\ell)}), \nonumber \\
\varepsilon^{(\ell)} &\sim \mathcal{N}(0, I) \nonumber
\label{eq:z-sample}
\end{align}
The training and inference steps for conditional VAEs can be simplified as shown in Box~\ref{box:train-infer-vae}.

\begin{boxes}[t]
\begin{tcolorbox}[title=Conditional VAE: Training and Inference Steps, colback=gray!5, colframe=black!75, fonttitle=\bfseries]

\textbf{Training Steps}
\vspace{1em}
\begin{enumerate}[leftmargin=*, itemsep=1em]
\item $\mu_{\text{post}}, \; \sigma^2_{\text{post}} = \text{Encoder}_\phi(x,y)$
\item $z = \mu_{\text{post}} + \sigma_{\text{post}} \odot \varepsilon, 
\quad \varepsilon \sim \mathcal{N}(0,I)$
\item $\mu_{\text{prior}}, \; \sigma^2_{\text{prior}} = \text{Decoder\_Prior}_\theta(x)$
\item $\hat{y}_{\text{prob}} = \text{Decoder}_\theta(x,z)$
\item $\mathcal{L} =
\begin{aligned}[t]
   &\;\; \text{CrossEntropy}(\hat{y}_{\text{prob}}, y) \\
   &+ \mathrm{KL}\!\Big(\mathcal{N}(\mu_{\text{post}},\sigma^2_{\text{post}})
   \;\|\; \mathcal{N}(\mu_{\text{prior}},\sigma^2_{\text{prior}})\Big)
\end{aligned}$
\item Update Encoder $\phi$ and Decoder $\theta$ parameters.
\end{enumerate}

\vspace{1em}
\textbf{Inference Steps}
\vspace{1em}
\begin{enumerate}[leftmargin=*, itemsep=1em]
\item $\mu_{\text{prior}}, \; \sigma^2_{\text{prior}} 
= \text{Decoder\_Prior}_\theta(x)$
\item $z = \mu_{\text{prior}}$
\item $\hat{y}_{\text{prob}} = \text{Decoder}_\theta(x,z)$
\item $\hat{y} = \arg\max(\hat{y}_{\text{prob}})$
\end{enumerate}
\end{tcolorbox}
\caption{Simplified Training and Inference Procedure for the Conditional VAE}
\label{box:train-infer-vae}
\end{boxes}

Kingma et~al.~\cite{kingma2014semi} approached conditional VAEs from a different perspective. They treat labels and attributes as conditioning variables. This lets them fix a label at test time and generate the corresponding image. Training has two parts: a supervised component on the $(x,y)$ pairs that minimizes the negative ELBO loss, and an unsupervised component on $x$ only that minimizes the negative ELBO by summing (marginalizing) over all $y$ via $q_{\phi}(y|x)$. A key benefit is dual use: classification (input $x$ to predict $y$) and generation (choose a desired $y$, then sample $x$ from $p_{\theta}(x|y,z)$ with $z \sim p(z)$).

\subsection{Other Variants of VAEs}
\subsubsection{IWAE: Importance Weighted Autoencoders}

In 2016, Burda et~al.~\cite{burda2015importance} introduced Importance Weighted Autoencoders (IWAE). Their model architecture is the same as the vanilla VAE, but they used a tighter log-likelihood lower bound which is derived from importance weighting. In vanilla VAE, one would estimate the expectation with the following 1-sample ELBO:

\begin{equation}
z^{(1)} = \mu_\phi(x) + \sigma_\phi(x) \odot \varepsilon^{(1)}, 
\quad \varepsilon^{(1)} \sim \mathcal{N}(0,I)
\label{eq:mu-sample}
\end{equation}
IWAE training replaces the objective itself and introduces the \(w_k\) weight as below:

\begin{equation}
w_k = \frac{p_\theta(x, z_k)}{q_\phi(z_k \mid x)}
\label{eq:p-to-q}
\end{equation}

\begin{equation}
\begin{split}
\mathcal{L}_K(x) 
&= \mathbb{E}_{z_{1:K} \sim q_\phi} 
   \left[ \log \frac{1}{K} \sum_{k=1}^K w_k \right], \\
\mathcal{L}_1 
&= \text{ELBO} \;\leq\; \mathcal{L}_K \;\leq\; \log p_\theta(x)
\end{split}
\end{equation}
where $K$ is the number of independent samples drawn from the model. As $K$ grows, $\mathcal{L}_K$ is a tighter lower bound than $\mathcal{L}_1$.

\subsubsection{DRAW: Deep Attention Recurrent Writer}
\begin{figure}
  \centering
  \includegraphics[width=\columnwidth]{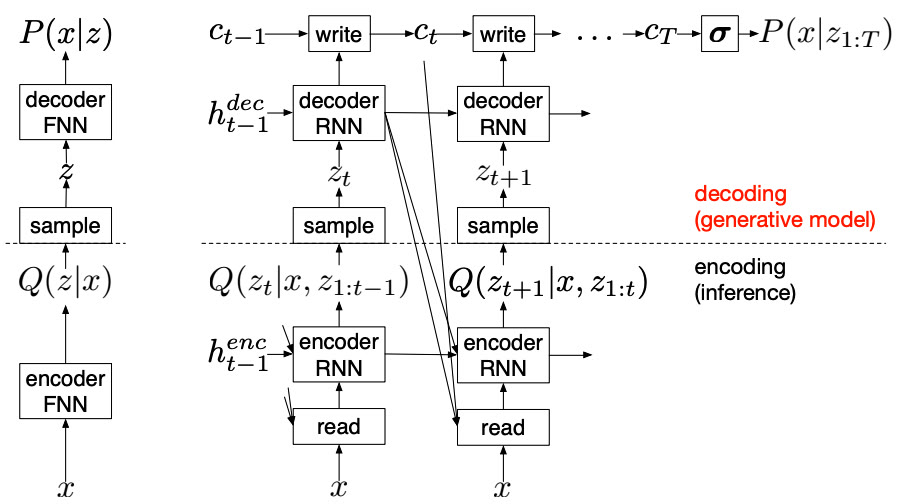}
  \caption{Left side: Conventional Variational Autoencoders. Right side: DRAW Network. Note the iterative process through which the model improves the output of the previous time step. Image adapted from~\cite{gregor2015draw}}
  \label{fig:draw-diagram}
\end{figure}

Gregor and coauthors~\cite{gregor2015draw} introduced Deep Attention Recurrent Writer (DRAW). A recurrent VAE that combines a spatial attention mechanism with a sequential VAE. The iterative process as shown in the right side of Fig.~\ref{fig:draw-diagram}, allows the model to go from coarse to fine image pixels.
\begin{figure}[t]
  \centering
  \includegraphics[width=\columnwidth]{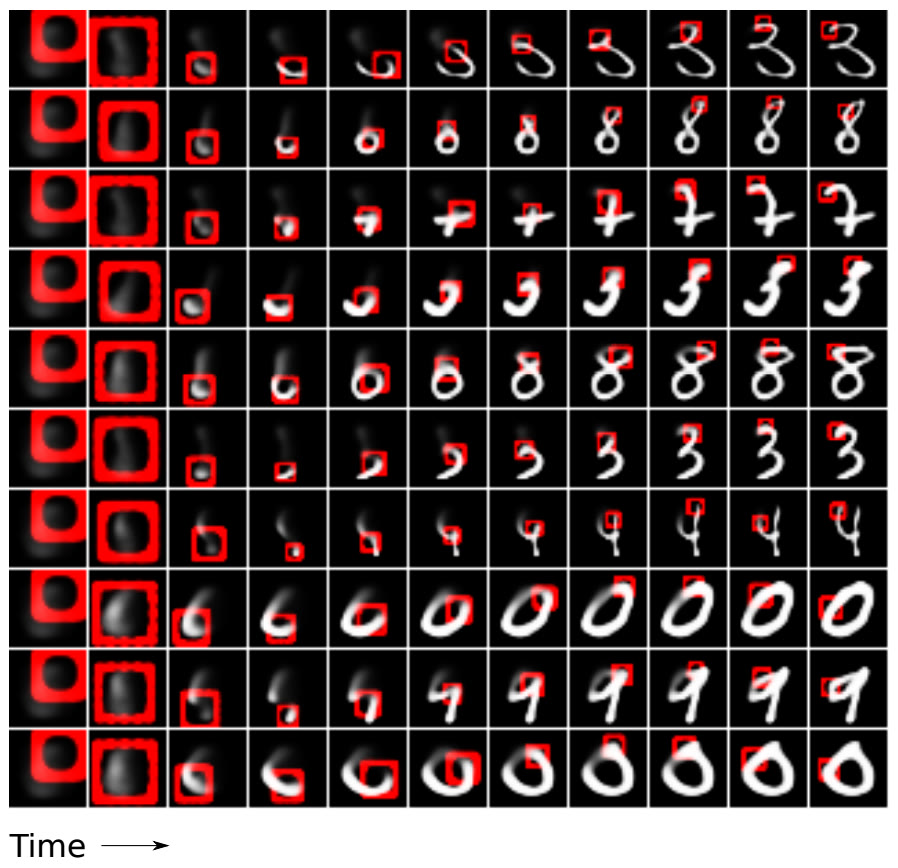}
  \caption{Iterative generation process utilized in DRAW networks. Note how the pixels become sharper through time. Red squares signify the area attended by the network at each time step. Image adapted from~\cite{gregor2015draw}}
  \label{fig:draw-plot}
\end{figure}
Fig.~\ref{fig:draw-plot} shows how pixels become sharper as we move through time. The ELBO loss can be represented as:

\begin{equation}
\text{ELBO} = 
\underbrace{\log p_\theta(x \mid z_{1:T})}_{\text{Bernoulli NLL on }\hat{x}}
-
\underbrace{\sum_{t=1}^T 
   \mathrm{KL}\!\left(q_t(z_t \mid x) \,\|\, p_t(z_t)\right)}_{\text{per-step KL}}
\label{eq:elbo-draw}
\end{equation}

\subsubsection{Vector Quantized VAE}\label{sec:vqvae}
\begin{figure}
  \centering
  \includegraphics[width=\columnwidth]{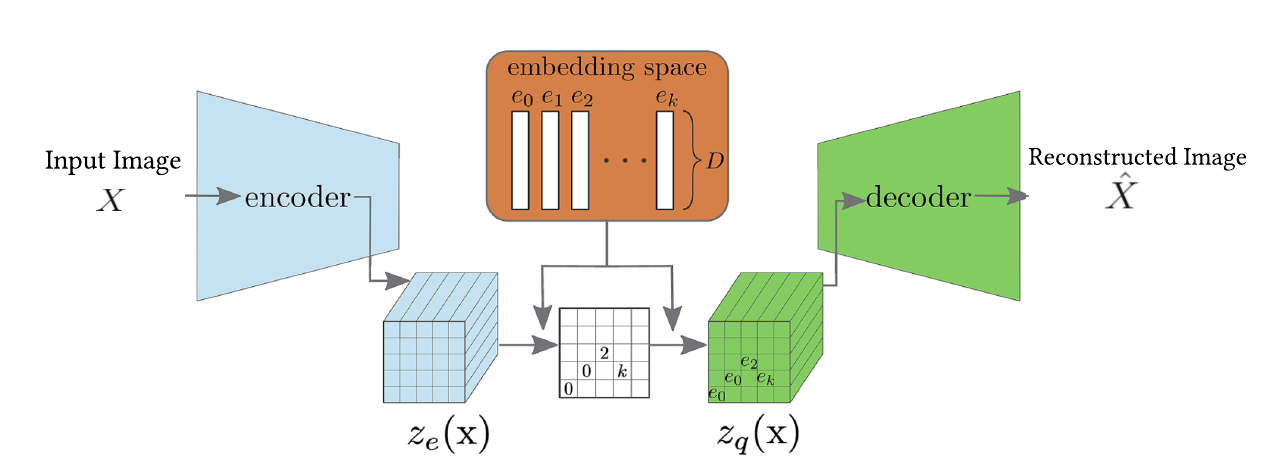}
  \caption{Simplified diagram of the end to end VQ-VAE model architecture. Image adapted from ~\cite{metzger2021deep}}
  \label{fig:vq-vae-diagram}
\end{figure}

In Vector Quantized Autoencoder (VQ-VAE), Oord et~al.~\cite{van2017neural} try to enhance VAEs in two ways. First, the encoder outputs latent vectors that are quantized by selecting entries from a learned codebook. This reduces the need for the encoder to average over fine details and instead it can learn to assign discrete codes to individual image attributes, resulting in sharper reconstructions with less blurry output images. Second, the prior $p(z)$ is learned and is no longer a fixed Gaussian distribution. As we will see in Section~\ref{sec:diffusion}, the overall concept of VQ-VAEs plays a central role in several successful diffusion models. Fig.~\ref{fig:vq-vae-diagram} shows the overall diagram.

% Intuitively, the encoder writes a continuous vector at each latent cell, and the codebook maps it to the nearest code. Since there are only $K$ discrete codes, each chosen code is a clear choice. This leads to a crisper and less fuzzy reconstruction.

As shown in Box~\ref{box:train-vq-vae}, training for VQ-VAE occurs in two stages. In stage 1, the encoder, codebook, and decoder are trained to find the closest value in the codebook that matches the output of the encoder. In stage 2, the encoder and decoder are frozen and the prior is trained over code indices.  An autoregressive prior $p_\psi(c)$ (e.g., PixelCNN, Transformer) can be trained on the code sequences:

\begin{boxes}[t]
\begin{tcolorbox}[title=Training Steps for VQ-VAE,
                  colback=gray!5, colframe=black!75, fonttitle=\bfseries]

\textbf{Stage 1}
\vspace{1em}
\begin{enumerate}[leftmargin=*, itemsep=0.8em]

\item $z_e = \text{Encoder}_\phi(x), \quad z_e \in \mathbb{R}^{D \times H \times W}$

\item Vector Quantization (nearest neighbor per location):  
\[
\begin{aligned}
k^*(u,v) &= \arg\min_k \left\lVert z_e(:,u,v) - e_k \right\rVert_2,\\
z_q(:,u,v) &= e_{k^*(u,v)} \in \mathbb{R}^D
\end{aligned}
\]

\item Straight-Through (ST) estimator for backpropagation: 
\[
z_q^{ST} = z_e + (z_q - z_e)_{\text{stop-grad}}
\]  
Forward path: $z_q^{ST} = z_q$  
Backward path: $\tfrac{\partial z_q^{ST}}{\partial z_e} = I, \quad \tfrac{\partial z_q^{ST}}{\partial z_q} = 0$

\item $\hat{x} = \text{Decoder}_\theta(z_q^{ST})$

\item Loss:  
\[
\mathcal{L} = \mathcal{L}_{\text{reconstruction}}(x, \hat{x}) 
+ \beta \lVert z_e - \mathrm{sg}[z_q] \rVert_2^2
\]  
where $\mathrm{sg}[\cdot]$ denotes the stop-gradient operator.

\item Update parameters $\theta, \phi$ via backpropagation, and update embeddings $E$ via exponential moving average (EMA).

\end{enumerate}
\vspace{2em}
\textbf{Stage 2}
\vspace{1em}
\begin{enumerate}[leftmargin=*, itemsep=0.8em]

\item Freeze the encoder/decoder networks and codebook.

\item Let \(c \in \{1,\dots,K\}^{H\times W} \) be the grid of code indices $c(u,v)=k^*(u,v)$ from Stage 1.

\item Train an autoregressive prior $p_\psi(c)$ (e.g., PixelCNN/Transformer) on the index sequences:
\[
p_\psi(c)\;=\;\prod_{i=1}^{HW} p_\psi\!\big(c_i \,\big|\, c_{<i}\big),
\]

\item Use cross-entropy loss and update the parameters \(\psi\).
\end{enumerate}
\end{tcolorbox}
\caption{Simplified Training Steps for the  VQ-VAE}
\label{box:train-vq-vae}
\end{boxes}

\begin{equation}
p_\psi(c) = \prod_i p_\psi(c_i \mid c_{<i}).
\label{eq:p-thi-c}
\end{equation}

During sampling, they would sample $c \sim p_\psi(c)$ and map the indices to embeddings $z_q$ via the codebook. Finally, use $z_q$ as input to the decoder:

\begin{equation}
\hat{x} = \text{Decoder}_\theta(z_q).
\label{eq:x-decoder}
\end{equation}
For conditional generation, the prior is conditioned on labels/text codes, and the condition is injected into the autoregressive prior.

% \begin{tcolorbox}[title=Generation Steps for VQ-VAE,
%                   colback=gray!5, colframe=black!75, fonttitle=\bfseries]

% \textbf{Reconstruction (given an image $x$)}
% \vspace{0.5em}
% \begin{enumerate}[leftmargin=*, itemsep=1em]
% \item $z_e = \text{Encoder}_\phi(x)$
% \item Quantize: $z_q = \mathrm{VQ}(z_e; E)$ by finding the nearest code
% \item $\hat{x} = \text{Decoder}_\theta(z_q)$
% \end{enumerate}

% \vspace{1em}
% \textbf{Unconditional Generation}
% \vspace{0.5em}
% \begin{enumerate}[leftmargin=*, itemsep=1em]
% \item Sample code indices $c \sim p_\psi(c)$
% \item Lookup embedding: $z_q(u,v) = e_c(u,v)$
% \item Decode: $\hat{x} = \text{Decoder}_\theta(z_q)$
% \end{enumerate}

% \vspace{1em}
% \textbf{Conditional Generation}
% \vspace{0.5em}
% \begin{enumerate}[leftmargin=*, itemsep=1em]
% \item Condition the prior $p_\psi(c \mid \text{cond})$ on label/text
% \item Then follow steps in unconditional generation (2 onward)
% \end{enumerate}

% \end{tcolorbox}
% \vspace{-1em}
% \captionsetup{type=boxes}  % makes it use "Box" counter
% \caption{Simplified Inference/Generation Steps for the  VQ-VAE}
% \vspace{1.5em}
% \label{box:inference-vq-vae}

\subsubsection{Deep Hierarchical VAEs}

Hierarchical Models were introduced by Ranganath, Tran, and Blei~\cite{ranganath2016hierarchical} as a way to capture complex structure for both discrete and continuous latent space variables. Instead of solving one giant hard problem, hierarchical variational models allow big picture latents to guide finer latents so that latents are dependent. 

\begin{figure}[!t]
  \centering
  \includegraphics[width=\linewidth]{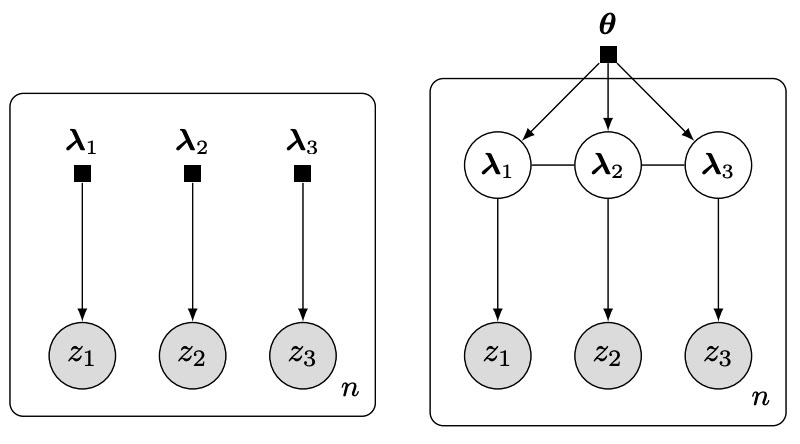}
  \caption{On the right, hierarchical models induce dependence among the latent variables. High-level latents (coarse) set the context for the low-level (fine-detail) latents. Image adapted from~\cite{ranganath2016hierarchical}}
  \label{fig:hier-ranganath-plot}
\end{figure}

In the context of VAEs, in order to increase expressiveness, the latent space ($z$) is partitioned~\cite{kingma2016improved,sonderby2016ladder,klushyn2019learning} into disjoint groups, $z=\{z_1, z_2, ...,z_L\}$ where $L$ is the number of groups. The prior is represented by
\begin{equation}
p(z) = \prod_{i=1}^L p(z_i \mid z_{<i}),
\label{eq:prior}
\end{equation}
and the posterior is represented by
\begin{equation}
q_\phi(z \mid x) = \prod_{l=1}^L q_\phi(z_l \mid z_{<l}, x).
\label{eq:posterior}
\end{equation}

The ELBO loss can be represented as
\begin{align}
\mathcal{L}(\theta, \phi; x) 
&= \mathbb{E}_{q_\phi(z \mid x)} 
   \left[ \log p_\theta(x \mid z) \right] \nonumber \\
&\quad + \mathrm{KL}\!\Big(q_\phi(z_1 \mid x) \,\|\, p(z_1)\Big) \nonumber \\
&\quad - \sum_{l=2}^L 
   \mathbb{E}_{q_\phi(z_{<l} \mid x)} 
   \Big[ \mathrm{KL}\!\big(q_\phi(z_l \mid z_{<l}, x) 
   \,\|\, p(z_l \mid z_{<l})\big) \Big]
\label{eq:elbo-hierarchical}
\end{align}

\begin{figure}[!tbp]
  \centering
  \includegraphics[width=\linewidth]{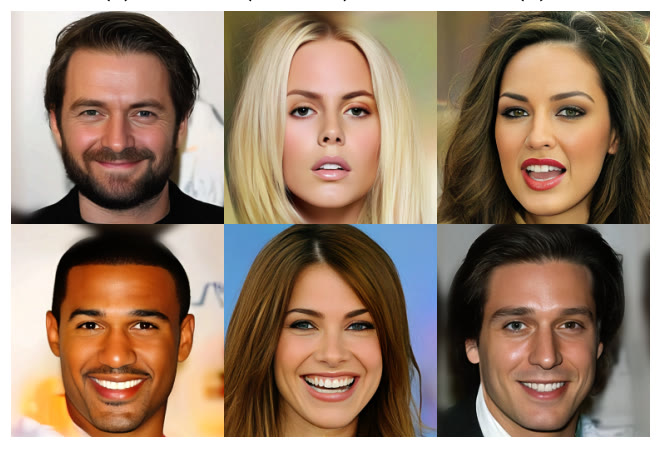}
  \caption{Sample random images generated by NVAE. Image adapted from~\cite{vahdat2020nvae}}
  \label{fig:celeb}
\end{figure}

Later, Vahdat and Kautz~\cite{vahdat2020nvae} introduced Nouveau VAE (NVAE) where they scale up the training into a large number of hierarchical groups and image sizes while keeping the training stable via regularization and reparameterization. Using this approach, they are able to generate high-quality images like Fig.~\ref{fig:celeb} on the CelebA dataset~\cite{liu2018large}

\begin{figure}[!tbp]
  \centering
  \includegraphics[width=\linewidth]{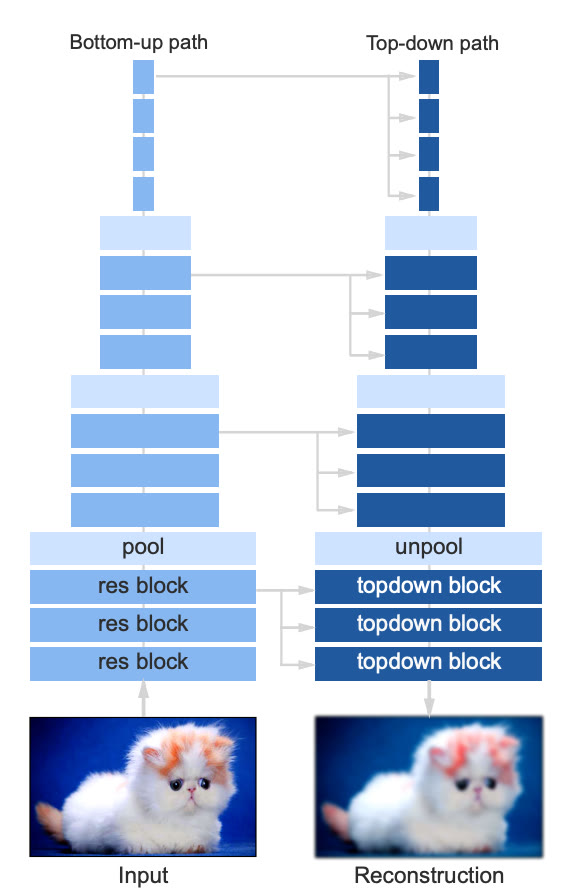}
  \caption{Diagram of VDVAE architecture with top-down VAE architecture with residual blocks. Image adapted from~\cite{child2011very}.}
  \label{fig:vdvae}
\end{figure}

Rewon Child~\cite{child2011very} took this one step further and introduced a Very Deep Variational Autoencoder (VDVAE). By making the hierarchy very deep like in Fig.~\ref{fig:vdvae}, the VAE's top-down chain of priors acts as an autoregressor over the latent space. 
This approach ended up being faster than pixel-space autoregressive versions like PixelCNN since decoding happens at the very end rather than autoregressively. One clever trick they use to make the training more stable is gradient skipping. VAEs are notorious for optimization and convergence issues. They address this by skipping gradient updates that are greater than a certain threshold.

\subsection{Conclusion}
Variational Autoencoders (VAEs) provided a probabilistic framework for generative modeling. They achieved this through a tractable loss function that allows one to pass gradients through stochastic layers.

Although VAEs offered a strong probabilistic foundation for generating high-quality images, they also suffer from several shortcomings. First, as discussed earlier, the ELBO loss is only an approximation of the true likelihood. Moreover, decoders can sometimes ignore the latent space, which can lead to posterior collapse. Lastly, despite the interpretability and controllability advantages of VAEs, they are often difficult to train and optimize. In particular, large gradients can destabilize the training regime. Later, better optimization techniques were utilized in order to make VAE training more stable. As we will see in Section~\ref{sec:diffusion} VQ-VAEs made their way through next generations of image synthesis such as diffusion models and played a crucial role in their success.

\section{Generative Adversarial Networks}

Generative Adversarial Networks (GANs) were introduced by Goodfellow et~al.~\cite{goodfellow2014generative} shortly after Variational Autoencoders. Although different in approach, the goal was similar, generating high-quality images. Their approach uses two models: a generative model \( G \) and a discriminative model \( D \). They used a multi-layer perceptron for both discriminator and generator.

\begin{figure}[h]
  \centering
  \includegraphics[width=\linewidth]{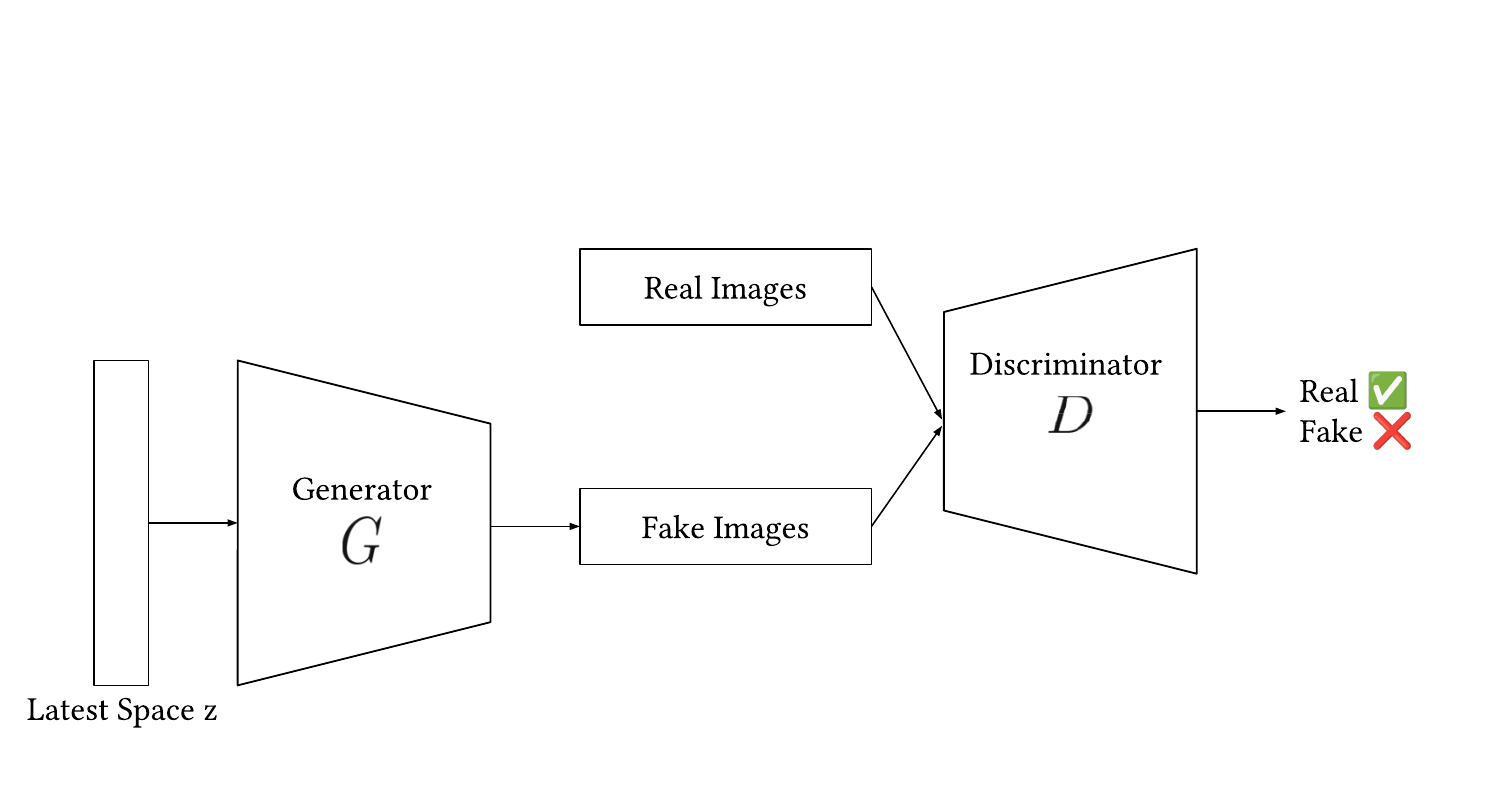}
  \caption{Training pipeline for Generative Adversarial Networks}
  \label{fig:gan-0}
\end{figure}

As shown in Fig.~\ref{fig:gan-0}, the goal of \( G \) is to capture the data distribution by generating images that resemble the training data, and the goal of \( D \) is to discriminate whether a sample came from training data or from the generator \( G \). \( G \) generates images and \( D \) tries to discriminate whether a sampled image is real (from training data) or fake (generated by \( G \)). This adversarial game provides a training signal for \( G \) through \( D \)'s gradient. During inference \( G \) is used to generate images while \( D \) is used only during training.

To borrow an analogy from the authors, the generative model acts as a team of counterfeiters that tries to produce fake currency without being detected. On the other hand, the discriminative model acts as a policeman, trying to detect the counterfeit money. As shown in Fig.~\ref{fig:gan-0}, the generator starts by sampling $z$ that is drawn from $p_z$. The choices for $p_z$ are distributions such as $\mathcal{N}(0, I)$  or uniform $\mathcal{U}(-1, 1)$ since they are easy to sample from and provide randomness. The generator then maps those points to the data space. \( D \) is trained to maximize the probability of assigning the correct label to both training examples and samples from \( G \). \( G \) is trained to minimize \( \log(1 - D(G(z))) \). \( G \) and \( D \) are trained simultaneously using a minimax value function \( V(G, D) \):

\begin{align}
\min_G \max_D V(G, D) 
&= \mathbb{E}_{x \sim p_{\text{data}}(x)}[\log D(x)] \notag \\
&\quad + \mathbb{E}_{z \sim p_z(z)}[\log(1 - D(G(z)))]
\end{align}
One can write the generator and discriminator's cost functions separately as in the following:\\

\noindent\textbf{Discriminator:}
\begin{equation}
\mathrm{Max}:\;\nabla_{\theta_d}\,\frac{1}{m}\sum_{i=1}^{m}
\Bigl[
  \log D(x^{(i)}) + \log\!\bigl(1 - D(G(z^{(i)}))\bigr)
\Bigr]
\end{equation}

\noindent\textbf{Generator:}\\
\begin{equation}
\mathrm{Min}:\;\nabla_{\theta_g}\,\frac{1}{m}\sum_{i=1}^{m}
\log\!\bigl(1 - D(G(z^{(i)}))\bigr)
\end{equation}

Early during training, \( G \) is weak, and \( D \) can easily reject fake samples. As a result, instead of training \( G \) to minimize \( \log(1 - D(G(z))) \), they train \( G \) to maximize \( \log D(G(z)) \). This small change provides a much stronger gradient early in training. Additionally, in training frameworks, the preference is to work with minimization functions; as a result, we can change the objective functions to minimization. Applying these two changes modifies the objective functions as shown in Eq.~\ref{eq:eq-d},~\ref{eq:eq-g}.

\begin{boxes}[t]
\begin{tcolorbox}[title=Training Algorithm for GANs,
                  colback=gray!5, colframe=black!75, fonttitle=\bfseries]

\textbf{for} $N$ training iterations \textbf{do}

\hspace{1.5em} \textbf{for} $k$ steps \textbf{do}

\begin{itemize}[leftmargin=4em]
    \item Sample minibatch of $m$ noise samples 
    $\{z^{(1)}, \ldots, z^{(m)}\}$ from noise prior $p_z(z)$.
    \item Sample minibatch of $m$ examples 
    $\{x^{(1)}, \ldots, x^{(m)}\}$ from data generating distribution $p_{\text{data}}(x)$
    \item Update the discriminator by descending its stochastic gradient:
\end{itemize}

\[
\nabla_{\!\theta_d} \frac{1}{m} 
\sum_{i=1}^{m} 
\Big[ 
-\log D(x^{(i)}) - 
\log \big(1 - D(G(z^{(i)}))\big)
\Big]
\]

\hspace{1.5em} \textbf{end for}

\begin{itemize}
    \item Sample minibatch of $m$ noise samples 
    $\{z^{(1)}, \ldots, z^{(m)}\}$ from noise prior $p_z(z)$.
    \item Update the generator by descending its stochastic gradient:
\end{itemize}

\[
\nabla_{\!\theta_g} \frac{1}{m} 
\sum_{i=1}^{m} 
-\log \big(D(G(z^{(i)}))\big)
\]

\textbf{end for}
\end{tcolorbox}
\caption{Simplified Training Steps for the original GAN paper}
\label{box:training-gan}
\end{boxes}

\noindent\textbf{Discriminator:}
\begin{equation}
\mathrm{Min}\;\nabla_{\theta_d}\,\frac{1}{m}\sum_{i=1}^{m}
\Bigl[
  -\log D(x^{(i)}) - \log\!\bigl(1 - D(G(z^{(i)}))\bigr)
\Bigr]
\label{eq:eq-d}
\end{equation}

\noindent\textbf{Generator:}
\begin{equation}
\mathrm{Min}\;\nabla_{\theta_g}\,\frac{1}{m}\sum_{i=1}^{m}
\bigl(-\log D(G(z^{(i)}))\bigr)
\label{eq:eq-g}
\end{equation}

\begin{figure}
  \centering
  \includegraphics[width=\linewidth]{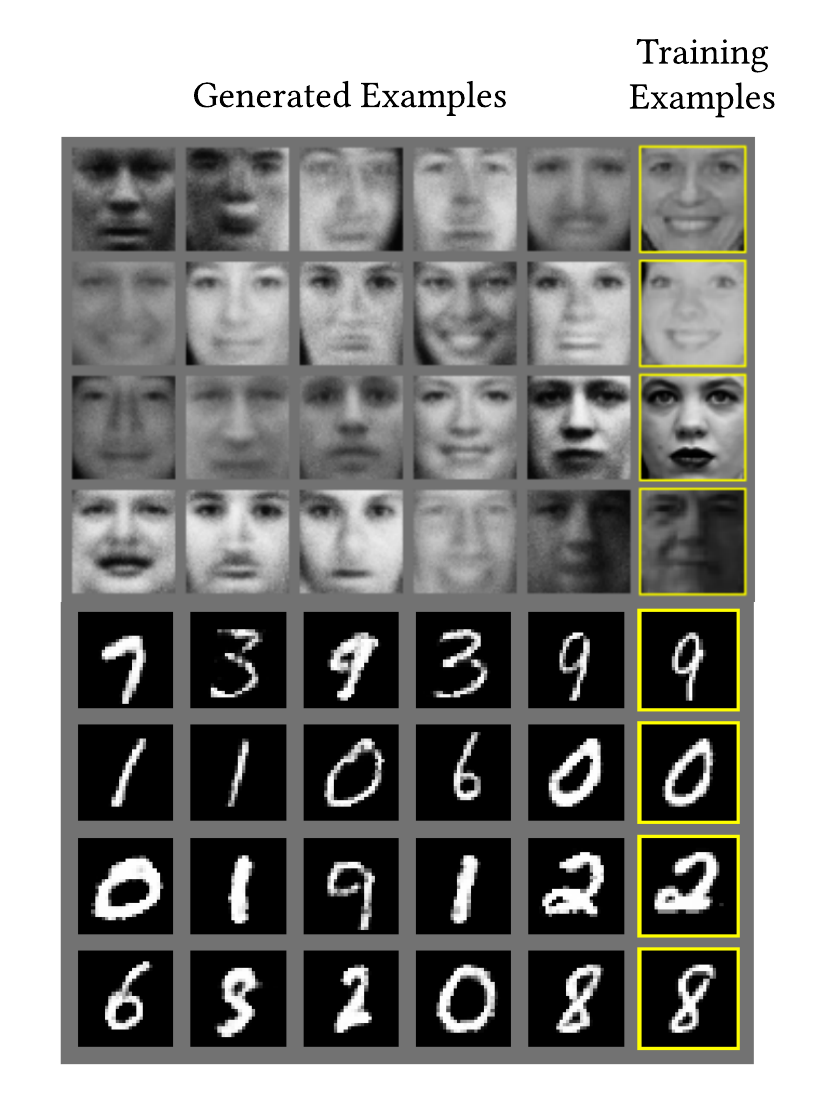}
  \caption{Sample Images Drawn from the Generator $G$. Image source:~\cite{goodfellow2014generative}}
  \label{fig:gan-1}
\end{figure}

One of the major advantages of GANs is the fact that the whole network is trained end-to-end using backpropagation. Another advantage of GANs is that the generator network is not directly updated from data examples, but only with gradients flowing through the discriminator network. This can help since the discriminator can turn data into a differentiable objective for the generator. One major challenge the authors faced was the training instability. For example, the discriminator network $D$ must be in sync with the generator network $G$ during training. If $G$ is trained too much without updating $D$, then $G$ collapses many values of $z$ to the same value of $x$. They remedy this by alternating between \( k \) optimization steps for \( D \) and one optimization step for \( G \). The training process can be simplified as shown in Box~\ref{box:training-gan}. Fig.~\ref{fig:gan-1} shows some of the sample images drawn from the generator $G$.

Building on the success of Convolutional Neural Networks (CNNs) and GANs, Radford et~al.~\cite{radford2015unsupervised} introduced Deep Convolutional Generative Adversarial Networks (DCGANs). They introduced a set of architectural changes to the topology of GANs that offered a more stable training. In fact, the architectural properties of the later GAN papers can be traced back to this paper. They replaced pooling layers with \emph{strided} convolutions in the $D$ and \emph{fractional-strided/transpose} convolutions in the $G$. This tweak allows the model to learn downsampling/upsampling; moreover, it helps with better gradient flow and spatial control. They also utilized batch normalization in intermediate layers of both $G$ and $D$ to stabilize training and prevent early collapse. They also replaced fully connected layers with stacking deep convolutional layers. They only used fully connected layers in the $G$ for projection to a 4D tensor. Lastly, They used \emph{ReLU} in the $G$ (except for \emph{tanh} in the output) and \emph{LeakyReLU} in $D$ for stronger gradients.

\begin{figure}
  \centering
  \includegraphics[width=\linewidth]{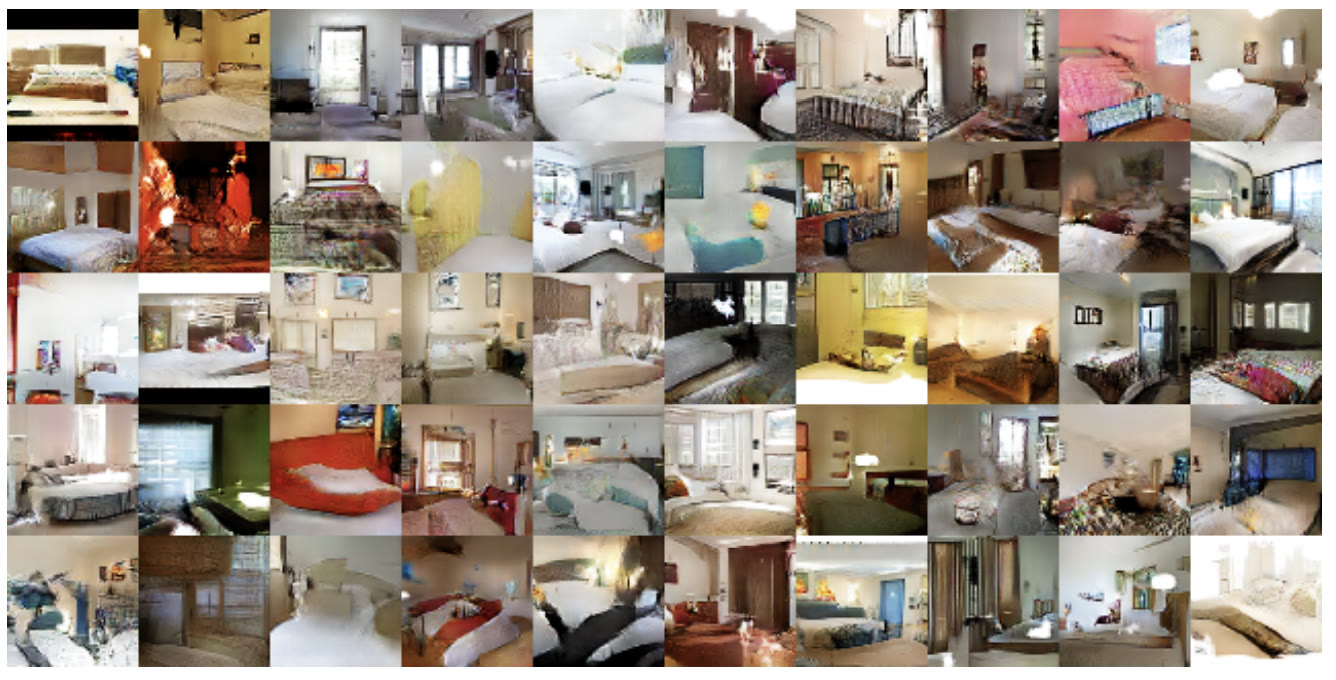}
  \caption{Sample Images Drawn from the Generator $G$ of DCGANs on LSUN dataset~\cite{lsundataset}}
  \label{fig:DCGAN}
\end{figure}

\begin{figure}
  \centering
  \includegraphics[width=\linewidth]{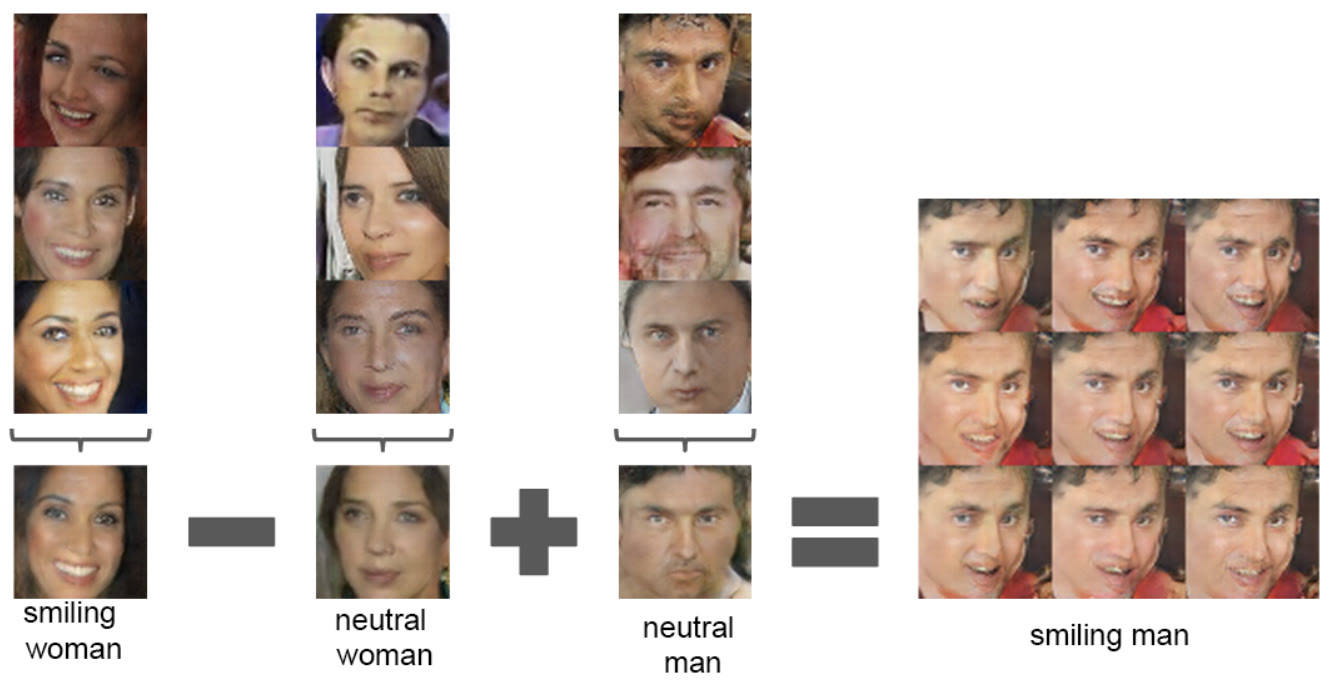}
  \caption{Results confirming that GAN's latent space is organized linearly}
  \label{fig:DCGAN-vector-math}
\end{figure}

Looking at the results of their work, they were able to generate stable $64 \times 64$ images on the LSUN Bedrooms, ImageNet-1k crops, and Faces datasets. Their results, as shown in Fig.~\ref{fig:DCGAN}, seem cleaner than the prior GAN results.
They also analyze latent-space arithmetic. They performed vector arithmetic in the latent space via Eq.~\ref{eq:lat-analyz}:

\begin{equation}
z_{\text{new}} = z_{\text{base}} + (z_{\text{concept A}} - z_{\text{concept B}})
\label{eq:lat-analyz}
\end{equation}
They then fed \( z_{\text{new}} \) to the generator. The result was images that semantically reflected the arithmetic, as shown in Fig.~\ref{fig:DCGAN-vector-math}. This was a strong hint that the GANs’ latent space is organized linearly enough that simple vector operations map to interpretable images. Moreover, it shows that GANs are able to learn features that are disentangled.

\subsection{Training and Objective Function}
A major challenge in working with GANs is achieving stable and reliable training. A lot of research has gone into making the training regime stable. For example, one recurring issue with training GANs is that as the discriminator gets better, the updates to the generator consistently deteriorate.
Salimans and coauthors~\cite{salimans2016improved} introduced techniques to improve the training of GANs. They introduced the \emph{Inception Score (IS)}. Using an inception style neural network~\cite{szegedy2015going}, IS measures how realistic and diverse a set of generated images are. In order to stabilize the generator, they used a virtual batch normalization layer where each example is normalized using a fixed reference batch and the example itself. Standard batch normalization can be problematic, since it makes each output depend on noisy minibatch neighbors. 

They also proposed another approach called \emph{minibatch discrimination} where during mode collapse, $G$ emits near-identical samples, and $D$ does not have a way to penalize the lack of diversity. To alleviate this, they give $D$ features that measure the closeness among samples in the same minibatch. Additionally, in order to damp the spiral of two-player gradient dynamics, they nudge the parameters towards the running average of the past values. Arjovsky et~al.~\cite{arjovsky2017towards} provide a comprehensive analysis of this problem and prove that the data distribution \( p_{\text{data}} \) and the model distribution \( p_g \) live on different low-dimensional manifolds with nearly disjoint support. As a result, an almost perfect discriminator exists. Their recommendation is to smooth the two distributions with noise so that the discriminator cannot perfectly separate real versus generated data.

Arjovsky and coauthors~\cite{arjovsky2017wassersteingan} subsequently introduced the Wasserstein GAN (WGAN). In it, they switched the GAN objective from popular probability distributions like Kullback–Leibler (KL) divergence to the Earth-Mover (EM) distance, or Wasserstein-1 distance. They consider \(P_r\) and \(P_g\) as the real data and generated data distributions and introduce the distance as:

\begin{equation}
W(P_r, P_g) = \inf_{\gamma \in \Pi(P_r, P_g)} \mathbb{E}_{(x, y) \sim \gamma} [ \| x - y \| ]
\end{equation}

where \( \Pi(P_r, P_g) \) is the set of all joint distributions \( \gamma(x, y) \) whose marginals are \( P_r \) and \( P_g \), respectively. 
Here, \( \gamma(x, y) \) denotes the amount of “mass” that has to be transported from \( x \) (real) to \( y \) (generated) such that the two distributions \( P_r \) and \( P_g \) are close. Wasserstein-1 shows the minimum effort required to change the generator’s distribution into the real data distribution.

\begin{figure}
  \centering
  \includegraphics[width=\linewidth]{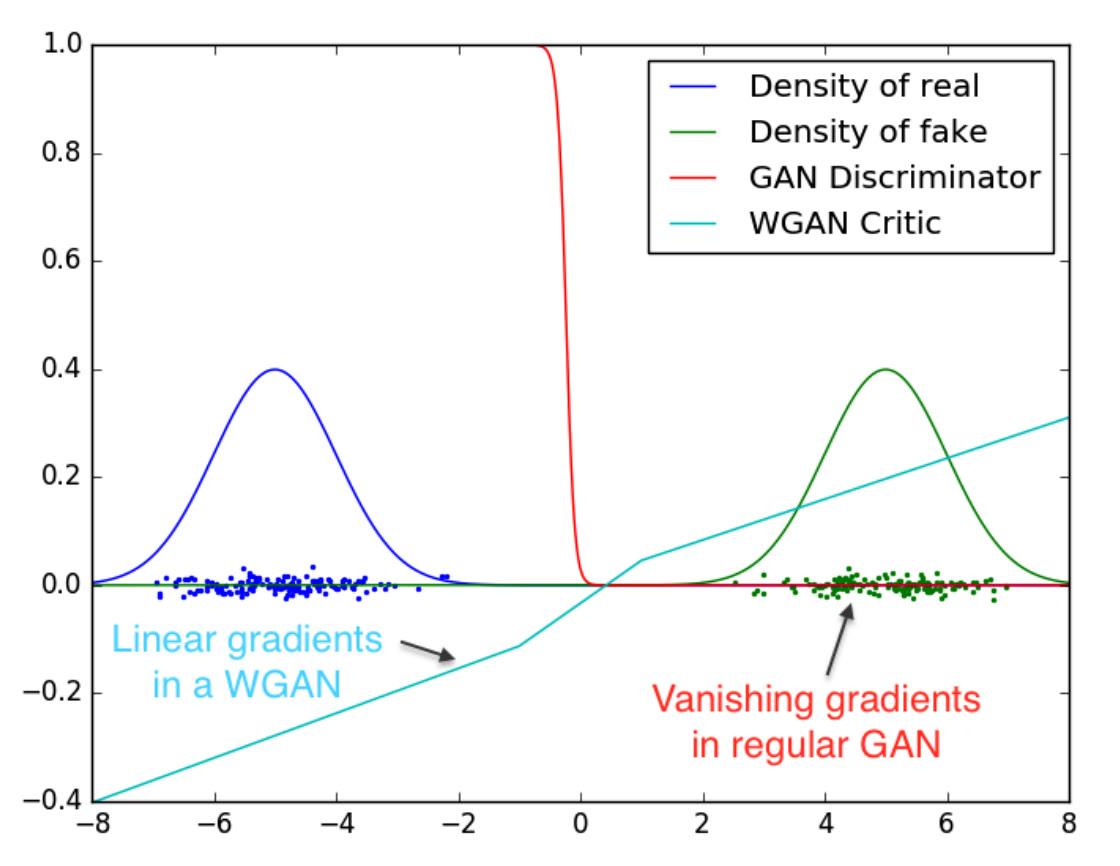}
  \caption{WGAN provides a clear and crisp gradient everywhere. Figure adapted from~\cite{arjovsky2017wassersteingan} }
  \label{fig:wgan}
\end{figure}

Moreover, instead of training a discriminative classifier, WGAN authors train a critic that gives higher scores to real images. The distance between the average real and fake scores is an estimation of the Wasserstein distance, where a smaller gap means closer distributions. They also prevent the critic from changing too fast by clamping every weight to a fixed box after each optimization step. To illustrate the impact of their approach, Fig.~\ref{fig:wgan} shows the non-vanishing gradients across a wider region of parameter space in WGAN compared to the vanishing gradients in regular GANs.

\begin{boxes}[t]
\begin{tcolorbox}[title=Training Algorithm for WGAN-GP,
                  colback=gray!5, colframe=black!75, fonttitle=\bfseries]

\textbf{Assumptions:} The gradient penalty coefficient $\lambda$, the number of critic iterations per generator iteration $n_{\text{critic}}$, the batch size m. Initial critic parameters $w_0$, initial generator parameters $\theta_0$.

\vspace{0.1em}
\textbf{while} $\theta$ has not converged \textbf{do}

\hspace{0.5em} \textbf{for} $t = 1, \ldots, n_{\text{critic}}$ \textbf{do}

\hspace{2em} \textbf{for} $i = 1, \ldots, m$ \textbf{do}

\hspace{3em} Sample real data $x \sim \mathbb{P}_r$

\hspace{7em} latent variable $z \sim p(z)$

\hspace{7em} and a random number $\epsilon \sim \mathcal{U}[0, 1]$.

\hspace{3em} $\tilde{x} \leftarrow G_\theta(z)$

\hspace{3em} $\hat{x} \leftarrow \epsilon x + (1 - \epsilon)\tilde{x}$

\hspace{3em} $L^{(i)} \leftarrow D_w(\tilde{x}) - D_w(x) + 
\lambda (\|\nabla_{\hat{x}} D_w(\hat{x})\|_2 - 1)^2$

\hspace{2em} \textbf{end for}

\hspace{2em} $w \leftarrow \text{Adam}\left(\nabla_w \frac{1}{m}\sum_{i=1}^{m} L^{(i)}\right)$

\hspace{0.5em} \textbf{end for}

\hspace{0.5em} Sample a batch of latent variables $\{z^{(i)}\}_{i=1}^{m} \sim p(z)$ 

\hspace{0.5em} $\theta \leftarrow \text{Adam}\left(\nabla_\theta \frac{1}{m}\sum_{i=1}^{m} -D_w(G_\theta(z^{(i)}))\right)$

\textbf{end while}
\end{tcolorbox}
\caption{Training Algorithm for Wasserstein GAN with Gradient Penalty (WGAN-GP) \cite{gulrajani2017improved}}
\label{box:training-wgan-gp}
\end{boxes}

Clamping weights in WGANs can result in undesirable behaviors like generating poor samples or failing to converge. Gulrajani et~al.~\cite{gulrajani2017improved} proposed an alternative, WGAN-GP. Instead of clipping weights, they penalize the norm of the gradient of the critic with respect to its input. With this small change, they are able to achieve stable training in a wide variety of GAN architectures with no hyperparameter tuning. Box~\ref{box:training-wgan-gp} shows their overall algorithm, which is a slight modification of WGAN.

Mescheder and coauthors~\cite{mescheder2018training} published their work where they analyzed how GANs converge under different scenarios. Their major finding was that unregularized GANs are not locally convergent. This happens because the discriminator keeps non-zero slopes, and the generator updates come from chaining through $D$:

\begin{equation}
\begin{split}
\nabla_{\theta} Loss_G
= \mathbb{E}_{z}\!\Big[
\underbrace{\nabla_{x} D(x)}_{\text{slope of $D$ in input space}}
\Big|_{x=G_{\theta}(z)} \times {}\\
\underbrace{\nabla_{\theta} G_{\theta}(z)}_{\text{how $x$ moves if $\theta$ moves}}
\Big]
\end{split}
\end{equation}

If the generator $G_\theta(z)$ reaches the data distribution (i.e., realistic generation), we would like the update to be zero so that training can stop. The easiest way for that to happen is when $\nabla_x D(x) = 0$ on real data. Then $\nabla_{\!\theta} G \approx 0$, and $G$ does not move. If $\nabla_x D(x) \neq 0$ near real data, then even when $G$ has matched the data, $\nabla_{\!\theta} G$ will not be zero. WGAN attempts to fix this, but it cannot guarantee convergence.

There are two ways to make the training schemes more stable and ensure local convergence. Adding noise to training samples~\cite{sonderby2016amortised} or using zero-centered gradient penalties~\cite{roth2017stabilizing}. Here we briefly go over the zero-centered gradient penalties. The $R_1$ penalty penalizes the squared norm of the input gradient of $D$ at real samples:

\begin{equation}
R_1(D) = \frac{\lambda}{2} \mathbb{E}_{x \sim p_{\text{data}}} 
\left[ \| \nabla_x D(x) \|_2^2 \right]
\end{equation}
Adding this to the discriminator loss function results in:

\begin{equation}
\mathcal{L}_D^{R_1} 
= \mathcal{L}_D 
+ \frac{\lambda}{2} 
\mathbb{E}_{x \sim p_{\text{data}}} 
\left[ \| \nabla_x D(x) \|_2^2 \right]
\end{equation}
By making this change, the training becomes a more tractable optimization. Large gradients mean that $G$ is far from $D$, while shrinking the gradients to zero means that $G$ is closer to $D$.

\subsection{Conditional GANs}

\begin{figure}
  \centering
  \includegraphics[width=\linewidth]{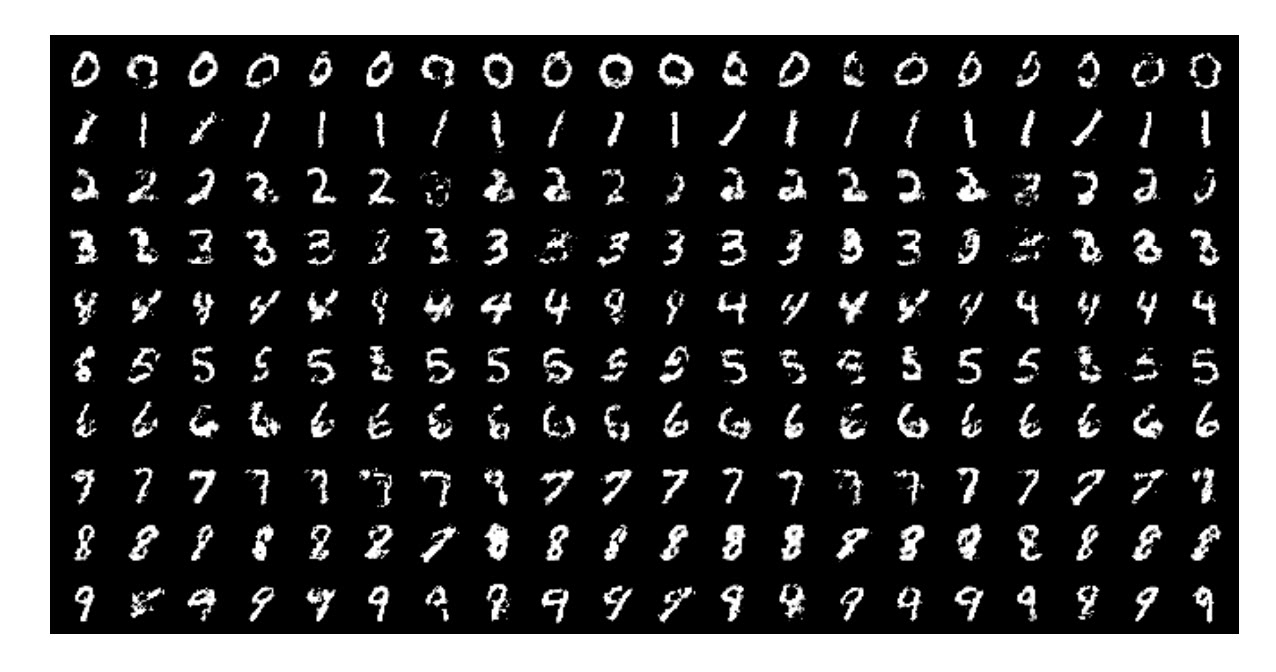}
  \caption{MNIST digits generated are conditioned on labels $0, 1, \ldots, 9$ respectively. Figure adapted from~\cite{mirza2014conditional} }
  \label{fig:cgan-mirza}
\end{figure}

Shortly after the original GAN paper, Mirza and Osindero~\cite{mirza2014conditional} introduced the \emph{Conditional GANs (CGANs)}. CGANs can be constructed by feeding the class label $y$ to both the generator and the discriminator, as shown in the top section of Fig.~\ref{fig:ACGAN-CGAN}. The objective function was represented as:

\begin{figure}
  \centering
  \includegraphics[width=\linewidth]{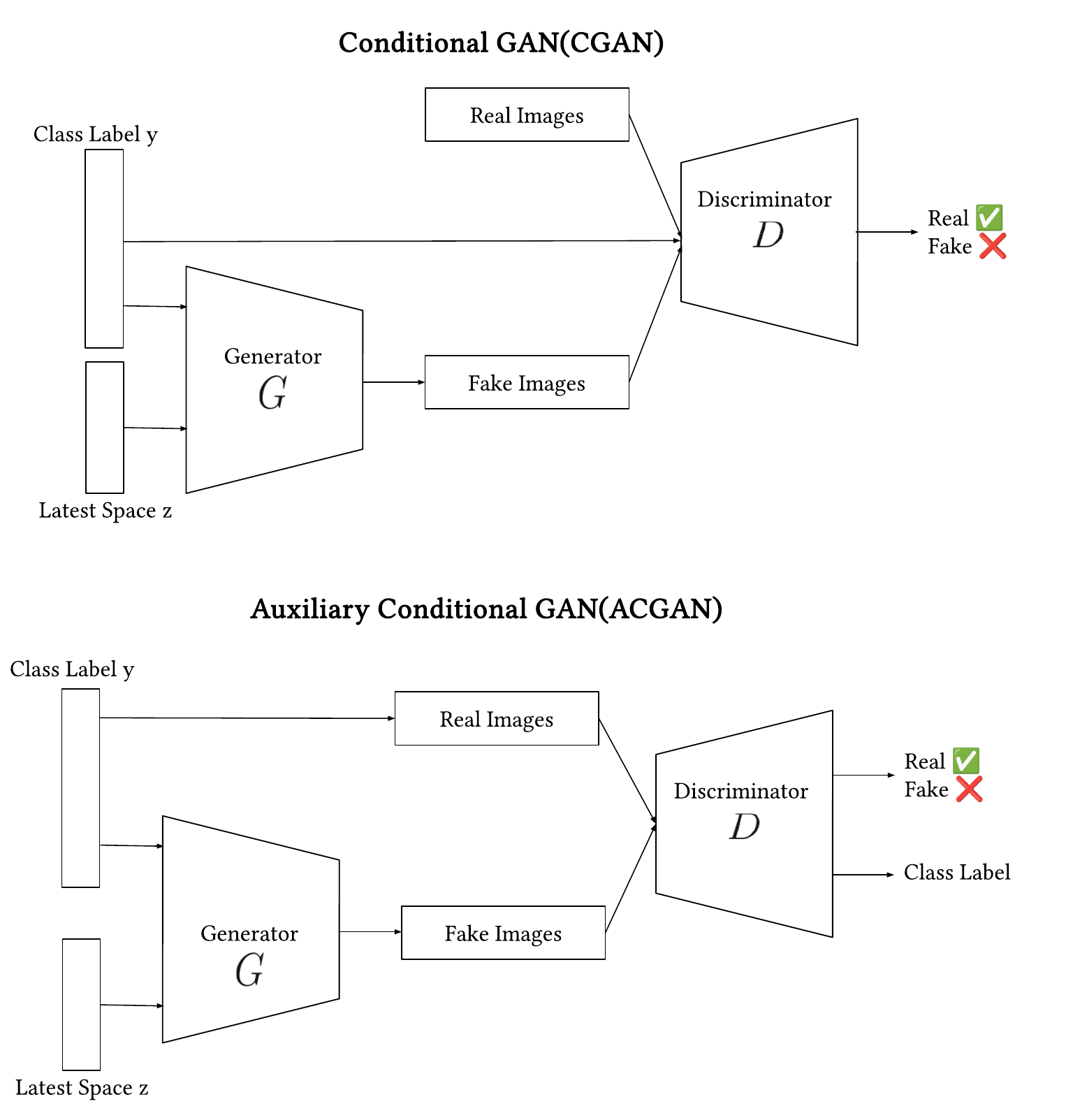}
  \caption{Overall Diagram for Conditional GANs (CGANs) and Auxiliary Classifier GANs (AC-GANs)}
  \label{fig:ACGAN-CGAN}
\end{figure}

\begin{equation}
\begin{split}
\min_G \max_D V(D,G) = 
\mathbb{E}_{x,y \sim p_{\text{data}}(x,y)} 
[\log D(x|y)] \\
+ \mathbb{E}_{z \sim p_z(z)} 
[\log (1 - D(G(z|y)))]
\end{split}
\end{equation}
Fig.~\ref{fig:cgan-mirza} shows sample images generated where each row is conditioned on a label.

Later, Odena et~al.~\cite{odena2017conditional} introduced \emph{Conditional Image Synthesis with Auxiliary Classifier GANs (AC-GANs)} as shown in the bottom section of Fig.~\ref{fig:ACGAN-CGAN}.  
They define $x_{\text{fake}}$ as below, where $c$ is the class label and $z$ is the input noise.

\begin{equation}
x_{\text{fake}} = G(c, z), \quad c \sim p_c, \, z \sim p_z
\end{equation}
The discriminator outputs two distributions for an image $x$: a source head $P(S|X)$ where $S \in \{\text{real}, \text{fake}\}$, and a class head $P(C|X)$. The objectives are defined as:

\begin{align}
L_S &=
\mathbb{E}[\log P(S=\text{real}\mid x_{\text{real}})] +
\mathbb{E}[\log P(S=\text{fake}\mid x_{\text{fake}})], \nonumber\\[1em]
L_C &=
\mathbb{E}[\log P(C=c\mid x_{\text{real}})] +
\mathbb{E}[\log P(C=c\mid x_{\text{fake}})]
\end{align}
During training, the goal is:
\begin{equation}
\max_D (L_S + L_C), 
\quad \max_G (L_C - L_S)
\end{equation}

The source head acts like it did in the original GAN, where $D$ wants to tell real versus fake images, and $G$ wants to fool it. For the auxiliary class, we would like both $G$ and $D$ to classify correctly. Using this approach, they are able to improve image quality as measured by the Inception Score (IS). They are able to achieve an IS of $8.25 \pm 0.07$ compared to $8.09 \pm 0.07$ in~\cite{salimans2016improved} on CIFAR-10.

\begin{figure}
  \centering
  \includegraphics[width=\linewidth]{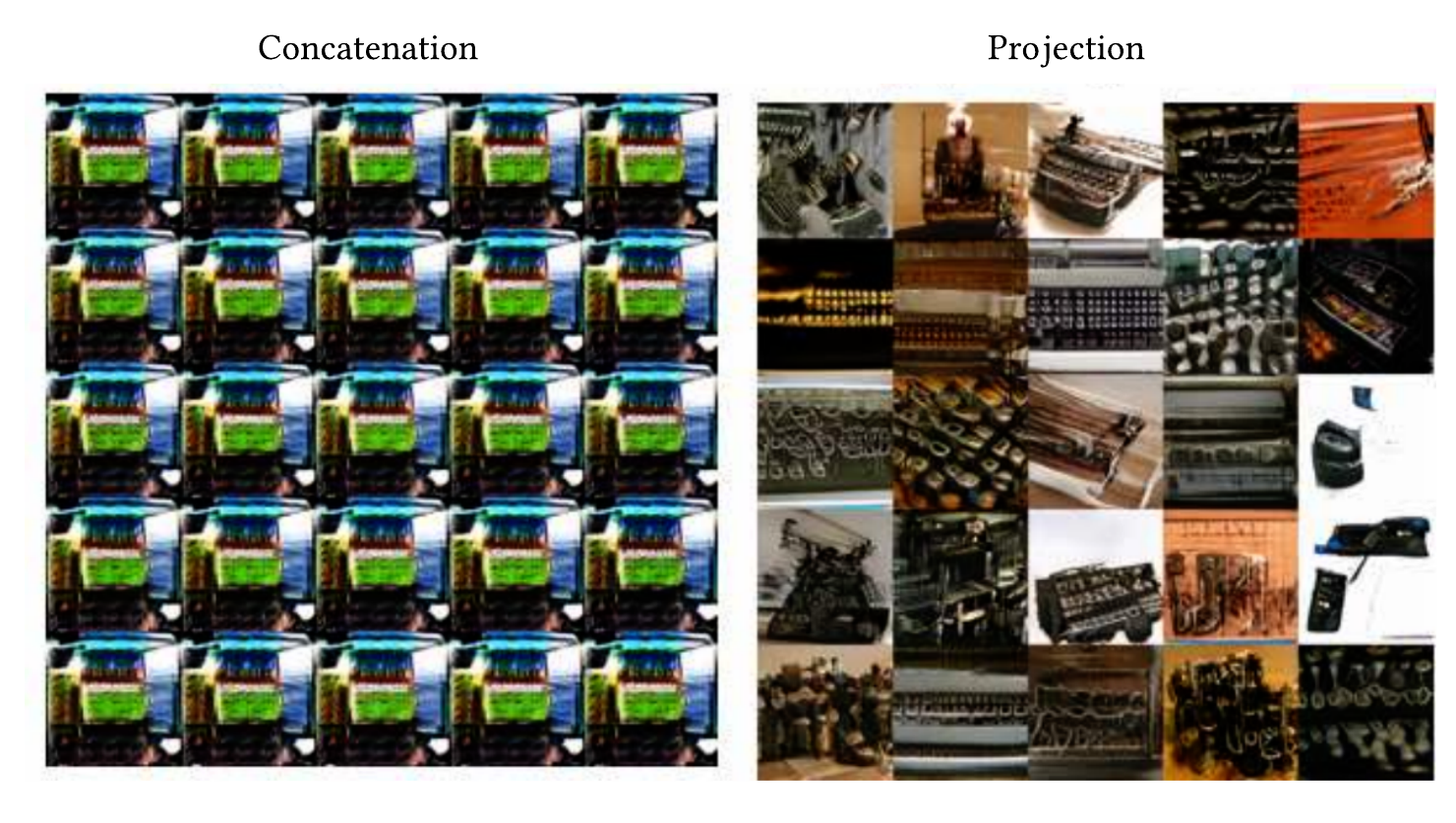}
  \caption{Sample "Typewriter" images generated using the concat and projection models. Note the higher quality of images in the projection model and how all the images in the concat model look the same. Figure adapted from~\cite{miyato2018cgans}.}
  \label{fig:concat-proj-cgan}
\end{figure}

Koyama and Miyato~\cite{miyato2018cgans} modified the prior conditional GAN framework. Instead of concatenating the label $y$ into the discriminator, they define a discriminator score that is the sum of $\langle v_y, \phi(x) \rangle$ and $\psi(\phi(x))$. The first term measures whether the image belongs to class \(y\) and the second term measures whether the image looks real.

\begin{equation}
D(x, y) = \underbrace{\langle v_y, \phi(x) \rangle}_{\text{label-image match}} + 
\underbrace{\psi(\phi(x))}_{\text{unconditional term}}
\end{equation}
where $\phi(x)$ denotes the input image ($x$) features derived from the discriminator network, $v_y$ is the embedding vector learned for class $y$, and $\psi(\cdot)$ is a small network that outputs a scalar measuring how realistic the image looks. Making this adjustment, they are able to achieve an Inception Score of $29.7 \pm 0.61$ compared to $28.5 \pm 0.2$ in AC-GANs on the ImageNet. Refer to Table~\ref{tab:imagenet-is-fid} for more detailed results. Fig.~\ref{fig:concat-proj-cgan} shows how they are able to generate a more realistic photo of a typewriter compared to the previous concatenation methods for labels.

\begin{table}[t]
\centering
\begin{tabular}{lc}
\toprule
\textbf{Method} & \textbf{Inception Score} \\
\midrule
AC-GANs                      & 28.5\,$\pm$\,0.20 \\
concat                       & 21.1\,$\pm$\,0.35 \\
\textbf{projection}          & \textbf{29.7\,$\pm$\,0.61} \\
\midrule
\textit{projection (850K iterations)} & \textbf{36.8\,$\pm$\,0.44} \\
\bottomrule
\end{tabular}
\caption{Comparing the Inception Scores of different conditional GAN models on the ImageNet}
\label{tab:imagenet-is-fid}
\end{table}

\subsection{High Quality Image Generation}
\begin{figure}[h]
    \centering
    \includegraphics[width=\linewidth]{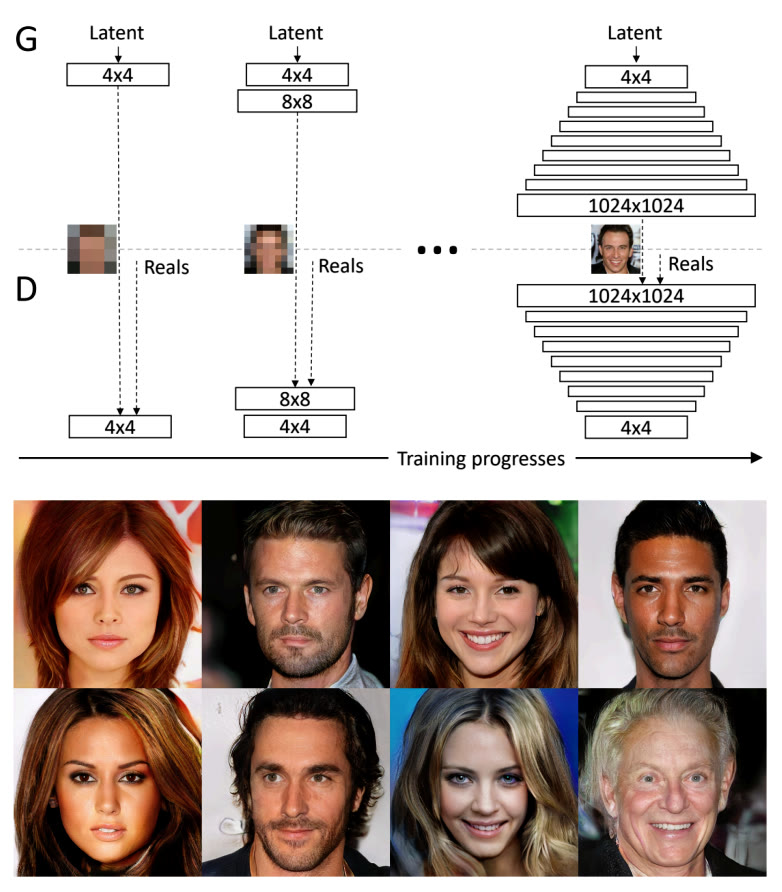}
    \caption{(Top) Karras et~al.~\cite{karras2017progressive} start from low-resolution images and progressively increase the complexity of the models and the size of the input images. (Bottom) Sample images generated from this approach. Image adapted from~\cite{karras2017progressive}.}
    \label{fig:progressive_gan}
\end{figure}

From around 2018 onward, a series of papers were introduced that dramatically increased the quality of generated images and provided more control over the style of the generated output. Karras et~al.~\cite{karras2017progressive} introduced a new way of training GANs. Their training starts with very low-resolution images of $4 \times 4$ pixels. They increase network complexity and image size progressively as training continues. Fig.~\ref{fig:progressive_gan} shows their approach and sample images generated. Following~\cite{salimans2016improved}, they used the minibatch standard deviation to provide a diversity score for images. Additionally, they incorporated pixel-wise vector normalization in the generator.

\begin{boxes}[t]
\begin{tcolorbox}[title=Style Mixing in StyleGAN1, colback=gray!5, colframe=black!75, fonttitle=\bfseries]
\begin{enumerate}
    \item Sample $z_1, z_2 \sim P(z)$.
    \item Pass them through the mapping network: 
    \[
    w_1 = f(z_1), \quad w_2 = f(z_2)
    \]
    \item Pick a random layer $L_c$:  
    \begin{itemize}
        \item Layers $\leq L_c$: use style $w_1$
        \item Layers $> L_c$: use style $w_2$
    \end{itemize}
    \item Repeat this for a random set of minibatches.
\end{enumerate}
\end{tcolorbox}
\caption{Style Mixing in StyleGAN1}
\label{box:stylemixing}
\end{boxes}

After this, a series of \textit{StyleGAN} papers, known as \textit{StyleGAN1}~\cite{karras2019style}, \textit{StyleGAN2}~\cite{karras2020analyzing}, and \textit{StyleGAN3}~\cite{karras2021alias}, were introduced that increasingly improved the quality of the generated images.

\begin{figure}[h]
    \centering
    \includegraphics[width=\linewidth]{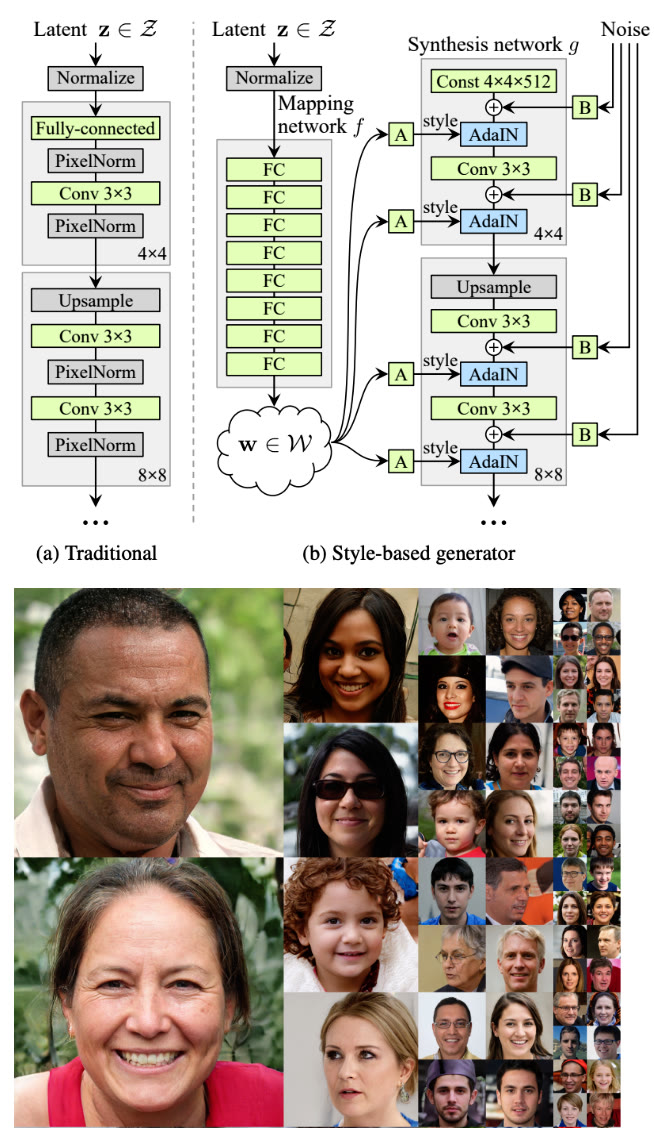}
    \caption{(Top) StyleGAN1 architecture where $z$ is mapped to an intermediate latent $w$. Early layers determine coarse structure, while later layers refine color and details. (Bottom) Sample pictures generated by StyleGAN. Image source:~\cite{karras2019style}}
    \label{fig:stylegan1}
\end{figure}

\begin{figure}[h]
    \centering
    \includegraphics[width=\linewidth]{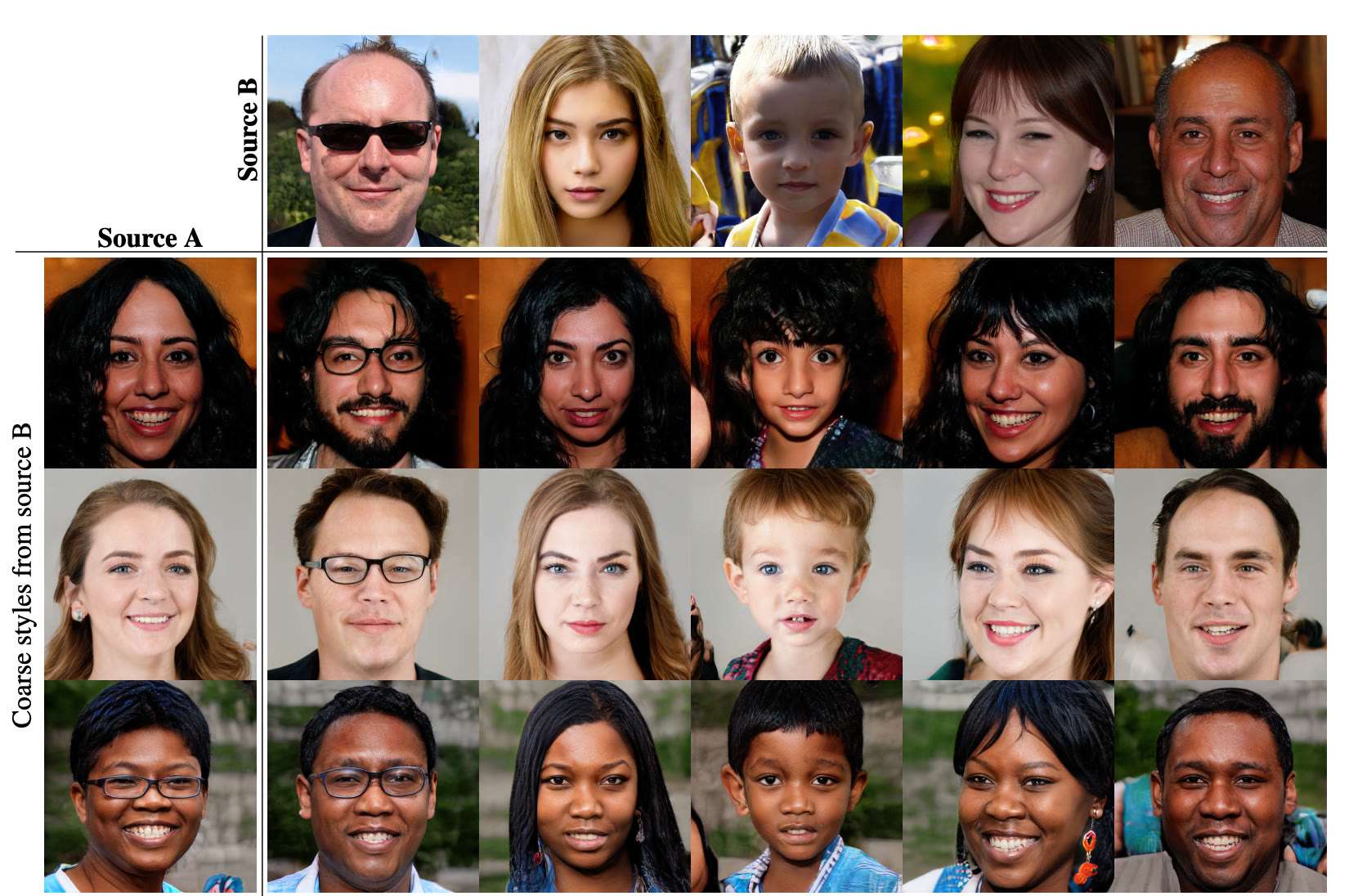}
    \caption{Style mixing in StyleGAN1 demonstrates high-quality image generation by separating coarse and fine styles from two different image styles. Image source:~\cite{karras2019style}}
    \label{fig:stylemixing}
\end{figure}

In StyleGAN1, Karras and coauthors~\cite{karras2019style} borrowed ideas from style transfer architectures. Instead of feeding $z$ directly into the first layer, they mapped $z$ to an intermediate latent vector $w$, as shown in Fig.~\ref{fig:stylegan1}. Early layers set the pose and shape in generated images, while later layers determine color and fine details. 
Another interesting concept they introduced is \emph{style mixing}, summarized in Box~\ref{box:stylemixing}. Fig.~\ref{fig:stylemixing} shows how they are able to generate high-quality variant of source \(A\) images by using images in source \(B\) for coarse style.

\begin{figure}[h]
    \centering
    \includegraphics[width=\linewidth]{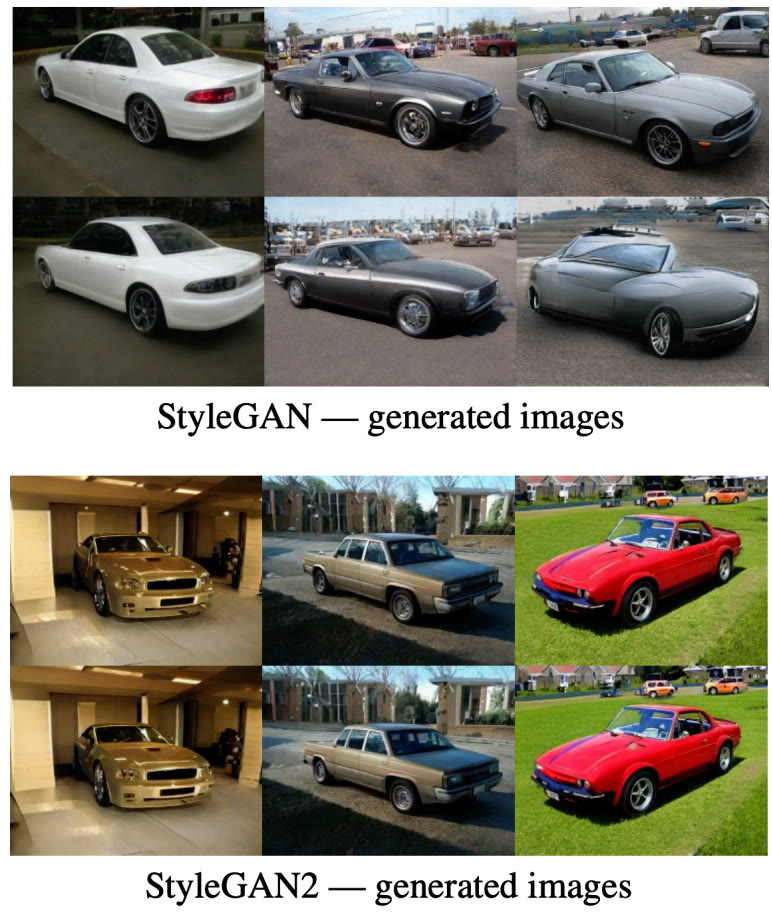}
    \caption{StyleGAN2 improves image quality and reduces artifacts compared to StyleGAN1. Image source:~\cite{karras2020analyzing}}
    \label{fig:stylegan2}
\end{figure}

Karras et~al.~\cite{karras2020analyzing} in \textit{StyleGAN2} proposed several architectural and training improvements to alleviate the artifacts observed in \textit{StyleGAN1}. Fig.~\ref{fig:stylegan2} shows how their method improves image quality and removes artifacts from generated samples. Incorporating these changes allowed them to stabilize and smooth the generator training process, replacing the progressive growing technique of \textit{StyleGAN1} with skip and residual connections.

\begin{figure}[h]
    \centering
    \includegraphics[width=\linewidth]{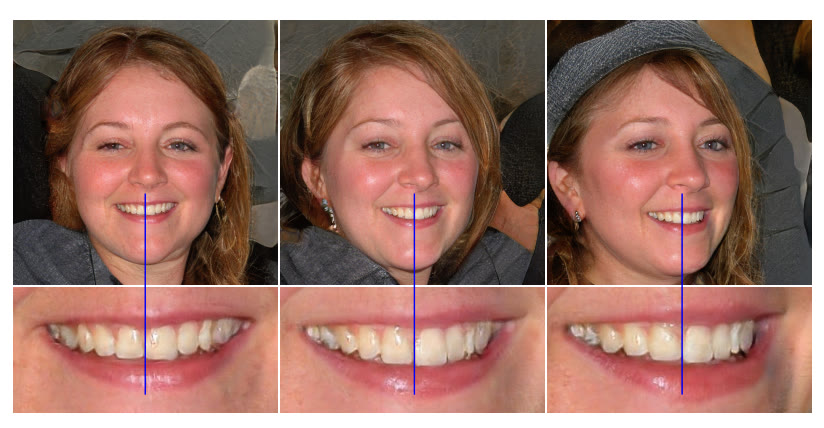}
    \caption{In this image, the teeth should follow the pose, but they stay fixed at a specific pixel coordinate. 
    Image adapted from StyleGAN2~\cite{karras2020analyzing}.}
    \label{fig:texture-sticking}
\end{figure}

\begin{table}[t]
\centering
\small
\renewcommand{\arraystretch}{1.2}
\setlength{\tabcolsep}{3pt}
\resizebox{\columnwidth}{!}{%
\begin{tabular}{@{}l l S[table-format=2.2] S[table-format=2.2] S[table-format=2.2]@{}}
\toprule
\textbf{Dataset} & \textbf{Config} & \textbf{FID $\downarrow$} & \textbf{EQ-T $\uparrow$} & \textbf{EQ-R $\uparrow$} \\
\midrule
FFHQ-U & StyleGAN2 & 3.79 & 15.89 & 10.79 \\
70,000 img, $1024^2$ & StyleGAN3-T (ours) & 3.67 & 61.69 & 13.95 \\
Train from scratch & \textbf{StyleGAN3-R (ours)} & \textbf{3.66} & \textbf{64.78} & \textbf{47.64} \\
\midrule
FFHQ & \textbf{StyleGAN2} & \textbf{2.70} & 13.58 & 10.22 \\
70,000 img, $1024^2$ & StyleGAN3-T (ours) & 2.79 & 61.21 & 13.82 \\
Trained from scratch & StyleGAN3-R (ours) & 3.07 & \textbf{64.76} & \textbf{46.62} \\
\midrule
METFACES-U & StyleGAN2 & 18.98 & 18.77 & 13.19 \\
1,336 img, $1024^2$ & \textbf{StyleGAN3-T (ours)} & \textbf{18.75} & 64.11 & 16.63 \\
ADA, from FFHQ-U & \textbf{StyleGAN3-R (ours)} & \textbf{18.75} & \textbf{66.34} & \textbf{48.57} \\
\midrule
METFACES & StyleGAN2 & 15.22 & 16.39 & 12.89 \\
1,336 img, $1024^2$ & \textbf{StyleGAN3-T (ours)} & \textbf{15.11} & \textbf{65.23} & 16.82 \\
ADA, from FFHQ & StyleGAN3-R (ours) & 15.33 & 64.86 & \textbf{46.81} \\
\midrule
AFHQv2 & StyleGAN2 & 4.62 & 13.83 & 11.50 \\
15,803 img, $512^2$ & \textbf{StyleGAN3-T (ours)} & \textbf{4.04} & 60.15 & 13.51 \\
ADA, from scratch & StyleGAN3-R (ours) & 4.40 & \textbf{64.89} & \textbf{40.34} \\
\midrule
BEACHES & StyleGAN2 & 5.03 & 15.73 & 12.69 \\
20,155 img, $512^2$ & \textbf{StyleGAN3-T (ours)} & \textbf{4.32} & 59.33 & 15.88 \\
ADA, from scratch & StyleGAN3-R (ours) & 4.57 & \textbf{63.66} & \textbf{37.42} \\
\bottomrule
\end{tabular}%
}
\caption{StyleGAN3 improving FID and EQ scores compared to StyleGAN2. Results adapted from~\cite{karras2021alias}.}
\label{tab:stylegan3-results}
\end{table}

In StyleGAN3, the authors faced a challenge known as \emph{texture sticking}, where image details were locked to a specific pixel coordinate as shown in Fig.~\ref{fig:texture-sticking}. To address this, the authors borrowed ideas from signal processing to remove aliasing effects. They incorporated anti-aliasing techniques inside the generator. For instance, they avoided input features that could encode pixel coordinates. Additionally, during upsampling or downsampling, they applied anti-aliasing filters. Lastly, when nonlinearities were applied, they first increased the internal resolution, applied the nonlinearity, and then filtered back. This ensured that image features did not get stuck at specific pixel coordinates. Using these techniques, they were able to improve both FID and EQ scores compared to StyleGAN2 set forth in Table~\ref{tab:stylegan3-results}. The Fréchet Inception Distance (FID) was introduced by Heusel et~al.~\cite{heusel2017gans} and it measures how close the distribution of generated images is to the distribution of real images. Equivariance scores, EQ-T~\cite{zhang2019making} and EQ-R~\cite{karras2021alias}, measure whether image features stick to the image or to the screen as the image rotates or translates. Ideally we would like the features to stick to the image as they are rotated or translated and not the screen(i.e., fixed pixel location) like in Fig.~\ref{fig:texture-sticking}.

\subsection{Other Applications of GANs}

\begin{figure}[t]
    \centering
    \includegraphics[width=\linewidth]{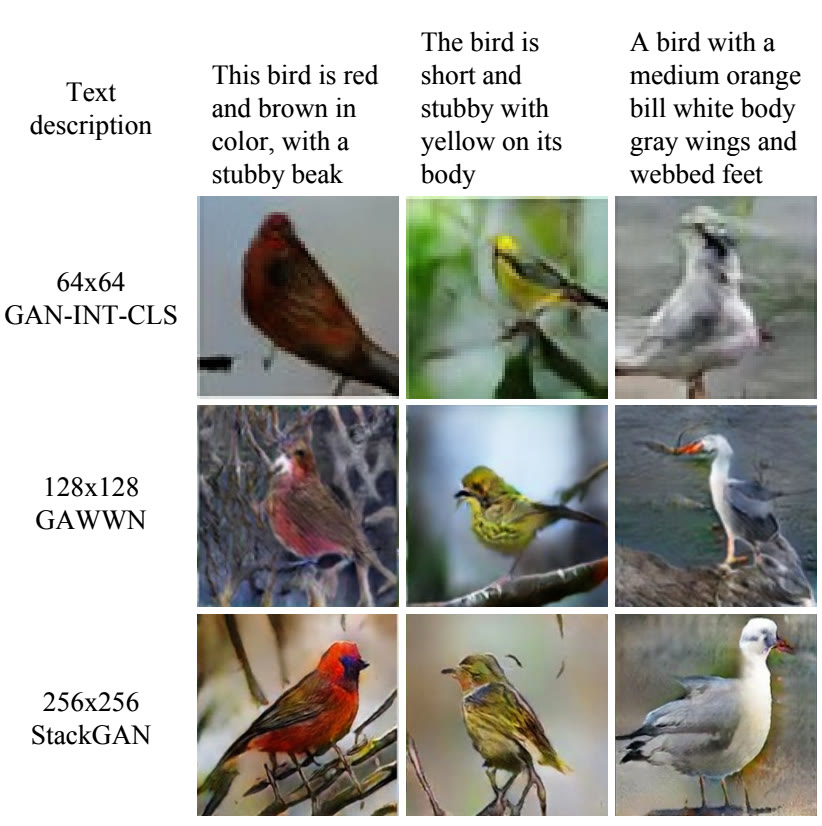}
    \caption{Example results by StackGAN, GAWWN~\cite{reed2016learning}, and GAN-INT-CLS~\cite{reed2016generative}.}
    \label{fig:stackgan_results}
\end{figure}

Thus far, when we talked about GANs, we mainly focused on generating images via random inputs $z$, or by concatenating a class label $y$ to $z$. Zhang et~al.~\cite{zhang2017stackgan} introduced StackGAN, where they synthesize high-quality images from text descriptions. They concatenate the text embeddings $\phi_t$ to the input noise $z \sim N(0,1)$ and pass that to the Stage-I generator. Conditioned on the results of Stage-I, Stage-II refines the details and generates a more realistic image. Fig.~\ref{fig:stackgan_results} shows some of the results of their work (last row) compared to GAWWN~\cite{reed2016learning} and GAN-INT-CLS~\cite{reed2016generative}.

\begin{figure}[h]
    \centering
    \includegraphics[width=\linewidth]{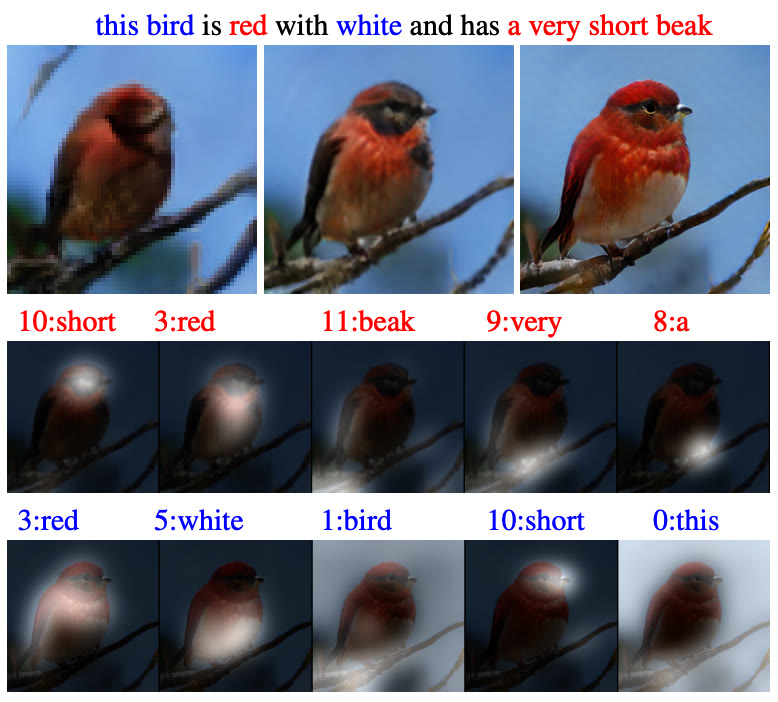}
    \caption{Results from AttnGAN. Images generated in 3 stages of generation. Note the most attended words in the 2nd and 3rd rows.}
    \label{fig:attngan_results}
\end{figure}

Xu and coauthors~\cite{xu2017attnganfinegrainedtextimage} utilized a similar multi-stage approach coupled with an attention mechanism in order to generate fine-grained images from input text. Fig.~\ref{fig:attngan_results} shows example results of their proposed AttnGAN. The first row shows images generated through multi-stage generation. The second and third rows show the most attended words by the attention layer.

\begin{figure}[h]
    \centering
    \includegraphics[width=\linewidth]{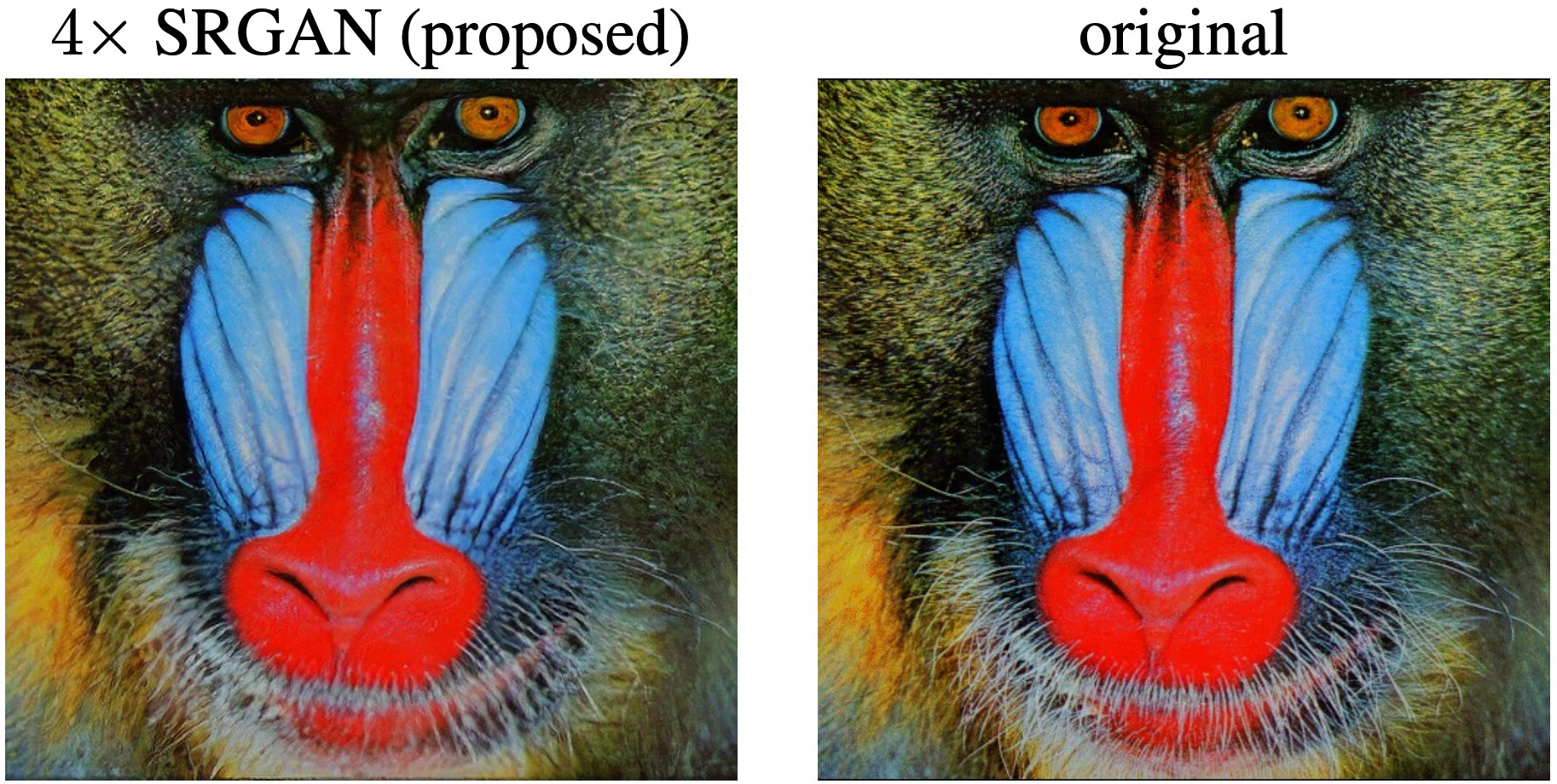}
    \caption{Image on the left is a 4x upsampled version of the image on the right. Image source: ~\cite{ledig2017photo}}
    \label{fig:srgan_results}
\end{figure}

Another area that GANs are useful is Super Resolution, increasing the quality and resolution of an image. During this process, the model adds realistic details that might not have been present in the original image. Ledig and coauthors~\cite{ledig2017photo} used GANs for super-resolution, they define the loss function as

\begin{equation}
\mathcal{L}_{\text{loss}} = 
\underbrace{\| \Phi(I_{HR}) - \Phi(G(I_{LR})) \|_2^2}_{\text{content loss}} 
+ \lambda \underbrace{(-\log D(G(I_{LR})))}_{\text{GAN term}}
\end{equation}

where $\Phi(\cdot)$ is a pre-trained VGG~\cite{simonyan2014very} network that acts as a feature extractor. $I_{LR}$ and $I_{HR}$ are low- and high-resolution images, respectively. The parameter $\lambda$ balances realism and faithfulness. Larger $\lambda$ corresponds to sharper details but comes with a higher risk of hallucination, while smaller $\lambda$ leads to smoother details but fewer hallucinations. Using this approach, they are able to recover photo-realistic textures from downsampled images, as shown in Fig.~\ref{fig:srgan_results}.

\subsection{Conclusion}
Generative Adversarial Networks (GANs) have had a long-lasting impact on image generation. They started in 2014 with low-quality (but still impressive) results, and over the next four to five years, the quality of images improved dramatically. The idea was original and quickly drew attention, but it came with downsides. First, as discussed earlier, they are hard to optimize, and the generator can fall into mode collapse, where many outputs look the same. Second, GAN training can be sensitive to optimizer and hyperparameter choices. Lastly, with limited data, the discriminator can overfit and the generator will suffer greatly from this.

% Lastly, metrics such as the Inception Score(IS) and Fréchet Inception Distance (FID) may not reliably correlate with the perceptual quality of the images.

\begin{figure}[h]
    \centering
    \includegraphics[width=\linewidth]{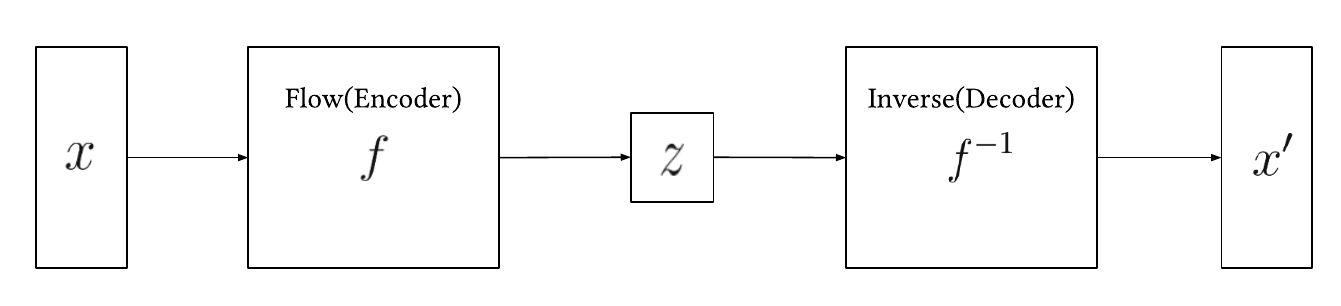}
    \caption{Overall diagram of a Normalizing Flow. $f$ and $f^{-1}$ share the same neural network parameters.}
    \label{fig:flow-diagram}
\end{figure}

\section{Normalizing Flows}
Around 2015, generative image models moved from niche to mainstream research interest. VAEs, GANs, and Normalizing Flows (NF) were introduced within two years of each other. The general idea behind normalizing flows is to transform data to a simple distribution (e.g., Gaussian) via an invertible mapping without losing any information~\cite{agnelli2010clustering,papamakarios2021normalizing}. To achieve this goal, a step by step transformation is learned in order to transform the input into a simple distribution such that each step is reversible. Dinh et~al.~\cite{dinh2014nice,dinh2016density} popularized the idea in the context of generative image modeling by passing the input through a transformation such that the resulting output fits a distribution \(p_Z\) that consists of $d$ independent components, $z_i$. The authors’ chosen distribution was Gaussian or logistic.

\begin{equation}
p_Z(z) = \prod_{i=1}^{D} p_{Z_i}(z_i)
\end{equation}

The transformation $z = f(x)$ is invertible where $x = f^{-1}(z)$. One of the key design choices they make is the choice of \(f\). It must be invertible and the dimension of $z$ is the same as $x$. Fig.~\ref{fig:flow-diagram} shows the simplified diagram of normalizing flows. $f$ and $f^{-1}$ share the same neural network parameters. Using the change of variable formula leads to (for proof refer to ~\cite{weng2018flow}):

\begin{equation}
p_X(x) = p_Z(f(x)) \left| \det \frac{\partial f(x)}{\partial x} \right|
\end{equation}
where \(\frac{\partial f(x)}{\partial x}\) is the Jacobian matrix of function \(f\) at \(x\). The transformation $f$ is chosen such that the determinant of the Jacobian and the inverse of $f$ can be easily calculated. The core idea behind this is that $x$ can be split into two parts \((x_{keep}, x_{shift})\) and applying transformation leads to \((y_{keep},y_{shift})\) as shown below:

\begin{equation}
y_{\text{keep}} = x_{\text{keep}}, \qquad
y_{\text{shift}} = x_{\text{shift}} + f(x_{\text{keep}})
\label{eq:eq-nice1}
\end{equation}
The inverse is obtained by:

\begin{equation}
x_{\text{keep}} = y_{\text{keep}}, \qquad
x_{\text{shift}} = y_{\text{shift}} - f(y_{\text{keep}})
\label{eq:eq-nice2}
\end{equation}
Here, \(f\) is a neural network and the same neural network parameters are used for the inverse process as well. Moreover, since the Jacobian is triangular, the determinant is tractable. They compose an \(L\)-layer transformation as follows in order to keep the Jacobian determinant tractable:

\begin{equation}
f = f_L \circ \cdots \circ f_2 \circ f_1
\end{equation}
The Jacobian determinant is the product of the Jacobian determinants of all layers.

\begin{equation}
\log \left| \det J_{f_{\theta}} \right| = \sum_{l=1}^{L} \log \left| \det J_{f_l}(z^{(l-1)}) \right|
\end{equation}
where \(z^{(0)} = x\) and \(z^{(l)} = f_l(z^{(l-1)})\). Finally, the exact log likelihood can be calculated via

\begin{equation}
\log p_X(x) = \log p_Z(f_\theta(x)) + \log \left| \det J_{f_\theta}(x) \right|
\end{equation}

\begin{figure}[t]
    \centering
    \includegraphics[width=\linewidth]{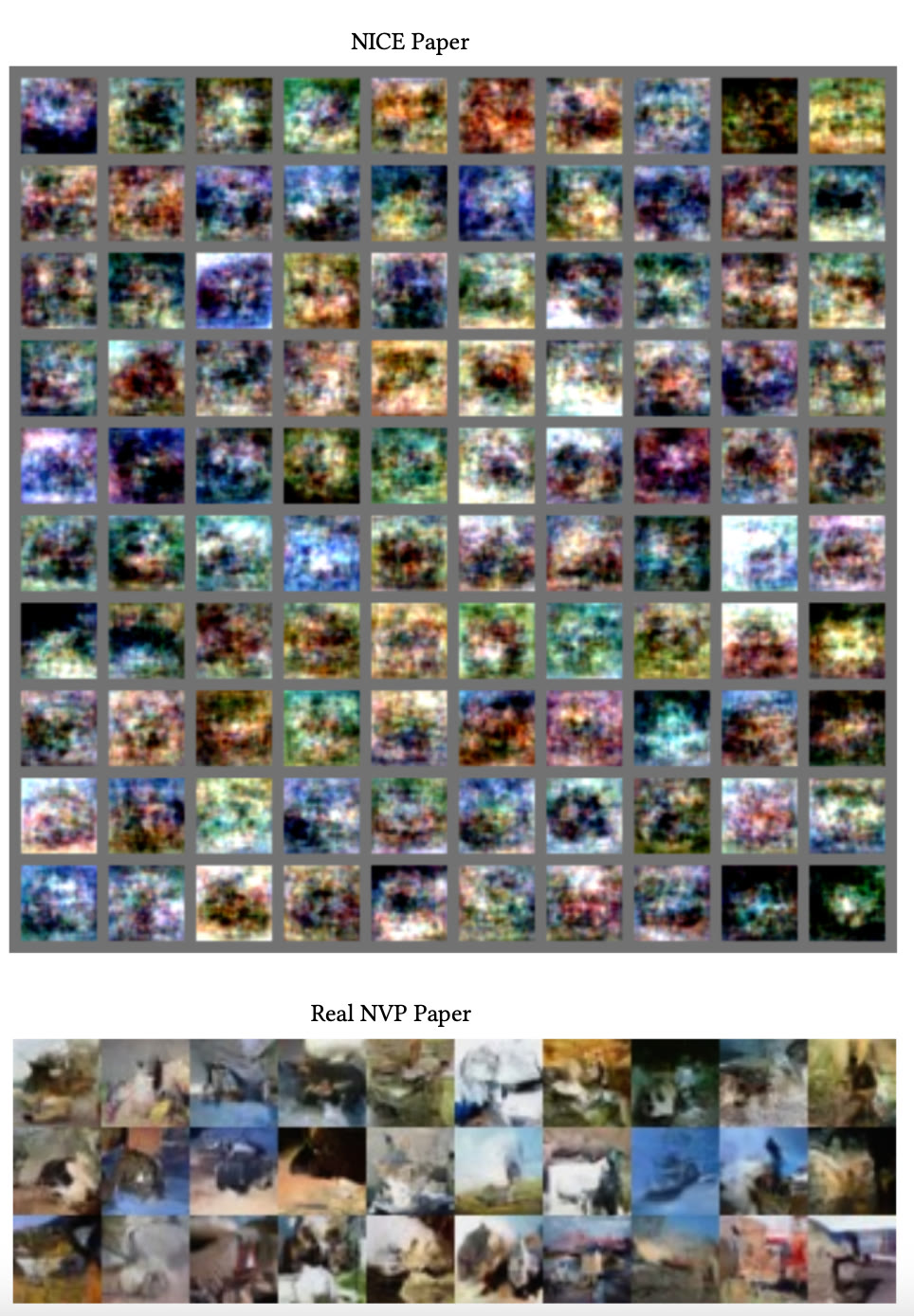}
    \caption{Unbiased samples from NICE(up) and Real NVP(down) trained on CIFAR-10. You can notice the improvement in image quality going from NICE to Real NVP.}
    \label{fig:NICE-RealNVP}
\end{figure}

In practice, since the determinant is 1, they add a scale at the end to allow the model to give more weight to some dimensions and less on others. The overall training and generation steps of the NICE~\cite{dinh2014nice} paper can be summarized as in Box~\ref{box:train-infer-flows}. Fig.~\ref{fig:NICE-RealNVP} shows unbiased samples from a NICE model. They sample $z \sim  p_Z(z)$ and output $x=f^{-1}(z)$.

\begin{boxes}[t]
\begin{tcolorbox}[title=Normalizing Flows: Training and Generation Steps, colback=gray!5, colframe=black!75, fonttitle=\bfseries]

\textbf{Training Steps}
\vspace{0.7em}

\begin{enumerate}[leftmargin=*, itemsep=1em]

\item Encode via invertible network layers and scale
\[
z = f_\theta(x)
\]

\item Compute prior log-density
\[
\log p_Z(z) = \sum_i \log \mathcal{N}(z_i; 0, 1)
\]

\item Compute log-Jacobian determinant
\[
\log \left| \det J_{f_\theta}(x) \right|
= \sum_{l} \log \left| \det J_{f_l}(z^{(l-1)}) \right|
\]
where \(z^{(0)} = x\) and \(z^{(l)} = f_l(z^{(l-1)})\).

\item Compute log-likelihood and loss over \(B\) batches
\[
\log p_X(x) 
= \log p_Z(z) 
+ \log \left| \det J_{f_\theta}(x) \right|
\]

\[
\mathcal{L}(\theta)
= -\frac{1}{|B|} \sum_{x \in B} \log p_X(x)
\]

\item Update parameters using Adam.
\end{enumerate}

\vspace{3em}
\textbf{Generation Steps}
\vspace{0.7em}

\begin{enumerate}[leftmargin=*, itemsep=1em]
\item Draw latent sample \(z \sim p_Z(z)\), e.g. \(\mathcal{N}(0, I)\).
\item Invert the flow to reconstruct data \(x = f_\theta^{-1}(z)\)
by reversing layers.
\end{enumerate}
\end{tcolorbox}
\caption{Training and Generation Procedure for Normalizing Flows}
\label{box:train-infer-flows}
\end{boxes}

Dinh et al.~\cite{dinh2016density} published a follow-up paper, "Density Estimation Using Real NVP," where they improve their previous results~\cite{dinh2014nice}. They made a few changes compared to their earlier work (NICE). They switched from additive coupling as shown in Eq.~\ref{eq:eq-nice1} to affine coupling, where

\begin{equation}
y_{1:d} = x_{1:d}
\end{equation}

\begin{equation}
y_{d+1:D} = x_{d+1:D} \odot \exp(s(x_{1:d})) + t(x_{1:d})
\end{equation}
where $s$ and $t$ are scale and translation, and $\odot$ is the element-wise product. To reduce computation, they incorporate a masking scheme both across spatial $(x,y)$ and channel dimensions. Moreover, they utilize residual CNNs that lead to a more stable training at scale. Adding batch normalization also helped them with training deeper networks. Applying these changes enabled them to generate higher quality images compared to NICE as shown in Fig.~\ref{fig:NICE-RealNVP}.

\subsection{Other Variants of Normalizing Flows}
Rezende and Mohamed~\cite{rezende2015variational} introduced Normalizing Flows in the context of VAEs to improve the expressiveness of the posterior distribution. They used flows to parameterize the posterior in VAEs. They started from a base density $q_0(z_0|x)$ and passed it through a sequence of invertible maps

\begin{equation}
z_k = f_k \circ \cdots \circ f_1(z_0)
\end{equation}
The result is a rich posterior $q(z_k|x)$. This leads to a tighter bound for the ELBO loss and improves end-to-end training. Moreover, this process is computationally cheap due to the invertible layers.

Kingma et al.~\cite{kingma2016improved} subsequently introduced Inverse Autoregressive Flows (IAF). They used an inverse autoregressive transform (such as PixelCNN) in order to capture the rich dependencies. Their autoregressive model is more expressive and results in a very rich latent space. They let an autoregressive model output two unconstrained real-valued vectors

\begin{equation}
[m_t, s_t] \leftarrow \text{AutoregressiveModel}_t(z_{t-1}, z; \theta)
\end{equation}
and compute \(z_t\) as

\begin{equation}
\begin{aligned}
\sigma_t &= \operatorname{sigmoid}(s_t) \\
z_t &= \sigma_t \odot z_{t-1} + (1 - \sigma_t) \odot m_t
\end{aligned}
\end{equation}
Update at each step allows the posterior to express itself more freely. To be consistent with Normalizing Flows, the log-determinant is a simple sum of \(\log \sigma\) where \(D\) is the dimensionality of the vector \(z_t\).

\begin{equation}
\log q(z_T|x)
= -\sum_{i=1}^{D} \left( \left( \frac{1}{2} \epsilon_i^2 + \frac{1}{2}\log (2\pi) \right)
+ \sum_{t=0}^{T} \log \sigma_{t,i} \right)
\end{equation}

\begin{figure}[h]
\centering
\includegraphics[width=\linewidth]{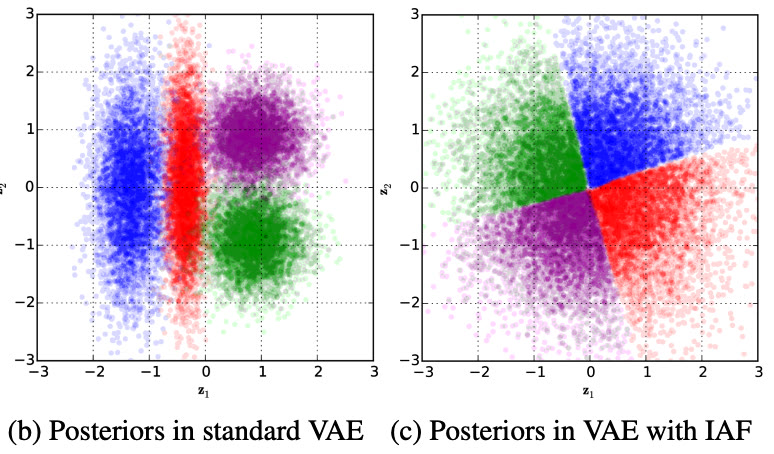}
\caption{IAF VAE results in a clearer fit of posterior distribution compared to normal VAE.}
\label{fig:iaf_vs_vae}
\end{figure}

Using this approach, they are able to scale to a rich, high-dimensional latent space. Fig.~\ref{fig:iaf_vs_vae} shows the flexibility of the posterior distribution in IAF VAE compared to the standard VAE. Following this work, Papamakarios and coauthors~\cite{papamakarios2017masked} introduce Masked Autoregressive Flows (MAF). They combined ideas from Masked Autoencoder for Distribution Estimation(MADE)~\cite{germain2015made} and Inverse Autoregressive Flows (IAF) and propose a normalizing flow model that is capable of density estimation, i.e. the underlying probability distribution of the dataset~\cite{huang2018neural}.

\begin{figure}[h]
\centering
\includegraphics[width=\linewidth]{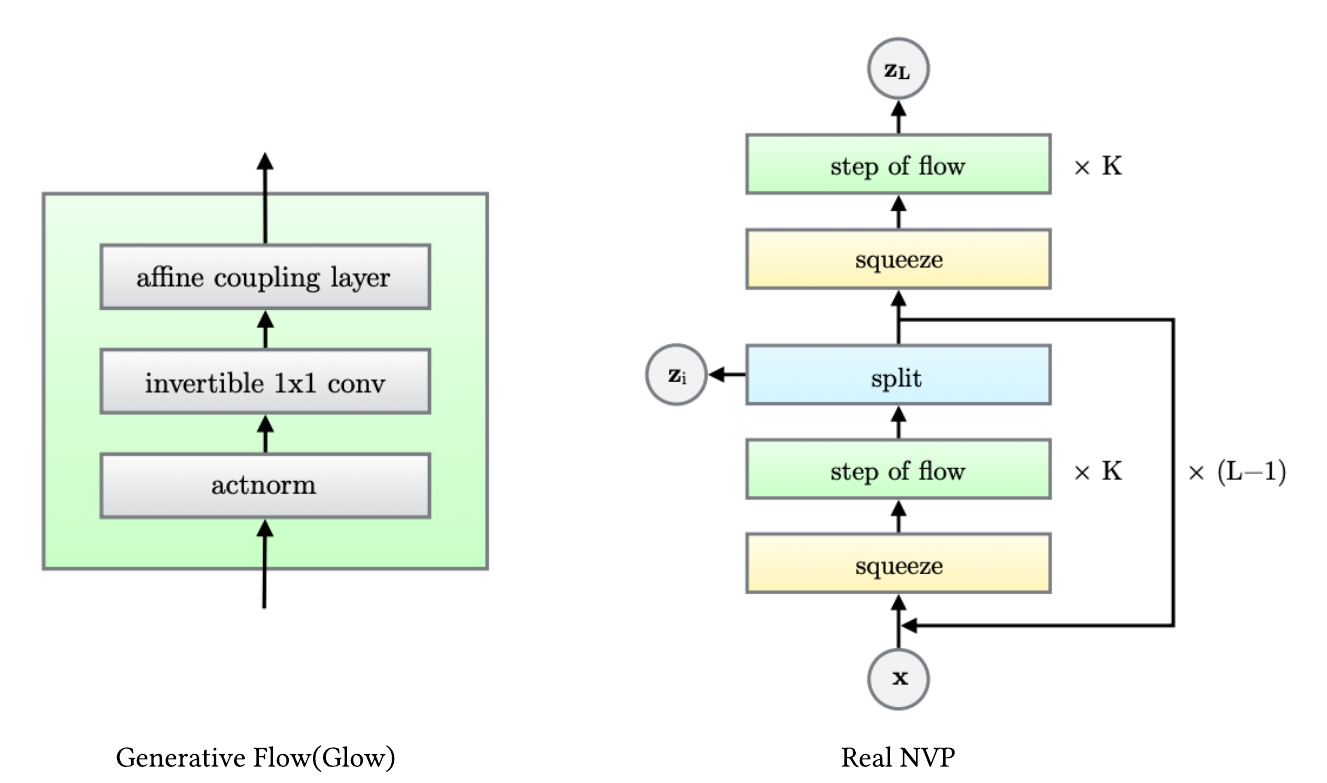}
\caption{Glow architecture compared to RealNVP.}
\label{fig:glow_arch}
\end{figure}

\begin{figure}[h]
\centering
\includegraphics[width=\linewidth]{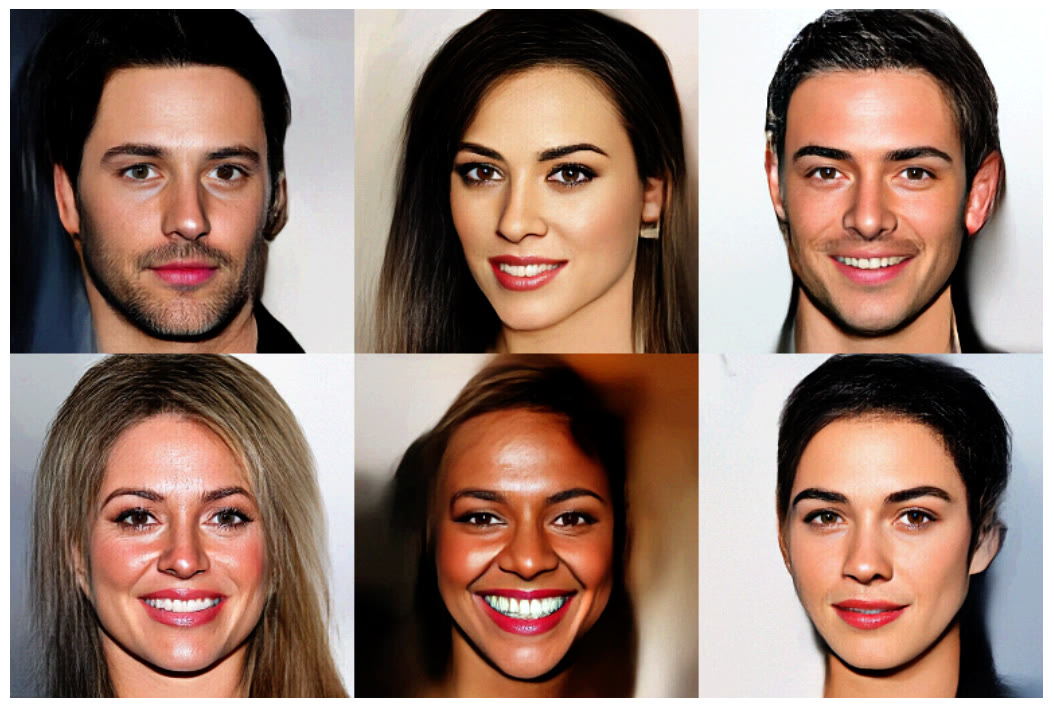}
\caption{Generated images of celebrities with Generative Flow (Glow).}
\label{fig:glow_samples}
\end{figure}

Later, Kingma and Dhariwal~\cite{kingma2018glow} improve the previous state of the art results in Normalizing Flows and produce high fidelity sample images. They propose three changes to Normalizing Flow architectures. First, they replace the batch normalization layer with activation normalization (ActNorm). ActNorm performs an affine transformation of activations using a per-channel scale and bias. They set it up such that activations are zero-mean and unit-variance on the first minibatch, afterwards the model learns. This change has a stabilizing impact on deep networks and large images. Second, they replace the fixed permutation used in prior work~\cite{dinh2016density} with a learnable invertible $1 \times 1$ convolution layer. Third, they maintain the affine coupling layer introduced by Nice/RealNVP~\cite{dinh2016density,dinh2014nice}, but zero-initialize the last convolution layer of each neural network. They perform splits only along the channel dimension to simplify the overall architecture. Fig.~\ref{fig:glow_arch} shows a comparison of their architecture compared to RealNVP~\cite{dinh2016density}. By making these changes, they are able to generate realistic images such as those shown in Fig.~\ref{fig:glow_samples}.

Ho, Chen et al.~\cite{ho2019flow++} introduced Flow++. They propose three key changes to the flow-based generative models. First, instead of adding uniform dequantization noise to pixels, Flow++ learns a conditional noise model $q(u|x)$ (parameterized by a small flow). They train this auxiliary flow with a tight variational bound, which leads to a better objective and reduces overfitting. Second, they incorporate more expressive coupling layers to better capture data complexity. Third, to improve long-range dependencies, they incorporate convolutional layers and multi-head self-attention inside the network that predicts the coupling parameters.

\begin{table}[h!]
\centering
\begin{tabular}{lccc}
\toprule
\textbf{Model} & \textbf{CIFAR-10} & \textbf{ImageNet 32$\times$32} & \textbf{ImageNet 64$\times$64} \\
\midrule
Real NVP & 3.49 & 4.28 & -- \\
Glow & 3.35 & 4.09 & 3.81 \\
IAF-VAE & 3.11 & -- & -- \\
Flow++ & \textbf{3.08} & \textbf{3.86} & \textbf{3.69} \\
\bottomrule
\end{tabular}
\caption{Unconditional image modeling results in bits/dim. Table source~\cite{ho2019flow++}.}
\label{tab:flowpp}
\end{table}

By making these changes, Flow++ achieves state-of-the-art performance among non-autoregressive generative models at the time (2019). They evaluate using \textbf{bits per dimension (bits/dim)}, defined as

\begin{equation}
\text{bits/dim} = \frac{\text{NLL}}{N \log(2)}
\label{eq:bits-per-dim}
\end{equation}
where $N$ is the number of pixels and NLL is the negative log likelihood. Lower bits/dim indicate a better generative model. Table~\ref{tab:flowpp} shows the results of Flow++ compared to other models discussed earlier.

Durkan, Bekasov et al.~\cite{durkan2019neural} introduced Neural Spline Flows, where they replaced the additive and affine transformations in coupling layers with a monotonic rational quadratic spline (RQS) transformation. RQS is a smooth, strictly increasing piecewise function. A bounded interval $[-B, B]$ is chosen and split into $K$ bins. Inside each bin, the transformation is defined as:

\begin{equation}
y = \frac{a z^2 + b z + c}{d z^2 + e z + 1},
\end{equation}

such that each bin is strictly increasing and has continuous derivatives. This ensures that the transformation remains invertible and smooth across all regions. Affine transformations are often too simple and cannot effectively model multimodal or complex densities. By switching to RQS couplings, the authors greatly improve the flexibility of the model while still retaining analytical invertibility.

By making these changes, Neural Spline Flows are able to match the expressiveness of autoregressive flow models while maintaining the efficiency and speed benefits of coupling-layer architectures. Moreover, this improvement is achieved with minimal computational overhead compared to prior work.

\begin{figure}[h]
\centering
\includegraphics[width=\linewidth]{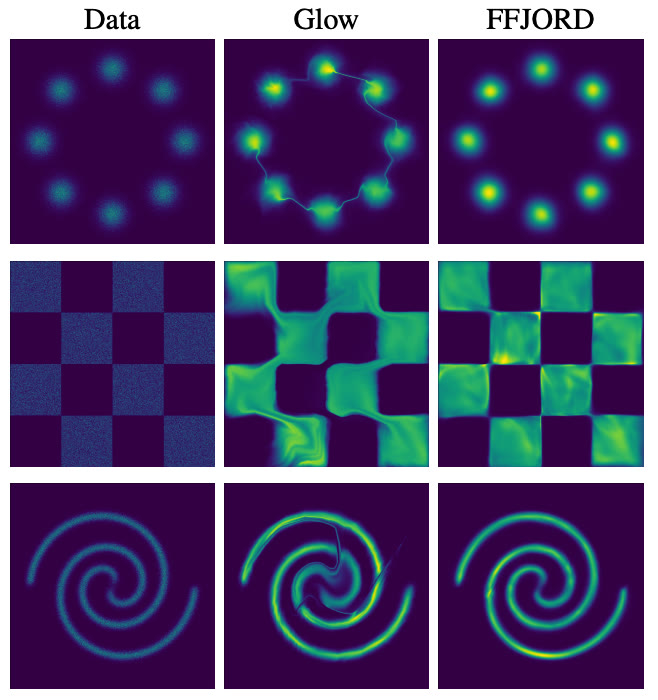}
\caption{Comparison between Glow and FFJORD in estimating distribution. Image source:~\cite{grathwohl2018ffjord}.}
\label{fig:ffjord}
\end{figure}

Grathwohl and coauthors~\cite{grathwohl2018ffjord} came up with a new substitute for Jacobian determinant. Normalizing flows like NICE~\cite{dinh2014nice}, RealNVP~\cite{dinh2016density}, and Glow~\cite{kingma2018glow} keep the determinant cheap by restricting the type of layers via partition variables and affine transformation. This in turn limits the expressiveness of those models. Instead, the authors suggest Free-Form Jacobian of Reversible Dynamics (FFJORD). Instead of computing Jacobian determinant, they use Hutchinson’s trace estimator to come up with a scalable unbiased estimate of the log-density. Making this change allows them to use unrestricted networks since the trace can be calculated cheaply. Moreover, because of using Hutchinson’s estimator, they are able to compute the trace of the Jacobian in $O(D)$ while previously computing exact determinant would require $O(D^3)$ where $D$ is width $\times$ height $\times$ channels. Fig.~\ref{fig:ffjord} shows how they are able to do a better job in estimating multi-modal and discontinuous distributions.

\begin{figure}[h]
\centering
\includegraphics[width=\linewidth]{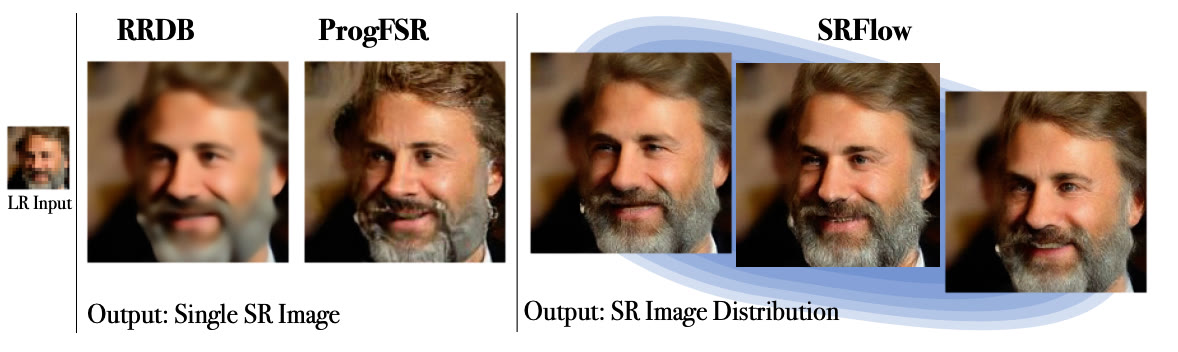}
\caption{Deterministic mapping in prior work versus SRFlow that learns the distribution of photorealistic images. Image source:~\cite{lugmayr2020srflow}.}
\label{fig:srflow}
\end{figure}

Lugmayr et al.~\cite{lugmayr2020srflow} introduce Super Resolution with Normalizing Flow (SRFlow). Their approach relies on normalizing flow where it learns the conditional distribution of the output given the low-resolution input. An invertible network $f_{\theta}(y; x)$ turns the high-resolution $y$ into a latent vector $z$. Given a simple latent prior $z$, the density is

\begin{equation}
p(y|z) = p(z) \left| \det \frac{\partial f_{\theta}(y; x)}{\partial y} \right| , \qquad z = f_{\theta}(y; x)
\end{equation}
During generation, they sample $z \sim p(z)$ and create the high-resolution image $y$ via

\begin{equation}
y = f_{\theta}^{-1}(z; x)
\end{equation}
Prior super-resolution works favor the average of many plausible high-resolution images, as a result the output is blurry. However, SRFlow learns a conditional density $p(y|x)$, as a result it can sample and generate multiple photorealistic images as shown in Fig.~\ref{fig:srflow}.

\subsection{Modern Directions in Normalizing Flows}
Papamakarios and coauthors~\cite{papamakarios2021normalizing} published a comprehensive review paper that provides a unified perspective on normalizing flows. They show that normalizing flows can represent any distribution $p_x(x)$ under reasonable conditions on $p_x(x)$. More specifically, for two well-behaved distributions, $p_x(x)$ (target) and $p_u(u)$ (base), there exists an invertible function that can turn $p_u(u)$ into $p_x(x)$. Moreover, they offer a unifying view on how to use normalizing flows for generative modeling versus density estimation.

\begin{figure}[h]
\centering
\includegraphics[width=\linewidth]{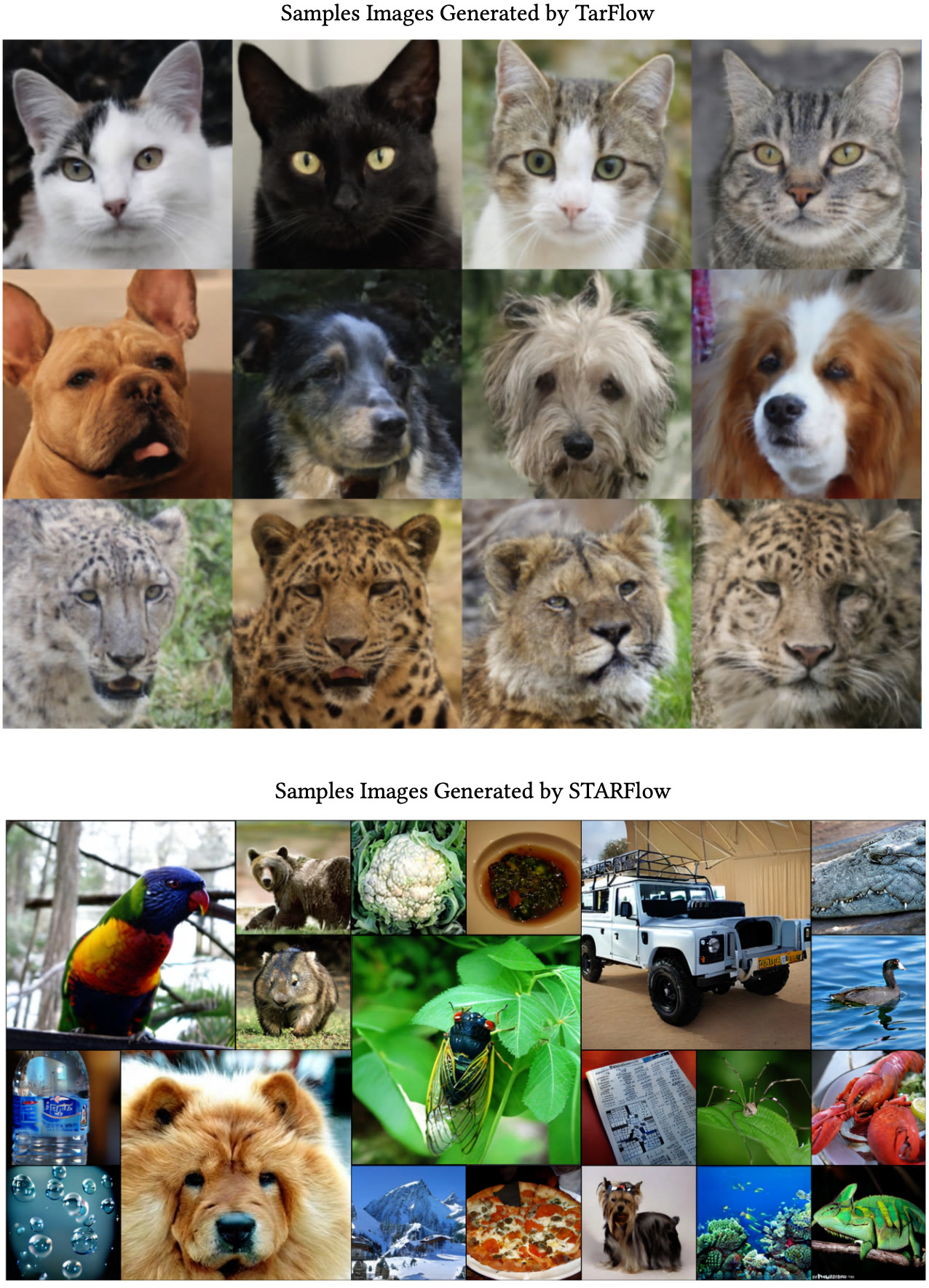}
\caption{TARFLOW (top) and STARFLOW(bottom) improve generation capabilities of Normalizing Flows. Image source:~\cite{zhai2024normalizing,gu2025starflow}.}
\label{fig:tarflow-starflow}
\end{figure}

Around 2021 to 2022, the interest in normalizing flows started to decline because of the success of diffusion and transformer-based generative image models. In 2025, a team at Apple (Zhai et al.~\cite{zhai2024normalizing}) brought renewed attention to it by introducing TARFLOW. TARFLOW is the transformer-based variant of Masked Autoregressive Flows (MAFs). TARFLOW consists of a stack of autoregressive transformer blocks on image patches, alternating the autoregressive direction between layers. They also offer training techniques that significantly improve the generation quality. Apple subsequently published STARFLOW (Gu et al.~\cite{gu2025starflow}), a high-resolution and more powerful generative model, building on ideas from TARFLOW. Fig.~\ref{fig:tarflow-starflow} shows sample images generated by TARFLOW (top) and STARFLOW (bottom).

\subsection{Conclusion}
Normalizing flows offer a simple but powerful formulation for generating images. They rely on simple transformations that are mathematically tractable. Every data point $x$ can be mapped to another point in the latent space, and this provides a powerful tool to inspect how small latent changes affect pixels. Moreover, they provide a simple one-step sampling, whereas diffusion or transformer based models require multiple steps to generate the output image. Around 2022, the community shifted the attention away from the normalizing flows, since they often underperform diffusion/autoregressive models on high-fidelity synthesis benchmarks. Because of their interpretable latents, they still remain useful in modern generative architectures.

\section{Transformer and Autoregressive Models}
This section focuses on image generation models that make predictions autoregressively. Early models achieved this with recurrent or convolutional networks, while later architectures adopted transformers~\cite{vaswani2017attention} and all of them follow the same next token (patch) prediction principle. In autoregressive models, we predict the next element conditioned on everything generated so far. In the case of images, an image is tokenized (via pixels or discrete latents) and generation conditions on previously generated tokens.

\subsection{RNN and CNN Autoregressive Image Generation}
In one of the early works, Oord et al.~\cite{van2016pixel} utilized recurrent neural networks in order to sequentially predict pixels in an image. The probability of each pixel $\mathbf{x}$ in an $n \times n$ image can be represented as
~\begin{equation}
p(\mathbf{x})=\prod_{i=1}^{n^{2}} p\!\left(x_i \mid x_1,\ldots,x_{i-1}\right),
\end{equation}
where $p\!\left(x_i \mid x_1,\ldots,x_{i-1}\right)$ is the probability of the $i$-th pixel $x_i$ given all previous pixels. In practice, they treat the three red, green, and blue channels as follows:
~\begin{equation}
p(\mathbf{x})
=\prod_{i=1}^{n^{2}}
p\!\left(x_{i,R}\mid x_{<i}\right)
p\!\left(x_{i,G}\mid x_{<i},x_{i,R}\right)
p\!\left(x_{i,B}\mid x_{<i},x_{i,R},x_{i,G}\right),
\end{equation}
where each pixel is dependent on the other channels as well as all the previously generated pixels. In the end, they used a simple softmax over 256 values to obtain the value for each pixel.

\begin{figure}[h]
\centering
\includegraphics[width=\linewidth]{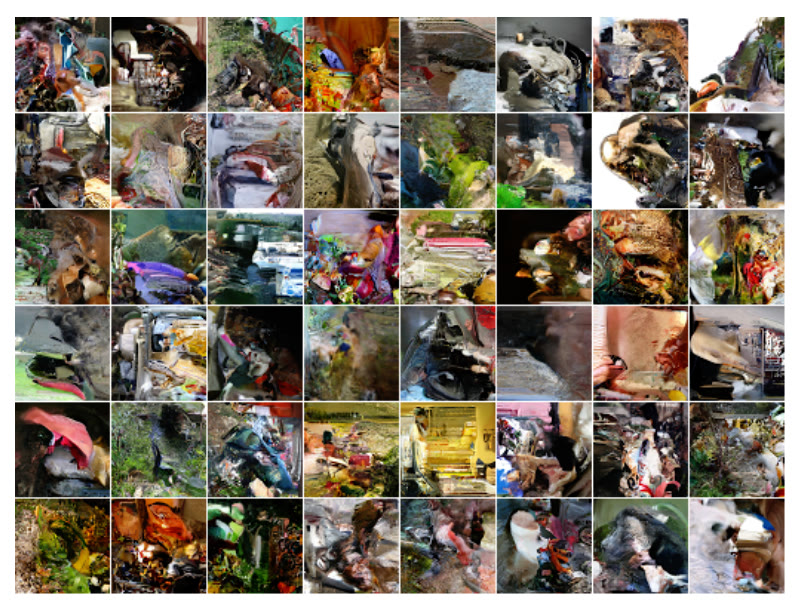}
\caption{Sample images generated by PixelRNN. Image source:~\cite{van2016pixel}.}
\label{fig:pixelrnn}
\end{figure}

They introduced two different architectures: PixelRNN and PixelCNN. PixelRNN is an LSTM-based architecture that utilizes masked convolution for computing features and LSTM to model dependencies across previously generated pixels. In order to provide parallelization during training, they also introduced PixelCNN. PixelCNN is a convolutional-based architecture that causally models the dependence among pixels via masked convolution and a stack of convolutional layers. During training, the model factorizes the likelihood in raster-scan order (left-to-right, top-to-bottom) using masked convolutions. While doing this, they use a masking mechanism to make sure that each pixel can see only the past rows and columns. The overall architecture for PixelCNN and LSTM-based PixelRNN is shown in Table~\ref{tab:arch-pixelrnn}. They train by minimizing the negative log-likelihood of a 256-way softmax for
each pixel:

\begin{table}[t]
\centering
\setlength{\tabcolsep}{8pt}
\renewcommand{\arraystretch}{1.25}
\begin{tabular}{|>{\centering\arraybackslash}m{0.27\linewidth}
                |>{\centering\arraybackslash}m{0.33\linewidth}
                |>{\centering\arraybackslash}m{0.33\linewidth}|}
\hline
\thead{PixelCNN} & \thead{Row LSTM} & \thead{Diagonal BiLSTM} \\
\hline
\multicolumn{3}{|c|}{$7 \times 7$ conv mask A} \\
\hline
\multicolumn{3}{|c|}{\textbf{Multiple residual blocks:}} \\
\hline
\makecell{Conv\\ $3 \times 3$ mask B} &
\makecell{Row LSTM\\ i-s: $3 \times 1$ mask B\\ s-s: $3 \times 1$ no mask} &
\makecell{Diagonal BiLSTM\\ i-s: $1 \times 1$ mask B\\ s-s: $1 \times 2$ no mask} \\
\hline
\multicolumn{3}{|c|}{ReLU followed by $1 \times 1$ conv, mask B (2 layers)} \\
\hline
\multicolumn{3}{|c|}{256-way Softmax for each RGB color (Natural images) \textit{or} Sigmoid (MNIST)} \\
\hline
\end{tabular}
\caption{Architecture summary of PixelCNN and LSTM based PixelRNN. Table source:~\cite{van2016pixel}.}
\label{tab:arch-pixelrnn}
\end{table}

\begin{equation}
p(\mathbf{x})=\prod_{i=1}^{n^2} \prod_{c\in\{R,G,B\}}
p\!\left(x_{i,c}\mid \mathbf{x}_{<i},\, x_{i,<c}\right).
\end{equation}
Fig.~\ref{fig:pixelrnn} shows sample images of a model trained on ImageNet $64\times64$ images.

\begin{figure}[h]
\centering
\includegraphics[width=\linewidth]{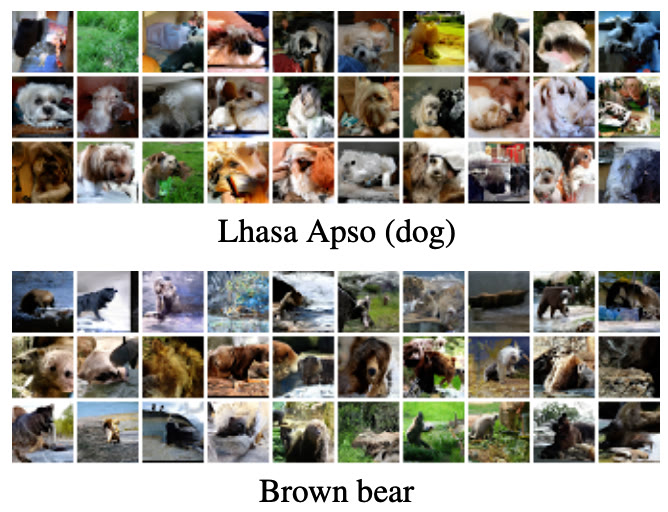}
\caption{Sample images generated by Gated PixelCNN conditioned on class labels Lhasa Apso dog (top) and Brown bear (bottom). Image source:~\cite{van2016conditional}.}
\label{fig:gated-pixelcnn}
\end{figure}

Shortly after, Oord et~al.~\cite{van2016conditional} tried to improve PixelCNN by introducing a gating mechanism. The authors discovered that PixelRNN tended to perform better than PixelCNN because of the gating mechanism in LSTMs. In order to achieve a similar behavior in PixelCNN, they replaced the ReLU block with a gated convolutional block:
\begin{equation}
y = \tanh(W_{k,f} * x) \odot \sigma(W_{k,g} * x).
\end{equation}
Here \(W_{k,f}\) and \(W_{k,g}\) are convolution filters in layer \(k\). The operator \(*\) is convolution, \(\sigma\) is sigmoid, and \(\odot\) is elementwise product. \(\tanh\) acts as a signal carrier and \(\sigma\) acts as a gate to decide how much of the signal goes through. Moreover, they eliminate the flawed masked convolution's blind spot by splitting the network
into two convolutional stacks: a vertical stack for the rows above the current row and a horizontal stack for the left columns in the current row. Using Gated PixelCNN architecture they match PixelRNN's performance (in bits per dimension) with much less compute. They also introduced conditional PixelCNN. Given an image description represented as a vector \(h\), they fit the conditional distribution
\begin{equation}
p(x \mid h) = \prod_{i=1}^{n^{2}} p(x_i \mid x_1,\ldots,x_{i-1}, h).
\end{equation}
They model this by adding \(h\) to the activation
\begin{equation}
y = \tanh(W_{k,f} * x + V_{k,f}^{\top} h)
    \odot
    \sigma(W_{k,g} * x + V_{k,g}^{\top} h).
\end{equation}
Fig.~\ref{fig:gated-pixelcnn} shows sample images generated conditioned on Lhasa Apso dog (top) and Brown bear (bottom). They also tried using PixelCNN as the decoder in a standard autoencoder. They used conditional PixelCNN for modeling (decoding) \(p(x \mid h)\), where \(h\) is the latent code generated by the encoder.

Salimans et al.~\cite{salimans2017pixelcnn++} introduced PixelCNN++, in which they made 5 changes to the original PixelCNN's model structure and training. Those changes include:
1) instead of a 256-way softmax, they use a discretized logistic
mixture model on pixels and round it to the nearest 8-bit value
$z \in \{0,\ldots,255\}$.
2) They simplify the masking logic by conditioning the network on
previous whole pixels. 3) They utilize down-sampling and up-sampling to capture structure at multiple resolutions.
4) They include long-range skip connections to recover fine details and help with optimization. 5) Apply dropout for regularization. By applying these changes, they are able to improve bits per sub-pixel from 3.00 in PixelCNN to 2.92 in PixelCNN++.

% Refer to Eq.~\ref{eq:bits-per-dim} for the definition of bits per sub-pixel.

\begin{figure*}[t]
  \centering
  \includegraphics[width=0.85\textwidth]{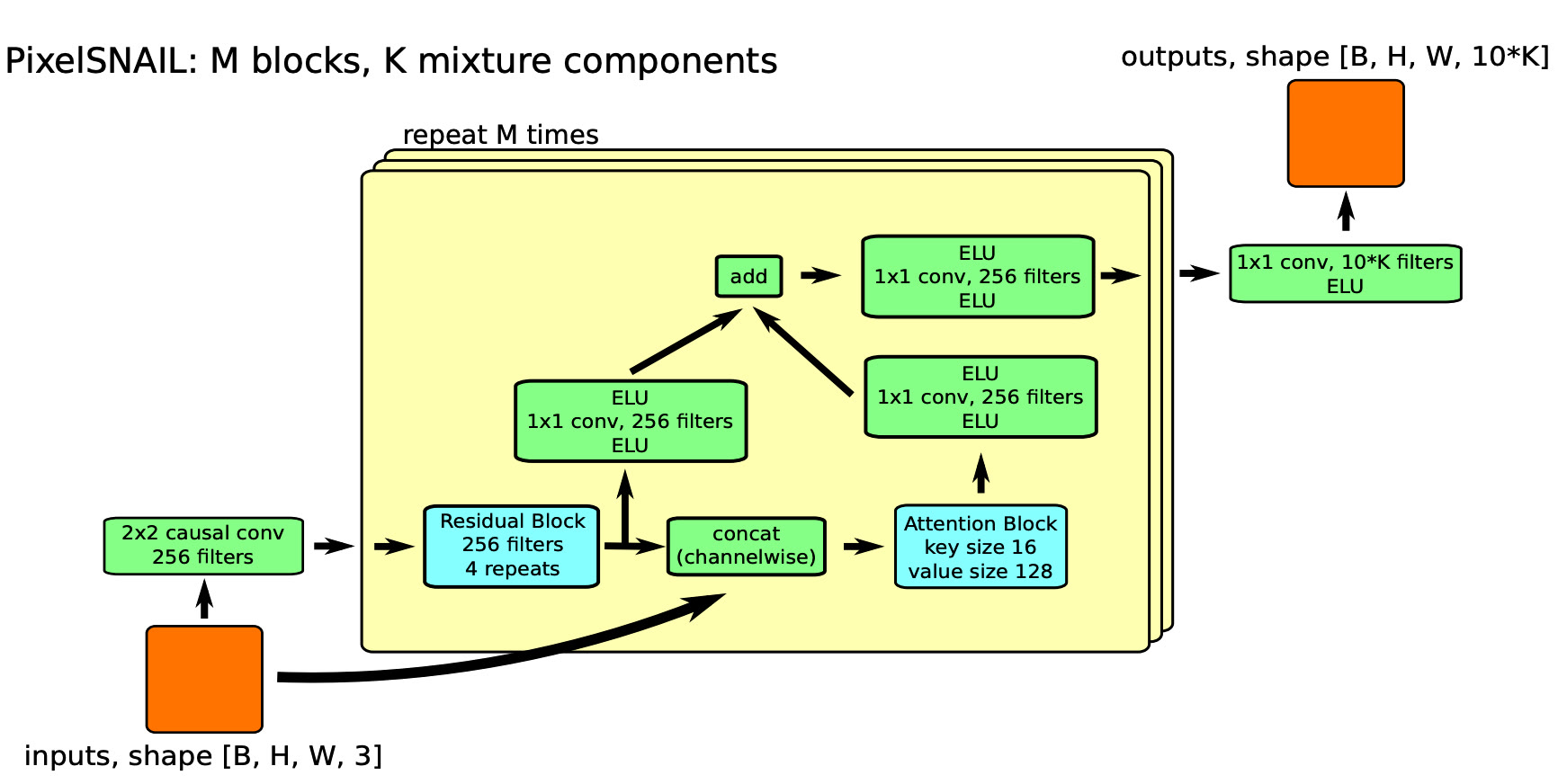}
  \caption{PixelSnail Model Architecture combining causal CNN with self attention. Architecture source:~\cite{chen2018pixelsnail}.}
  \label{fig:pixelsnail}
\end{figure*}

Chen and coauthors~\cite{chen2018pixelsnail} introduced PixelSNAIL. They combine causal convolution with self-attention in order to better capture long-range dependencies in autoregressive generation models like PixelCNN. One issue with convolutions is that they have a finite receptive field and cannot capture information from distant pixels. They use self-attention to counteract this issue. They interleave causal convolutions and self-attention so that each pixel can read from local context as well as distant pixels. Fig.~\ref{fig:pixelsnail} shows the overall architecture of their model. Using this approach, they are able to improve bits per sub-pixel from 2.92 in PixelCNN++ to 2.85.

\subsection{Transformer Autoregressive Image Generation}

After the Transformer architecture~\cite{vaswani2017attention} was introduced in 2017, it was shown that these architectures have strong ability to capture long term dependency as well as training parallelization. Although the transformer architecture was originally proposed for text and natural language applications, the architecture was eventually adopted for image applications. Parmar et~al.~\cite{parmar2018image} utilized the transformer architecture for auto-regressive image generation, Image Transformer. The underlying principle is the same, given the pixels generated so far, what is the most likely next pixel. 

\begin{equation}
p(\mathbf{x}) = \prod_{t=1}^{T} p(x_t \mid x_{<t}), \quad \text{where } T = H \times W \times 3
\end{equation}

\begin{figure}[h]
  \centering
  \includegraphics[width=0.8\linewidth]{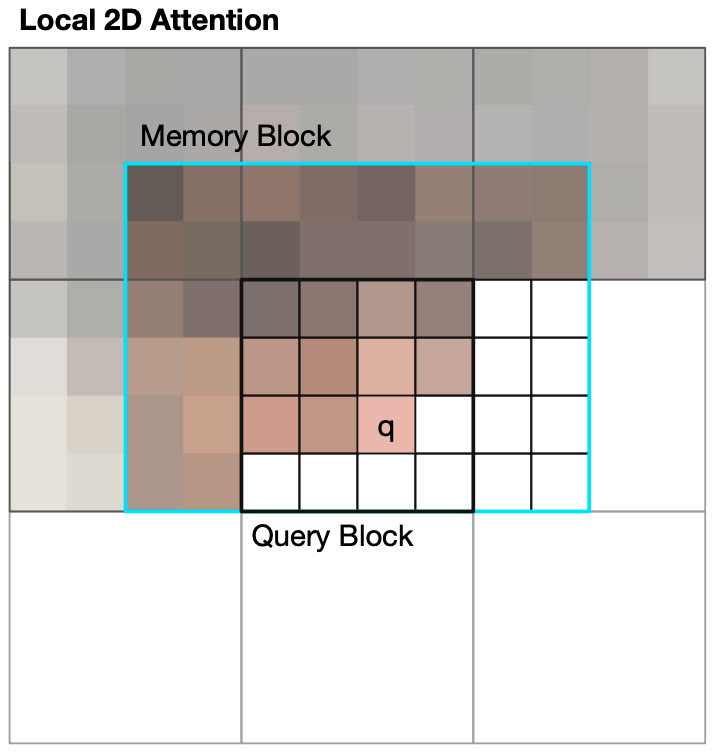}
  \caption{Query block size of $h_q\times w_q$. Memory block extends the query block upward by $h_m$ and left/right by $w_m$ pixels. $h_m=w_m=2$ and $h_q=w_q=4$. Note how the current pixel, $q$, is attending the prior pixels. The future (masked) pixels are shown in white color. Image source:~\cite{parmar2018image}.}
  \label{fig:image-trans}
\end{figure}

Each image is broken into query blocks with the size $h_q \times w_q$ as shown in Fig.~\ref{fig:image-trans}. These form the query tokens for attention. There is also a memory block that extends the query block on the left, right, and above sides. The query and memory blocks only include the pixels that have been visited so far. The simplified training steps are shown in Box~\ref{box:train-image-transformer}.

\begin{boxes}
\begin{tcolorbox}[title=Training for Autoregressive Image Transformer, colback=gray!5, colframe=black!75, fonttitle=\bfseries]
\begin{enumerate}[leftmargin=*, itemsep=1em]
\item Per-pixel embedding that maps 3-channel input $x$ to a $d$-dimensional
vector. This can be a $1 \times 1$ or $1 \times 3$ conv layer.
\[
E \in \mathbb{R}^{B \times H \times W \times d}.
\]

\item 2D positional encoding for each batch $b$, row $r$, and column $c$:
\[
P_{\text{row}}[r] \in \mathbb{R}^{d/2}, 
P_{\text{col}}[c] \in \mathbb{R}^{d/2},
\]
\[
P = \operatorname{concat}(P_{\text{row}}, P_{\text{col}}).
\]
\[
X^{(0)}_{b,r,c} = e_{b,r,c} + P_{r,c} \in \mathbb{R}^{d},
\]

Then $X \in \mathbb{R}^{B \times H \times W \times d}$ becomes input to the transformer.

\item Partition the $H \times W$ grid into $h_q \times w_q$ query blocks as shown in Fig.~\ref{fig:image-trans}. The memory block is
extended by $h_m, w_m$ pixels upward, left, and right. Extract features for query and memory as
\[
X_Q \in \mathbb{R}^{B \times L_Q \times d},
\qquad
X_M \in \mathbb{R}^{B \times L_M \times d},
\]
where $L_Q = h_q \times w_q$ and $L_M$ is the memory window.

\item Compute self-attention scores via $Q$, $K$, and $V$:
\[
Q = X_Q W_Q^{(\ell)} \in \mathbb{R}^{B \times L_Q \times d},
\]
\[
K = X_M W_K^{(\ell)} \in \mathbb{R}^{B \times L_M \times d},
\]
\[
V = X_M W_V^{(\ell)} \in \mathbb{R}^{B \times L_M \times d}.
\]

\item After applying projection, residual, layer norm, and feedforward layer, we will have:
\[
Z^{(\ell)} \in \mathbb{R}^{B \times H \times W \times d}.
\]
Repeat this process for $L$ transformer layers.

\item Pass $Z$ to discretized mixture of logistics (DMoL) in order to get
the likelihood for an image:
\[
\log p_\theta(x)
= \sum_{\text{row},\,\text{col},\,\text{ch}}
    \log P\bigl(x_{\text{row},\text{col},\text{ch}}
                 \mid z_{\text{row},\text{col}}\bigr).
\]

\item Compute the training loss over each batch and update the weights $\theta$ via gradient updates:
\[
\mathcal{L}(\theta)
= -\frac{1}{B} \sum_{b=1}^{B} \log p_\theta\bigl(x^{(b)}\bigr).
\]

\end{enumerate}
\end{tcolorbox}
\caption{Training and Generation Procedure for an Autoregressive Image Transformer}
\label{box:train-image-transformer}
\end{boxes}

Table~\ref{tab:cifar10-nll} shows the negative log-likelihood scores of a wide variety of auto-regressive image generation models including the Image Transformers. Image Transformer is able to improve all the prior results except PixelSNAIL. The authors mention that they could see performance enhancement if they increase the size of the memory block.

\begin{table}[t]
\centering
\begin{tabular}{l c}
\toprule
Model              & NLL \\
\midrule
PixelCNN           & 3.14 \\
Row PixelRNN       & 3.00 \\
Gated PixelCNN     & 3.03 \\
PixelCNN++         & 2.92 \\
PixelSNAIL         & 2.85 \\
Image Transformer  & 2.9 \\
\bottomrule
\end{tabular}
\caption{Negative log-likelihood (NLL) on CIFAR-10 (test).}
\label{tab:cifar10-nll}
\end{table}

\begin{figure}[t]
  \centering
  \includegraphics[width=0.6\linewidth]{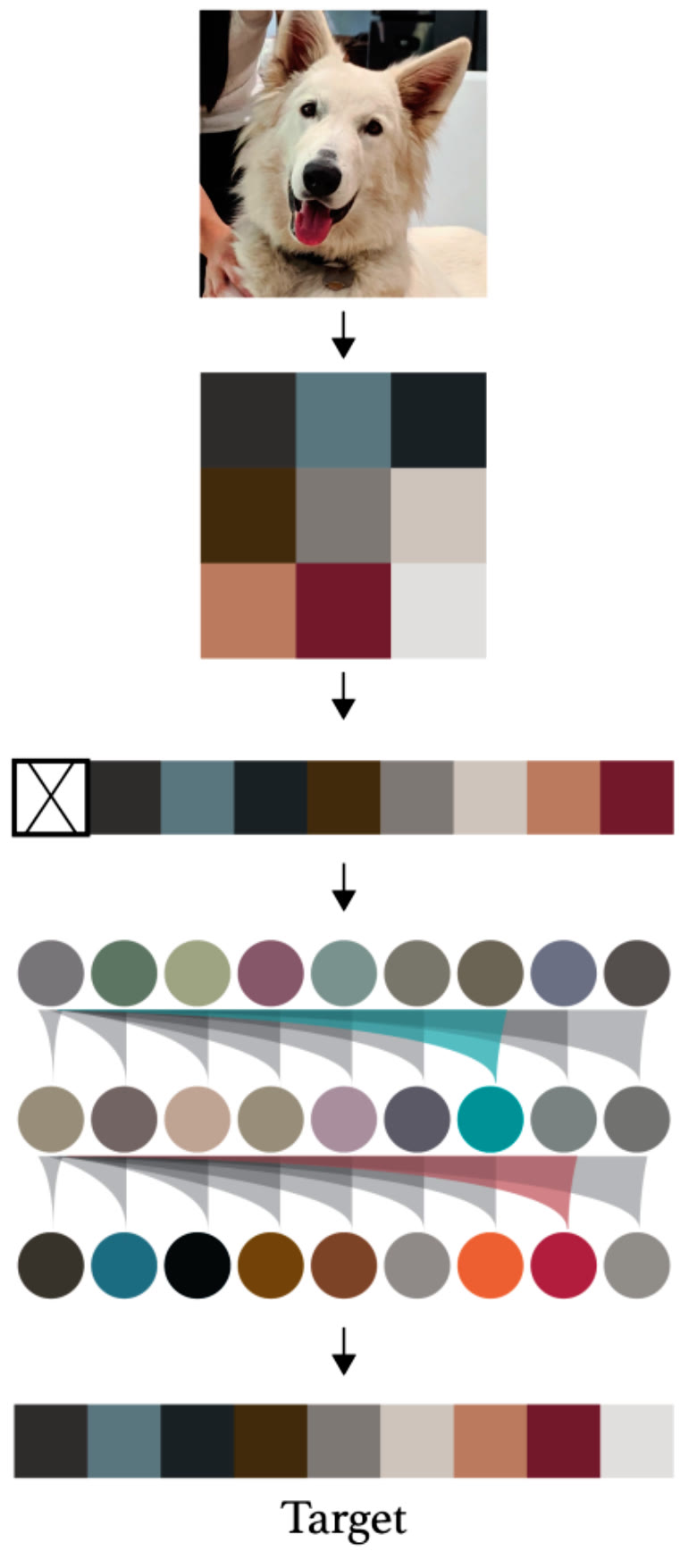}
  \caption{The simplified pre-training pipeline for iGPT. Image source:~\cite{chen2020generative}.}
  \label{fig:igpt}
\end{figure}

Inspired by autoregressive unsupervised learning for natural language data, Chen et al.~\cite{chen2020generative} introduced iGPT. They train a transformer-style model to autoregressively predict pixels by treating the image as a 1-D token
sequence, as shown in Fig.~\ref{fig:igpt}. Each pixel's RGB is mapped to the nearest of 512 learned palette colors. The image is first downsampled to $48\times48$ which results in a sequence of length $T=2304$ with $x_t \in \{0,\ldots,511\}$ for $t=1,\ldots,T$. Using the low-resolution and palette trick, they were able to decrease the context from $(224\times224\times3)$ to $64\times64$ at most. The rest of the process is similar to how a transformer-based model is trained. The process includes token/position embeddings, normalization layers, transformer blocks, and output heads. Using a causal mask, they minimize the negative log-likelihood:
\begin{equation}
\mathcal{L}=-\sum_{t=1}^{T}\log p(x_t \mid x_{<t}).
\end{equation}
They pretrained an autoregressive transformer on pixels and evaluated representations via linear probing/fine-tuning for classification. They normalized features across positions and then trained a linear classifier on the final feature. Using this, they were able to beat state-of-the-art unsupervised transfer methods at the time.

\subsection{Scaling Up Text to Image Transformers}
Although Chen et al.~\cite{chen2020generative} did not illustrate how iGPT can be utilized for image generation, they provided a great foundation for the image generation research that followed. Ramesh et al.~\cite{ramesh2021zero} introduced zero-shot text-to-image generation (DALL-E 1). Applying transformers at the pixel level is computationally intractable, especially for large images. Instead, they decided to represent an image as a composition of rich image constituents. Each of these constituents can be treated in the transformer model by its own token. Using this approach, the transformer model is able to capture the global correlation between different blocks within the image. For training, they utilize a two-stage approach that combines VQ-VAE~\cite{van2017neural} in stage 1 with a transformer in stage 2. Their approach relies on learning codebook vectors, $Z$, in stage 1 that act as a tokenizer for image partitions. In stage 1, they train a VQ-VAE~\cite{van2017neural} to compress the $256\times256$ RGB image into a $32\times32$ grid of image tokens. In stage 1, the model learns a codebook, an encoder, and a decoder. For a more detailed description of VQ-VAEs, please refer to section~\ref{sec:vqvae} in the VAE section of this article. For capturing local dependencies they utilize CNN in the encoder and decoder parts of the model. In stage 2, they freeze the VQ-VAE and concatenate up to 256 text tokens (from image captions) with the $32\times32=1024$ image tokens. Using this setup, they train an autoregressive transformer architecture to model the joint distribution between the image text description and the image itself. During generation (inference), they start with the text description. Using text tokens, they autoregressively generate $32\times32$ image tokens. These tokens are then passed to the codebook and then the VAE decoder for the final generated image.

\begin{figure}[h]
  \centering
  \includegraphics[width=\linewidth]{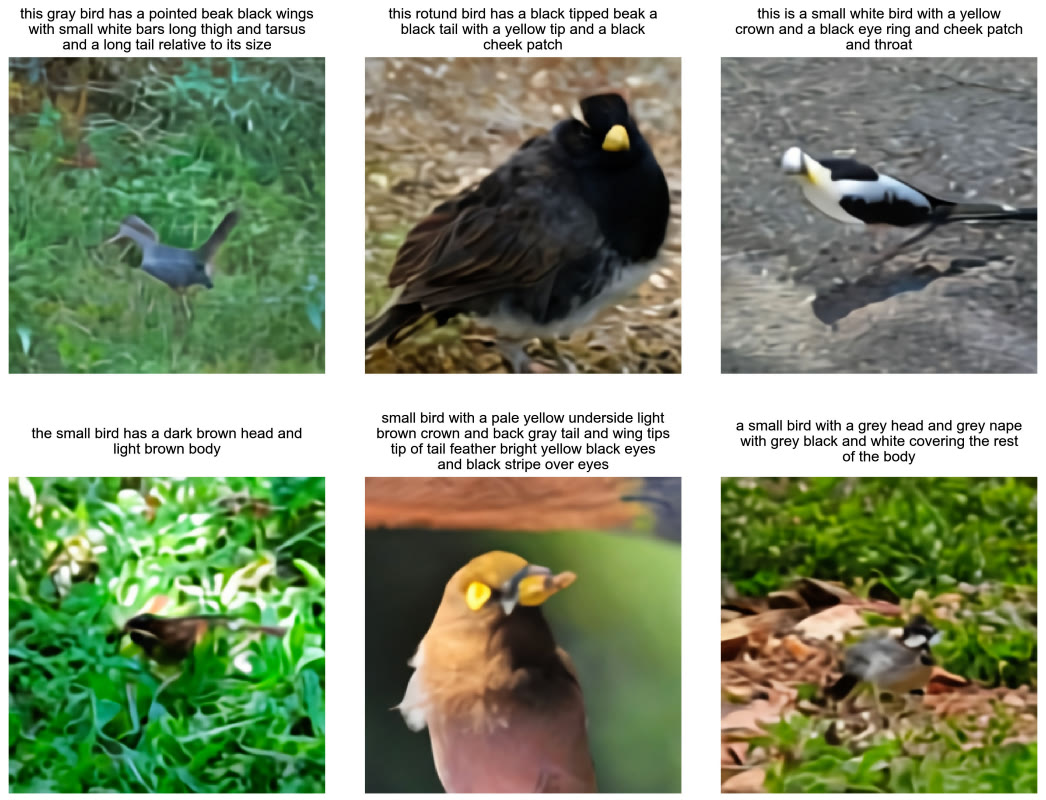}
  \caption{Zero-shot samples from DALL-E 1 on the CUB dataset. Image source:~\cite{ramesh2021zero}.}
  \label{fig:dalle-1}
\end{figure}

The transformer model used in stage 2 was a massive 12B-parameter model. In order to expedite training for this massive scale, they utilized mixed-precision coupled with distributed training. Fig.~\ref{fig:dalle-1} shows some of the sample images generated from text captions on the CUB dataset~\cite{WahCUB_200_2011}. Note that these examples are zero-shot since the model was never trained on this dataset.

\begin{figure}[h]
  \centering
  \includegraphics[width=\linewidth]{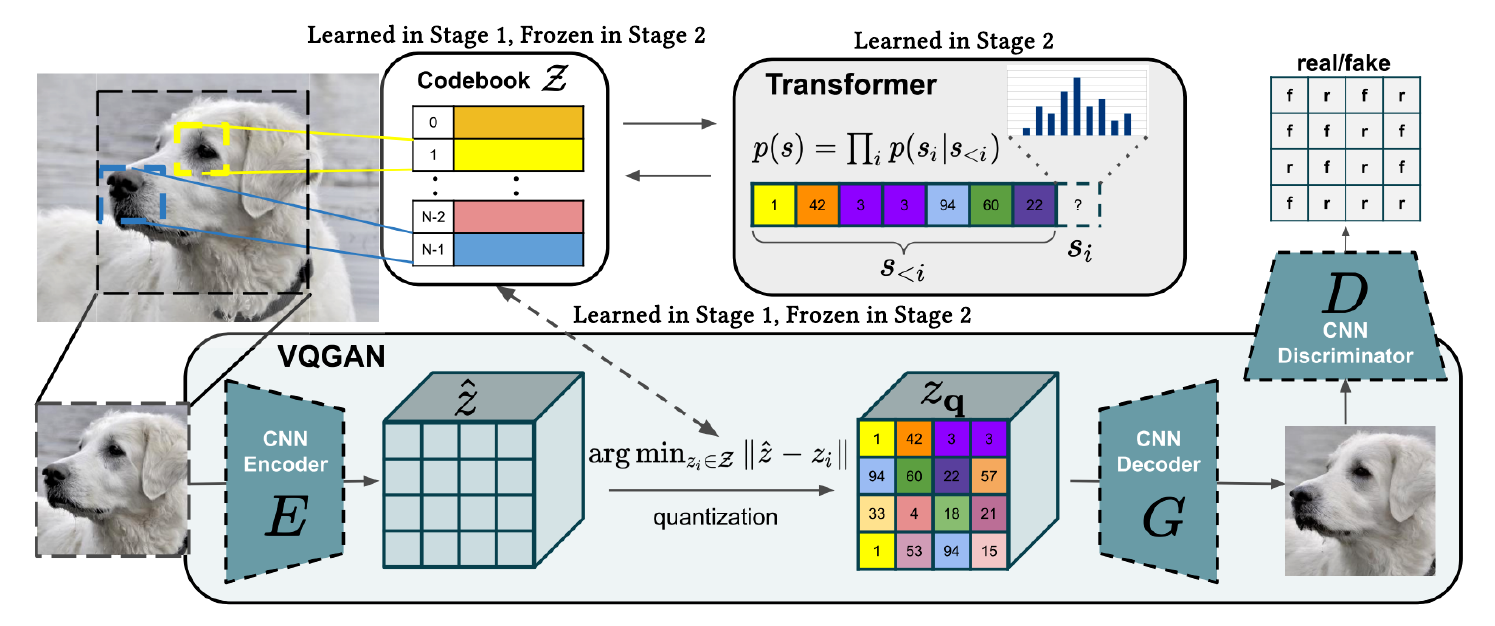}
  \caption{The training architecture for the Taming Transformers for High-Resolution Image Synthesis. Note how each part of the model is frozen during different training phases. Diagram source:~\cite{esser2021taming}.}
  \label{fig:vqgan-gpt}
\end{figure}

Inspired by DALL-E 1~\cite{ramesh2021zero}, Esser et al.~\cite{esser2021taming} used a similar two stage approach but replaced the VAE with GAN for high-resolution image generation. Stage 1 consists of a GAN variant of the VQ-VAE~\cite{van2017neural} where a convolutional encoder–decoder is trained end-to-end to reconstruct the image while learning the codebook \(Z=\{z_k\}_{k=1}^{K}\), as shown in the lower part of Fig.~\ref{fig:vqgan-gpt}.

In phase 2, the VQGAN is frozen and image \(x\) is passed to the encoder and then vector quantizer to get code indices
\(s \in \{0,\ldots,|Z|-1\}^{h\times w}\), where \(h,w\) are downsampling factors (e.g., \(16\times16\)). Now a decoder-only autoregressive transformer is trained on these sequences \(s\), where $N$ is the total number of patches (e.g. $16\times16=256$):
\begin{equation}
\log p(s)=\sum_{i=1}^{N}\log p(s_i \mid s_{<i}).
\end{equation}

\begin{figure}[h]
  \centering
  \includegraphics[width=0.9\linewidth]{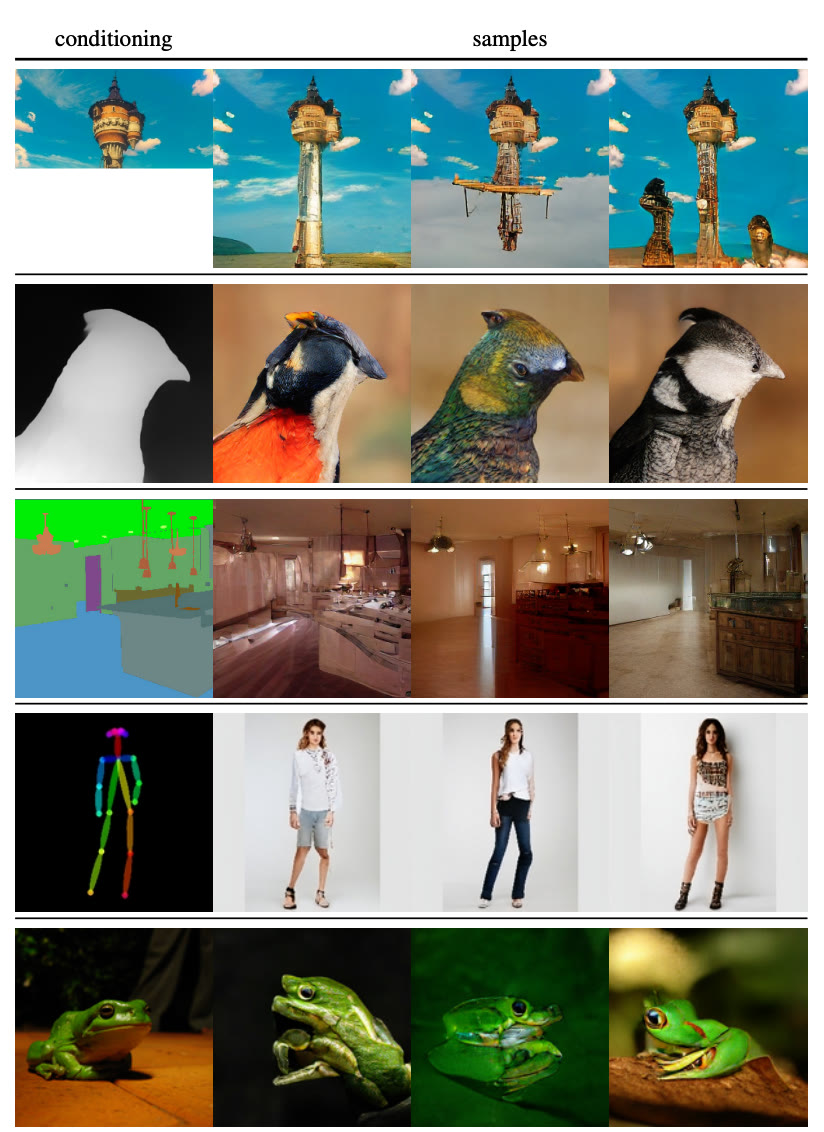}
  \caption{Image completion for unconditional, depth to image, guided synthesis, and pose-guided. Image source:~\cite{esser2021taming}.}
  \label{fig:vq-gpt-conditional}
\end{figure}

\noindent During generation, the transformer runs causally to generate
\(s_1,\ldots,s_{256}\) indices. Using the codebook \(Z\), indices are mapped to codes \(z_q\). The code \(z_q\) is passed to the frozen decoder \(G\) to get the generated output \(\hat{x}=G(z_q)\) as illustrated in Fig.~\ref{fig:vqgan-gpt}. For conditional generation, the transformer starts generating from the class label prior. Each row in Fig.~\ref{fig:vq-gpt-conditional} shows results from a different setup. Top row is unconditional training, second row is depth to image training, third row is semantically guided synthesis, fourth row is pose-guided person generation on DeepFashion~\cite{liu2016deepfashion}, and the fifth row is class-conditional samples.

\begin{figure}[h]
  \centering
  \includegraphics[width=\linewidth]{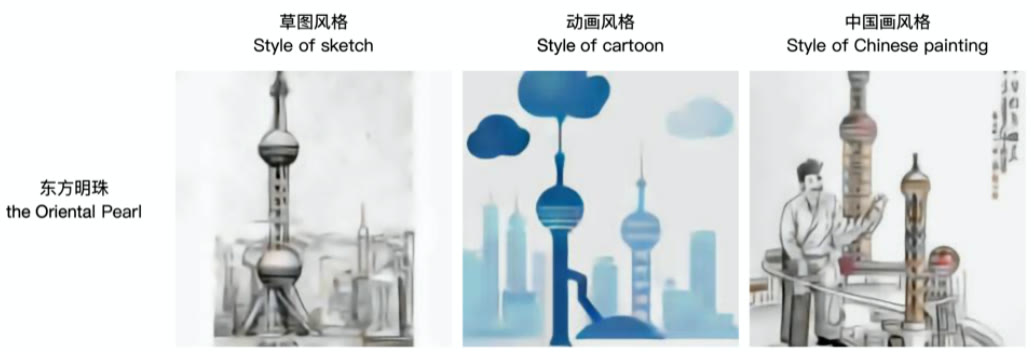}
  \caption{Sample images generated by the CogView style learning model. Image source:~\cite{ding2021cogview}.}
  \label{fig:cogview-style}
\end{figure}

Following DALL-E 1, Ding et al.~\cite{ding2021cogview} introduced CogView. Similar to DALL-E 1, they combine VQ-VAE tokenization with a GPT-style joint text–image language model. The GPT-style decoder transformer consists of \(4\mathrm{B}\) parameters, almost twice as large as DALL-E 1. Most of their innovations lie in the stability tricks applied to their training scheme. For example, they place layer norm before and after each residual branch to prevent exploding values in deep layers. They also reorder and shift attention scores to prevent Float16 overflow. Additionally, they finetuned the final model for other downstream tasks such as super-resolution, image captioning, style learning, and industrial design. In the case of style learning, they finetune the model on four styles: Chinese traditional drawing, oil painting, sketch, and cartoon. When finetuning, the corresponding text for the image is “An image of \{style\} style.” During generation, the input text is passed to the model in the form of “A \{object\} of \{style\} style,” where \{object\} is the object of interest to generate. Fig.~\ref{fig:cogview-style} shows some of the sample images generated by the finetuned style learning model.

\begin{figure}[h]
  \centering
  \includegraphics[width=\linewidth]{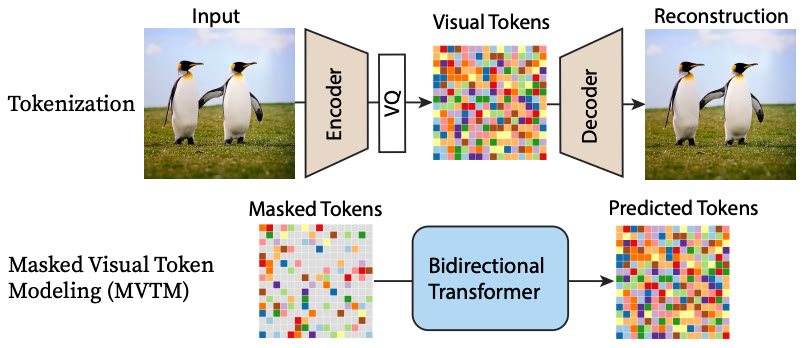}
  \caption{MaskGIT training pipeline. Image source:~\cite{chang2022maskgit}.}
  \label{fig:maskgit-pipeline}
\end{figure}

\indent When training image transformers, images are treated in a sequential manner like text. The autoregressive sequence length grows quadratically with the number of patches, and this can make image generation slow; e.g., generating a single image on a GPU can take 30 seconds. Moreover, unlike language, image is not sequential, and there are other alternatives to capture correlation between image patches. In order to overcome these challenges, Chang et al.~\cite{chang2022maskgit} introduced a bidirectional transformer for image generation and called it Masked Generative Image Transformer (MaskGIT). Instead of a left to right causal mask, they randomly mask a fraction of tokens in the grid and pass the masked grid to the transformer. Fig.~\ref{fig:maskgit-pipeline} shows the pipeline overview.

\begin{figure}[h]
  \centering
  \includegraphics[width=\linewidth]{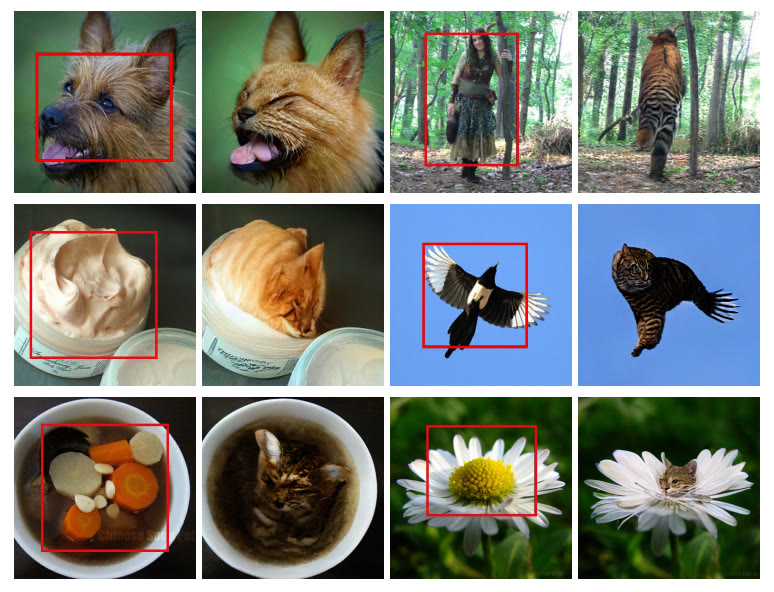}
  \caption{Given the images on the left and target class, MaskGIT replaces the bounding box region with the target class. Image source:~\cite{chang2022maskgit}.}
  \label{fig:maskgit-results}
\end{figure}

The output tokens from the VQ encoder can be denoted by
\(\mathbf{Y}=[y_i]_{i=1}^{N}\), where \(N=32\times32\).
They introduce a binary mask \(\mathbf{M}=[m_i]_{i=1}^{N}\), where they sample a subset of tokens and replace them with a special \texttt{[MASK]} token. The token \(y_i\) is replaced with \texttt{[MASK]} if \(m_i=1\), otherwise it is left unchanged if \(m_i=0\). The training objective is to minimize the negative log-likelihood of predicting masked tokens. When the decoder starts all patches are unknown and it gradually predicts the masked patches. This decoding strategy reduces the sampling steps (8 vs. 1024 tokens), and the authors report strong results on class-conditional editing, as shown in Fig.~\ref{fig:maskgit-results}.

\begin{figure}[h]
  \centering
  \includegraphics[width=\linewidth]{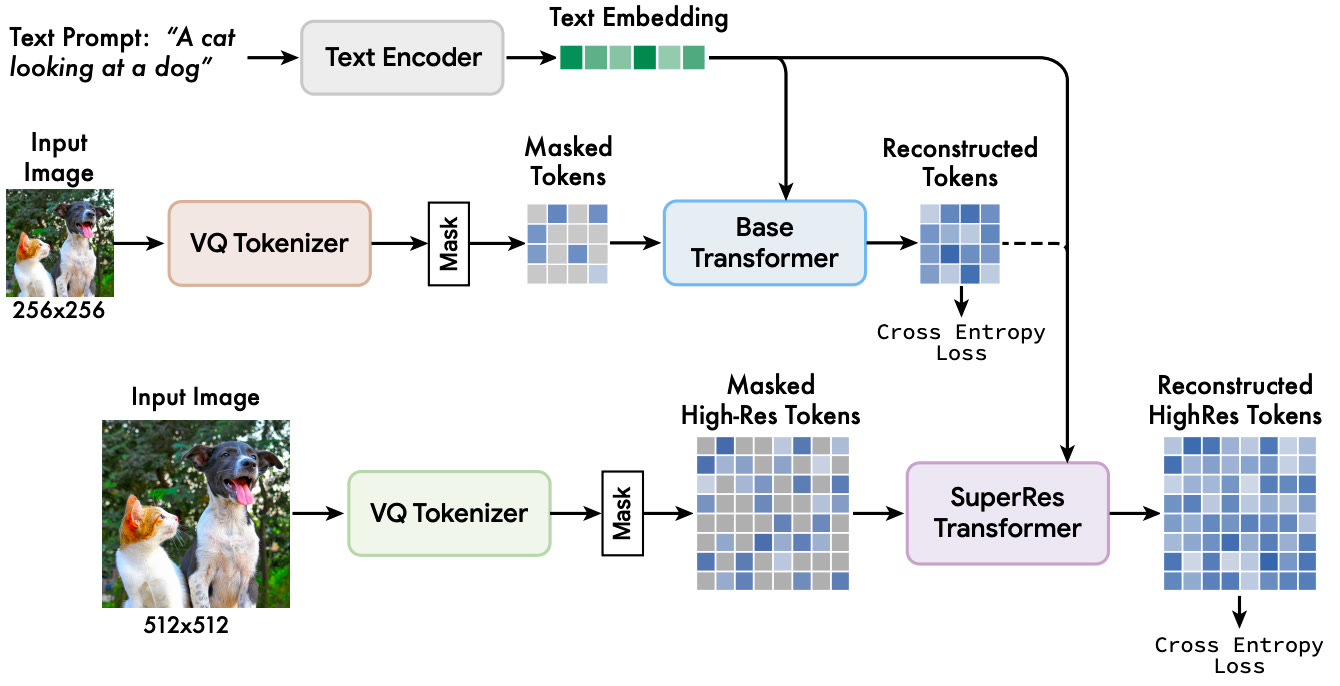}
  \caption{Training pipeline in Muse. They added a second VQ tokenizer for increasing the resolution of the image. They brought in a pretrained language model (T5) for encoding input text. Image source:~\cite{chang2023muse}.}
  \label{fig:muse-pipeline}
\end{figure}

The same group of authors also introduced a follow up improvement and named it Muse~\cite{chang2023muse}. They  used a similar masked approach to MaskGIT~\cite{chang2022maskgit}, but made several architectural changes, as shown in Fig.~\ref{fig:muse-pipeline}. Instead of training a separate text encoder, they start from a pretrained language model that has good text understanding, T5~\cite{raffel2020exploring}. Moreover, they increased the number of VQ-GAN tokenizers from one to two: a base version to fill $16\times16$ masked tokens and a high resolution model to map $16\times16$ to $64\times64$ tokens.

\begin{figure}[h]
  \centering
  \includegraphics[width=\linewidth]{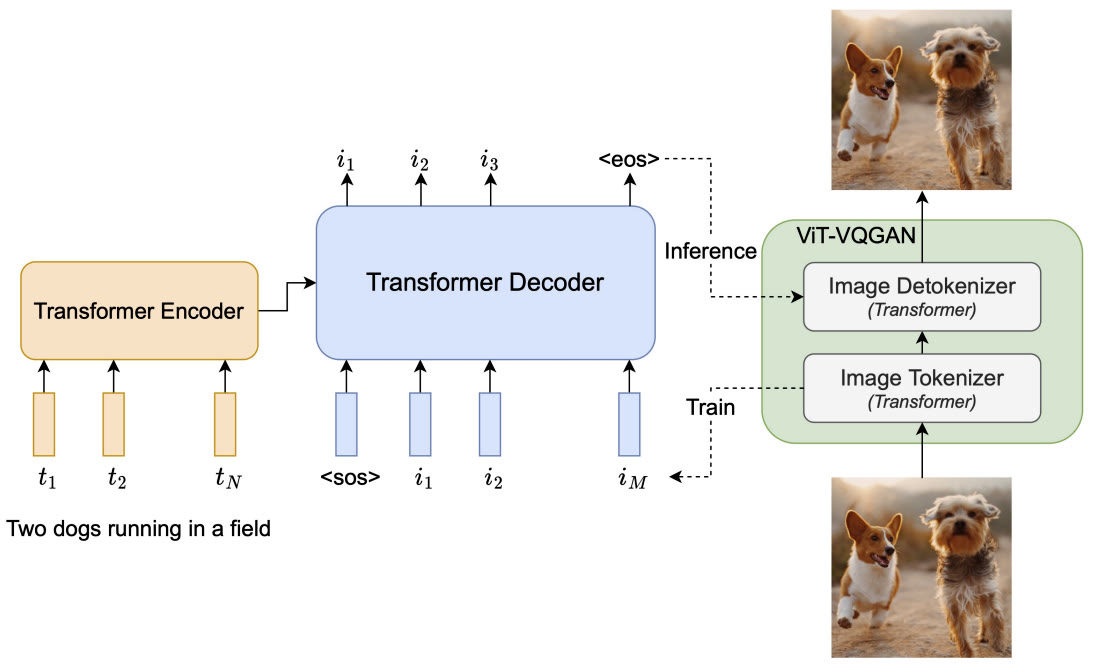}
  \caption{Training pipeline for Parti. Left side shows the autoregressive transformer based encoder-decoder. Right side is the ViT-VQGAN for image tokenizer/detokenizer. The super-resolution section at the end of ViT-VQGAN is omitted for simplicity. Diagram source~\cite{yu2022scaling}.}
  \label{fig:parti-overview}
\end{figure}

\begin{figure}[h]
  \centering
  \includegraphics[width=\linewidth]{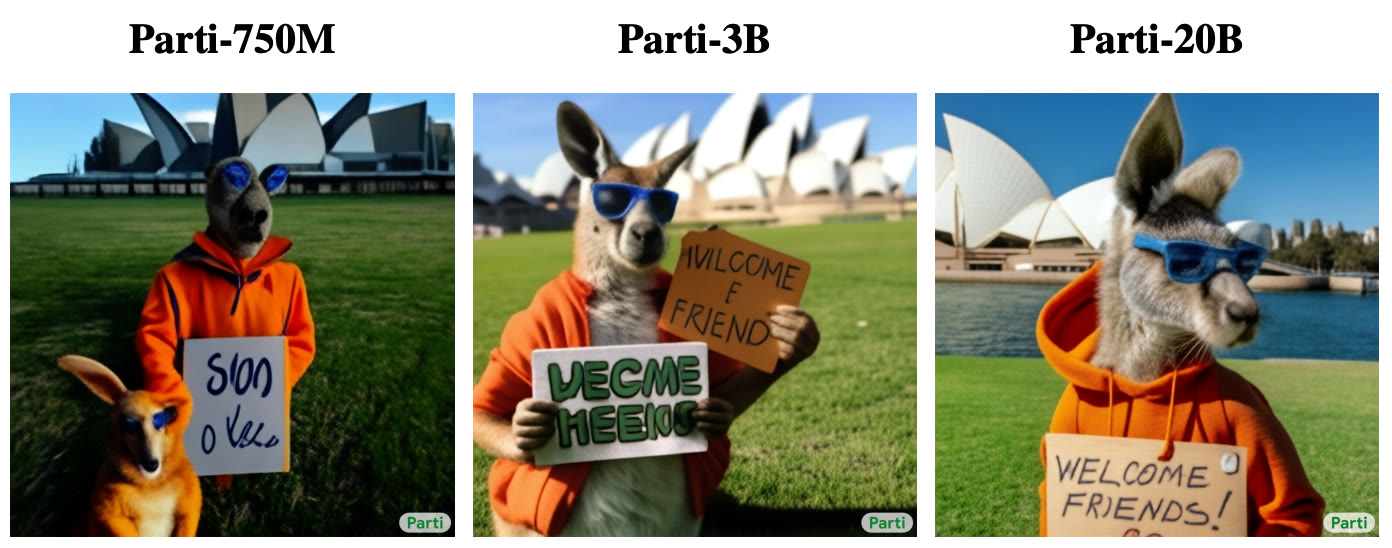}
  \caption{Improved sample quality in Parti as the number of parameters in the autoregressive encoder–decoder increases from 750M to 20B. The input prompt is ``A portrait photo of a kangaroo wearing an orange hoodie and blue sunglasses standing on the grass in front of the Sydney Opera House holding a sign on the chest that says Welcome Friends!''. Image source:~\cite{yu2022scaling}.}
  \label{fig:parti-scaling}
\end{figure}

\indent Majority of the models introduced so far rely on a convolutional network in stage 1, i.e. encoder or decoder part of VQ-VAE. Yu et al.~\cite{yu2021vector} proposed a new approach, where they replace the CNN based encoder–decoder with a Vision Transformer. Shortly after, Yu et al.~\cite{yu2022scaling} introduced Parti. As shown in the right side of Fig.~\ref{fig:parti-overview}, the encoder in ViT-VQGAN transforms the input image into 1024 ($32\times32$) tokens. The decoder reconstructs the original image from these tokens. Furthermore, they add a trainable super-resolution section after the decoder to upsample the image. For the sequence to sequence part of the model (left side of Fig.~\ref{fig:parti-overview}), the text encoder processes the input text description and the image decoder processes both the text encoder output and the image tokens as they arrive. They note that using an encoder–decoder variant of the architecture beats decoder-only autoregressive architectures in both training loss and sample quality. They also show that, as they increase the number of parameters in the autoregressive encoder–decoder, sample quality increases, as shown in Fig.~\ref{fig:parti-scaling}.

\subsection{Conclusion}
Autoregressive image generation predicts each pixel from the previously generated pixels. Unlike VAEs and GANs, these models are trained by maximum likelihood with a tractable objective. Moreover, they support conditioning naturally via the autoregressive factorization and masking. These architectures can easily incorporate pretrained text encoders for conditional generation, which makes them strong at contextual understanding. Despite all of these benefits, they come with downsides such as \(O(n^2)\) attention cost in token length. In the next section, we are going to review diffusion-based models and how they can generate even higher-quality images.

\section{Diffusion Based Models}\label{sec:diffusion}
Unlike image generation models that we have investigated so far (VAEs, GANs, Normalizing Flows, Transformers), diffusion models draw inspiration from physics. For example, when a sugar cube is dropped into a cup of tea, the sugar molecules gradually diffuse into the water, and this diffusion process can be described as a stochastic process. Diffusion models learn to reverse such a stochastic diffusion process where they transform noise into data. In the original image diffusion model proposed by Sohl-Dickstein et~al.~\cite{sohl2015deep}, a similar idea is used: starting from a clean image, a small amount of Gaussian noise is added at consecutive time steps until the image becomes nearly indistinguishable from isotropic Gaussian noise $\mathcal{N}(0, I)$. The goal of a diffusion model is to learn the reverse of this noising process, transforming pure noise back into the original image through a learned reverse diffusion procedure.

If one starts from an input image $x^{(0)}$ and gradually adds noise to it at consecutive time steps,
\begin{equation}
x_{(0)} \rightarrow x_{(1)} \rightarrow \cdots \rightarrow x_{(T)}
\end{equation}
The reverse diffusion process starts with pure noise $x_{(T)}$, and a diffusion model $p_\theta$ learns to remove the noise:
\begin{equation}
x_{(T)} \rightarrow x_{(T-1)} \rightarrow \cdots \rightarrow x_{(0)}
\end{equation}

The forward process is known, and fixed and the goal is to train a model that learns the reverse process. If we assume a Markov chain process, then the forward process distribution can be written as
\begin{equation}
q(x_{(0:T)}) = q(x_{(0)}) \prod_{t=1}^{T} q\bigl(x_{(t)} \mid x_{(t-1)}\bigr)
\end{equation}
After $T$ steps, the distribution of $x_{(T)}$ becomes a simple distribution (e.g., an isotropic Gaussian):
\begin{equation}
q\bigl(x_{(T)}\bigr) \approx \pi\bigl(x_{(T)}\bigr).
\end{equation}

\begin{figure}[h]
  \centering
  \includegraphics[width=\linewidth]{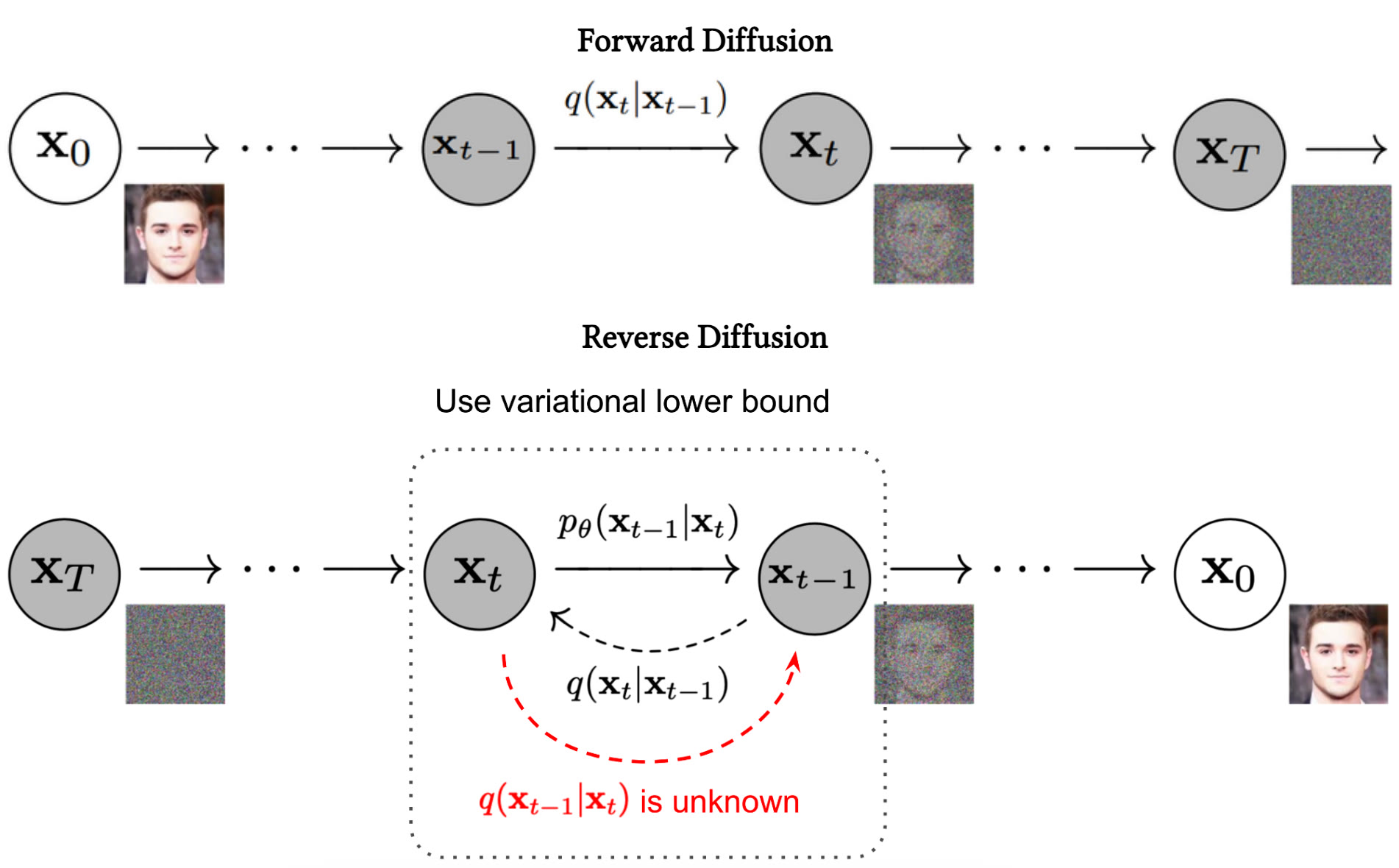}
  \caption{Top: Systematically destroying the structure in the image by gradually adding noise to the image. Bottom: Learned reverse diffusion process to restore the original image. Image source:~\cite{ho2020denoising,weng2021diffusion}.}
  \label{fig:diffusion-process}
\end{figure}

With a Gaussian assumption, if we add a small variance $\beta_t$ at each time step and multiply the input by $\sqrt{1-\beta_t}$ to shrink it, the forward process becomes
\begin{equation}
q\bigl(x_{(t)} \mid x_{(t-1)}\bigr)
= \mathcal{N}\!\bigl(x_{(t)}; \sqrt{1-\beta_t}\,x_{(t-1)}, \beta_t I\bigr)
\end{equation}
After $T$ steps, the image will turn into pure Gaussian noise, as shown in Fig.~\ref{fig:diffusion-process}.

The reverse diffusion process follows a similar Markov chain process. We start from pure noise (isotropic Gaussian) and go backward, starting from $p\bigl(x_{(T)}\bigr) = \pi\bigl(x_{(T)}\bigr)$,
and defining the reverse joint distribution as
\begin{equation}
p_\theta(x^{(0:T)}) = p\bigl(x^{(T)}\bigr)\prod_{t=1}^{T} p_\theta\bigl(x^{(t-1)} \mid x^{(t)}\bigr)
\end{equation}
Our goal is to train a model (with parameters $\theta$) that learns the denoising transition process $p_\theta\bigl(x_{(t-1)} \mid x_{(t)}\bigr)$. By taking the log-likelihood of the generative model and applying Jensen's inequality to the expectation over trajectories, we can obtain a lower bound $K$ on the log-likelihood:
\begin{equation}
L = \mathbb{E}_{q(x_{(0)})}\bigl[\log p_\theta(x_{(0)})\bigr] \ge K,
\end{equation}
where

\begingroup
\scriptsize
\begin{equation}
\begin{split}
K
&= -\sum_{t=2}^{T} \mathbb{E}_{q(x_{(0)}, x_{(t)})}
    \Bigl[
      D_{\mathrm{KL}}\bigl(
        q(x_{(t-1)} \mid x_{(t)}, x_{(0)})
        \,\big\|\,
        p_\theta(x_{(t-1)} \mid x_{(t)})
      \bigr)
    \Bigr] \\
&+ C
\end{split}
\end{equation}
\endgroup
The term $q\bigl(x_{(t-1)} \mid x_{(t)}, x_{(0)}\bigr)$ can be derived analytically from the Gaussian forward process. The distribution $p_\theta\bigl(x_{(t-1)} \mid x_{(t)}\bigr)$ is the output of the model. $D_{\mathrm{KL}}$ is the KL divergence between two Gaussians, and $C$ is an entropy term. In simple terms, the training objective is to make each learned reverse step, $p_\theta\bigl(x_{(t-1)} \mid x_{(t)}\bigr)$, as close as possible to the true reverse of the forward diffusion, $q\bigl(x_{(t-1)} \mid x_{(t)}, x_{(0)}\bigr)$.

Although Sohl-Dickstein et~al.~\cite{sohl2015deep} introduced a strong mathematical foundation for diffusion models, diffusion models lagged behind GANs in terms of quality of the generated images. It wasn't until the introduction Denoising Diffusion Probabilistic Models (DDPM) by Ho et~al.~\cite{ho2020denoising} that things started to really take off for diffusion based models. In DDPM, the authors make a few architectural design choices that help with generating higher-quality images. First, they reparameterize the reverse mean so that the network predicts the noise $\varepsilon$ that was added in the forward process. In Eq.~\ref{eq:forward-marginal}, the signal term \(\sqrt{\bar\alpha_t}\) decays with $t$, while the noise term \(\sqrt{1-\bar\alpha_t} \epsilon \) grows, so \(x_t\) becomes increasingly noisy.

\begin{equation}
x_t = \sqrt{\bar{\alpha}_t}\, x_0 + \sqrt{1 - \bar{\alpha}_t}\,\varepsilon,
\qquad
\varepsilon \sim \mathcal{N}(0, I)
\label{eq:forward-marginal}
\end{equation}
where
\[
\alpha_t = 1 - \beta_t, \qquad
\bar{\alpha}_t = \prod_{s=1}^{t} \alpha_s
\]
and $\beta_t$ is the forward noise variance at step $t$. They train a model $\varepsilon_\theta(x_t, t)$ to approximate $\varepsilon$. As a result of this reparameterization, the training loss simply becomes the mean squared error (MSE) of the noise:
\begin{equation}
L(\theta) = \mathbb{E}_{x_0, t, \varepsilon}
\bigl[\lVert \varepsilon - \varepsilon_\theta(x_t, t) \rVert^2\bigr].
\label{eq:loss-ddpm}
\end{equation}

They also make significant architectural improvements. They use a UNet style CNN~\cite{ronneberger2015u}, coupled with downsampling and upsampling paths. Additionally, they incorporate self-attention at the lower feature-map resolution $(16 \times 16)$. Box~\ref{box:ddpm} shows the overall training and inference algorithm for a diffusion model trained via DDPM.

\begin{boxes}
\begin{tcolorbox}[title=Simplified Training Steps for Diffusion Model (DDPM), colback=gray!5, colframe=black!75, fonttitle=\bfseries\footnotesize]

\textbf{Assumptions.}

\begin{itemize}[leftmargin=*, itemsep=0.25em]
\item Clean image $x_0 \in \mathbb{R}^{H \times W \times C}$.
\item $T$: total number of diffusion steps.
\item $\{\beta_t\}_{t=1}^{T}$: variance schedule, $0 < \beta_t < 1$.
\item $\alpha_t = 1 - \beta_t,\quad
      \bar{\alpha}_t = \prod_{s=1}^{t} \alpha_s$.
\item $\varepsilon_\theta(x_t, t)$: model for predicting noise at step $t$.
\end{itemize}

\vspace{1em}
\textbf{Training Steps}
\vspace{1em}
\begin{enumerate}[leftmargin=*, itemsep=1em]

\item Start with dataset $\{x_0\}$.

\item Sample a random diffusion time
$t \sim \mathrm{Uniform}\{1,\dots,T\}$.

\item Sample Gaussian noise with the same shape as $x_0$,
$\varepsilon \sim \mathcal{N}(0, I)$.

\item Construct the noisy image at step $t$, where $\bar{\alpha}_t$ is
looked up using the sampled $t$:
\[
x_t = \sqrt{\bar{\alpha}_t}\, x_0
      + \sqrt{1 - \bar{\alpha}_t}\,\varepsilon
\]

\item Using the model, predict the noise
\[
\hat{\varepsilon} = \varepsilon_\theta(x_t, t)
\]

\item Compute the loss $L(\theta)$ and take a gradient step on $\theta$ to
minimize $L(\theta)$:
\[
L(\theta)
= \mathbb{E}_{x_0, t, \varepsilon}
  \bigl[\lVert \varepsilon - \hat{\varepsilon} \rVert^2\bigr]
\]

\end{enumerate}

\vspace{1em}
\textbf{Generation Steps}
\vspace{1em}
\begin{enumerate}[leftmargin=*, itemsep=1em]

\item Sample noise $x_T \sim \mathcal{N}(0, I)$ with shape $[H, W, C]$.

\item Reverse diffusion process.

For $t = T, T-1, \dots, 1$:
\begin{itemize}[leftmargin=*, itemsep=0.4em]
\item Predict noise at step $t$, $\hat{\varepsilon} = \varepsilon_\theta(x_t, t)$.

\item Compute the mean of the reverse distribution:
\[
\mu_\theta(x_t, t)
= \frac{1}{\sqrt{\alpha_t}}
  \left(
    x_t
    - \frac{1 - \alpha_t}{\sqrt{1 - \bar{\alpha}_t}}
      \,\hat{\varepsilon}
  \right)
\]

\item Sampling $z \sim \mathcal{N}(0, I)$ and using reverse variance $\tilde{\beta}_t$ to compute
\[
x_{t-1} = \mu_\theta(x_t, t) + \sqrt{\tilde{\beta}_t}\, z
\]

\item If $t = 1$, return the generated image $x_0$.
\end{itemize}

\end{enumerate}

\end{tcolorbox}
\caption{Training and Generation Procedure for a Denoising Diffusion Probabilistic Model (DDPM).}
\label{box:ddpm}
\end{boxes}

\begin{figure}[t]
  \centering
  \includegraphics[width=\linewidth]{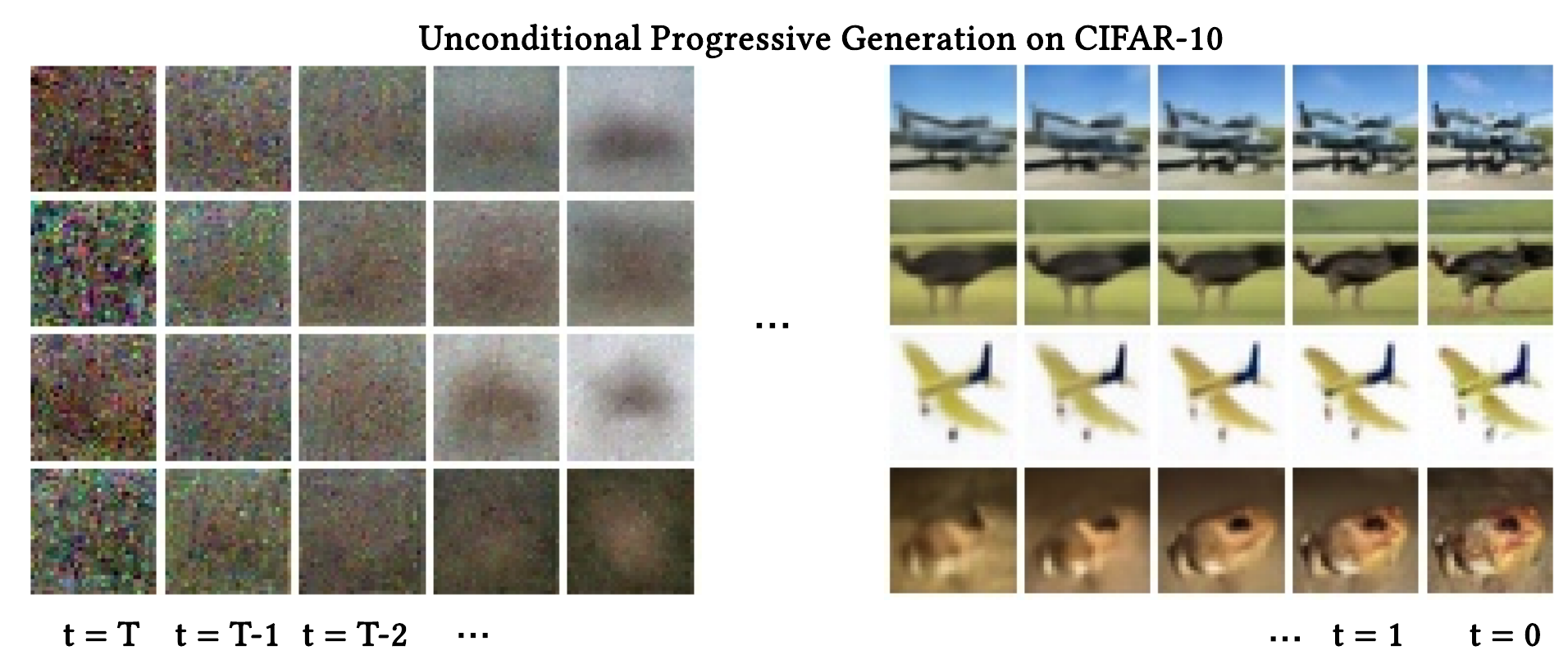}
  \caption{This figure shows unconditional progressive generation on CIFAR-10 using a diffusion model. Image source: adapted from~\cite{ho2020denoising}.}
  \label{fig:diffusion-generated}
\end{figure}

Their approach is able to match inception score (IS) and Fréchet Inception Distance (FID) of other state-of-the-art image generation models at the time. Fig.~\ref{fig:diffusion-generated} shows some of the sample images generated during the denoising process (reverse diffusion).

\subsection{Further Improvements to Diffusion Models}
While the DDPM was able to generate high-quality images on par with GANs, it had to go through thousands of time steps in order to generate a single sample. In order to accelerate this process, Song et al.~\cite{song2020denoising} introduced denoising diffusion implicit models (DDIM). They keep the training intact and their major innovation comes into play during generation process. They noticed that the loss in DDPM (Eq.~\ref{eq:loss-ddpm}) only depends on $x_0$ and the marginal $q(x_t)$. Since the network has learned how to denoise at each level $t$ and arrive at $x_0$, they do not necessarily have to move sequentially from $t = 1000$ to $t = 0$. They proposed a reverse sampling process that uses the same model, $\varepsilon_\theta(x_t, t)$, but changes how they update $x_t \rightarrow x_{t-1}$. They also introduce controllable stochastic and deterministic sampling procedures. They first use the model to predict the noise via $\hat{\varepsilon} = \varepsilon_\theta(x_t, t)$. They then use the forward marginal formula (Eq.~\ref{eq:forward-marginal}) to calculate $x_0$:
\begin{equation}
\hat{x}_0
= \frac{x_t - \sqrt{1-\bar{\alpha}_t}\,\hat{\varepsilon}}{\sqrt{\bar{\alpha}_t}}.
\end{equation}
In the stochastic version, using $\hat{x}_0$ and a free randomness parameter $\sigma_t$, they derive
\begin{equation}
x_{t-1}
= \sqrt{\bar{\alpha}_{t-1}}\,\hat{x}_0
  + \sqrt{1-\bar{\alpha}_{t-1}-\sigma_t^2}\,\hat{\varepsilon}
  + \sigma_t z
\end{equation}
where \(z \sim \mathcal{N}(0, I)\). $\sigma_t$ controls the randomness at each step. A fully stochastic version corresponds to the classic DDPM, and a less stochastic version is DDIM. Two consequences of these changes are that $\sigma_t$ acts as a control knob for sampling stochasticity. Because the model is trained to denoise any $t$, DDIM can also sample using a subset of timesteps (skipping steps) to accelerate generation. During training, the model only sees the forward noising process
\begin{equation}
x_t = \sqrt{\bar{\alpha}_t}\, x_0 + \sqrt{1-\bar{\alpha}_t}\,\varepsilon,
\qquad
\varepsilon \sim \mathcal{N}(0, I),
\end{equation}
which means that the model can denoise at any level $t$. As a result, some of the sampling steps can be skipped. Image quality does drop if the skip happens too aggressively, but empirically it works well, and they are able to generate high-quality
images $10\times$ to $50\times$ faster in wall-clock time compared to the classic DDPM.

Despite strong sample quality, many diffusion models lagged behind GANs in terms of log-likelihood. Log-likelihood captures something more fundamental: how well the model matches the data distribution. Nichol and Dhariwal~\cite{nichol2021improved} added a variational lower-bound term to the loss and gave the model more freedom (via learned variance) in order to achieve a tighter bound. Their model outputs an extra variance head $v = v_\theta(x_t, t)$, and they turn this into variances as follows:
\begin{equation}
\Sigma_\theta(x_t, t)
= \exp\bigl(v \log \beta_t + (1 - v)\log \bar{\beta}_t\bigr).
\end{equation}
The DDPM loss as shown in Eq.~\ref{eq:loss-ddpm} does not depend on the variance, and they modify it via
\begin{equation}
L_{\text{hybrid}} = L_{\text{simple}} + \lambda L_{\text{vlb}}
\end{equation}
where $L_{\text{vlb}}$ is the variational lower-bound loss. In practice, they set $\lambda$ to a small value so that $L_{\text{vlb}}$ guides the variance, but $L_{\text{simple}}$ is still the main source of influence on the output image. As a result
of this change, they directly optimize the log-likelihood bound (VLB). Additionally, they introduce a cosine scheduling scheme for $\beta_t$ instead of linear scheduling. With this change, the noise addition is spread more smoothly across timesteps and this leads to more robust sampling with fewer time steps. Table~\ref{tab:nll-comparison} shows the comparison between negative log-likelihood in their work and several other generative models on ImageNet and CIFAR datasets. Simultaneously, Song et~al.~\cite{song2020score} introduced a score based continues approach to generative modeling that relied on Stochastic Differential Equations (SDEs). They utilize SDE to transform the data distribution into a known prior distribution (noise) and a corresponding SDE for the reverse process. They were able to achieve a competitive log-likelihood of 2.99 on CIFAR as presented in the table below.

\begin{table}[t]
\centering
\begin{tabular}{lcc}
\hline
Model & ImageNet & CIFAR \\
\hline
Improved DDPM~\cite{nichol2021improved}        & 3.53 & 2.94 \\
DDPM (cont.\ flow)~\cite{song2020score}   & --   & 2.99 \\
DDPM~\cite{ho2020denoising}                 & 3.77 & 3.70 \\
Sparse Transformers~\cite{child2019generating}  & 3.44 & 2.80 \\
Routing Transformers~\cite{roy2021efficient} & 3.43 & --   \\
Pixel SNAIL~\cite{chen2018pixelsnail}          & 3.52 & 2.85 \\
\hline
\end{tabular}
\caption{Negative log-likelihood (bits per dimension) of improved DDPM compared to other generative models on ImageNet and CIFAR.}
\label{tab:nll-comparison}
\end{table}

\begin{figure}[t]
  \centering
  \includegraphics[width=\linewidth]{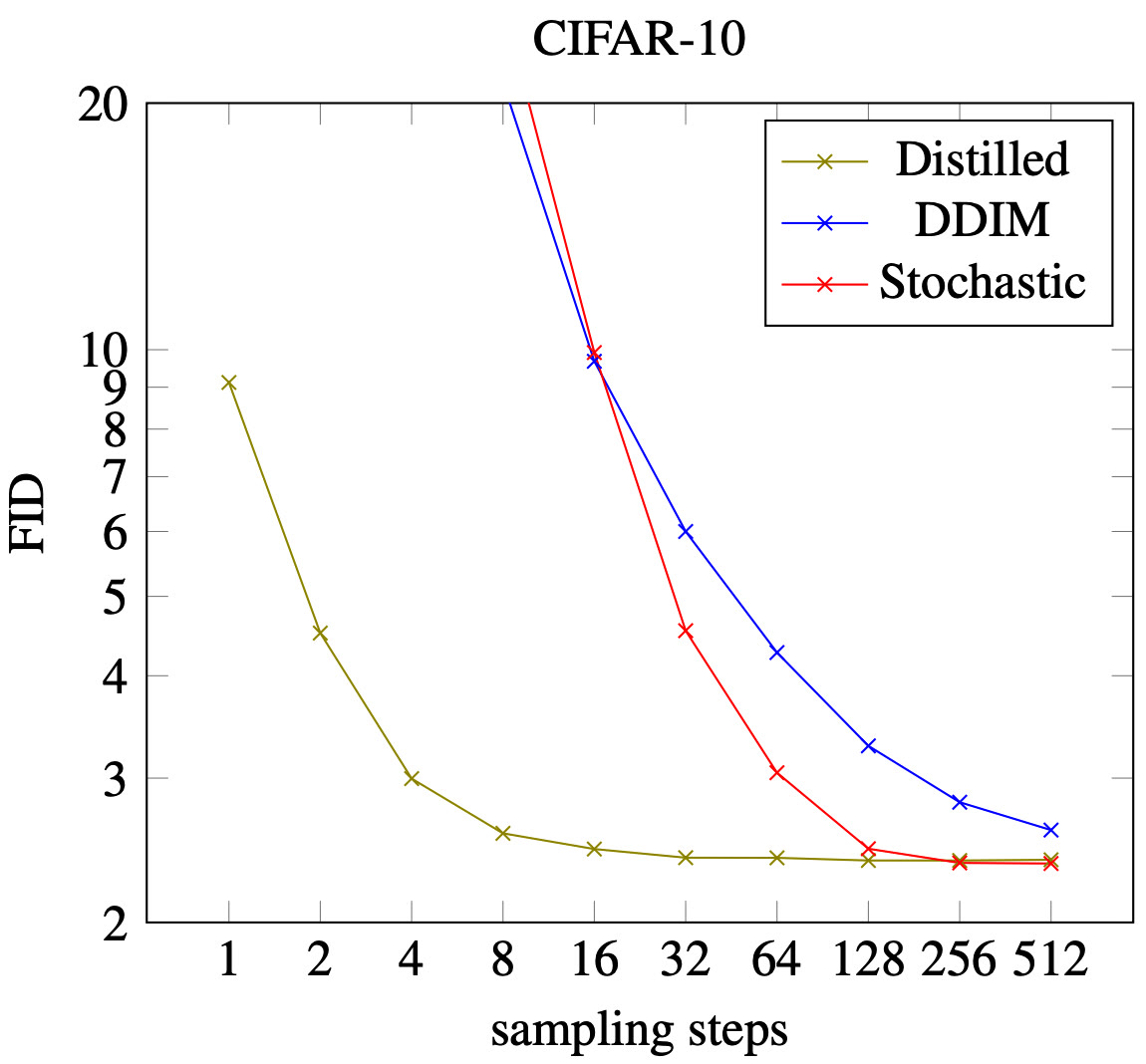}
  \caption{Progressive distillation by Salimans and Ho enables a diffusion model to generate final outputs in a considerably smaller number of steps. Image source:~\cite{salimans2022progressive}.}
  \label{fig:progressive-distillation}
\end{figure}

Salimans and Ho~\cite{salimans2022progressive} introduced the idea of
distillation for diffusion models with the goal of dramatically reducing the number of denoising steps. They started from a pretrained teacher diffusion model that must be sampled for $2N$ steps to go from pure noise to an image. The goal is to train a student model with the exact same architecture that can produce the same image as the teacher in only $N$ steps; i.e., each student step imitates two teacher steps.

\begin{equation}
\begin{aligned}
Teacher: \tau_{2N} &\rightarrow \tau_{2N-1} \rightarrow \tau_{2N-2} \rightarrow \cdots \rightarrow \tau_{0}\\
Student: t_{N} = \tau_{2N} &\rightarrow t_{N-1} = \tau_{2N-2} \rightarrow \cdots \rightarrow t_{0} = \tau_{0}
\end{aligned}
\end{equation}
In the next iteration, they replace the original teacher model with the student model and distill again by halving the number of steps. At inference time, the final student model can
be treated as a normal diffusion model that has the ability to generate a clean image in only a few steps. Fig.~\ref{fig:progressive-distillation} shows how they are able to match the FID score of DDIM or the stochastic method with considerably fewer steps.

\begin{table}[t]
\centering
\small
\resizebox{0.48\textwidth}{!}{%
\begin{tabular}{lcccc}
\hline
Model & NFE & FID $\downarrow$ & Precision $\uparrow$ & Recall $\uparrow$ \\
\hline
PD~\cite{salimans2022progressive}   & 1 & 15.39 & 0.59 & 0.62 \\
DFNO~\cite{zheng2023fast} & 1 &  8.35 &   -- &   -- \\
CD   & 1 &  6.20 & 0.62 & 0.63 \\
PD~\cite{salimans2022progressive}   & 2 &  8.95 & 0.63 & \textbf{0.65} \\
CD   & 2 &  \textbf{4.70} & \textbf{0.69} & 0.64 \\
\hline
\end{tabular}%
}
\caption{ImageNet $64 \times 64$ results comparing progressive distillation
(PD), Diffusion Fourier Neural Operator (DFNO), and consistency distillation (CD). NFE denotes the number of function evaluations.}
\label{tab:consistency-distillation}
\end{table}

\begin{figure}[t]
  \centering
  \includegraphics[width=\linewidth]{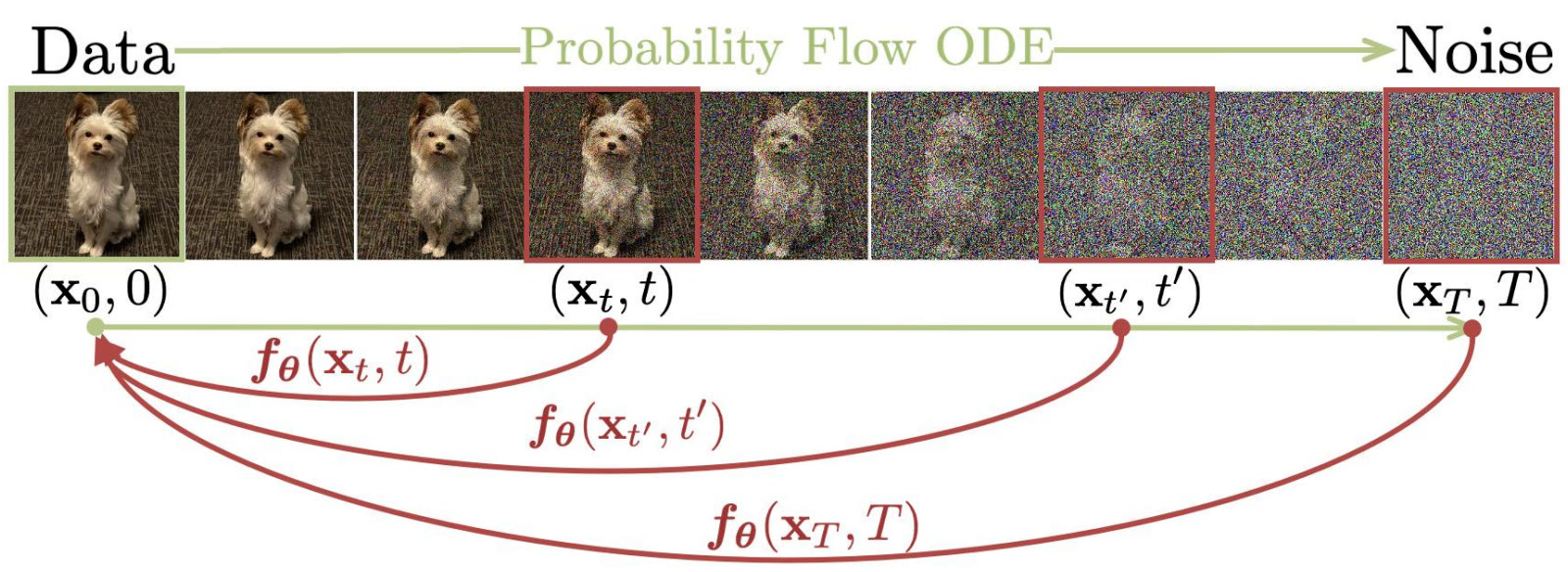}
  \caption{Consistency distillation learns to map any noisy point back to its origin $x_0$ along the generative trajectory. Image source:~\cite{song2023consistency}.}
  \label{fig:consistency-distillation}
\end{figure}

Yin et~al.~\cite{yin2024one} achieve a $100\times$ speed-up in generation by introducing Distribution Matching Distillation (DMD), which approximately minimizes a KL divergence between two diffusion models, one tracking the generator’s "fake" distribution and a fixed teacher model for the "real" distribution. Later, Song et~al.~\cite{song2023consistency} introduced consistency models for fast one step generation while still allowing multi-step generation for more quality. As shown in Fig.~\ref{fig:consistency-distillation}, they train a function $f_\theta$ such that any noisy sample $x_t$ on the same trajectory maps back to the origin, i.e., $f_\theta(x_t, t) \approx x_0$. This enables them to do one-step generation by sampling $x_T \sim \mathcal{N}(0, I)$ and outputting $\hat{x}_0 = f_\theta(x_T, T)$, where $\hat{x}_0$ is treated as the generated image.

In multi-step sampling, they start from $\hat{x}_0 = f_\theta(x_T, T)$ and continue the denoising process for a sequence of time points $t_N, t_{N-1}, \dots, t_0$ along the same trajectory. Using this approach, they are able to achieve competitive FID and IS scores compared to other diffusion models that require significantly more sampling steps (i.e., more compute time). Table~\ref{tab:consistency-distillation} shows a comparison between consistency distillation and several other diffusion models on ImageNet $64 \times 64$.

\begin{figure}[h]
  \centering
  \includegraphics[width=\linewidth]{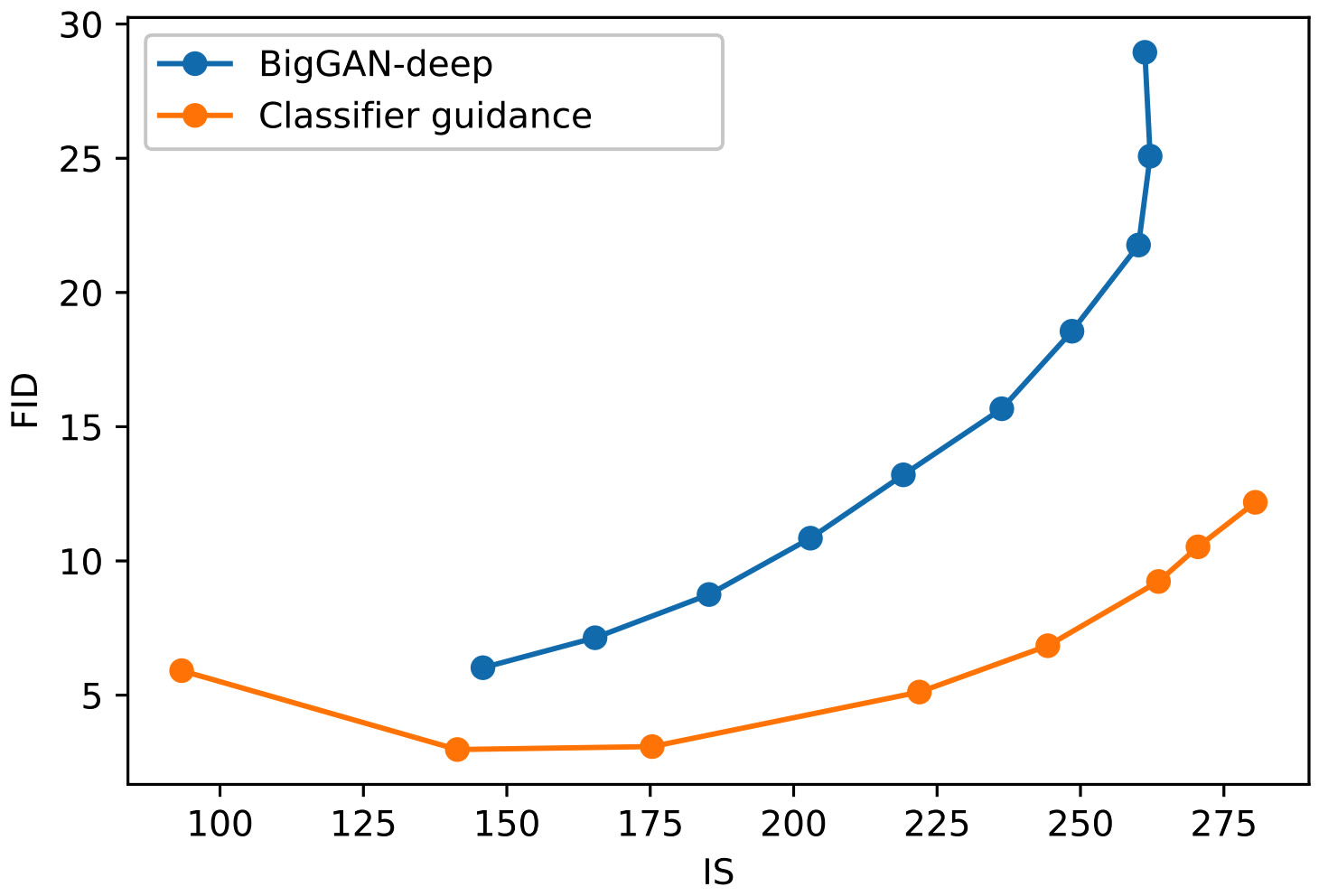}
  \caption{Classifier-guided diffusion models beating BigGAN-deep in FID scores. Lower FID score indicates more realistic images, while a higher IS score indicates more diverse images. Image source:~\cite{dhariwal2021diffusion}.}
  \label{fig:classifier-guidance}
\end{figure}

\subsection{Conditional Diffusion Models}
Dhariwal and Nichol~\cite{dhariwal2021diffusion} make further improvements to DDPM models by updating the model architecture and classifier guidance for conditional image generation. For the model, they perform a large ablation study and come up with the Ablated Diffusion Model (ADM). They modify the model architecture in the following ways: increase the model depth versus width, increase the number of attention heads, use attention heads at $32 \times 32$, $16 \times 16$, and $8 \times 8$ resolutions rather than only at $16 \times 16$, and make a few more architectural changes to the residual blocks. For conditional image generation, they exploit a separate classifier model $p_\phi(y \mid x_t)$ to further improve diffusion generation. During sampling, they modify the reverse diffusion step to shift the mean along the classifier gradient
$\nabla_{x_t} \log p_\phi(y \mid x_t)$. To condition on label $y$, they want to sample from
\begin{equation}
p_{\theta,\phi}(x_t \mid x_{t+1}, y)
\propto
p_\theta(x_t \mid x_{t+1}) \, p_\phi(y \mid x_t)
\end{equation}
Using a local Taylor approximation, this can be approximated as
\begin{equation}
x_t \sim
\mathcal{N}\Bigl(
  \mu_\theta(x_{t+1})
  + s \,\Sigma_\theta(x_{t+1}) \nabla_{x_t} \log p_\phi(y \mid x_t),
  \;\Sigma_\theta(x_{t+1})
\Bigr)
\end{equation}
where $s$ is a scaling parameter. In other words, the mean is shifted along the classifier gradient with scale $s$. Higher $s$ leads to more class consistency. With classifier guidance, they are able to beat BigGAN-deep's~\cite{brock2018large} FID and IS scores, as shown in Fig.~\ref{fig:classifier-guidance}. Note that a lower FID indicates more realistic images, while a higher IS indicates more diverse images.

\begin{figure}[h]
  \centering
  \includegraphics[width=\linewidth]{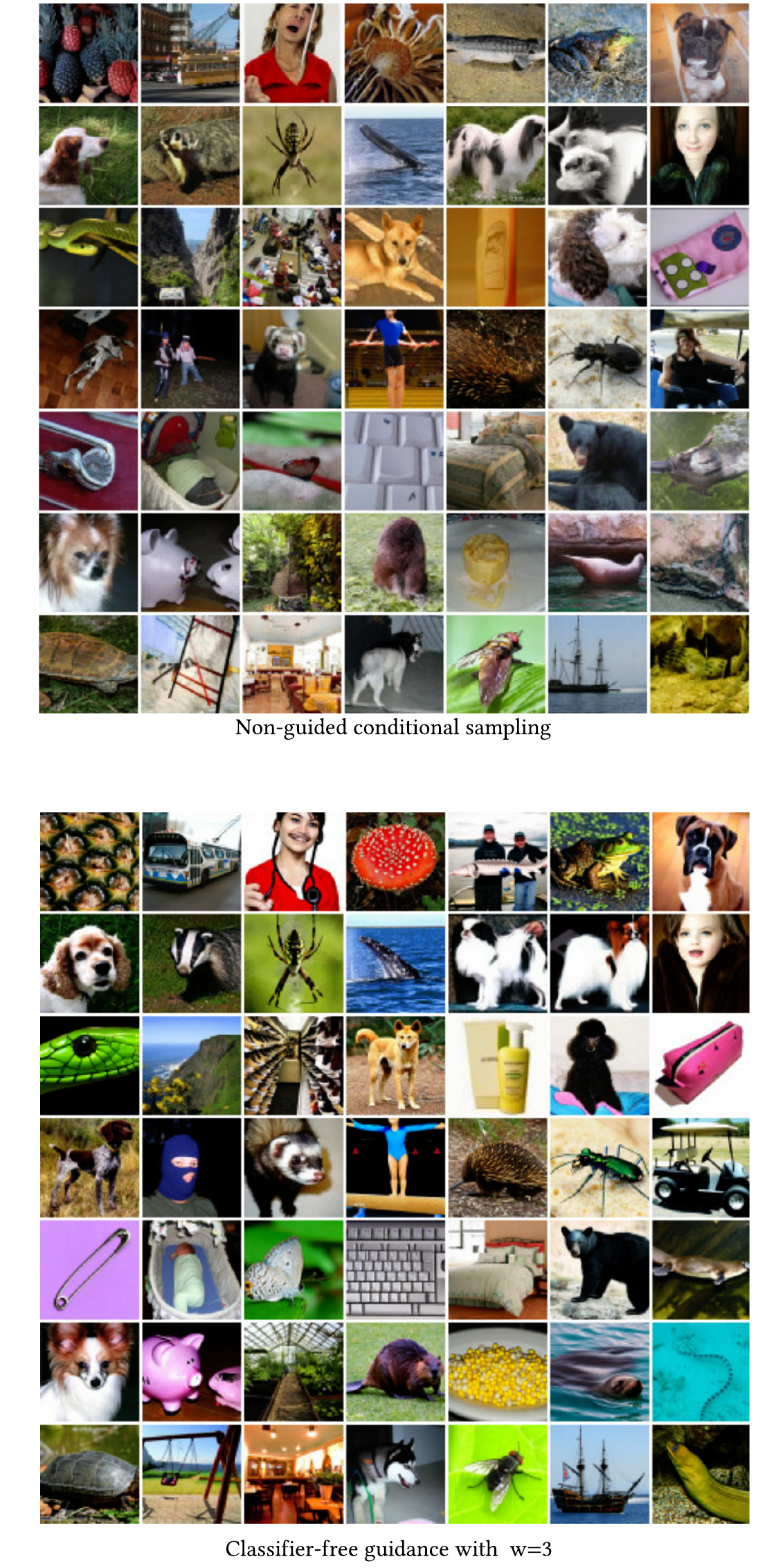}
  \caption{Top: samples from non-guided conditional sampling ($w=0$).
Bottom: samples from classifier-free guidance with $w=3$. Image source:
\cite{ho2022classifier}.}
  \label{fig:cf-guidance-samples}
\end{figure}

Unlike Dhariwal and Nichol~\cite{dhariwal2021diffusion}, Ho and Salimans~\cite{ho2022classifier} decided to eliminate the classifier and jointly train a conditional and unconditional diffusion model using only a single network. At sampling time, scores are combined as below
\begin{equation}
\tilde{\epsilon}_\theta(x_t, c)
= (1 + w)\,\epsilon_\theta(x_t, c)
  - w\,\epsilon_\theta(x_t, \varnothing)
\end{equation}
where $w$ is the guidance strength and $\varnothing$ is the null condition. During training, they sample $(x, c)$ from the dataset and, with probability $p_{\text{uncond}}$, drop the label and replace it with a special null token $\varnothing$. When $w = 0$, this is equivalent to an ordinary conditional diffusion model. Small $w$ corresponds to slightly sharper images and better FID scores. Large $w$ leads to very sharp images and high IS scores, but worsened FID. The top section of Fig.~\ref{fig:cf-guidance-samples} shows samples from non-guided conditional sampling ($w=0$), and the bottom section shows samples from classifier-free guidance with $w=3$.

Later, Nichol and Dhariwal~\cite{nichol2021glide} introduced GLIDE, where they used CLIP guidance and classifier-free guidance for text conditioning, and showed that classifier-free guidance beats CLIP guidance on different composition metrics, including human evaluations. CLIP models~\cite{radford2021learning} were first introduced as an approach to learning joint representations between text and images. CLIP provides a score for how close an image is to a text description. GLIDE applies the same idea to classifier guidance and replaces the classifier with a CLIP model. They perturb the reverse-process mean with the gradient of the dot product between the image and text embeddings:
\begin{equation}
\hat{\mu}_\theta(x_t \mid c)
= \mu_\theta(x_t \mid c)
  + s \,\Sigma_\theta(x_t \mid c)
    \nabla_{x_t}
    \sum_i f_i(x_t)\,g_i(c),
\end{equation}
where $c$ is the image caption and $f$ and $g$ are the image and text encoders from the CLIP model, respectively. This mirrors the approach in classifier guidance but replaces the classifier log-probability with CLIP image–text similarity. When they visually compare CLIP guidance to classifier-free guidance, they find that classifier-free guidance produces more realistic images. For classifier-free guidance, GLIDE uses a similar approach to Ho and Salimans~\cite{ho2022classifier}, except that GLIDE uses the image caption as the text condition and encodes it using a transformer model. Ho and Salimans~\cite{ho2022classifier} use the one-hot encoding of image labels as the condition.

Karras et al.~\cite{karras2022elucidating} realized that training diffusion models is often unnecessarily convoluted and consists of ad-hoc collections of recipes. Instead, they present a design space that tries to separate different design choices from the core concept of the model. They show that many diffusion-based models such as DDPM, DDIM, ADM, and VP/VE SDEs can be written as a denoiser for a Gaussian noisy image, plus a set of independent design choices on the noise schedule $\sigma(t)$, optimal scaling $s(t)$, preconditioning of inputs/outputs, and how to weight the loss. By combining these pieces, they improve the previous state-of-the-art FID scores on CIFAR-10 and ImageNet-64. They also show that, without training new networks, older models such as ADM can achieve better results simply by using a better sampler.

\subsection{Diffusion Models Transition into Latent Space}
\begin{figure}[t]
  \centering
  \includegraphics[width=\linewidth]{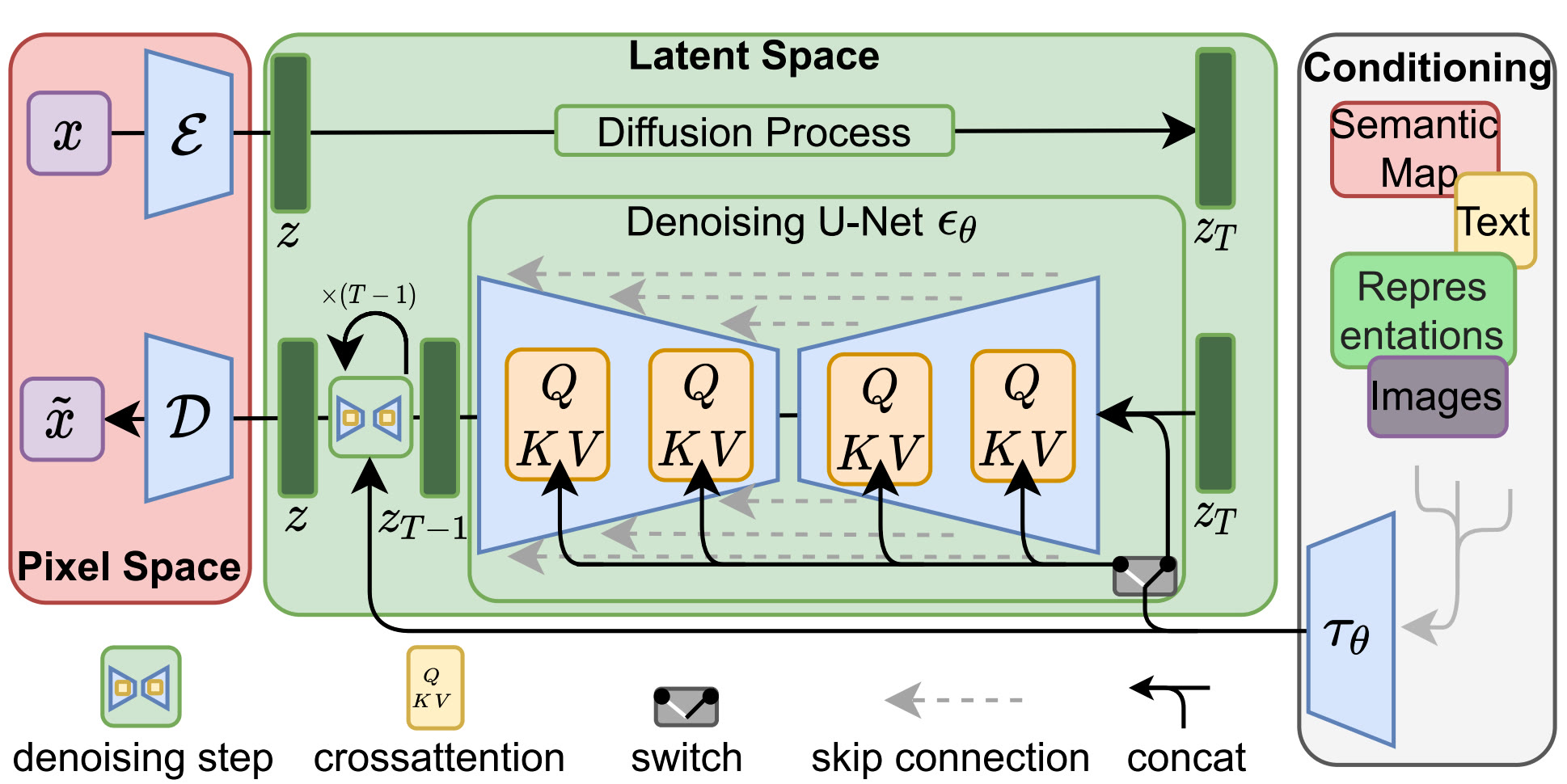}
  \caption{Overall training pipeline of latent space diffusion model where the diffusion happens on the latent space. Image source:~\cite{rombach2022high}.}
  \label{fig:ldm-arch}
\end{figure}

So far we talked about the diffusion models that operate directly on the pixel space, and this can make them costly and slow. Rombach et al.~\cite{rombach2022high} instead decided to apply diffusion models in the latent space. Like many successful image generation models, their approach consists of two stages. In stage 1, they train a perceptual autoencoder that learns to reconstruct the input image. The encoder down-samples by a factor $f$ (e.g., $4$ or $8$) to generate a latent $z$:
\begin{equation}
z = E(x) \in \mathbb{R}^{h \times w \times c},
\
h = H/f,\; w = W/f,
\end{equation}
and the decoder learns to reconstruct $\hat{x} = D(z)$. The left side of Fig.~\ref{fig:ldm-arch} depicts the autoencoder architecture.

In stage 2, they train a diffusion model in the latent space, a latent diffusion model (LDM). On the latent domain $z$, they train a classic $\epsilon$-prediction diffusion model. They also add more flexibility to diffusion models by augmenting the UNet backbone with a cross-attention mechanism for conditional image generation applications. They preprocess the condition(s) $y$ from various modalities via a domain-specific encoder $T_\theta$ that projects $y$ into an intermediate representation $\tau_\theta(y) \in \mathbb{R}^{M \times d_c}$. This is mapped to intermediate layers of the UNet via cross-attention. The conditional latent diffusion loss becomes
\begin{equation}
L_{\mathrm{LDM}}
= \mathbb{E}_{x,\, y,\, \epsilon \sim \mathcal{N}(0, I),\, t}
\Bigl[
  \bigl\|
    \epsilon - \epsilon_\theta(z_t, t, \tau_\theta(y))
  \bigr\|^2
\Bigr]
\end{equation}

\begin{figure}[h]
  \centering
  \includegraphics[width=\linewidth]{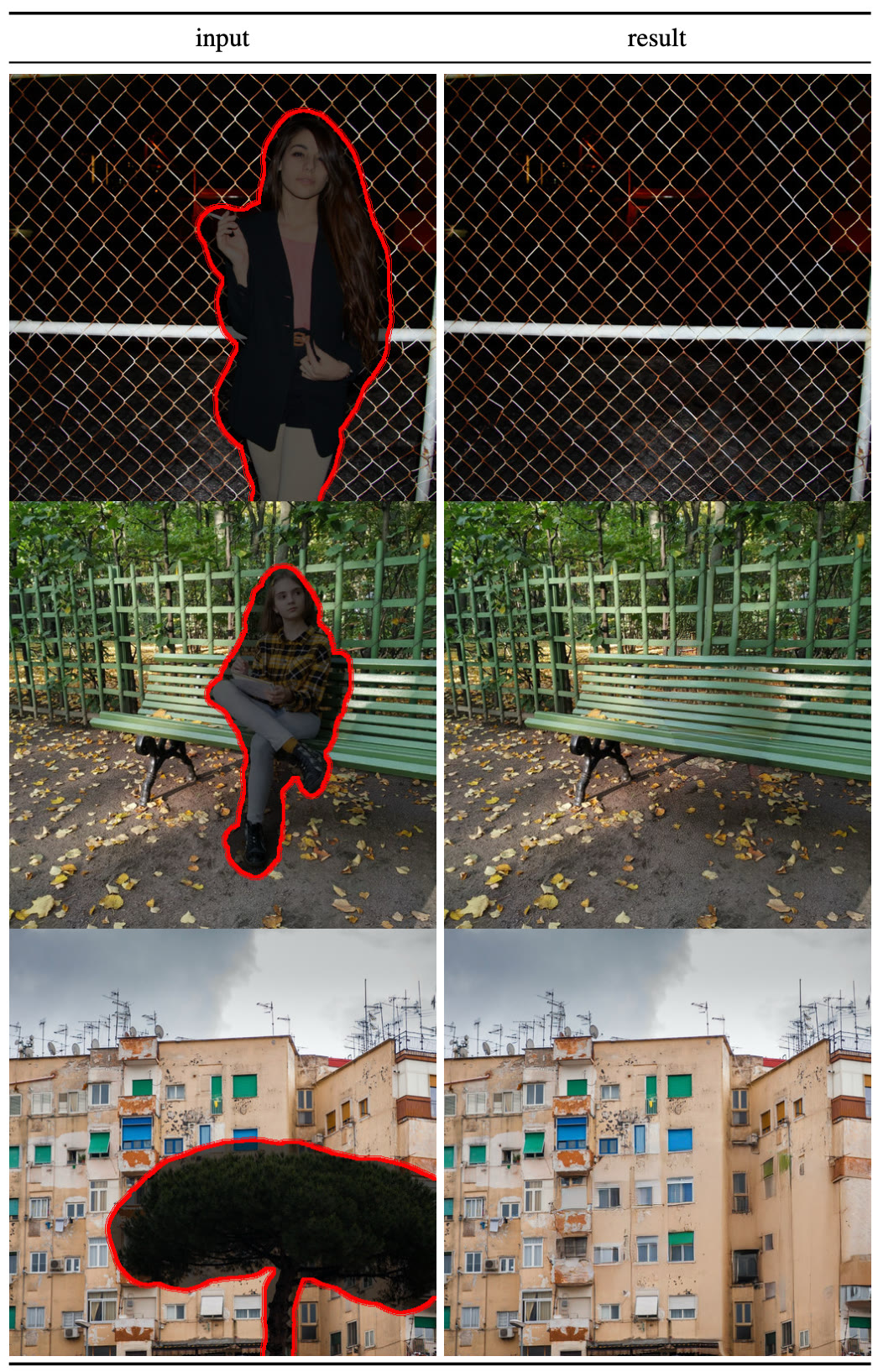}
  \caption{Latent diffusion model (LDM) maps the input condition to the intermediate layers of UNet via cross attention and generates output images based on input condition. Here the input is conditioned on removing the highlighted parts of the image. Image source:~\cite{rombach2022high}.}
  \label{fig:ldm-samples}
\end{figure}

Using this two stage approach, they are able to reduce the overall complexity of the pixel space diffusion models while preserving image quality and details. Fig.~\ref{fig:ldm-samples} shows samples of this model when conditioned on removing a highlighted part of an image.

\begin{figure}[h]
  \centering
  \includegraphics[width=\linewidth]{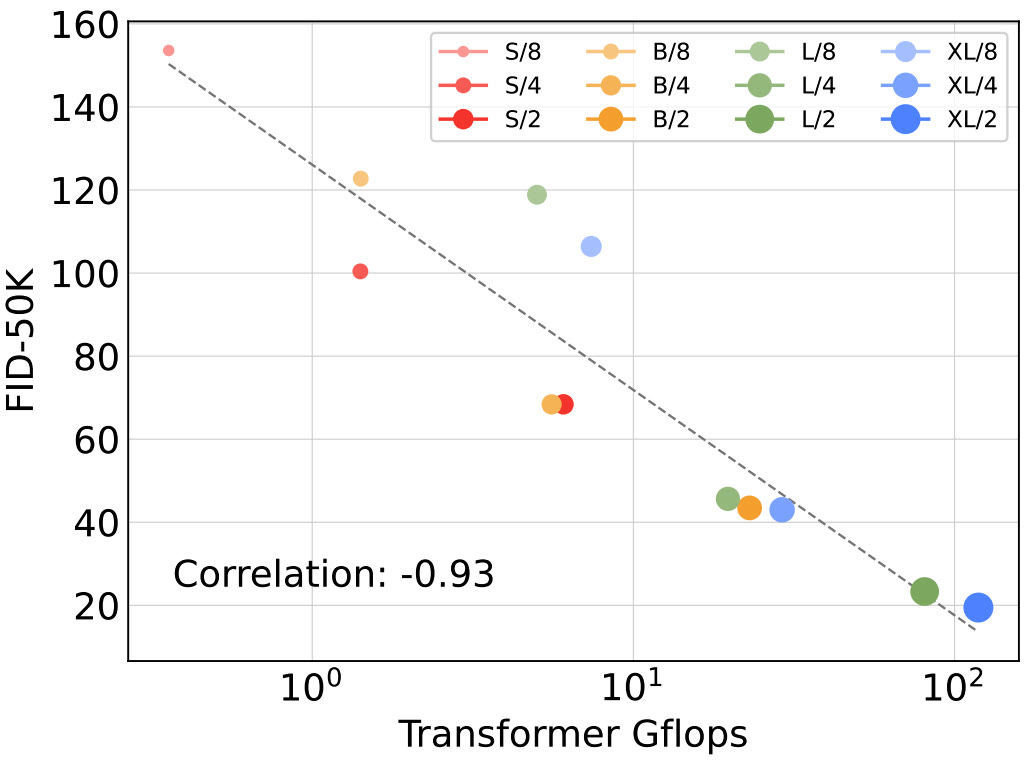}
  \caption{Transformer GFLOPs vs.\ FID for DiT models. Larger transformers achieve lower (better) FID scores. Image source:~\cite{peebles2023scalable}.}
  \label{fig:dit-gflops}
\end{figure}

\begin{figure}[h]
  \centering
  \includegraphics[width=\linewidth]{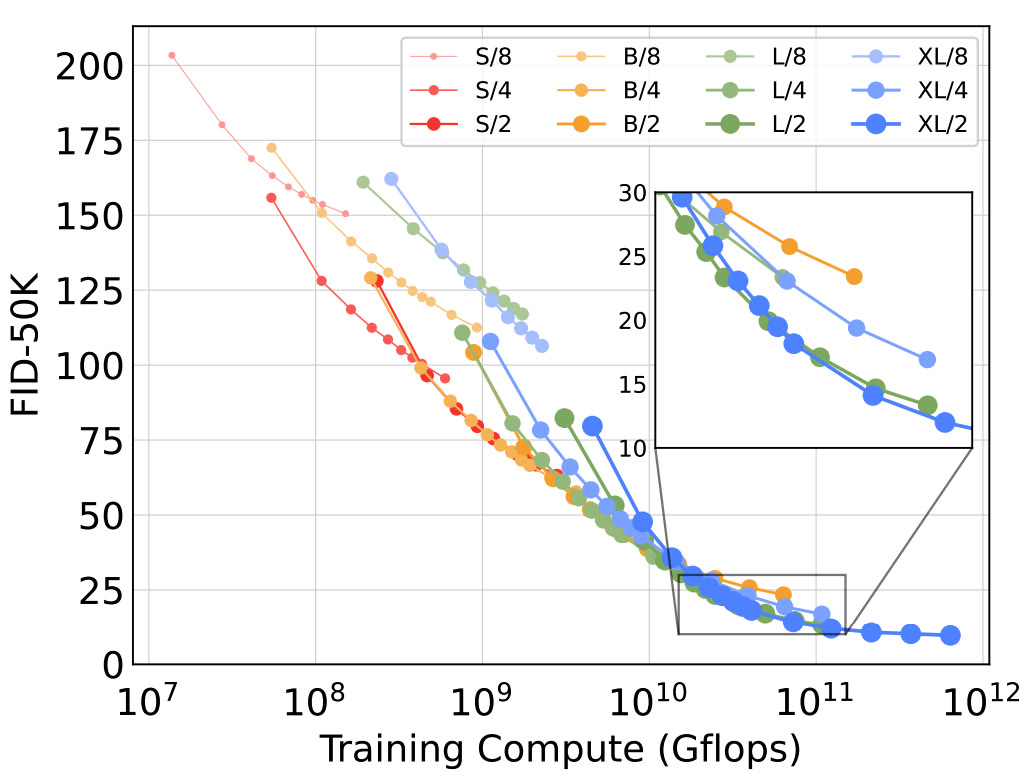}
  \caption{Compute efficiency of DiT, i.e., larger models reach better FID scores with less training. Image source:~\cite{peebles2023scalable}.}
  \label{fig:dit-efficiency}
\end{figure}

Peebles and Xie~\cite{peebles2023scalable} introduced Diffusion Transformers (DiT), where they replace the usual UNet backbone with a vision-transformer(ViT) style model that operates on latent patches. This replaces the UNet style models
that were common in diffusion-based image generation. They first encode the input image with a frozen pretrained VAE into a low-resolution latent (e.g., $32 \times 32 \times 4$ for a $256 \times 256 \times 3$ image). They then add noise in the latent space and patchify the noisy output before feeding the patches to the transformer model, along with timestep and class conditioning. The transformer is trained to predict the noise at each time step. At generation time, they denoise the latent iteratively and finally decode it back to an RGB image with the VAE decoder. After making these architectural changes, they find that increasing transformer GFLOPs (via depth or width of the transformer) monotonically improves FID scores, as shown in Fig.~\ref{fig:dit-gflops}. In other words, transformer
capacity is strongly correlated with FID. They also demonstrate that larger models are more compute efficient, as shown in Fig.~\ref{fig:dit-efficiency}. For the same amount of compute, larger DiT models are able to reach better FID scores.

\begin{figure}[h]
  \centering
  \includegraphics[width=\linewidth]{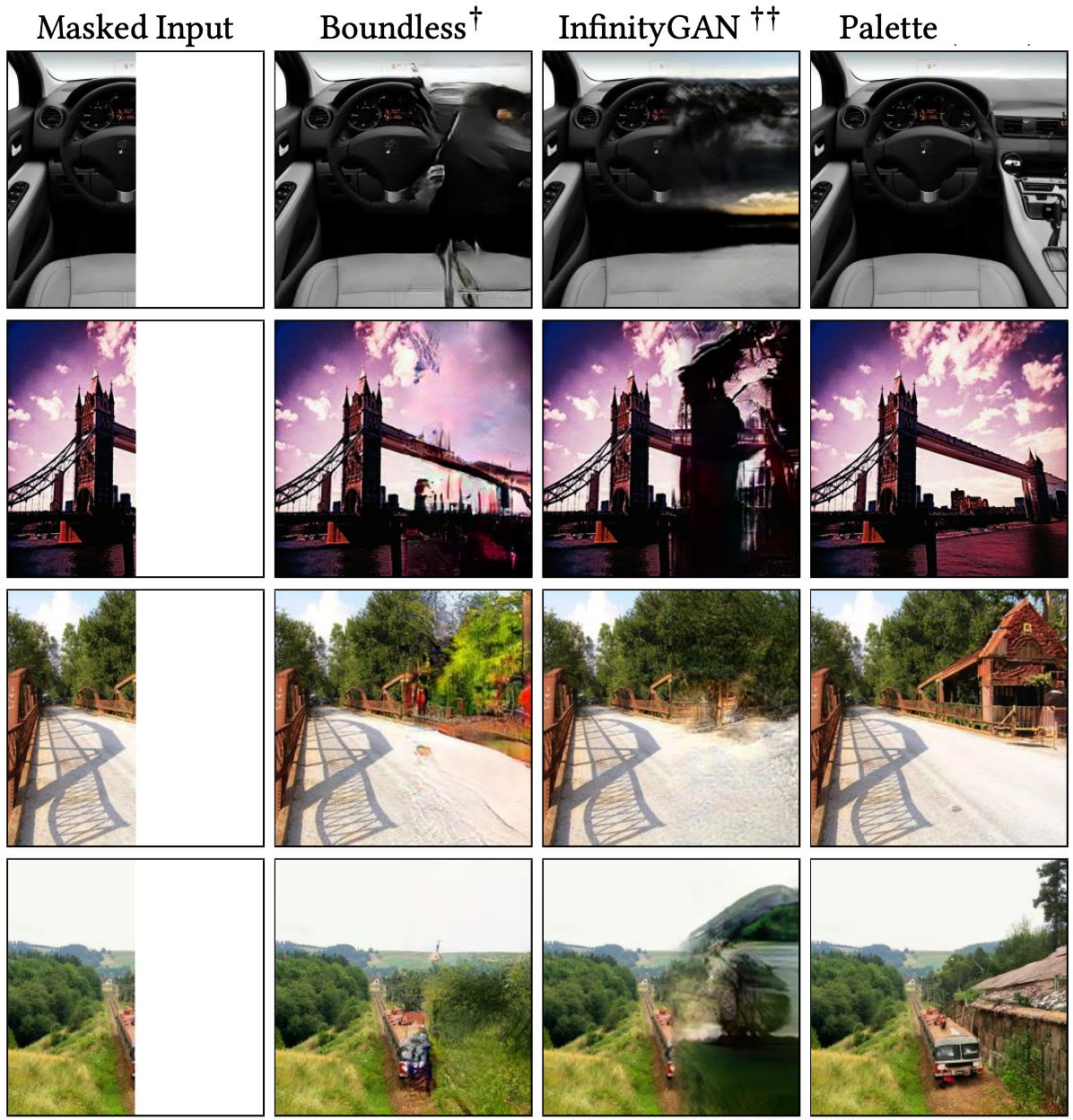}
  \caption{Image uncropping results comparing Palette, Boundless, and InfinityGAN on the Places2~\cite{zhou2017places} dataset. Image source:~\cite{saharia2022palette}.}
  \label{fig:palette-uncrop}
\end{figure}

Saharia et al.~\cite{saharia2022palette} introduced Palette, a unified diffusion-based generative model for different conditional image-to-image translation tasks, including colorization, inpainting, uncropping, and JPEG restoration. Given a training output image $y$, they add noise to it and generate $\tilde{y}$ as
\begin{equation}
\tilde{y} = \sqrt{\gamma}\, y + \sqrt{1-\gamma}\, \epsilon,
\qquad
\epsilon \sim \mathcal{N}(0, I)
\end{equation}
They then train an ADM-style UNet~\cite{dhariwal2021diffusion} $f_\theta$ to predict the noise $\epsilon$ from $x$, $\tilde{y}$, and the noise level $\gamma$. The loss for the neural network is denoted as
\begin{equation}
\mathbb{E}_{(x, y)} \,
\mathbb{E}_{\gamma,\,\epsilon \sim \mathcal{N}(0, I)}
\
  \| f_\theta(x, \tilde{y}, \gamma) - \epsilon \|^p
\
\end{equation}
where $p$ corresponds to an $L_1$ or $L_2$ norm. Fig.~\ref{fig:palette-uncrop} shows a comparison between their approach and other techniques such as Boundless~\cite{teterwak2019boundless} and InfinityGAN~\cite{lin2021infinitygan}.

\subsection{Scaling up Diffusion Models}
Ramesh et al.~\cite{ramesh2022hierarchical} introduced hierarchical
text-conditional image generation with CLIP latents, also known as DALL-E 2. They leverage a large pretrained CLIP model in order to guide the generation process in diffusion models, trained on roughly 650M image and caption pairs. They then freeze the CLIP model and train two subsequent models, $\theta$ and $\phi$. Given an input image $x$, the CLIP model produces $z_{\text{img}}$ and $z_{\text{txt}}$ for the corresponding image–caption pairs. The two components of the model consist of a prior $p_\phi(z_{\text{img}} \mid y)$ and a decoder $p_\theta(x \mid z_{\text{img}}, y)$ that produces images $x$ given the embedding of the image $z_{\text{img}}$ and optionally caption $y$:
\begin{equation}
p(x \mid y)
= p(x, z_{\text{img}} \mid y)
= p_\theta(x \mid z_{\text{img}}, y)\, p_\phi(z_{\text{img}} \mid y)
\end{equation}

\begin{figure}[h]
  \centering
  \includegraphics[width=\linewidth]{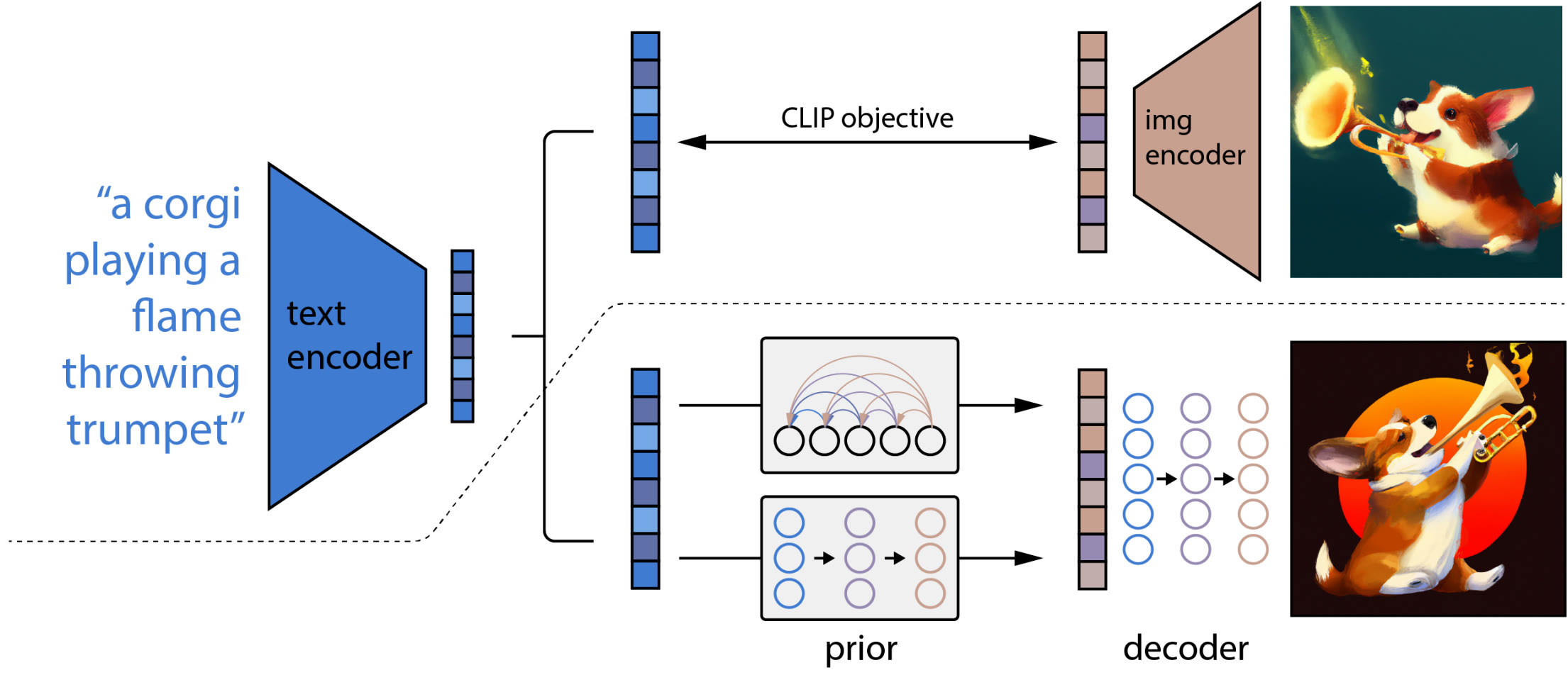}
  \caption{The overall architecture for DALL-E 2. The authors gave it the name un-CLIP since they are effectively generating an image from an image embedding (the reverse direction of CLIP). Image source:~\cite{ramesh2022hierarchical}.}
  \label{fig:unclip-arch}
\end{figure}

Note that $p(x \mid y) = p(x, z_{\text{img}} \mid y)$ since $z_{\text{img}}$ can be obtained from $x$ deterministically via the CLIP model. They also give the name un-CLIP to their approach, since they are able to generate images from image embeddings—the reverse of what CLIP does. Fig.~\ref{fig:unclip-arch} shows their overall architecture and Box~\ref{box:unclip-training} shows their overall algorithm.

\begin{boxes}
\begin{tcolorbox}[title=Simplified Training Algorithm for DALL-E 2 (un-CLIP), colback=gray!5, colframe=black!75, fonttitle=\bfseries\footnotesize]

\begin{enumerate}[leftmargin=*, itemsep=1em]

\item Train a CLIP model on image–caption pairs $(x, y)$.

\item Freeze the CLIP model and compute image and text embeddings via
\[
z_{\mathrm{img}} = E_{\mathrm{CLIP,img}}(x),
\qquad
z_{\mathrm{txt}} = E_{\mathrm{CLIP,txt}}(y).
\]

\item Train the diffusion decoder $p_\theta(x \mid z_{\mathrm{img}}, y)$ for $N$ iterations.

For each mini-batch (size $m$):

\begin{itemize}[leftmargin=*, itemsep=0.4em]
\item Compute CLIP embeddings $z_{\mathrm{img}}^{(i)}, z_{\mathrm{txt}}^{(i)}$ from the input images $x^{(i)}$.
\item Sample diffusion time steps and noise:
\[
t \sim \mathrm{Uniform}\{1,\dots,T\},
\qquad
\epsilon^{(i)} \sim \mathcal{N}(0, I).
\]
\item Perform the forward noising process to obtain $x_t^{(i)}$.
\item Do a forward pass through the decoder UNet:
\[
\hat{\epsilon}^{(i)}
= \epsilon_\theta\bigl(x_t^{(i)}, t, z_{\mathrm{img}}^{(i)}, z_{\mathrm{txt}}^{(i)}\bigr)
\]
\item Compute the noise-prediction loss and update weights using
\[
L_{\mathrm{decoder}}
= \frac{1}{m} \sum_{i}
  \left\lVert \epsilon^{(i)} - \hat{\epsilon}^{(i)} \right\rVert_2^2
\]
\end{itemize}

\item Train the diffusion prior $p_\phi(z_{\mathrm{img}} \mid y)$ for $N'$ iterations.

For each mini-batch (size $m$):

\begin{itemize}[leftmargin=*, itemsep=0.4em]
\item Compute CLIP embeddings $z_{\mathrm{img}}^{(i)}, z_{\mathrm{txt}}^{(i)}$.
\item Sample diffusion time steps and noise:
\[
t \sim \mathrm{Uniform}\{1,\dots,T\},
\
\epsilon^{(i)} \sim \mathcal{N}(0, I)
\]
\item Generate noisy CLIP image embeddings via the forward noising process
\[
z_{\mathrm{img},t}^{(i)}
= \text{forward\_noising}\bigl(z_{\mathrm{img}}^{(i)}, t, \epsilon^{(i)}\bigr)
\]
\item Run the reverse diffusion process with the prior model
\[
\hat{z}_{\mathrm{img}}^{(i)}
= f_\phi\bigl(z_{\mathrm{img},t}^{(i)}, t; z_{txt}^{(i)}\bigr)
\]
\item Compute the denoising loss and update weights by $\nabla_\phi L_{\mathrm{prior}}$:
\[
L_{\mathrm{prior}}
= \frac{1}{m} \sum_{i}
  \left\lVert \hat{z}_{\mathrm{img}}^{(i)} - z_{\mathrm{img}}^{(i)} \right\rVert_2^2
\]
\end{itemize}

\end{enumerate}
\end{tcolorbox}
\caption{Simplified training steps for the DALL-E 2 unCLIP model.}
\label{box:unclip-training}
\end{boxes}

\begin{figure}[h]
  \centering
  \includegraphics[width=0.6\linewidth]{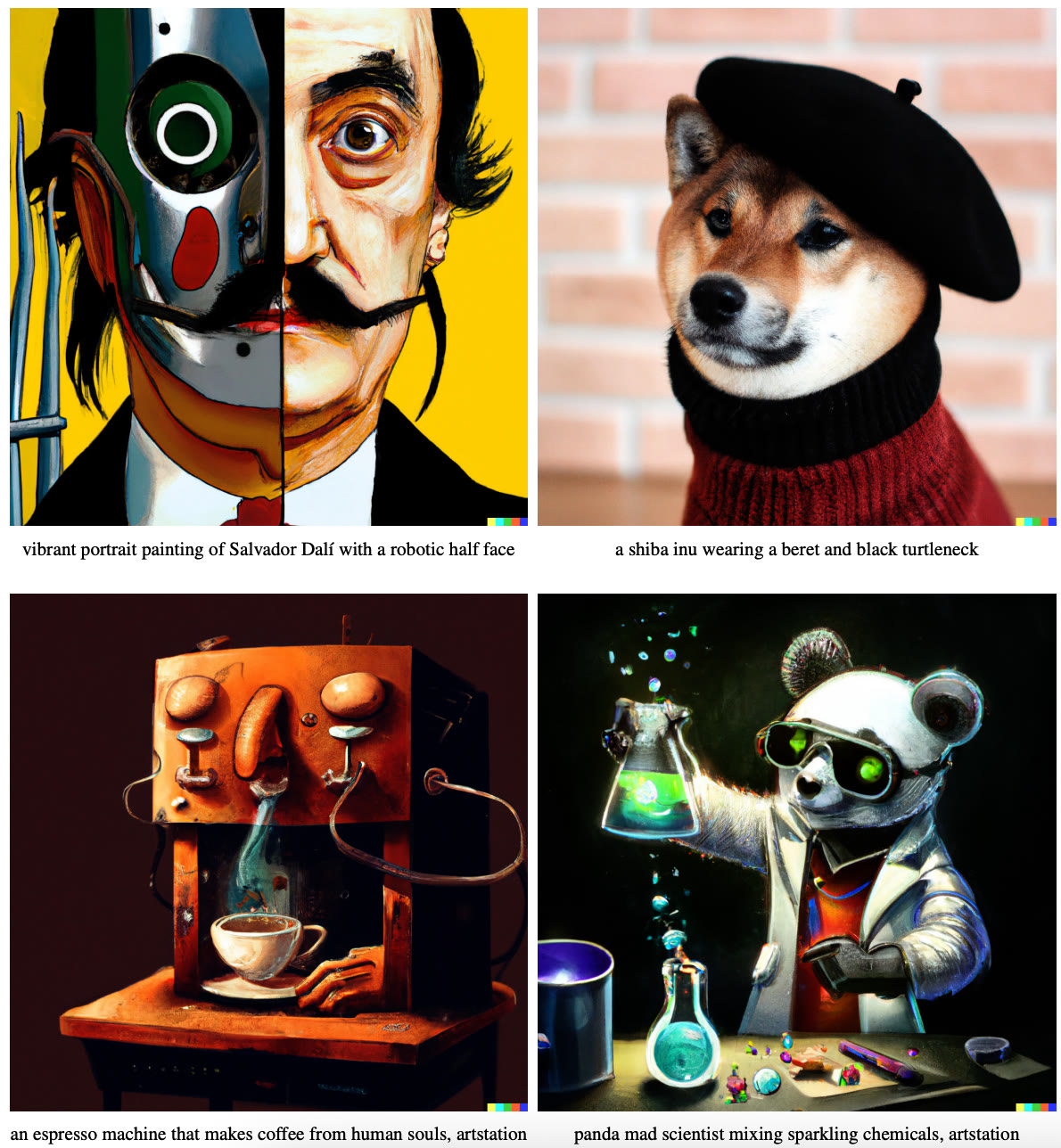}
  \caption{A few sample $1024 \times 1024$ images from the un-CLIP (DALL-E 2) model. Image source:~\cite{ramesh2022hierarchical}.}
  \label{fig:unclip-sample}
\end{figure}

Fig.~\ref{fig:unclip-sample} shows some of the sample images generated by un-CLIP (DALL-E 2). Note the high amount of control over the output images made possible by the combination of CLIP models and a pipeline of diffusion models. In order to be able to generate high quality images, the authors also utilized upsampling diffusion models at the end.

\begin{figure}[h]
  \centering
  \includegraphics[width=0.8\linewidth]{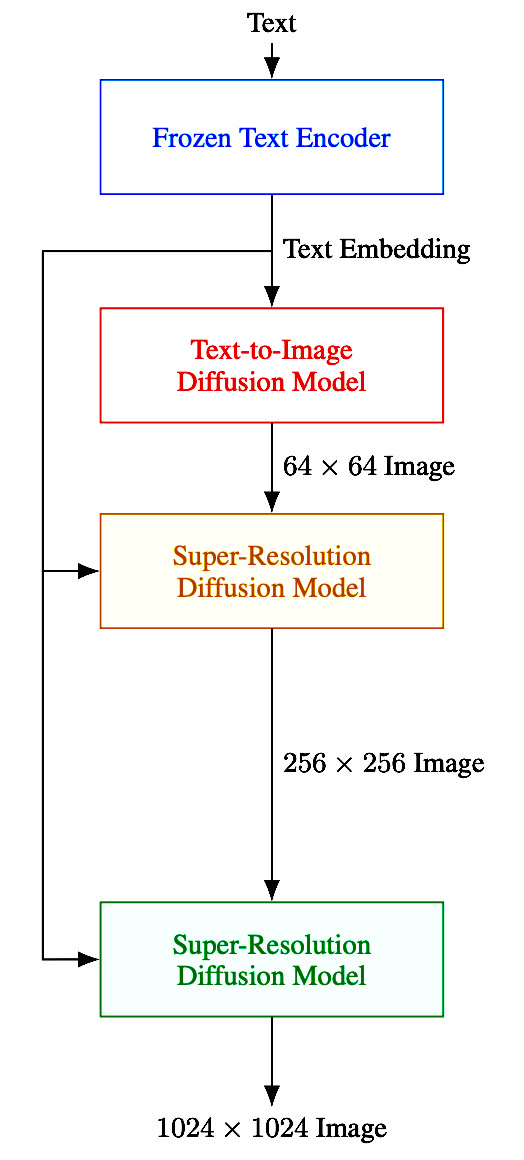}
  \caption{Oversimplified version of the Imagen model pipeline. Image source:~\cite{saharia2022photorealistic}.}
  \label{fig:imagen-arch}
\end{figure}

\begin{figure}[h]
  \centering
  \includegraphics[width=\linewidth]{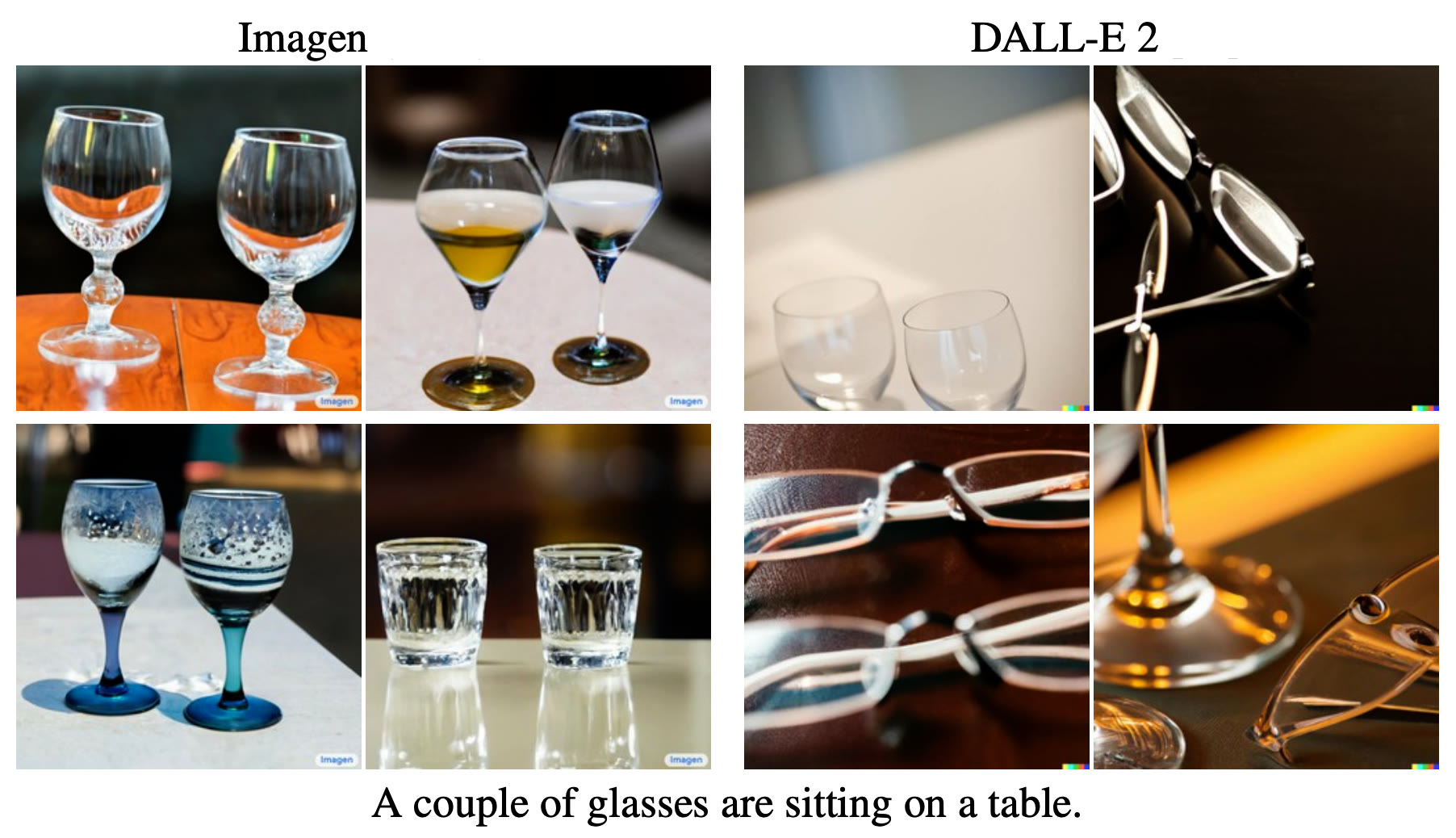}
  \caption{Image samples generated by Imagen (left) and DALL-E 2 (right). Image source:~\cite{saharia2022photorealistic}.}
  \label{fig:imagen-samples}
\end{figure}

Later, Saharia et al.~\cite{saharia2022photorealistic} introduced Imagen, where the model is architecturally simpler than DALL-E~2 but uses a very large pretrained T5~\cite{raffel2020exploring} model (T5-XXL) as the text encoder. They use three diffusion models as depicted in Fig.~\ref{fig:imagen-arch}: a base diffusion model to generate a $64 \times 64$ image, a second model to upsample images to $256 \times 256$, and a final diffusion model to upsample to $1024 \times 1024$. The base diffusion model consists of an efficient UNet–style noise predictor
\begin{equation}
\hat{\varepsilon}
= \varepsilon_\theta(z_t, t, e)
\end{equation}
where $z_t$ is the noisy $64 \times 64$ RGB image, $t$ is the timestep, and $e$ is the text embedding. They compute a reweighted MSE loss
\begin{equation}
L
= \mathbb{E}_{t,z_t,\varepsilon}
  \bigl[w(t)\,\lVert \varepsilon - \hat{\varepsilon} \rVert_2^2\bigr]
\end{equation}
where $w(t)$ is a cosine-based weight schedule. At the end, they use the two additional diffusion models to further upsample the generated $64 \times 64$ image to  $1024 \times 1024$. Table~\ref{tab:imagen-fid} compares zero-shot FID scores between Imagen and other state of the art models. As you can see, Imagen significantly improves the FID scores on the MS-COCO dataset compared to other models. Fig.~\ref{fig:imagen-samples} compares a few sample images generated by Imagen and DALL-E~2.

\begin{table}[t]
\centering
\small
\begin{tabular}{lc}
\hline
Model   & Zero-shot FID-30K \\
\hline
DALL-E  & 17.89 \\
GLIDE   & 12.24 \\
DALL-E 2& 10.39 \\
Imagen  & \textbf{7.27} \\
\hline
\end{tabular}
\caption{Zero-shot FID-30K scores on MS-COCO comparing Imagen to other
text-to-image models.}
\label{tab:imagen-fid}
\end{table}

\begin{figure}[h]
  \centering
  \includegraphics[width=\linewidth]{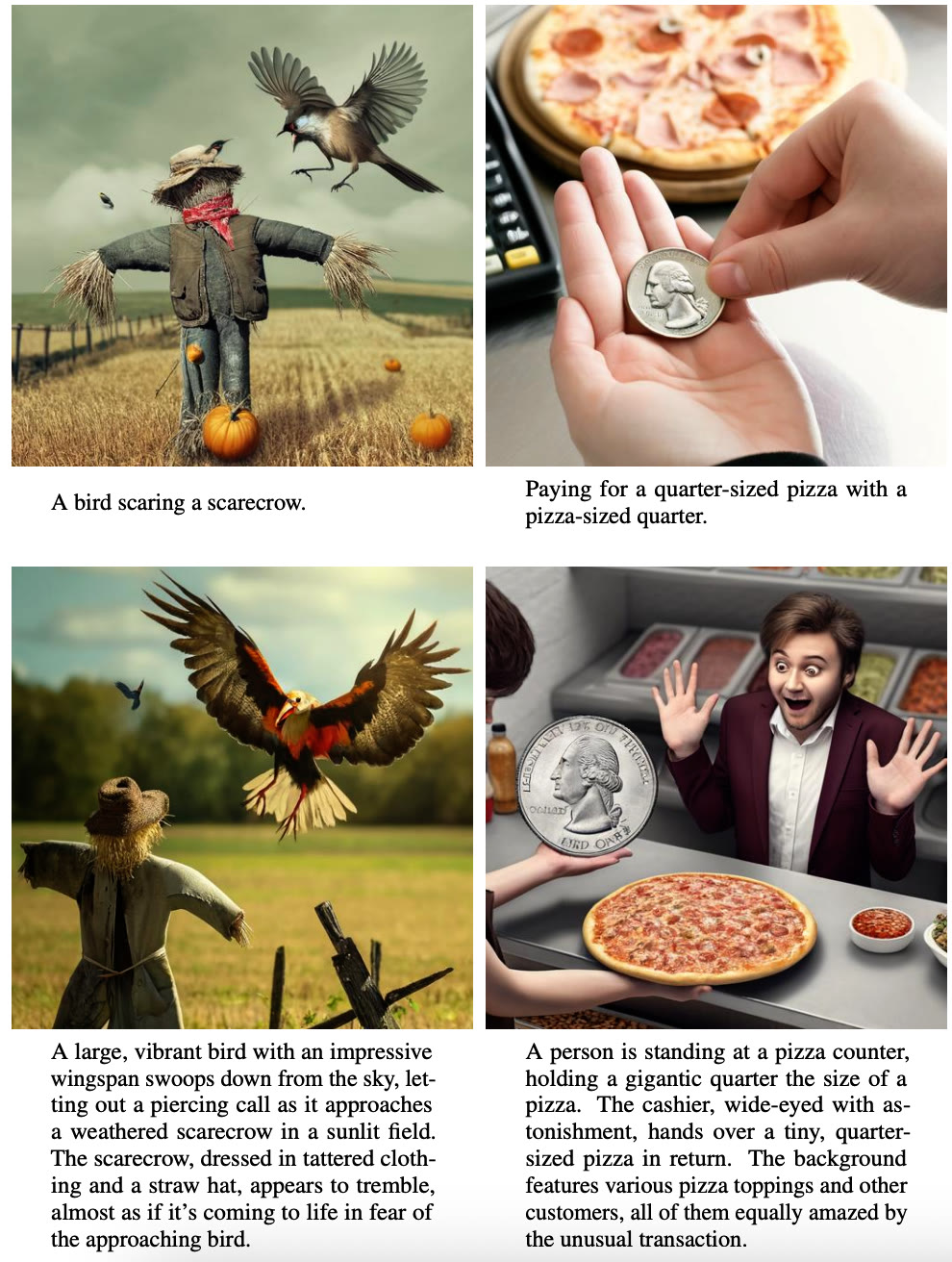}
  \caption{Top and bottom images show the original and upsampled image
captions, respectively. Images generated by DALL-E~3. Image source:~\cite{betker2023improving}.}
  \label{fig:dalle3-upsampled}
\end{figure}

Betker et al.~\cite{betker2023improving} introduced DALL-E 3. The authors realized that existing text-to-image generators struggled to follow prompts and image descriptions. They suspected that this might be caused by inaccurate image captions in the training dataset. To alleviate this problem, they trained a separate image captioner and re-captioned the training set. They then train a text-to-image latent diffusion model. They are able to generate output images that are consistently preferred by human evaluators.
Table~\ref{tab:dalle3-comparison} summarizes human evaluation results comparing DALL-E 3 versus other state of the art image generation models. Fig.~\ref{fig:dalle3-upsampled} shows the effects of using “upsampled’’ captions.

\begin{table}[t]
\centering
\small
\setlength{\tabcolsep}{3pt} % tighter column spacing
\resizebox{\columnwidth}{!}{%
\begin{tabular}{lcccc}
\hline
Dataset & DALL-E 3 & Midjourney 5.2 & Stable Diffusion XL & DALL-E 2 \\
\hline
DALL-E 3 Eval (prompt following) & \textbf{153.3} & -104.8 & -189.5 & -- \\
DALL-E 3 Eval (style)           & \textbf{74.0}  & 30.9   & -95.7  & -- \\
MSCOCO (coherence)              & \textbf{71.0}  & 48.9   & -84.2  & -- \\
Drawbench                       & \textbf{61.7}  & --     & -34.0  & -79.3 \\
\hline
\end{tabular}%
}
\caption{Comparing DALL-E 3 human evaluation performance against different image generation models across several datasets. Table from~\cite{betker2023improving}}
\label{tab:dalle3-comparison}
\end{table}

\begin{figure}[!t]
  \centering
  \includegraphics[width=\linewidth]{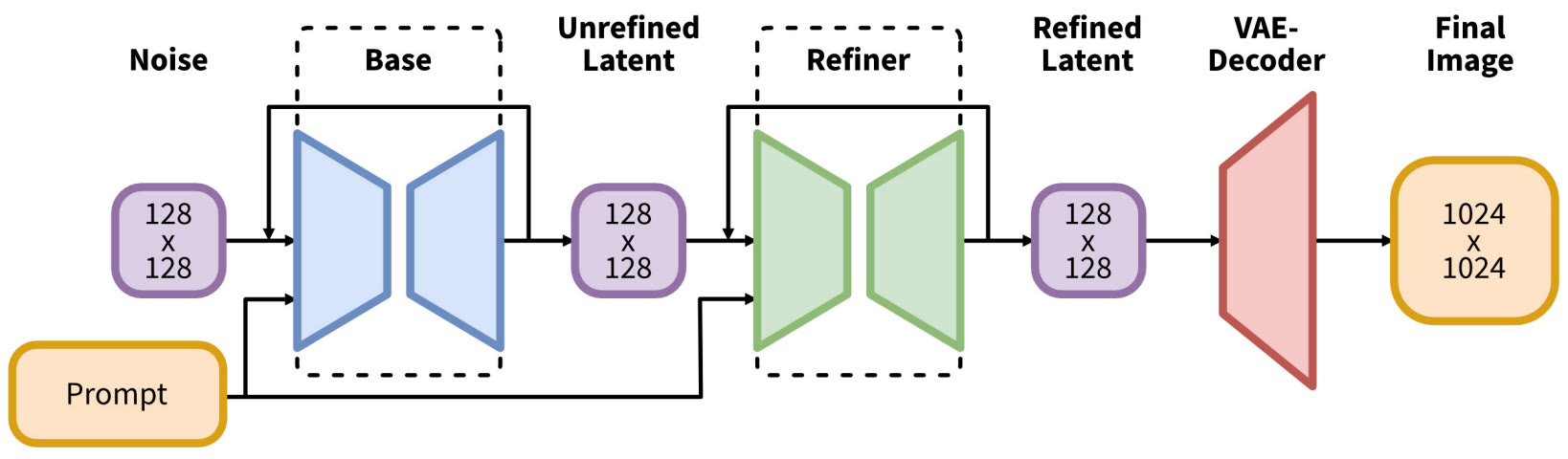}
  \caption{Two-stage pipeline for SDXL: a base model followed by a
refinement model for high-resolution image synthesis. Image source:~\cite{podell2023sdxl}.}
  \label{fig:sdxl-arch}
\end{figure}

\begin{figure}[!t]
  \centering
  \includegraphics[width=\linewidth]{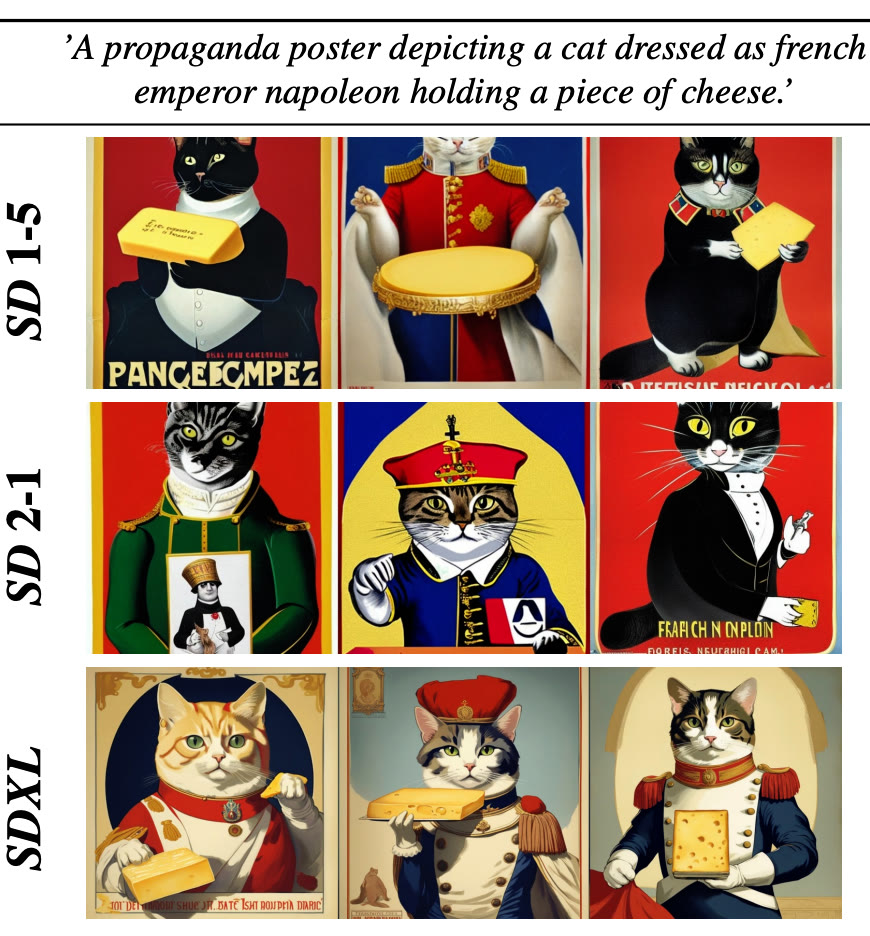}
  \caption{Sample images generated by SDXL versus Stable Diffusion
(SD) v1.5 and v2.1. Image source:~\cite{podell2023sdxl}.}
  \label{fig:sdxl-samples}
\end{figure}

At approximately the same time, Podell and coauthors~\cite{podell2023sdxl} introduced Stable Diffusion XL (SDXL) for high-resolution image synthesis. Compared to Stable Diffusion v1 or v2, they utilize a UNet backbone that is roughly three times larger. They also use richer text conditioning by concatenating the outputs of two encoders: CLIP ViT-L~\cite{radford2021learning} and OpenCLIP ViT-bigG. In addition, they include a refinement model at the end of the pipeline that is responsible for generating the final high-resolution images. Fig.~\ref{fig:sdxl-arch} shows the high-level diagram of their approach. Table~\ref{tab:sdxl-size} shows a comparison between SDXL and older Stable Diffusion (SD) models v1.x and v2.x. Fig.~\ref{fig:sdxl-samples} shows sample images generated by SDXL and compares them with Stable Diffusion (SD) v1.5 and v2.1.

\begin{table}[!t]
\centering
\small
\resizebox{0.48\textwidth}{!}{%
\begin{tabular}{lccc}
\hline
Model & SDXL & SD 1.4/1.5 & SD 2.0/2.1 \\
\hline
\# of UNet params   & 2.6B & 860M & 865M \\
Transformer blocks   & [0, 2, 10] & [1, 1, 1, 1] & [1, 1, 1, 1] \\
Channel mult.        & [1, 2, 4] & [1, 2, 4, 4] & [1, 2, 4, 4] \\
Text encoder         & CLIP ViT-L \& OpenCLIP ViT-bigG & CLIP ViT-L & OpenCLIP ViT-H \\
Context dim.         & 2048 & 768 & 1024 \\
Pooled text emb.     & OpenCLIP ViT-bigG & N/A & N/A \\
\hline
\end{tabular}%
}
\caption{Key architectural differences between SDXL and earlier Stable Diffusion (SD) models.}
\label{tab:sdxl-size}
\end{table}

\subsection{Conclusion}
Diffusion models started almost ten years ago as image denoising Gaussian models and have since evolved into a dominant choice for high quality image generation. Over time, the training objective evolved into models that gradually remove noise from the input via a sampling process. The training objectives of these models were continually refined, and more efficient sampling schemes were introduced without the need to retrain the models. The diffusion models also shifted from models that denoise the input image directly to models that denoise the latent space of the input image. The core model backbone has also shifted to transformer based models. Over the past five years, we have seen a proliferation of a wide variety of diffusion models. Recently, many state of the art diffusion systems have been released primarily as products, with fewer technical details disclosed. The big labs and companies shared fewer details about their image generation models. However, the field and community have remained strong and research continues to progress rapidly.

% The field of image generation is moving at an exciting rapid pace and as a result there is a vast volume of published work that try to enhance the current state of the art results. Here, we try to offer some insights into the recent trends in research.

% - Rectified Flow
% - Other topics that are kind of new and never made it into the production grade models

\section{Recent Developments in Image Generation}

Many top image generation systems have shifted from classic diffusion based formulations and moved toward Rectified Flow and Flow Matching, motivated by cleaner continuous time dynamics, improved training stability at scale, and image generation in fewer sampling steps. Moreover, the overall system has become more complicated and now consists of many pieces, including a large language model providing strong text understanding and better instruction following, more capable text encoders for precise conditioning, and a strong image backbone (often a Transformer) that converts those rich semantics into visual structure and detail. In the remainder of this section, we focus on the modeling shift itself: Rectified Flow and Flow Matching. Recently, they have emerged as compelling alternatives to diffusion objectives. We review the core formulation, and the key training and sampling choices that make these methods work well in practice. We also highlight the most impactful recent results that demonstrate their advantages for high quality and efficient image generation.

\subsection{Background}
\begin{figure}[h]
  \centering
  \includegraphics[width=\linewidth]{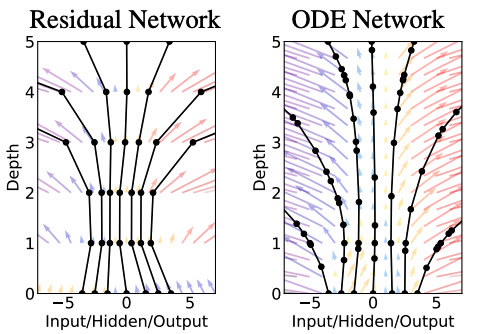}
  \caption{Similarity between residual networks and ODE-based networks. A residual network composes a sequence of discrete transformations of the hidden state, whereas an ODE network defines a continuous vector field whose flow transports the state over time. Image source:~\cite{chen2018neural}.}
  \label{fig:ode}
\end{figure}

Some classes of neural networks (such as residual networks, recurrent neural networks, and normalizing flows) apply a sequence of transformations to a state via the update
\begin{equation}
    x_{t+1} = x_t + f\!\left(x_t, \theta_t\right)
\end{equation}
where $t \in \{0,\dots, T-1\}$ and $x_t \in \mathbb{R}^D$. Chen et al.~\cite{chen2018neural} point out that this recursion resembles a forward Euler discretization of a continuous-time transformation (Fig.~\ref{fig:ode}). In particular, the continuous dynamics of the state can be modeled by an ordinary differential equation (ODE),
\begin{equation}
    \frac{d x(t)}{dt} = f\!\left(x(t), t\right).
\end{equation}
Starting from an initial condition $x(0)$ (e.g., the input layer), the network output at time $T$ is given by the ODE solution $x(T)$. In generative image modeling, we similarly start from noise \(x(0) \sim \mathcal{N}(0, I)\) and move forward by following a learned vector field until $t=1$, at which point $x(1)$ is the generated image. They call these models continuous normalizing flows (CNFs).

\subsection{Rectified Flow}
Building on the foundation of CNFs, Liu et al.~\cite{liu2022flow} introduced Rectified Flow, which uses an ODE vector field to transport samples from a source distribution $\pi_0$ to a target distribution $\pi_1$. They frame image generation, image translation, and domain transfer as learning a transport map from $\pi_0$ to $\pi_1$. In unconditional image generation, $\pi_0$ is typically standard Gaussian noise, while $\pi_1$ is the data distribution. For training, they draw independent pairs \((X_0, X_1) \sim \pi_0 \times \pi_1,\) and consider the straight line interpolation between endpoints,
\begin{equation}
    X_t = (1-t)\,X_0 + t\,X_1,
    \qquad t \in [0,1]
\end{equation}
The corresponding velocity along this path is $X_1 - X_0$. However, this velocity is not causal at time $t$ because it depends on the unknown point $X_1$. Rectified Flow therefore fits a causal drift field $v_\theta(X_t,t)$ via:
\begin{equation}
    \hat{\theta}
    =
    \arg\min_{\theta}\,
    \mathbb{E}\left[
        \left\|
            (X_1 - X_0) - v_\theta((1-t)X_0+tX_1, t)
        \right\|^2
    \right]
\end{equation}

\noindent where $t \sim \mathrm{Uniform}([0,1])$. During inference, one samples $Z_0 \sim \pi_0$ (e.g., Gaussian noise) and evolves it according to the ODE
\begin{equation}
    \frac{d Z_t}{dt} = v_\theta\!\bigl(Z_t, t\bigr)
\end{equation}
so that $Z_1$ is a generated sample. This ODE generally has no closed-form solution, so it is approximated by numerical integration (e.g., forward Euler), as summarized in Box~\ref{box:rectifiedflow-inference}.

\begin{boxes}[t]
\begin{tcolorbox}[title=Simplified Inference Algorithm for Rectified Flow, colback=gray!5, colframe=black!75,fonttitle=\bfseries\footnotesize,]
\begin{enumerate}[leftmargin=*, itemsep=1em]
\item Sample an initial noise \(Z_0 \sim \pi_0.\)
\item Choose $N$ as the number of solver steps and set the step size $\Delta t = 1/N$.

\item For $i = 0, 1, \dots, N-1$:
\begin{itemize}[leftmargin=*, itemsep=0.4em]
\item Current time $t_i = i/N$.
\item Evaluate the velocity using the neural network:
\[
u_i = v_\theta(Z, t_i).
\]
\item Update the state with a forward Euler step:
\[
Z = Z + \Delta t \, u_i.
\]
\end{itemize}

\item Return $Z \approx Z_1$ as the final generated sample.
\end{enumerate}
\end{tcolorbox}
\caption{Forward Euler inference procedure for Rectified Flow: start from $Z_0 \sim \pi_0$ and integrate $\frac{dZ_t}{dt}=v_\theta(Z_t,t)$ to obtain $Z_1$.}
\label{box:rectifiedflow-inference}
\end{boxes}

\begin{figure}[t]
    \centering
    \includegraphics[width=\linewidth]{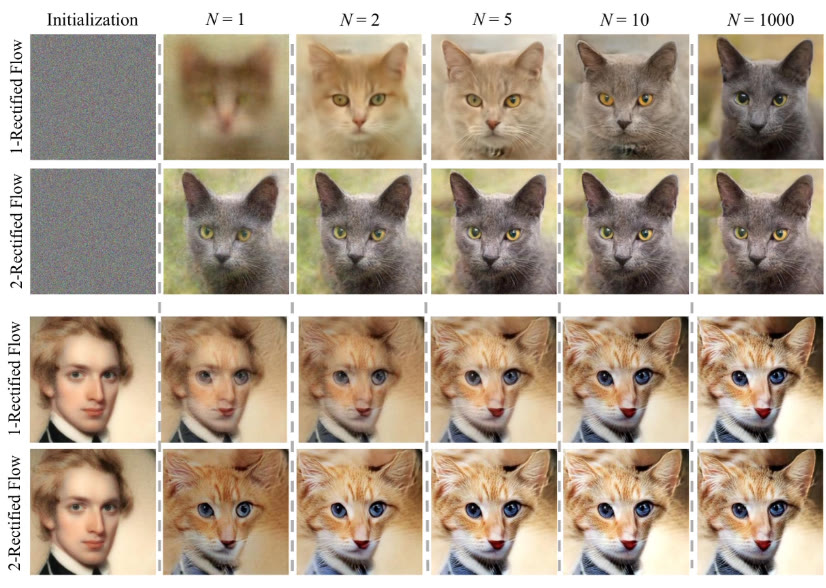}
    \caption{Rectified Flow and reflow trajectories. The top two rows illustrate noise to image transport under Rectified Flow. The second row shows reflow that produces straighter paths that can be traversed with fewer numerical integration steps. The bottom rows show an analogous effect for image to image transport, where reflow similarly reduces curvature and improves sampling efficiency. Image source: ~\cite{liu2022flow}}
    \label{fig:rectifiedflow-reflow}
\end{figure}

A key motivation for Rectified Flow is efficiency. If the transport trajectory from noise $Z_0$ to an image $Z_1$ is highly curved, accurate generation may require many small steps. If the trajectory is closer to a straight path, the sampler can take larger steps and reach $Z_1$ in fewer iterations. To encourage straighter trajectories, one can train an initial Rectified Flow model and use the trained model to generate paired samples $(Z_0, Z_1)$ and then retrain on these pairs. These reflow steps tend to straighten the trajectories and enable sampling with fewer solver steps. Fig.~\ref{fig:rectifiedflow-reflow} shows different rectified flow trajectories for image generation. The upper two rows show the mapping from noise to image, with the second row showing the impact of reflow. With reflow, fewer steps are needed to generate the final image. The lower two rows show a similar behavior but with image to image transport.

\begin{figure}[t]
  \centering
  \includegraphics[width=\linewidth]{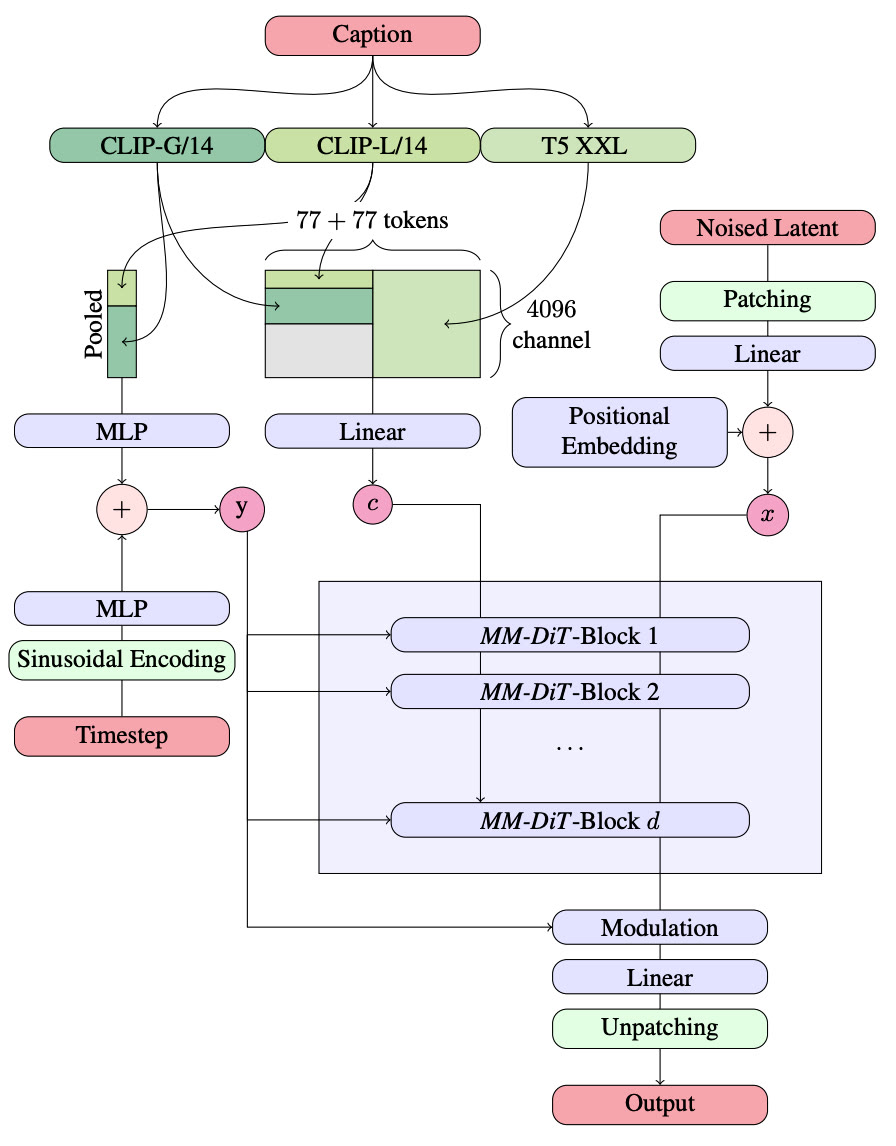}
  \caption{Model architecture for Scaling Rectified Flows that are based on Multimodal Diffusion Transformers (MM-DiT). Image source:~\cite{esser2024scaling}}
  \label{fig:rf-dit}
\end{figure}

\begin{figure}[h]
  \centering
  \includegraphics[width=\linewidth]{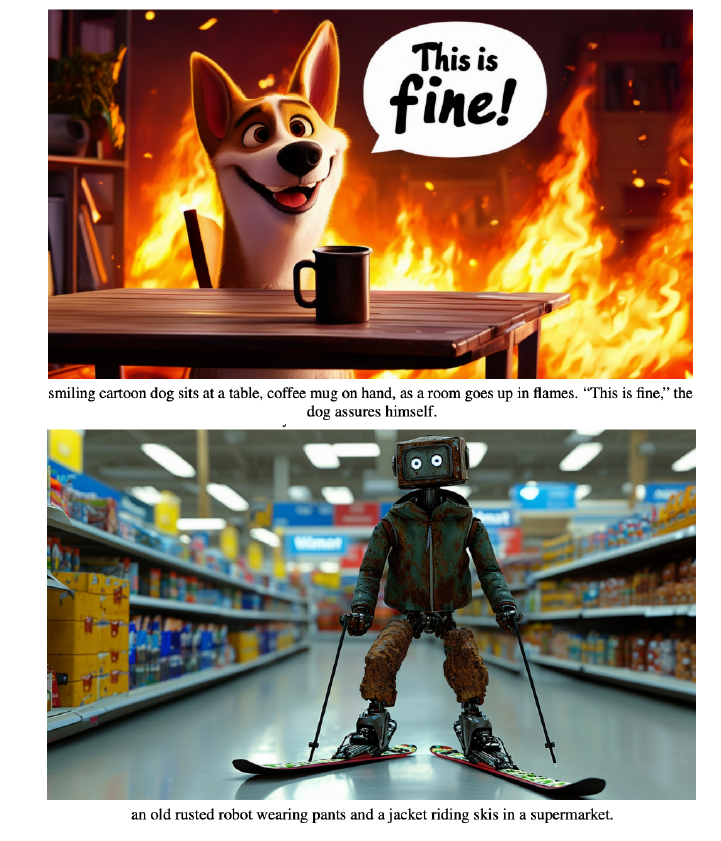}
  \caption{Sample images generated by Scaling Rectified Flow Transformers, along with their corresponding prompts. Image source:~\cite{esser2024scaling}}
  \label{fig:scale-rf-result}
\end{figure}

Later, Esser et al.~\cite{esser2024scaling} scaled Rectified Flow for high-resolution image generation and were able to achieve superior performance compared to established diffusion models. While their training objective follows Rectified Flow, for the text to image architecture they use a multimodal diffusion backbone, DiT~\cite{peebles2023scalable}. Fig.~\ref{fig:rf-dit} shows the overall architecture of their model. With the Rectified Flow training setup, they are able to achieve competitive results and outperform the best open-source models and DALL-E~3 on GenEval~\cite{ghosh2023geneval}. Fig.~\ref{fig:scale-rf-result} shows some of the sample images generated by their model with the corresponding prompts.

Lee et al.~\cite{lee2024improving} try to improve the training of Rectified Flows while keeping the number of function evaluations (NFEs) low. They propose techniques to improve 1-round training of Rectified Flows, hence making it unnecessary to use multiple iterations of reflow. With these techniques, they are able to improve the FID of the previous 2-rectified flow by up to $75\%$ in the single evaluation setting on CIFAR-10.

As discussed earlier in this section, RFs assume specific boundary conditions at $t=0$ and $t=1$. However, in practice, the neural network velocity field $v_\theta(x,t)$ often violates these boundary conditions. Hu et al.~\cite{hu2025improving} propose a boundary-enforced Rectified Flow parameterization so that the boundary constraints are satisfied for any neural-network weights. The velocity is reparameterized such that the boundary conditions are met:
\begin{equation}
v_\theta(x,t) \;=\; g(t)\,(C - x) \;+\; f(t)\,x \;+\; h(t)\,m(x,t)
\end{equation}
where $C=\mathbb{E}[X_1]$, and the scalar functions $f,g,h$ are chosen such that \(v(x,0)=C-x\) and \(v(x,1)=x\).

\subsection{Flow Matching}

Similar to Rectified Flow, Lipman et al.~\cite{lipman2022flow} use continuous normalizing flows (CNFs) as a foundation and introduce Flow Matching (FM). Both FM and Rectified Flow learn an ODE velocity field by regressing onto a target direction along a path. The difference, however, lies in what path they assume.

In FM, the authors try to iteratively improve the coupling between $z_0$ and $z_1$ and find a straight trajectory between points. They choose a conditional path family (diffusion or OT) and learn a vector field that realizes that marginal path. Instead of using an ODE solver during training, they train a neural-network vector field $v_\theta(x,t)$ for a designated probability path. This approach supports the standard diffusion path as a special case, and also supports non-diffusion paths such as optimal transport (OT), which they find to be simpler and faster.

Here we briefly summarize their approach. If we have a target probability path for the data distribution $p_t(x)$ and a vector field $u_t(x)$ that generates it, then FM trains a neural network $v_\theta(x,t)$ via regression:
\begin{equation}
\mathcal{L}_{\mathrm{FM}}(\theta)
=
\mathbb{E}_{t\sim \mathrm{U}[0,1],\, x\sim p_t}
\left[
\left\| v_\theta(x,t) - u_t(x) \right\|^2
\right]
\end{equation}

In practice, however, we do not have a closed form for $u_t$ for a useful marginal path $p_t$. Instead, they define the marginal path as a mixture over data. Given a real data point $x_1 \sim q(x_1)$,
\begin{equation}
p_t(x)
=
\int p_t(x \mid x_1)\, q(x_1)\, dx_1
\end{equation}
and they define the (marginal) vector field by averaging the conditional vector fields:
\begin{equation}
u_t(x)
=
\int
u_t(x \mid x_1)\,
\frac{p_t(x \mid x_1)\, q(x_1)}{p_t(x)}
\, dx_1
\end{equation}
In their implementation, they choose Gaussian conditional paths for $p_t(x\mid x_1)$:
\begin{equation}
p_t(x \mid x_1)
=
\mathcal{N}\!\bigl(x \mid \mu_t(x_1),\, \sigma_t(x_1)^2 I\bigr)
\end{equation}
with boundary conditions $\mu_0=0$, $\sigma_0=1$ at $t=0$ and $\mu_1=x_1$, $\sigma_1=\sigma_{\min}$ at $t=1$. They define an affine conditional map $\psi_t(\cdot)$ so that it pushes a standard Gaussian to a conditional Gaussian $p_t(x\mid x_1)$

\begin{equation}
\psi_t(z; x_1) = \sigma_t(x_1)\, z + \mu_t(x_1).
\end{equation}
This lets them reparameterize the loss as
\begin{equation}
\mathcal{L}_{\mathrm{FM}}(\theta)
=
\mathbb{E}_{t,\,x_1,\,z_0\sim \mathcal{N}(0,I)}
\left\|
v_\theta\!\bigl(\psi_t(z_0; x_1), t\bigr)
-
\frac{d}{dt}\psi_t(z_0; x_1)
\right\|^2
\end{equation}
where $\frac{d}{dt}\psi_t(z_0; x_1)$ is computed analytically from the definition of $\psi_t$. The authors show that, depending on the choice of the Gaussian path (i.e., $\mu_t,\sigma_t$), the resulting probability path can correspond to either a diffusion path or an optimal transport path. The simplified training and inference algorithm for FM is shown in Box.~\ref{box:flowmatching}.

They compare results using the same model architecture but trained with different objectives: one trained with optimal transport via Flow Matching, and another trained with a diffusion objective (DDPM~\cite{ho2020denoising}). As shown in Table~\ref{tab:fm-results}, Flow Matching beats the diffusion based model in likelihood (BPD), quality of generated samples (FID), and evaluation time (NFE).

Later, Kornilov et al.~\cite{kornilov2024optimal} came up with Optimal Flow Matching, which allows recovering the Optimal Transport map in a single Flow Matching step. They only consider vector fields restricted by some conditions, and perform a single Flow Matching optimization within that restricted condition.

\subsection{Conclusion}

Rectified Flows (RF) and Flow Matching (FM) allow researchers to go beyond diffusion based models, where an ordinary differential equation (ODE) vector field transports a simple noise distribution to the data distribution. RF tries to train a straight-line interpolation and improves sampling efficiency through reflow. On the other hand, FM provides a general framework that matches a target probability path using conditional objectives. Moreover, looking at these modern architectures, one can realize that they often consist of a stack of text conditioning, transformer backbones, and complex sampling techniques.

\begin{boxes}[t]
\begin{tcolorbox}[title=Simplified Training and Inference for Flow Matching,colback=gray!5,colframe=black!75,fonttitle=\bfseries\footnotesize]
\textbf{Training}
\begin{itemize}[leftmargin=*, itemsep=0.6em]
\item Sample a data point $x_1 \sim q(x_1)$.
\item Sample base noise $x_0 \sim \mathcal{N}(0,I)$.
\item Sample time $t \sim \mathrm{Uniform}([0,1])$.
\item Construct a point on the conditional path:
\[
x_t = \psi_t(x_0; x_1)
= \bigl(1-(1-\sigma_{\min})t\bigr)\,x_0 + t\,x_1
\]
\item Compute the target conditional velocity:
\[
u_t = x_1 - (1-\sigma_{\min})\,x_0
\]
\item Compute the MSE loss and update parameters:
\[
\mathcal{L}(\theta) = \left\| v_\theta(x_t,t) - u_t \right\|^2
\]
\item Update $\theta$ with Adam or SGD.
\end{itemize}
\vspace{1.5em}
\textbf{Inference (discretized Euler sampler).}
\begin{itemize}[leftmargin=*, itemsep=0.6em]
\item Sample $x \sim \mathcal{N}(0,I)$
\item For $i = 0, 1, \dots, N-1$:
\begin{itemize}[leftmargin=*, itemsep=0.6em, label=\textbullet]
\item $t = i/N$
\item \(x \leftarrow x + \frac{1}{N}\, v_\theta(x,t)\)
\end{itemize}

\item Return $x$ as the generated image.
\end{itemize}

\end{tcolorbox}
\caption{A simplified overview of Flow Matching training (via regression to a target velocity along a chosen conditional path) and inference (via an Euler sampler).}
\label{box:flowmatching}
\end{boxes}

\begin{table}[t]
\centering
\small
\setlength{\tabcolsep}{4pt}
\renewcommand{\arraystretch}{1.15}
\begin{tabular}{lccc|ccc}
\toprule
& \multicolumn{3}{c|}{\textbf{CIFAR-10}} & \multicolumn{3}{c}{\textbf{ImageNet $32\times 32$}} \\
\textbf{Model} & BPD$\downarrow$ & FID$\downarrow$ & NFE$\downarrow$
              & BPD$\downarrow$ & FID$\downarrow$ & NFE$\downarrow$ \\
\midrule
\textit{Ablations} \\
DDPM           & 3.12 & 7.48  & 274 & 3.54 & 6.99  & 262 \\
Score Matching & 3.16 & 19.94 & 242 & 3.56 & 5.68  & 178 \\
ScoreFlow      & 3.09 & 20.78 & 428 & 3.55 & 14.14 & 195 \\
\midrule
\textit{Ours} \\
FM w/ Diffusion & 3.10 & 8.06 & 183 & 3.54 & 6.37 & 193 \\
FM w/ OT        & \textbf{2.99} & \textbf{6.35} & \textbf{142}
                & \textbf{3.53} & \textbf{5.02} & \textbf{122} \\
\bottomrule
\end{tabular}
\caption{Flow Matching results compared to diffusion baselines. FM with Optimal Transport achieves better likelihood (BPD), sample quality (lower FID), and fewer function evaluations (lower NFE) than diffusion based DDPM.}
\label{tab:fm-results}
\end{table}

\section{Video Generation}
Video generation can be formulated as image generation expanded over time while keeping temporal consistency across frames. As image generation models continued to improve, researchers started to transfer ideas to the video generation domain.

\subsection{Video Generation with GANs}
Vondrick et~al.~\cite{vondrick2016generating} start from a vanilla GAN and utilize it for 32 frames video generation, and call it VideoGAN (VGAN). Similar to GANs, they start with a min–max game where the generator tries to fool the discriminator and the discriminator tries to distinguish between fake and real videos. In order to model the stationary world (background) and the moving objects (foreground) around it, they come up with a two-stream architecture:

 % They start from the latent code \(z\) is the latent code sampled from a normal distribution and \(x \sim p(x)\) are samples from the data distribution. 
% \begin{equation}
% \min_G \max_D \; \mathbb{E}_{x \sim p_{\text{real}}} \left[ \log D(x; w_D) \right]
% + \mathbb{E}_{z \sim p(z)} \left[ \log(1 - D(G(z; w_G); w_D)) \right]
% \end{equation}

\begin{equation}
G(z) = m(z) \odot f(z) + \left(1 - m(z)\right) \odot b(z)
\end{equation}
where \(f(z)\) and \(b(z)\) are the foreground and background videos with shape \((T, H, W, 3)\). \(b(z)\) is a static image replicated across time. \(m(z)\) is the mask tensor with values in \((0, 1)\) (after a sigmoid activation), with shape \((T, H, W, 1)\). \(T\) is the number of video frames, chosen as \(32\) in their work. They also add a sparsity prior on the mask \( \lambda \| m(z) \|_1 \) with \(\lambda = 0.1\) to encourage the mask to be mostly 0 (background) and only turn it on when necessary.. The intuition behind this is to encourage \(m\) to be mostly 0 (background). The generator loss becomes:

\begin{equation}
L_G = \mathbb{E}_z \left[ -\log D(G(z)) \right] + \lambda \, \mathbb{E}_z \| m(z) \|_1
\end{equation}

\begin{figure}[h]
  \centering
  \includegraphics[width=\linewidth]{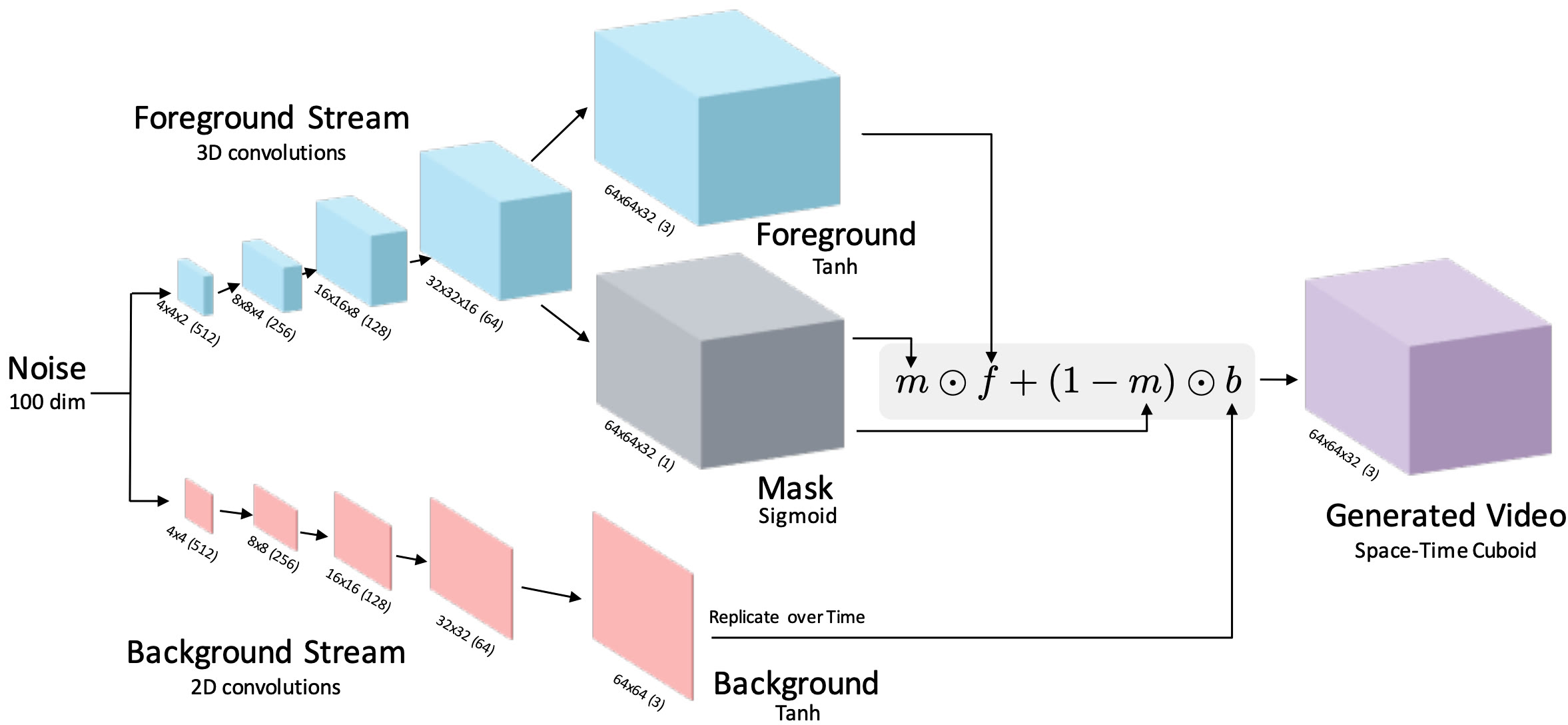}
  \caption{The overall pipeline for VideoGAN. The two foreground and background streams are combined to generate the output video. Image source:~\cite{vondrick2016generating}}
  \label{fig:vgan_pipeline}
\end{figure}

\begin{figure}[h]
  \centering
  \includegraphics[width=\linewidth]{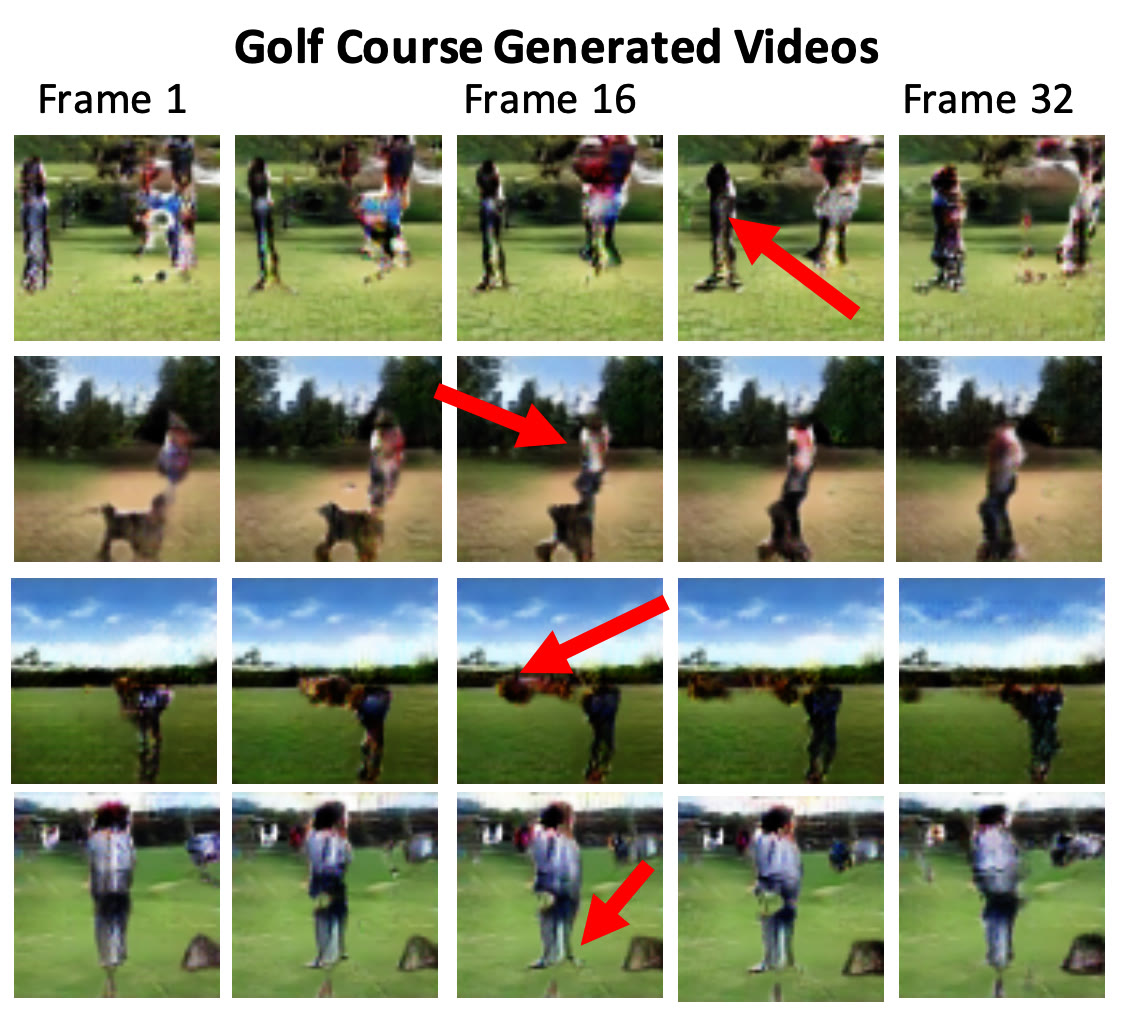}
  \caption{Sample generated video frames from VGAN showing people walking on a golf course. Image source:~\cite{vondrick2016generating}}
  \label{fig:vgan-result}
\end{figure}

Fig.~\ref{fig:vgan_pipeline} shows the overall pipeline of VGAN and Fig.~\ref{fig:vgan-result} shows some of the sample frames generated by their model. The model captures coarse motion patterns. For example, the golf video frames in Fig.~\ref{fig:vgan-result} show people walking on grass.

\begin{figure}[h]
  \centering
  \includegraphics[width=\linewidth]{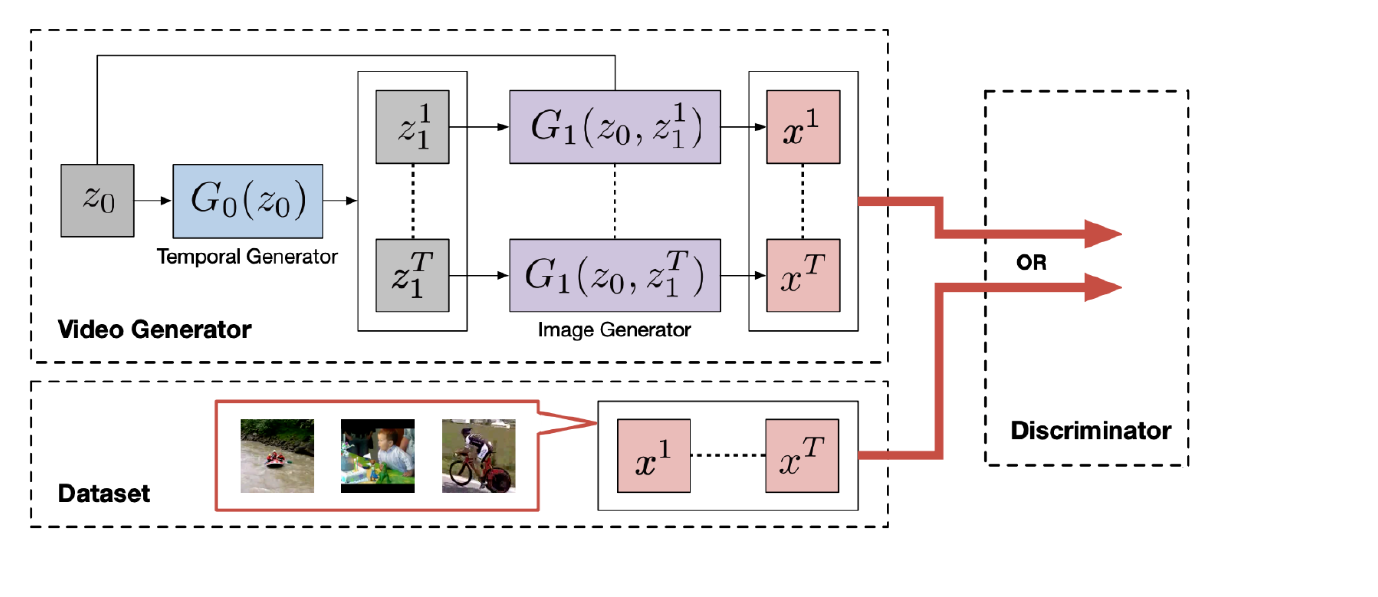}
  \caption{Temporal video generation (TGAN) model pipeline. Image source:~\cite{saito2017temporal}}
  \label{fig:tgan-pipeline}
\end{figure}

Saito et al.~\cite{saito2017temporal} introduced Temporal GAN (TGAN) video models. As shown in Fig.~\ref{fig:tgan-pipeline}, they introduced a temporal generator \(G_0\) that maps a single latent \(z_0\) into a sequence of per-frame latents \([z_1^{1}, \ldots, z_1^{T}]\). Subsequently, generator 
\(G_1\) maps \((z_0, z_1^{t})\) into an image frame \(x^{t}\):

\begin{equation}
z_0 \xrightarrow{G_0} (z_1^{1}, \ldots, z_1^{T}), \qquad x^{t} = G_1(z_0, z_1^{t}).
\end{equation}

\begin{figure}[h]
  \centering
  \includegraphics[width=\linewidth]{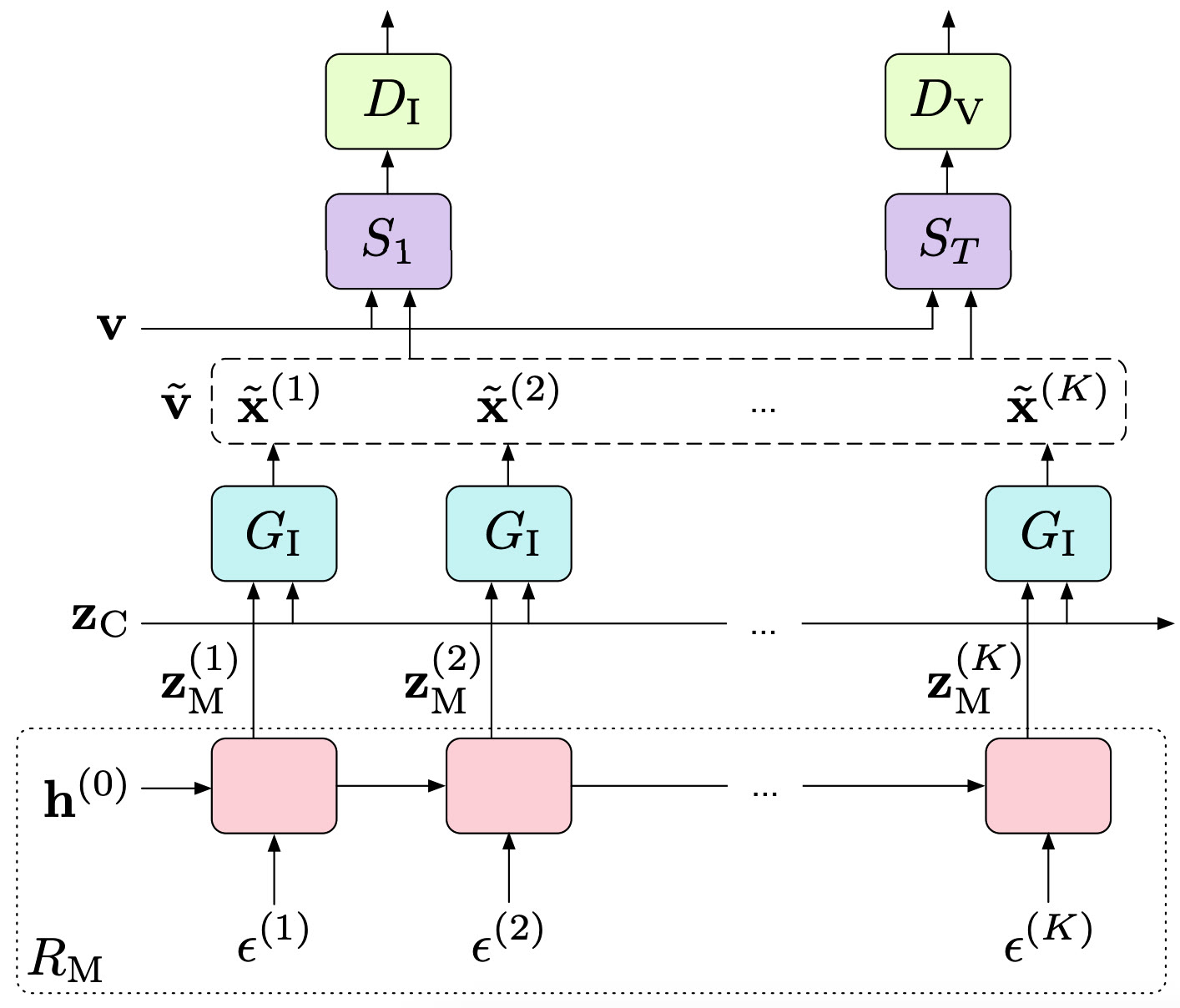}
  \caption{The overall model pipeline for MoCoGAN. Image source:~\cite{tulyakov2018mocogan}}
  \label{fig:mocogan-pipeline}
\end{figure}

Later, Tulyakov et~al.~\cite{tulyakov2018mocogan}  introduce MoCoGAN by making the assumption that video generation consists of two parts: motion and content. They treat a video as a trajectory, and formulate the latent space \(Z_I\) in GAN as \([z_I^{(1)}, \ldots, z_I^{(K)}]\) for \(K\) time steps. Each time step is then decomposed into \(z_C \in Z_C\) and \(z_M^{(k)} \in Z_M\) for content and motion subspaces. \(z_C\) is fixed across the video, and \(z_M^{(k)}\) changes over time as depicted in Fig.~\ref{fig:mocogan-pipeline}. The motion codes \(z_M^{(k)}\) are the output of a recurrent neural network (RNN) at each time step. This helps with correlation of motion at each time step. The generator network \(G_I\) is a standard DCGAN style deconvolutional network. For each time step the output of the generator is

\begin{equation}
\hat{x}^{(k)}_I = G_I\left(z_I^{(k)}\right), \qquad 
\hat{x}^{(k)}_I \in \mathbb{R}^{64 \times 64 \times 3}
\end{equation}

\noindent The generated video can be formed by a stack of these frames:

\begin{equation}
\hat{v} = \{ \hat{x}^{(1)}_I, \ldots, \hat{x}^{(K)}_I \} 
\in \mathbb{R}^{K \times 64 \times 64 \times 3}.
\end{equation}

They incorporate two discriminators in their architecture, \(D_I\) and \(D_V\). \(D_I\) sees individual frames and tries to enforce per-frame realism. \(D_V\) sees \(T\) consecutive frames and enforces temporal coherence. Both of these models operate on real and generated videos with random sampling. If \(p_r\) and \(\tilde{p}_r\) are the distributions of real and generated videos respectively,  
then the adversarial objective can be written as

\begin{equation}
\max_{G_I, R_M} \min_{D_I, D_V} \; F_V(D_I, D_V, G_I, R_M),
\end{equation}

where

\begin{equation}
\begin{aligned}
F_V &= \mathbb{E}_{v \sim p_v} \left[ -\log D_I(S_I(v)) -\log D_V(S_T(v)) \right] \\
&\quad
+ \mathbb{E}_{\tilde{v} \sim \tilde{p}_v} 
\left[ -\log \left( 1 - D_I(S_I(\tilde{v})) \right) -\log \left( 1 - D_V(S_T(\tilde{v})) \right) \right]
\end{aligned}
\end{equation}

% \begin{equation}
% \begin{aligned}
% F_V &= \mathbb{E}_{v \sim p_v} \left[ -\log D_I(S_1(v)) \right]
% + \mathbb{E}_{\tilde{v} \sim \tilde{p}_v} 
% \left[ -\log \left( 1 - D_I(S_1(\tilde{v})) \right) \right] \\
% &\quad + \mathbb{E}_{v \sim p_v} \left[ -\log D_V(S_T(v)) \right]
% + \mathbb{E}_{\tilde{v} \sim \tilde{p}_v} 
% \left[ -\log \left( 1 - D_V(S_T(\tilde{v})) \right) \right].
% \end{aligned}
% \end{equation}
\noindent The \(D_I\) terms are the losses for the image discriminator, and the \(D_V\) terms are the losses for the video discriminator. \(R_M\) denotes the RNN parameters responsible for generating the motion codes \(z_M\). \(S_1\) and \(S_T\) are random samplers for selecting 1 and \(T\) consecutive samples, respectively. Using this approach, they produce video outputs that score a much higher preference by the user, as shown in Table~\ref{tab:table-mocogan}.

\begin{table}[t]
\centering
\begin{tabular}{lcc}
\toprule
\textbf{User preference, \%} & \textbf{Facial Exp.} & \textbf{Tai-Chi} \\
\midrule
MoCoGAN / VGAN & \textbf{84.2} / 15.8 & \textbf{75.4} / 24.6 \\
MoCoGAN / TGAN & \textbf{54.7} / 45.3 & \textbf{68.0} / 32.0 \\
\bottomrule
\end{tabular}
\caption{User preference comparison between the generated videos by MoCoGAN, VGAN, and TGAN~\cite{saito2017temporal} on Facial Expression~\cite{aifanti2010mug} and Tai-Chi~\cite{cao2017realtime} datasets. Table source:~\cite{tulyakov2018mocogan}.}
\label{tab:table-mocogan}
\end{table}

\subsection{Transformer Based Video Generation}
\begin{figure}[h]
  \centering
  \includegraphics[width=\linewidth]{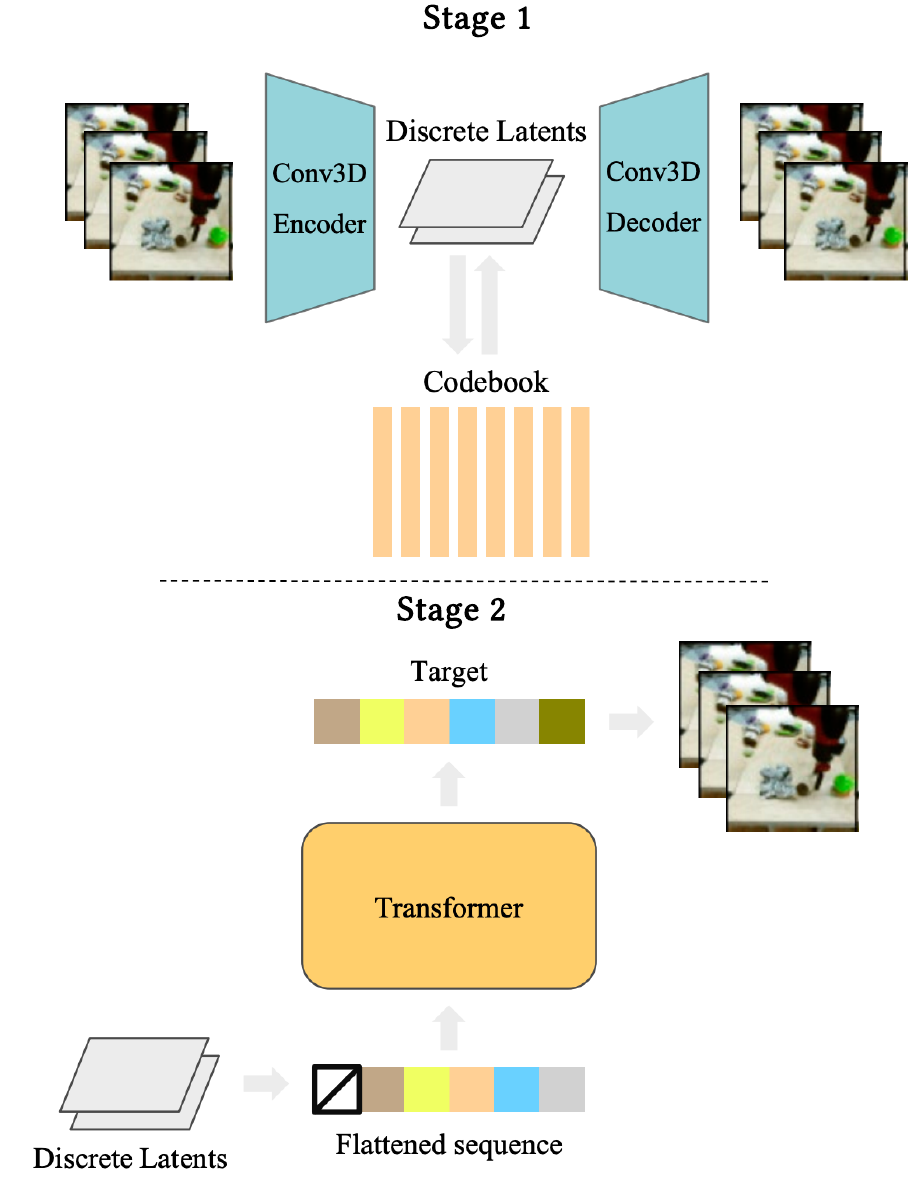}
  \caption{Two stage training pipeline for VideoGPT. In stage 1, the VQ-VAE is trained and in stage 2 a GPT-style transformer is trained over latent codebooks. Image source:~\cite{yan2021videogpt}}
  \label{fig:videogpt-pipeline}
\end{figure}

Yan et al.~\cite{yan2021videogpt} scale a VQ-VAE and GPT for generating videos. As depicted in Fig.~\ref{fig:videogpt-pipeline}, the training consists of two stages. In stage 1, a standard VQ-VAE is trained for learning discrete video latents. Given a video clip \(x \in \mathbb{R}^{T \times H \times W \times C}\), where \(T, H, W, C\) are the number of frames, height, width, and channel, the encoder produces a lower-resolution latent volume via 3D convolutional and attention blocks:

\begin{equation}
h = E(x) \in \mathbb{R}^{T' \times H' \times W' \times d}
\end{equation}
Each spatiotemporal partition is then quantized to the nearest code in a codebook of \(K\) embeddings (e.g., \(K = 1024\)):

\begin{equation}
e_{t',h',w'} = \text{nearest\_code}(h_{t',h',w'})
\end{equation}
The decoder then reconstructs the original video via 3D transposed convolution and attention blocks:

\begin{equation}
\hat{x} = D(e) \in \mathbb{R}^{T \times H \times W \times C}
\end{equation}
The model is trained via standard reconstruction and commitment loss for VQ-VAE.

In stage 2, an autoregressive transformer is trained over the flattened discrete latents \(e_{t',h',w'}\). The discrete latents are flattened into a one-dimensional sequence of indices

\begin{equation}
s = (k_1, k_2, \ldots, k_L), \qquad L = T' \times H' \times W'
\end{equation}
where \(k\) is the codebook index. The GPT-style transformer is trained via

\begin{equation}
p(s) = \prod_{i=1}^{L} p(k_i \mid k_{<i}).
\end{equation}

Using this approach, they achieve an inception score (IS) of 24.69. Table~\ref{tab:table-video} shows a comparison between their inception score and other state-of-the-art models on the UCF-101~\cite{soomro2012ucf101} dataset. UCF-101 is an action classification dataset with 13{,}320 videos across 101 classes.

\subsection{Diffusion Based Video Generation}
Ho et al.~\cite{ho2022video} introduced video diffusion models, where they train a diffusion model that directly models blocks of \(T\) frames as a single object. The objective is similar to how a diffusion model is trained for image generation. Given an input video \(x \in \mathbb{R}^{T \times H \times W \times C}\), noise gets added using continuous Gaussian diffusion:

\begin{equation}
q(z_t \mid x) = \mathcal{N}(\alpha_t x,\; \sigma_t^2 I),
\end{equation}
where \(\alpha_t\) and \(\sigma_t\) are differentiable noise schedules. Their core innovation lies in the 3D UNet architecture, i.e., the denoiser \(\epsilon_\theta\). Each block in the 3D UNet can process 4D tensors with axes labeled as frames \(\times\) height \(\times\) width \(\times\) channels. They factorize the 3D processing in space and time. In space, they replace the 2D 3×3 convolutional layers in an image UNet with 1×3×3 convolutions, where the time dimensions are passed through unchanged.
They include two types of attention blocks: spatial and temporal. In spatial attention, they apply attention over spatial positions (height and width), and frames (time) are treated as batch elements. In temporal attention, for each spatial location \((h, w)\), the \(T\) frames are treated as consecutive tokens. Relative positional embeddings are used to encode frame order. Since temporal attention is separate, they can mask (disable) it so that some frames can be treated as independent image samples.

\begin{figure*}[t]
  \centering
  \includegraphics[width=\textwidth]{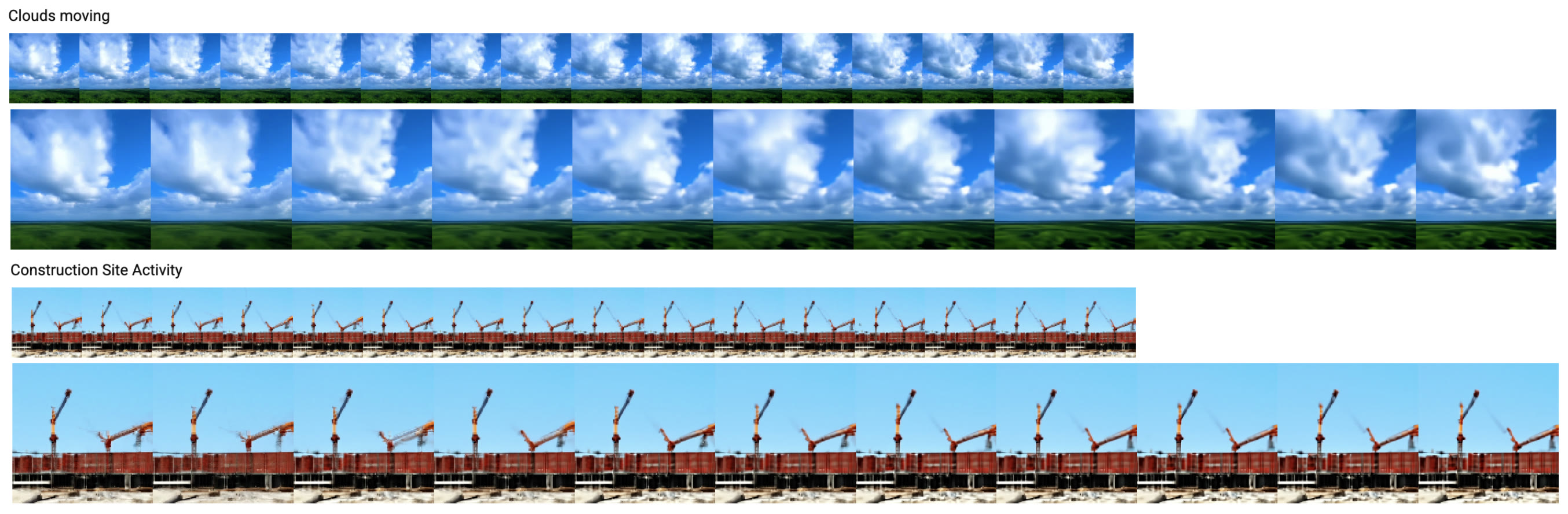}
  \caption{Text-conditioned video frames generated by the diffusion video model via a two stage diffusion model. Stage 1 generates the \(16 \times 64 \times 64\) lower resolution version and stage 2 uses a second model for simultaneous super-resolution and autoregressive extension to \(64 \times 128 \times 128\). Image source:~\cite{ho2022video}}
  \label{fig:diffusion-frames}
\end{figure*}

Using this approach, they are able to dramatically increase the inception score for unconditional video generation on the UCF-101 dataset, as shown in Table~\ref{tab:table-video}. Fig.~\ref{fig:diffusion-frames} shows text-conditioned video samples by cascading two models. First, samples are generated using a lower-resolution \(16 \times 64 \times 64\) model. Then the samples are passed to a second model for simultaneous super-resolution and autoregressive extension to \(64 \times 128 \times 128\).

\begin{table}[t]
\centering
\begin{tabular}{lccc}
\toprule
\textbf{Method} & \textbf{Resolution} & \textbf{FID} (\(\downarrow\)) & \textbf{IS} (\(\uparrow\)) \\
\midrule
MoCoGAN~\cite{tulyakov2018mocogan}      & 16x64x64     & 26998 $\pm$ 33      & 12.42 \\
TGAN-F~\cite{kahembwe2020lower}       & 16x64x64     & 8942.63 $\pm$ 3.72  & 13.62 \\
TGAN-ODE~\cite{gordon2021latent}     & 16x64x64     & 26512 $\pm$ 27      & 15.2 \\
TGAN-F~\cite{kahembwe2020lower}       & 16x128x128   & 7817 $\pm$ 10       & 22.91 $\pm$ .19 \\
VideoGPT~\cite{yan2021videogpt}     & 16x128x128   & ---                 & 24.69 $\pm$ 0.30 \\
TGAN-v2~\cite{saito2020train}      & 16x64x64     & 3431 $\pm$ 19       & 26.60 $\pm$ 0.47 \\
TGAN-v2~\cite{saito2020train}      & 16x128x128   & 3497 $\pm$ 26       & 28.87 $\pm$ 0.47 \\
DVD-GAN~\cite{clark2019adversarial}      & 16x128x128   & ---                 & 32.97 $\pm$ 1.7 \\
\midrule
\textbf{Video Diffusion} & 16x64x64 & \textbf{295 $\pm$ 3} & \textbf{57 $\pm$ 0.62} \\
\midrule
real data              & 16x64x64     & ---                  & 60.2 \\
\bottomrule
\end{tabular}
\caption{Comparing the inception score (IS) and Fréchet Inception Distance (FID) of different video generation methods on UCF-101 dataset. Table source:~\cite{ho2022video}.}
\label{tab:table-video}
\end{table}

\subsection{Scaling up Video Diffusion Models}
Blattmann et al.~\cite{blattmann2023stable} introduced one of the first widely used open-source latent video diffusion models and called it Stable Video Diffusion (SVD). Most works prior to SVD only trained the video generation models on a small video dataset. Moreover, prior works mostly focused on temporal architectural changes rather than data and training strategy. The training pipeline consists of three stages. In stage I, they perform image pretraining with text–image pairs. Their experiments show that humans strongly prefer image pretrained models. In stage II, they pretrain the base video model. They preprocess a video clip and select \(T\) frames and resize each frame to \(H \times W = 320 \times 576\). They build conditioning by passing the text caption of the input video to a text encoder. The frame rate information is also embedded into a vector. The conditioning embeddings are then injected into the UNet block. The UNet block consists of both spatial (height, width) and temporal (frame) attention layers. They utilize the Elucidated Diffusion Model (EDM)~\cite{karras2022elucidating} loss for training the UNet backbone.

\begin{figure}[h]
  \centering
  \includegraphics[width=\linewidth]{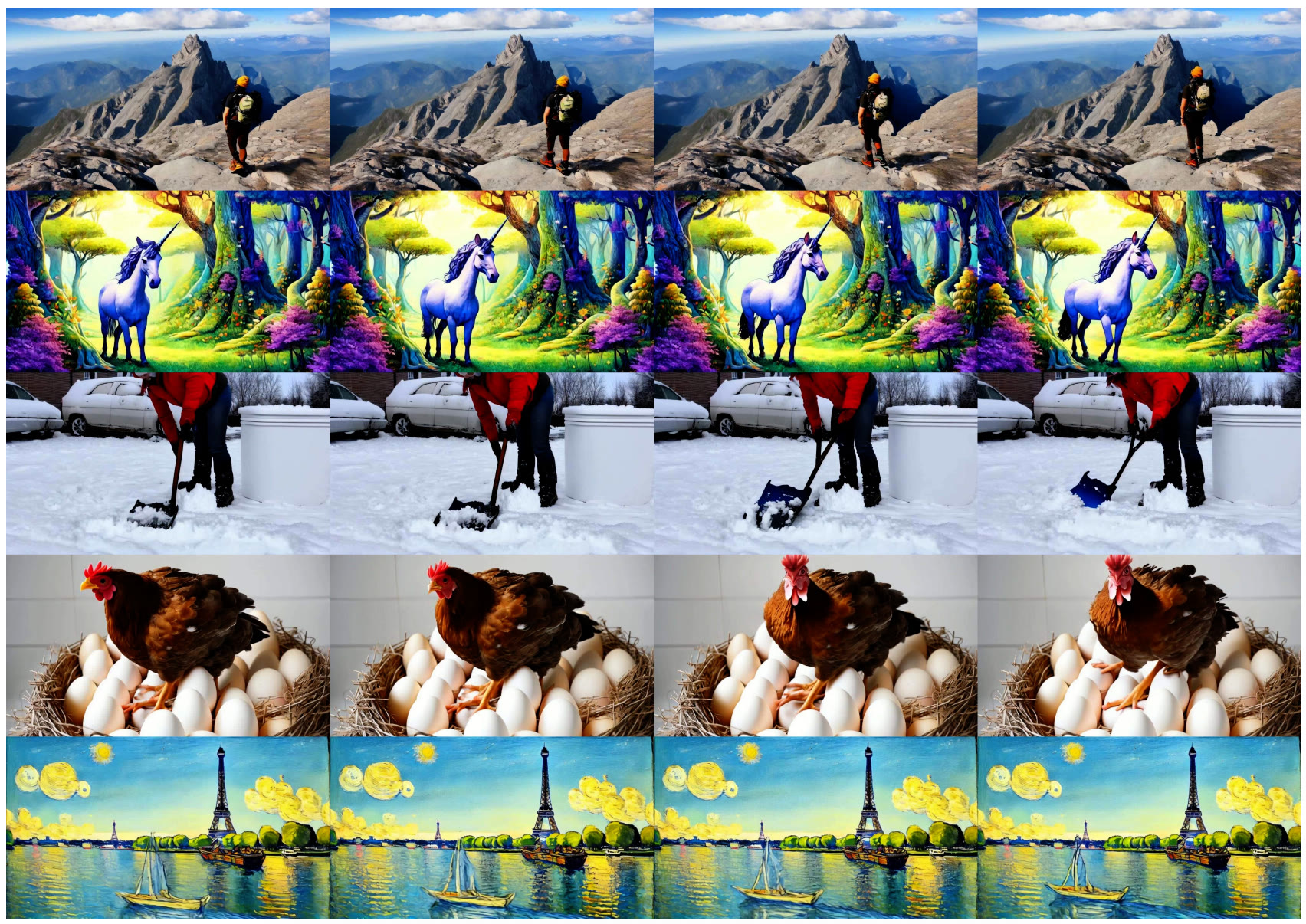}
  \caption{Text-to-video samples from the stable video diffusion (SVD) model. Captions from top to bottom are: “A hiker is reaching the summit of a mountain, taking in the breathtaking panoramic view of nature.”, “A unicorn in a magical grove, extremely detailed.”, “Shoveling snow”, “A beautiful fluffy domestic hen sitting on white eggs in a brown nest, eggs are under the hen.”, and “A boat sailing leisurely along the Seine River with the Eiffel Tower in background by Vincent van Gogh”. Image source:~\cite{blattmann2023stable}}
  \label{fig:stable-video-diffusion}
\end{figure}

Stage III has the same pipeline as stage II but they use 1 million high-quality text-to-video pairs that consist of lots of
object motion, clean steady camera paths, and good captions. The resolution for this stage is \(576 \times 1024\) compared
to \(320 \times 576\) at stage II. Box~\ref{box:svd-training} shows their overall training pipeline. The authors also train an image-to-video variant of the model. For those, they start from the same base UNet and replace the text embedding with CLIP image embeddings and a few more tweaks. Fig.~\ref{fig:stable-video-diffusion} shows some of the sample video frames generated by Stable Video Diffusion for different input text prompts.

\begin{boxes}
\begin{tcolorbox}[title=Simplified Training Algorithm for Stable Video Diffusion, colback=gray!5, colframe=black!75, fonttitle=\bfseries\footnotesize]

\begin{enumerate}[leftmargin=*, itemsep=1em]

\item Pretrain an \emph{image} diffusion model (stage I). A text-to-image latent diffusion model on image–caption pairs. This is a 2D UNet denoiser $\epsilon_\theta^{\mathrm{img}}$ conditioned on input text.

\item Inflate the 2D UNet model into a spatio–temporal model $u_\theta$.

\begin{itemize}[leftmargin=*, itemsep=0.4em]
\item Insert temporal conv/attention layers after each spatial block.
\item Adopt 3D positional encodings over $(\text{time}, \text{height}, \text{width})$.
\item Use the image model weights and random weights to initialize spatial and temporal
      layers respectively.
\end{itemize}

\item Train text-to-video diffusion models $u_\theta$ (stage II).

For each mini-batch of size $m$:

\begin{itemize}[leftmargin=*, itemsep=0.4em]

\item Sample and preprocess videos to a target frame rate,
      \[
      x^{(i)} \in \mathbb{R}^{T \times H \times W \times 3}
      \]

\item Encode frames to latents using an encoder VAE,
      \[
      z_0^{(i)} \in \mathbb{R}^{T \times H' \times W' \times 4}
      \]

\item Sample EDM noise levels and corrupt latents,
      \[
      \sigma^{(i)} \sim \mathcal{N}(P_{\text{mean}}, P_{\text{std}}^2), 
      \qquad
      \epsilon^{(i)} \sim \mathcal{N}(0, I),
      \]
      \[
      z^{(i)} = z_0^{(i)} + \sigma^{(i)} \epsilon^{(i)}
      \]

\item Build conditioning for each caption $y^{(i)}$,
      \[
      c_{\text{text}}^{(i)} = \mathrm{Embedding}_{\text{text}}\!\bigl(y^{(i)}\bigr)
      \]
      and embed frame-rate and aspect-ratio into $c_{\text{micro}}^{(i)}$.

\item Run the video UNet denoiser $u_\theta$,
      \[
      \hat{z}^{(i)} = u_\theta\bigl(z^{(i)}, c_{\text{text}}^{(i)}, c_{\text{micro}}^{(i)}, c_\sigma^{(i)}\bigr)
      \]
      where $c_\sigma^{(i)}$ is the noise-level embedding.

\item Update parameters $\theta$ via the per-sample loss with weight $w(\sigma^{(i)})$,
      \[
      L^{(i)} = w\bigl(\sigma^{(i)}\bigr) 
                \left\lVert \hat{z}^{(i)} - z_0^{(i)} \right\rVert_2^2,
      \]
      \[
      L_{\text{video}} = \frac{1}{m} \sum_{i=1}^{m} L^{(i)}
      \]
      Take a gradient step on $\theta$ using $\nabla_\theta L_{\text{video}}$.
\end{itemize}

\item High quality fine-tuning of text-to-video model (stage III) using the same UNet and procedure of stage II (step 3 above) but on a smaller, higher quality video set at $576\times1024$ resolution.

\end{enumerate}

\end{tcolorbox}
\caption{Simplified training algorithm for Stable Video Diffusion.}
\label{box:svd-training}
\end{boxes}

\begin{figure}[h]
  \centering
  \includegraphics[width=\linewidth]{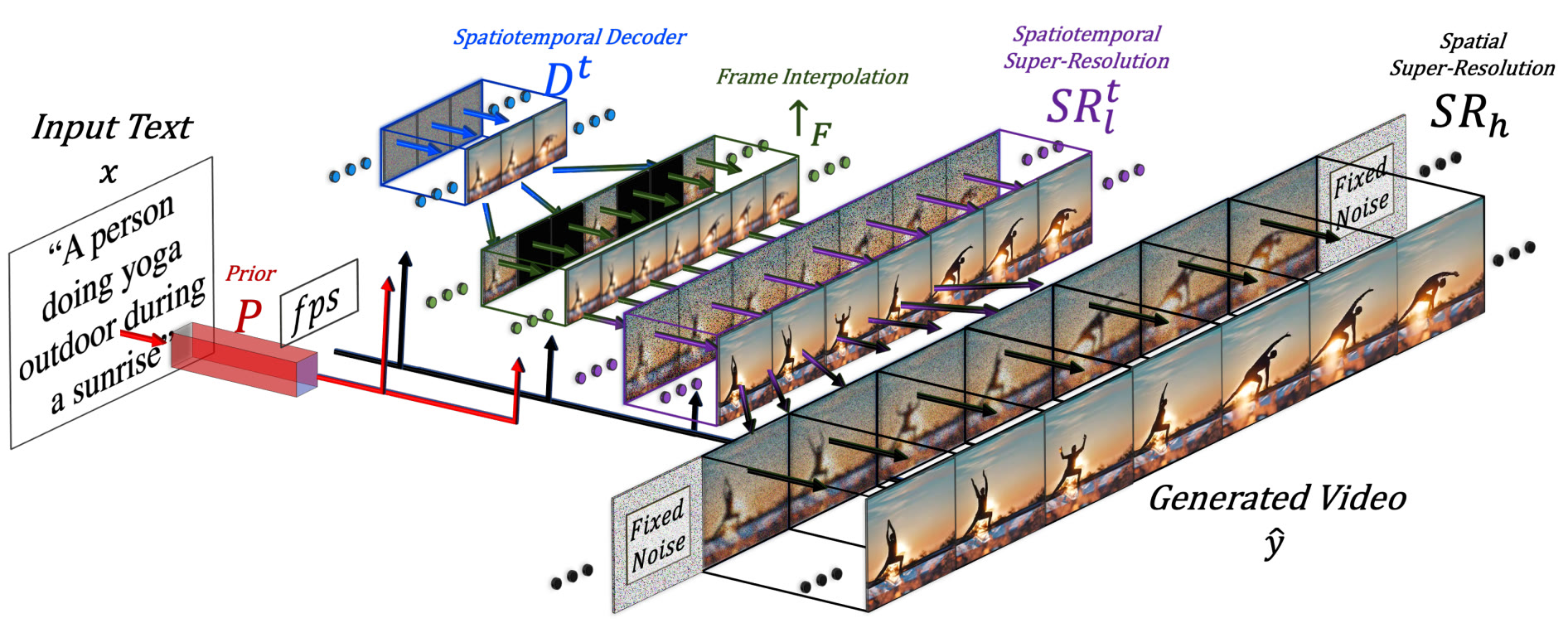}
  \caption{Simplified training pipeline for Make-A-Video. Input text gets mapped to image embedding and a desired fps by the prior \(P\). It then goes through decoder \(D_t\) that generates \(16 \times 64 \times 64\) RGB frames. Toward the end, it goes through spatial and temporal resolution blocks. Image source:~\cite{singer2022make}}
  \label{fig:architecture-make-a-video}
\end{figure}

\begin{figure}[h]
  \centering
  \includegraphics[width=\linewidth]{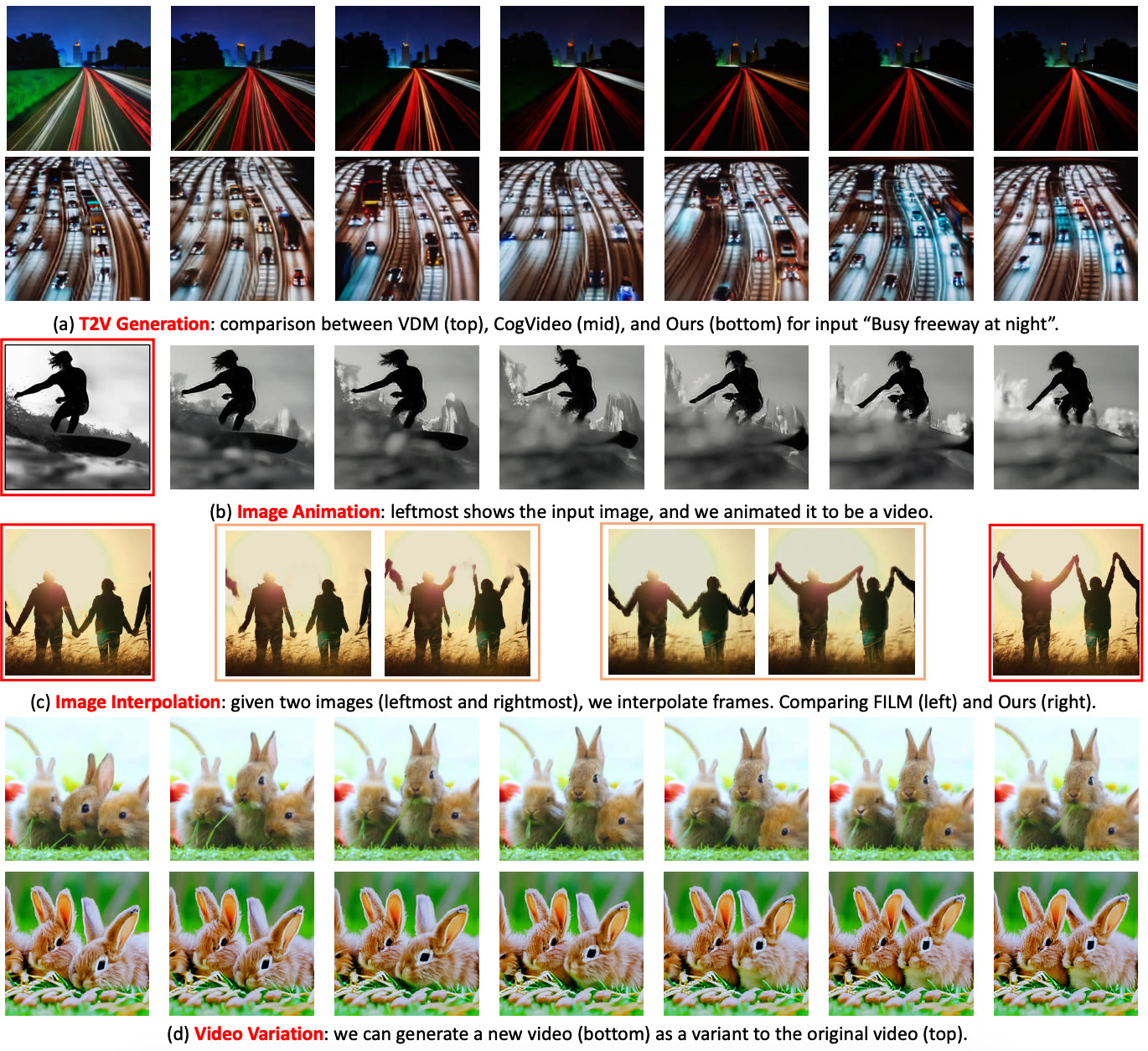}
  \caption{Sample video frames generated by Make-A-Video model for different applications such as generation, animation, and interpolation. Image source:~\cite{singer2022make}}
  \label{fig:results-make-a-video}
\end{figure}

Singer et al.~\cite{singer2022make} approached the video generation problem from a slightly different angle and introduced Make-A-Video. They learn text to appearance mapping from a huge text–image pairs, and don’t use any text–video pairs in their training. The text-to-video model is a text-to-image model extended in the time axis and further fine-tuned on unlabeled videos. Their model architecture consists of three major building blocks. In the first block, similar to DALL·E 2, they have a diffusion prior that maps text to CLIP image embeddings and a desired frame rate (fps). In the second block, they have a UNet style diffusion model that maps image embeddings to \(16 \times 64 \times 64 \times 3\) frames. The third block consists of a spatio–temporal network \(SR_l^{t}\), a spatial-only network \(SR_h\), and a frame interpolation network \(\uparrow F\), as depicted in Fig.~\ref{fig:architecture-make-a-video}. The final text-to-video inference depicted in Fig.~\ref{fig:architecture-make-a-video} can be formulated as:

\begin{equation}
\hat{y}_t = SR_h \circ SR_l^{t} \circ \uparrow F \circ D^t \circ P \circ (\hat{x},C_{txt}(x))
\end{equation}
where \(\hat{y}_t\) is the final generated video and $x$, \(\hat{x}\), and \(C_{\text{txt}}\) are the input text, BPE-encoded text, and CLIP text encoder, respectively. Fig.~\ref{fig:results-make-a-video} shows sample qualitative results for various applications such as generation, animation, and interpolation.

\begin{figure}[h]
  \centering
  \includegraphics[width=\linewidth]{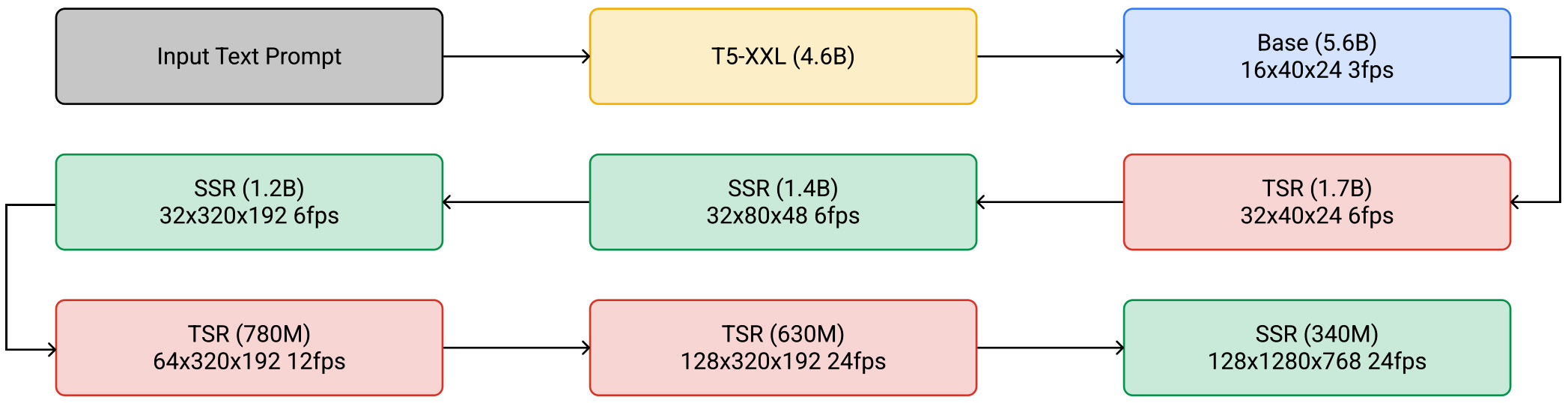}
  \caption{Overall simplified training pipeline for Imagen Video. In practice, the text encoder is injected into all the models. SSR and TSR are spatial and temporal super-resolution diffusion models. Image source:~\cite{ho2022imagen}}
  \label{fig:imagen-video-diffusion}
\end{figure}

\begin{figure}[h]
  \centering
  \includegraphics[width=\linewidth]{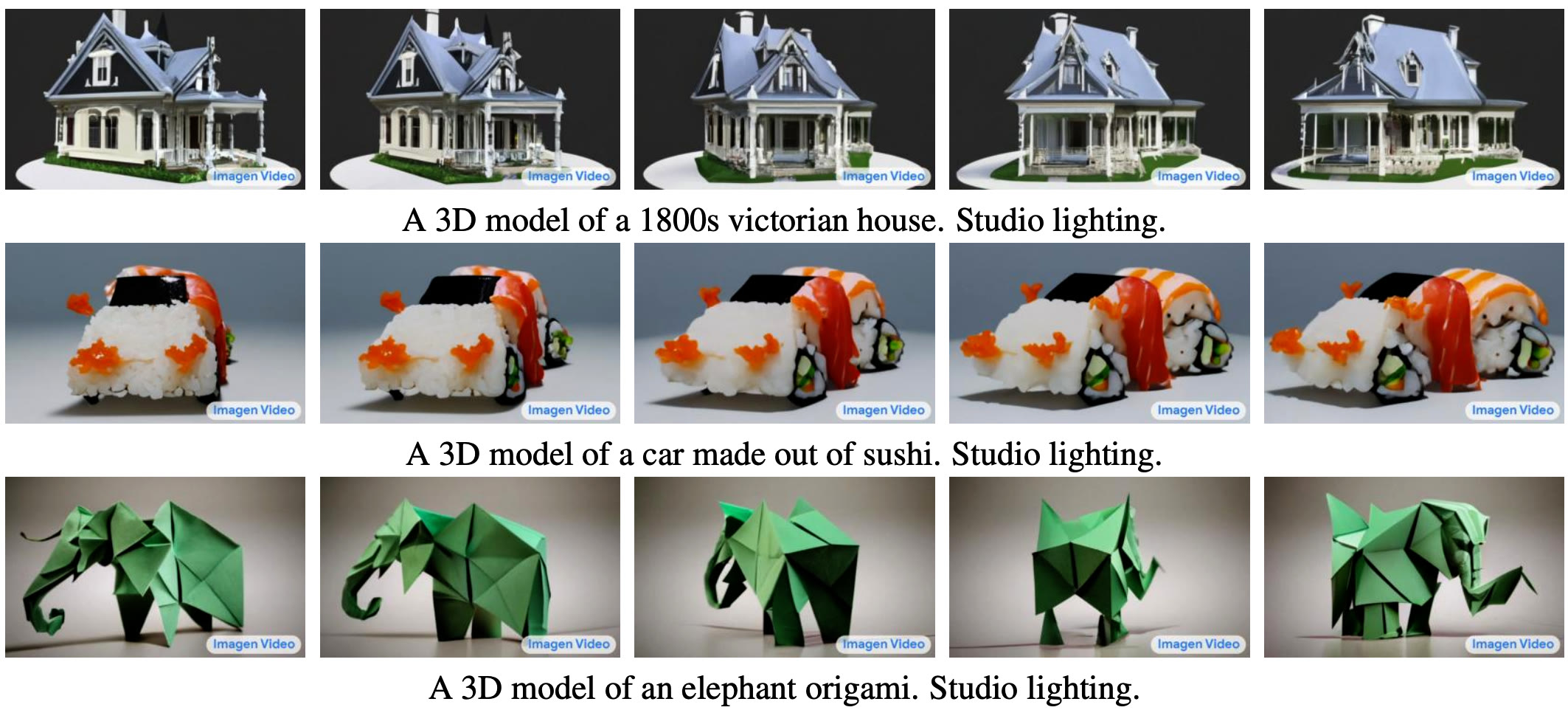}
  \caption{Sample video frames generated by Imagen Video. Note the model's understanding of 3-dimensional structure. Image source:~\cite{ho2022imagen}}
  \label{fig:imagen-video-result}
\end{figure}

Up until now, the video generation models that we have reviewed were modest both in terms of resolution and duration. Ho et al.~\cite{ho2022imagen} introduced Imagen Video, which consists of a cascade of video diffusion systems for scaling diffusion to HD videos. Fig.~\ref{fig:imagen-video-diffusion} shows the cascade pipeline, where they start from a text prompt and end with a 5.3-second video clip at \(1280 \times 768\) resolution at 24 fps. In practice, the text embedding from T5-XXL is injected into all of the models. ``SSR'' and ``TSR'' denote the spatial and temporal super-resolution models. In total, they utilize 8 different models: a text encoder, a base video model, 3 spatial super-resolution models, and 3 temporal super-resolution models. As you can imagine, going through 8 models and performing multiple sampling steps for each diffusion model can be very slow. To alleviate this problem, they apply progressive distillation with classifier-free guidance. In order for the model to absorb a variety of image properties, they train jointly on images and videos. They achieve this by bypassing temporal attention/convolution blocks and passing individual images through the model. Fig.~\ref{fig:imagen-video-result} shows snapshots of a video generated by Imagen Video, showcasing the model's understanding of 3-dimensional structure.

Villegas et al.~\cite{villegas2022phenaki} introduced Phenaki for variable length video generation. They are one of the first publications that can generate arbitrarily long videos by auto-regressively extending in time. They allow the prompt to change over time, where they generate some frames from prompt \(p_0\), then continue from the last \(k\) frames with a new prompt \(p_1\), and so on.

\begin{figure}[h]
  \centering
  \includegraphics[width=\linewidth]{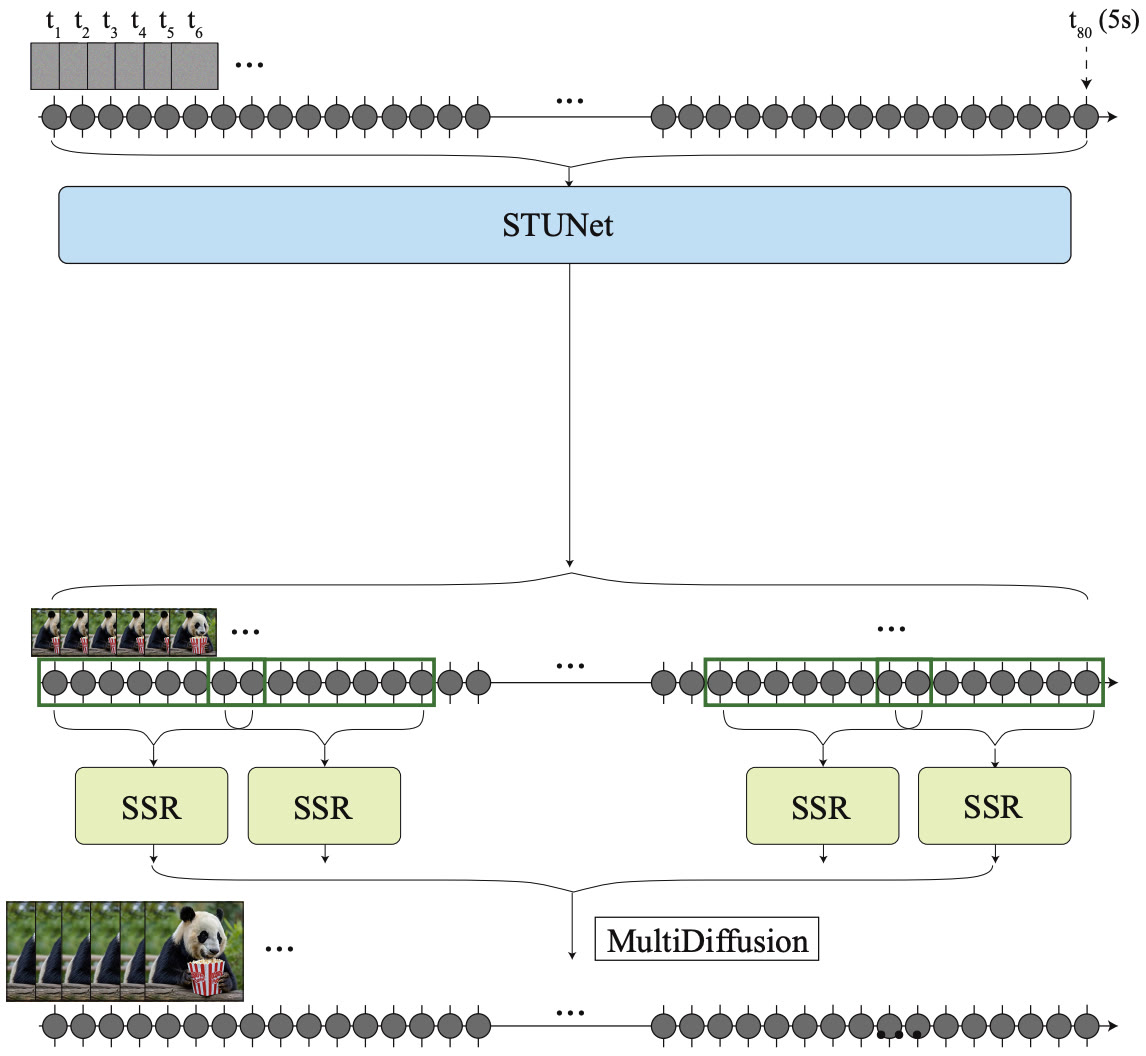}
  \caption{Lumiere model pipeline. Unlike Imagen Video, the Lumiere pipeline processes all the incoming frames at once without a cascade of models which allows it to learn a globally coherent motion. Image source:~\cite{bar2024lumiere}}
  \label{fig:Lumiere-arch}
\end{figure}

\begin{figure}[h]
  \centering
  \includegraphics[width=\linewidth]{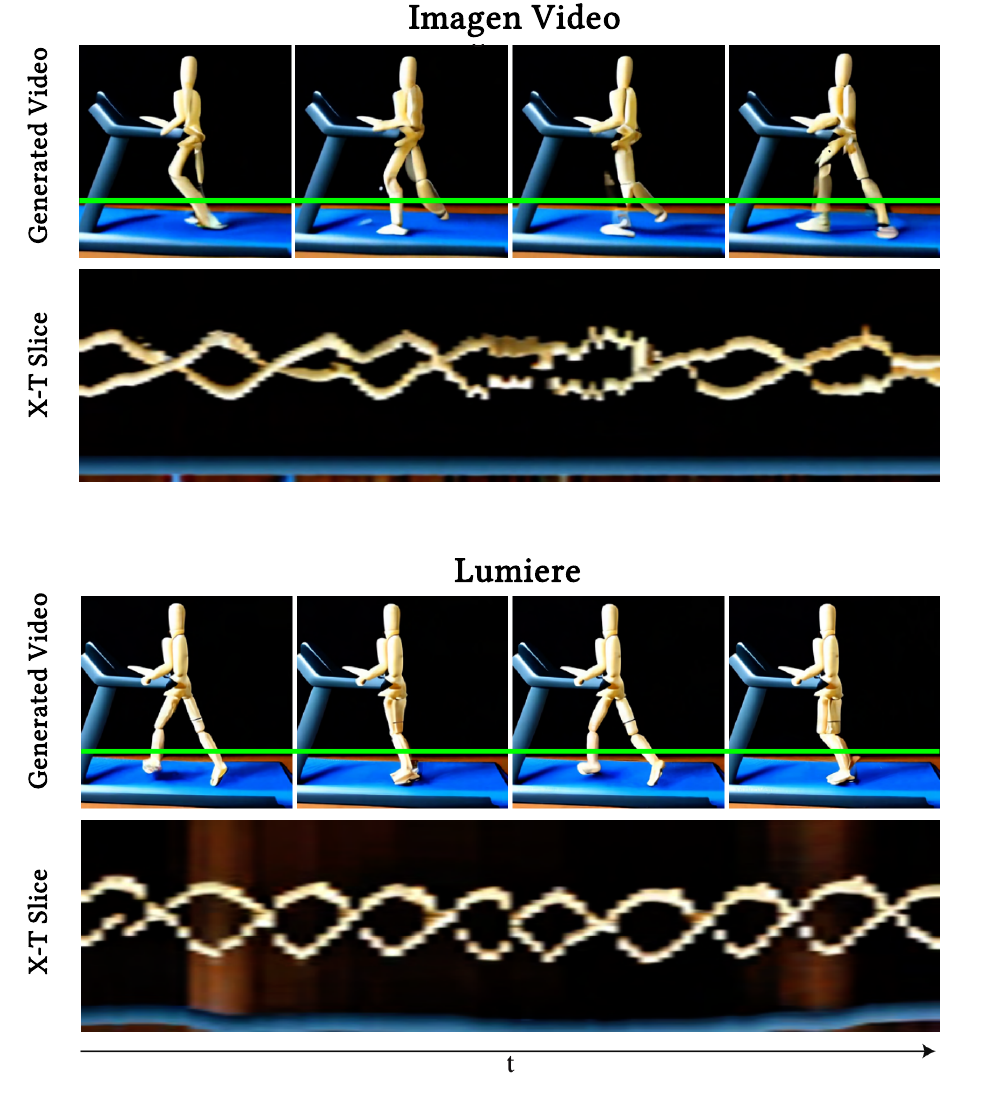}
  \caption{Lumiere applied to the first few frames of video from Imagen Video. Top row shows the X-T slice from the original Imagen Video and bottom row shows the X-T slice from the Lumiere. Note how Lumiere has a more coherent X-T slice. Image source:~\cite{bar2024lumiere}}
  \label{fig:lumiere-result}
\end{figure}

So far, we have seen how researchers utilize a cascade of a base video model, temporal super-resolution (TSR), and spatial super-resolution (SSR) to reach the final high-resolution video. Bar-Tal et al.~\cite{bar2024lumiere} introduce Lumiere by arguing that this can hurt the global motion coherence because of temporal aliasing. Instead, as shown in Fig.~\ref{fig:Lumiere-arch}, they introduce a Space-Time UNet (STUNet) that is capable of generating the full temporal duration at once via a base model. They utilize temporal down- and up-sampling inside the  UNet to compress both space and time simultaneously. They then use an SSR model at the end to reach the high-resolution video. Fig.~\ref{fig:lumiere-result} shows a comparison between their approach and Imagen Video. Note how the frames generated by Lumiere have a more coherent repetitive motion in the X-T slice domain. The X-T slice is created by taking the pixel values along a single horizontal line across all frames in a time sequence (X-axis across frames, T-axis across time).

\subsection{Conclusion}

As we have covered so far, the foundation of video generation models relies on similar core ideas as image generation: strong pretraining, diffusion process as an efficient form of representation, and large-scale data. Video generation is advancing at an unprecedented speed, similar to image generation. Some areas of active research include working towards long-range temporal coherence, physical causality, finer grain details, and compute efficiency.

\section{Fake Image Generation and Societal Impacts}
Like every new breakthrough, as image generation models improve there is growing concern about their use. Examples of these areas of concern include the following. \textbf{Deepfakes:} fake photos or videos of public figures that seem real and can be used to manipulate the public. \textbf{Copyright:} images that try to closely mimic the style of another artist. \textbf{Bias:} models that may produce a harmful and unfair representation of a specific gender, race, nationality, culture, etc. \textbf{Job market:} growing pressure on artists, creators, photographers, and designers alike. \textbf{Fraud:} using AI-generated images for phishing purposes, e.g., creating a fake receipt or fake proof. \textbf{Privacy:} creating fake images of someone based on their publicly available images without their permission.

\subsection{Societal Impact}

\begin{figure}[h]
  \centering
  \includegraphics[width=\linewidth]{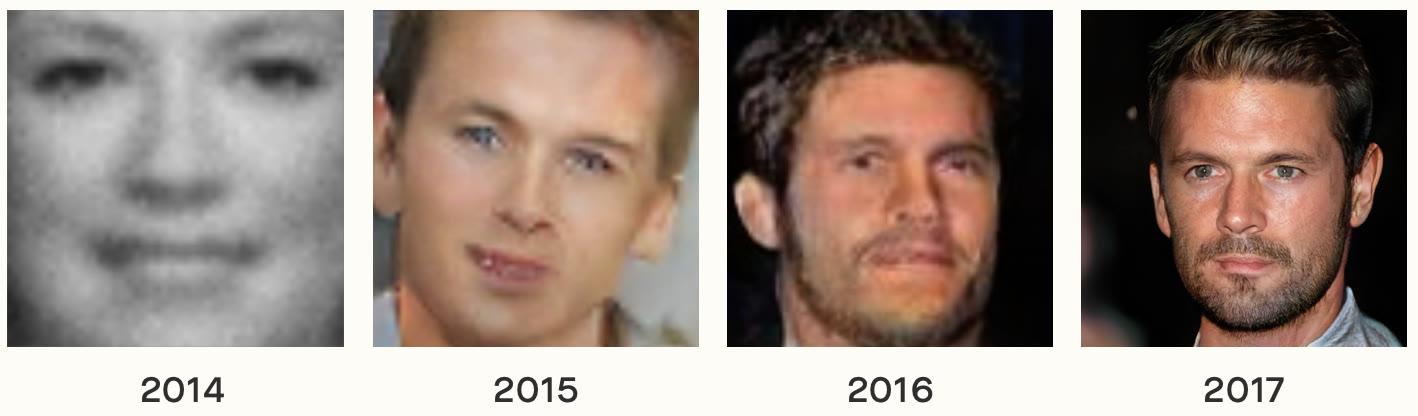}
  \caption{AI generated images have become increasingly photorealistic over time. Image source:~\cite{brundage2018malicious}}
  \label{fig:img-gen-history}
\end{figure}

As shown in Fig.~\ref{fig:img-gen-history}, image generation models have improved significantly in a short span of time. As the images generated by these models become indistinguishable from natural (real) images, their potential misuse can have increasingly severe impacts. Given the current popularity of video streaming and social network applications such as YouTube, Instagram, and TikTok, fake videos and images can more readily become viral. The dangers of fake viral images and videos include swaying public opinion, spreading fake news with realistic fabricated videos and audio, and enabling propaganda campaigns. For example, Goss et al.~\cite{Goss2024DeepfakesSocietalImpacts} discuss how AI-generated videos caused confusion and provoked public reactions during the Russo--Ukrainian war. Priti et al.~\cite{Priti2025DeepfakesDevelopingSocieties} discuss the impact of fake AI-generated content in vulnerable, low-resource communities. Moreover, bad actors can generate convincing videos of real individuals from as little as a single photograph and using them for harassment, blackmail, or manipulation purposes. Hameleers~\cite{Hameleers2024CheapDeep} goes beyond theory and measures how visual manipulation actually affects people and elicits an emotional reaction.

In one of the early major legal and societal surveys of deepfakes, Chesney and Citron~\cite{Chesney2019DeepFakes} zoom out from technical architectures of models and focus on the impact of these capabilities on society. They describe how applications like FakeApp are diffusing into the public sphere, effectively democratizing access to powerful but potentially harmful tools. They also lay out a legal perspective on civil liabilities for creators and platforms, as well as the criminal treatment of malicious uses.

Paris and Donovan~\cite{Paris2019DeepfakesCheapFakes} argue that AI-generated images are more computationally intensive versions of earlier image-editing, cutting, and re-staging tools. They also note that so-called cheapfakes (non-AI manipulations) are already doing significant damage, and deepfakes amplify their impact. They point out that mere technical fixes are not enough and we need social, cultural, and structural changes. For example, we must ask who is interpreting the evidence, whose safety we are prioritizing, and how we are curbing the power of existing elites. They call for a combination of technical and social remedies in order to mitigate these risks. Bender et~al.~\cite{Bender2021StochasticParrots} also talk about the impact of these large models on the environment and suggest that developers of these models should report the energy usage of their model when reporting the performance. Moreover, bias in training data is another area that they cover since most of the data is scraped from the web.

\subsection{Image Manipulation Detection}

Given all of these societal concerns, many efforts have been made to distinguish whether an image was generated or manipulated by a computer. As early as 2004, Popescu and Farid~\cite{popescu2004exposing} utilized principal component analysis (PCA) on small fixed-size image blocks to detect duplicated image regions, where portions of an image are copied and pasted to conceal a person or object in the scene. After applying PCA to the image blocks and lexicographically sorting all of the image blocks, they were able to detect duplicate regions, since similar blocks will end up next to each other in the sorted list.

Later, Farid~\cite{Farid2009ImageForgeryDetection} published a survey that presents a systematic way to detect the manipulation of digital images. Editing an image always leaves behind statistical and physical inconsistencies, and this can be utilized as the basis for detecting digital image manipulation. The author also warns that as fake image manipulation methods improve, we will need more advanced ways to detect them.

Luk\'a\v{s} et al.~\cite{lukas2006digital} establish photo-response non-uniformity (PRNU). PRNU is a noise pattern that is unique to cameras and can be a powerful way to detect forgery. They take many pictures with the same camera and run a denoising filter $F(\cdot)$. The result is the noise residuals:

\begin{align}
r^{(k)} &= p^{(k)} - F\!\left(p^{(k)}\right),\,\, P_c \approx \frac{1}{N_p}\sum_{k=1}^{N_p} r^{(k)}
\end{align}
where $p^{(k)}$ is the $k$th image, $N_p$ is the number of images for a specific camera, and $P_c$ is the reference camera pattern. For a new image $p$ at test time, they compute the noise residual $n$ and the correlation between $n$ and $P_c$ via

\begin{align}
n &= p - F(p),\,\, \rho_c(p) = \mathrm{corr}(n, P_c)
\end{align}
They model the correlation distribution for ``same camera'' versus ``different camera'' and choose a threshold value that balances false acceptance and false rejection. If someone claims an image to be from camera $C$ but the residuals do not correlate with $P_c$, this can be used as evidence of manipulation.

\subsection{Face Manipulation and Detection}

\begin{figure}[t]
  \centering
  \includegraphics[width=\linewidth]{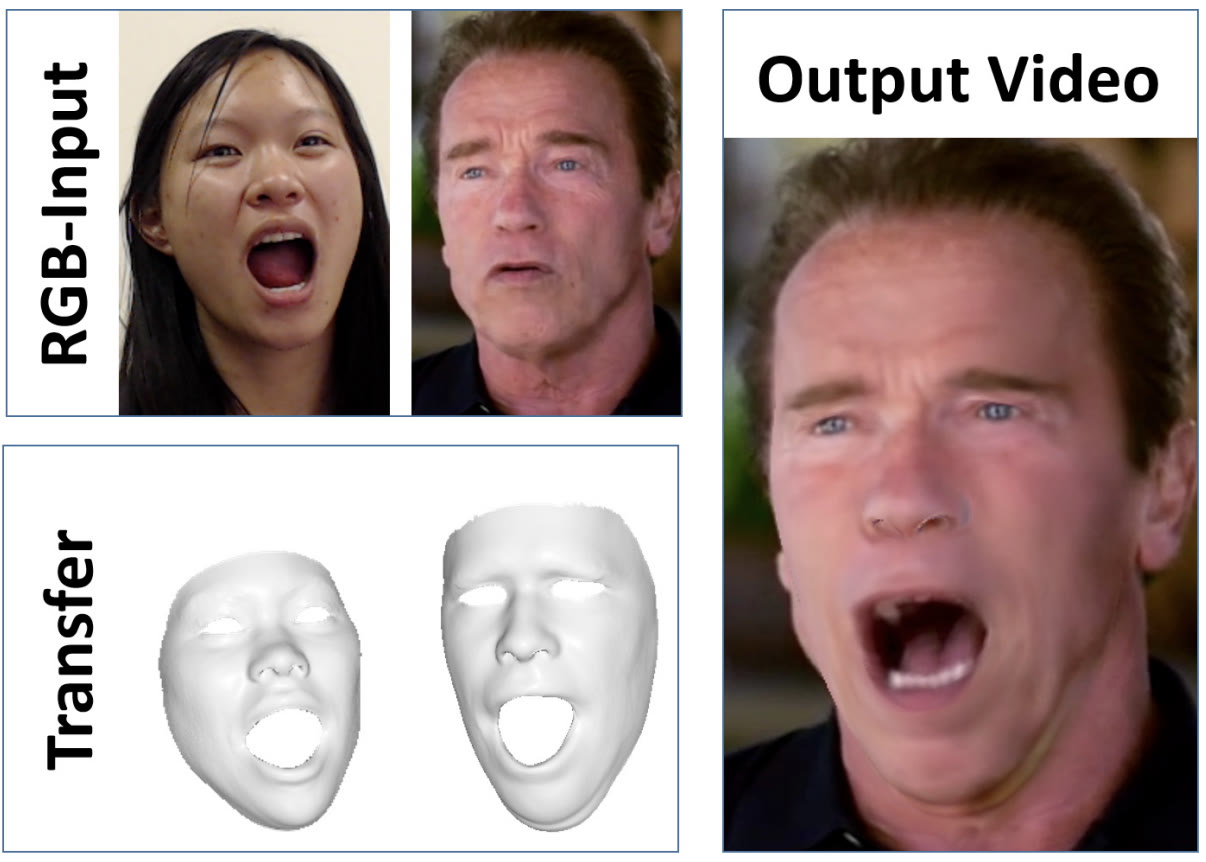}
  \caption{Transferring the source to the target expression in facial reenactment. Image source:~\cite{thies2016face2face}}
  \label{fig:thies_example}
\end{figure}

\begin{figure}[t]
  \centering
  \includegraphics[width=\linewidth]{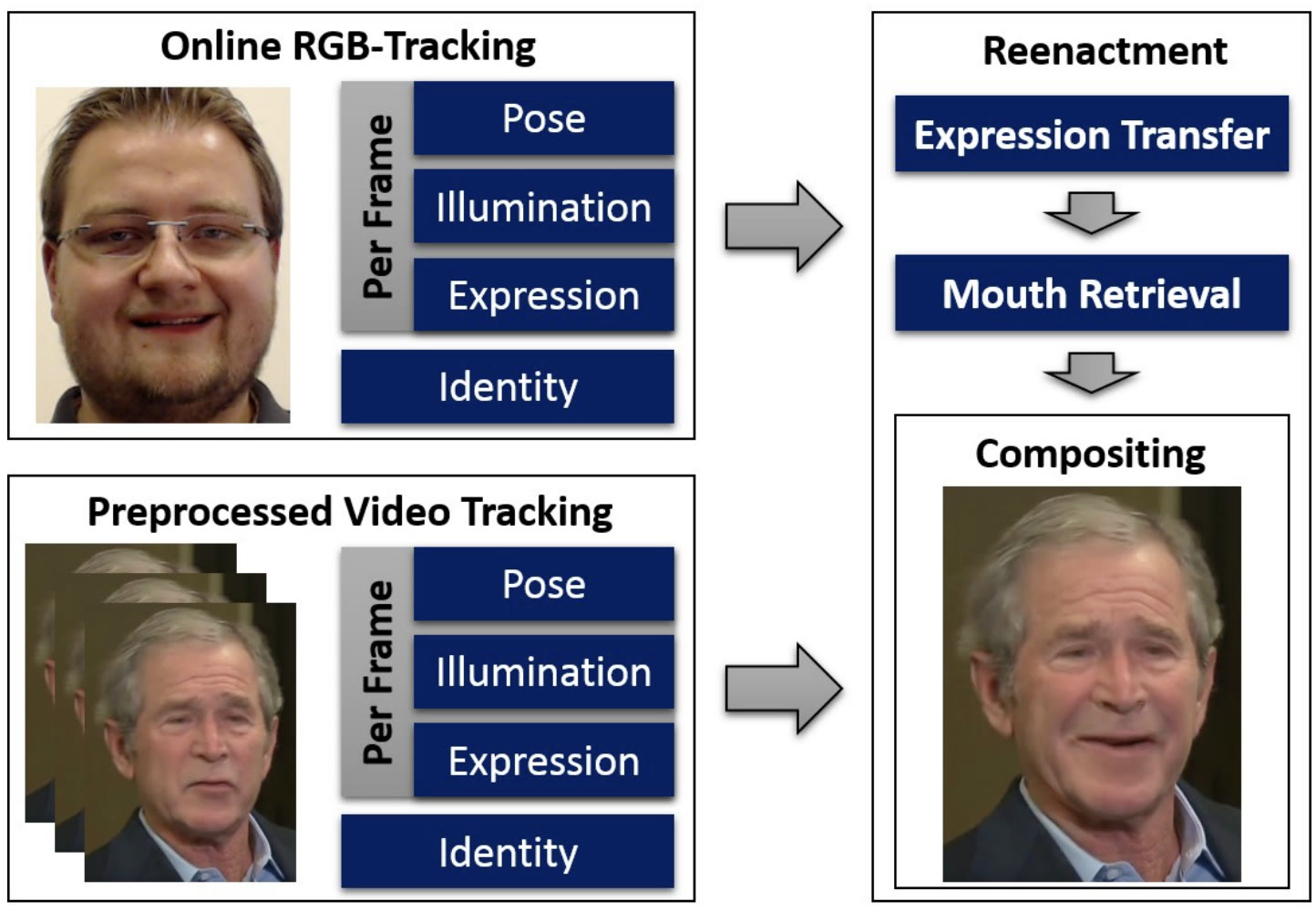}
  \caption{Overall pipeline for facial reenactment. Image source:~\cite{thies2016face2face}}
  \label{fig:thies_overview}
\end{figure}

Starting in 2016, researchers began to publish convincing results in real-time face manipulation. Most notably, Thies et al.~\cite{thies2016face2face} and Kim et al.~\cite{kim2018deep} published two important papers in which they were able to transfer the facial expression of a source image to a target image. Fig.~\ref{fig:thies_example} shows the results of Thies et al.~\cite{thies2016face2face}. Their approach is called facial reenactment. Earlier reenactment systems could only perform this offline, and many could only perform self-reenactment or less realistic mouth movements, while their approach had none of these limitations. Their approach relies on a multi-linear PCA face model that separates identity from expression. Fig.~\ref{fig:thies_overview} shows an overview of their approach.

\begin{figure}[h]
  \centering
  \includegraphics[width=\linewidth]{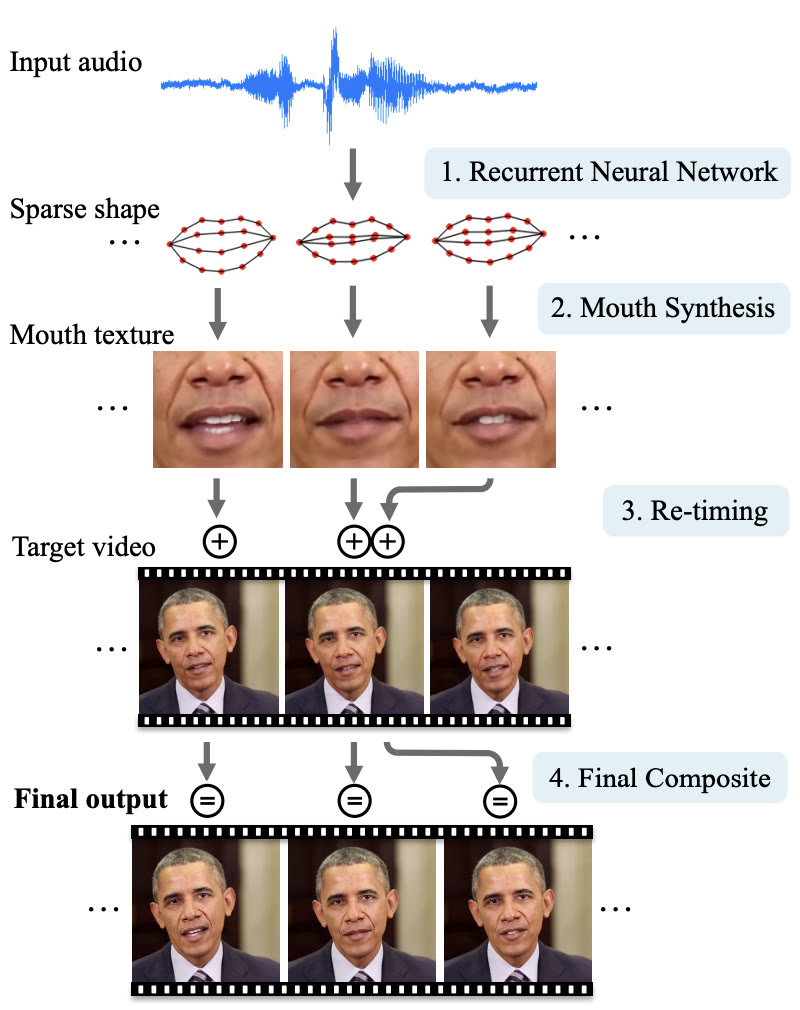}
  \caption{They first convert audio into a sparse, time-varying mouth shape, then generate a photorealistic mouth texture and composite it onto the target video. They also match and re-time the mouth sequence to the video so head motion looks natural and fits the speech. Image source:~\cite{Suwajanakorn2017Obama}}
  \label{fig:obama-sync}
\end{figure}

Suwajanakorn et~al.~\cite{Suwajanakorn2017Obama} showed that by training on a large video corpus of a single speaker, one can generate photorealistic mouth regions of a talking speaker from audio alone. Their result looked both convincing and alarming in that with enough data for one identity, audio-to-video lip-sync can look convincingly real. Fig.~\ref{fig:obama-sync} shows the overall pipeline of their approach.

\subsection{DeepFake Detection and Benchmarks}

\begin{figure}[h]
  \centering
  \includegraphics[width=\linewidth]{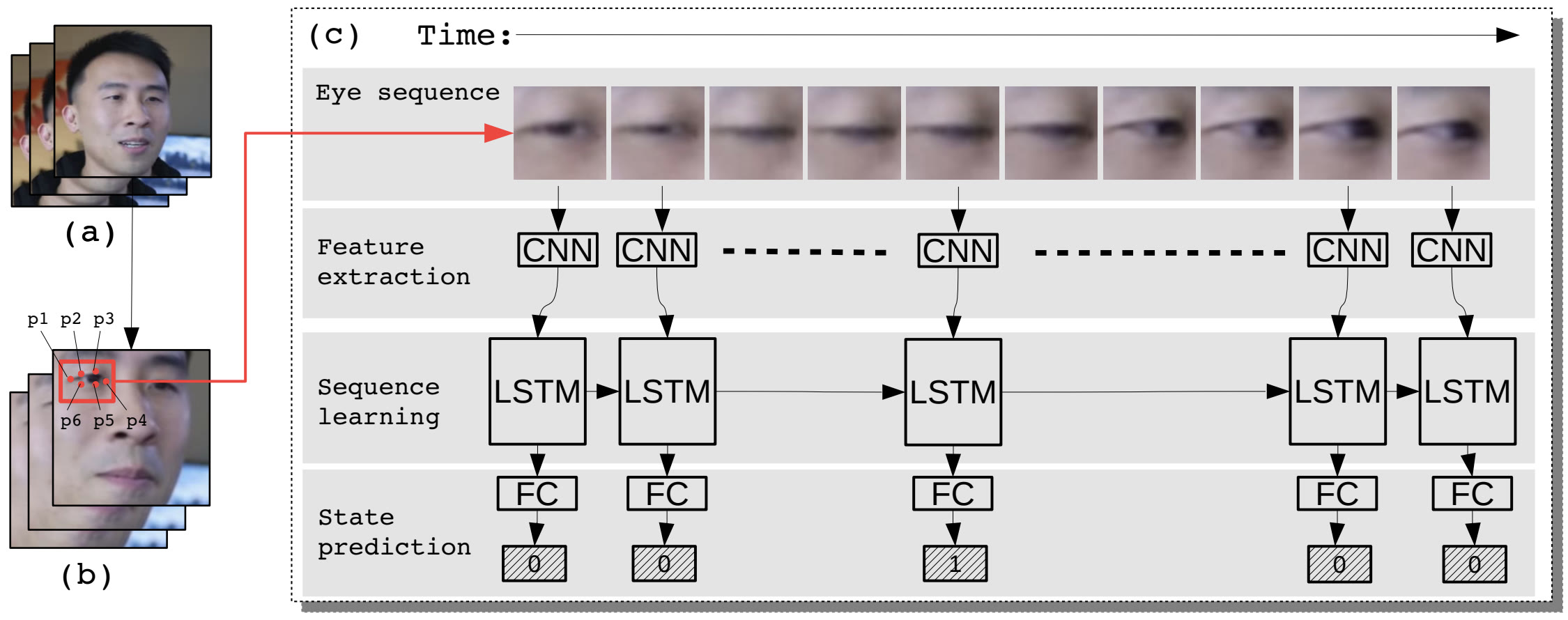}
  \caption{Overall pipeline for detecting blinking in fake face videos. (a) original sequence (b) sequence after alignment (c) LRCN: feature extraction, sequence learning, and state prediction. Image source:~\cite{li2018ictu}}
  \label{fig:li_blink_pipeline}
\end{figure}

After the surge of fake AI generated videos, a lot of efforts have gone into detecting such cases. Li et al.~\cite{li2018ictu} proposed a new method for detecting fake face videos by analyzing eye blinking. Fig.~\ref{fig:li_blink_pipeline} shows an overview of their pipeline, which relies on Long-term Recurrent Convolutional Neural Networks (LRCN). As shown in the figure, they first detect and align faces in each frame to reduce head pose and motion noise. Then they cropped the eye regions and passed them to a VGG-based CNN for feature extraction. The consecutive frame features are then fed to an LSTM to model the temporal dynamics of blinking. Finally, a fully connected layer predicts whether the eyes are open or closed. By reliably predicting whether an eye is open or closed, they can measure whether the blink pattern appears human or not.

Later, Li and Lyu~\cite{li2018exposing} observed that the DeepFake algorithm generates images at a specific resolution, and these need to be further warped to match the original faces in the source video. This transformation leaves behind distinctive artifacts that can be used for detecting DeepFake videos. Here is how their detection algorithm works. They first generate positive (real) and negative (fake) images. For the positive class, they collect the data from the internet, and for the negative class they generate fake images on the fly by simulating the warping artifact rather than running a DeepFake model. Their model consists of common backbones like VGG16 and ResNet50/101/152, which start from ImageNet weights and are then fine-tuned. For the input to the model, they crop the input image such that it contains the face and some surrounding area, since the classifier has to detect the inconsistency between the two regions. At the end, they use a hard-mining stage to define the decision boundary for fake versus real images. Table~\ref{tab:li_lyu_comparison} shows an AUC performance comparison between their methods and others on the UADFV~\cite{yang2019exposing} and DeepfakeTIMIT~\cite{korshunov2018deepfakes} datasets.
\begin{table}[t]
\centering
\begin{tabular}{l c cc}
\hline
\textbf{Methods} & \textbf{UADFV} & \multicolumn{2}{c}{\textbf{DeepfakeTIMIT}} \\
 &  & \textbf{LQ} & \textbf{HQ} \\
\hline
Two-stream NN~\cite{zhou2017two}     & 85.1 & 83.5 & 73.5 \\
Meso-4~\cite{afchar2018mesonet}            & 84.3 & 87.8 & 68.4 \\
MesoInception-4~\cite{afchar2018mesonet}   & 82.1 & 80.4 & 62.7 \\
HeadPose~\cite{yang2019exposing}          & 89.0 & --   & --   \\
\hline
VGG16-backbone        & 84.5 & 84.6 & 57.4 \\
ResNet50-backbone     & \textbf{97.4} & \textbf{99.9} & \textbf{93.2} \\
ResNet101-backbone    & 95.4 & 97.6 & 86.9 \\
ResNet152-backbone    & 93.8 & 99.4 & 91.2 \\
\hline
\end{tabular}
\caption{DeepFake detection AUC performance comparison on UADFV~\cite{yang2019exposing} and DeepfakeTIMIT~\cite{korshunov2018deepfakes} datasets. Table source:~\cite{li2018exposing}}
\label{tab:li_lyu_comparison}
\end{table}

\begin{figure}[t]
  \centering
  \includegraphics[width=\linewidth]{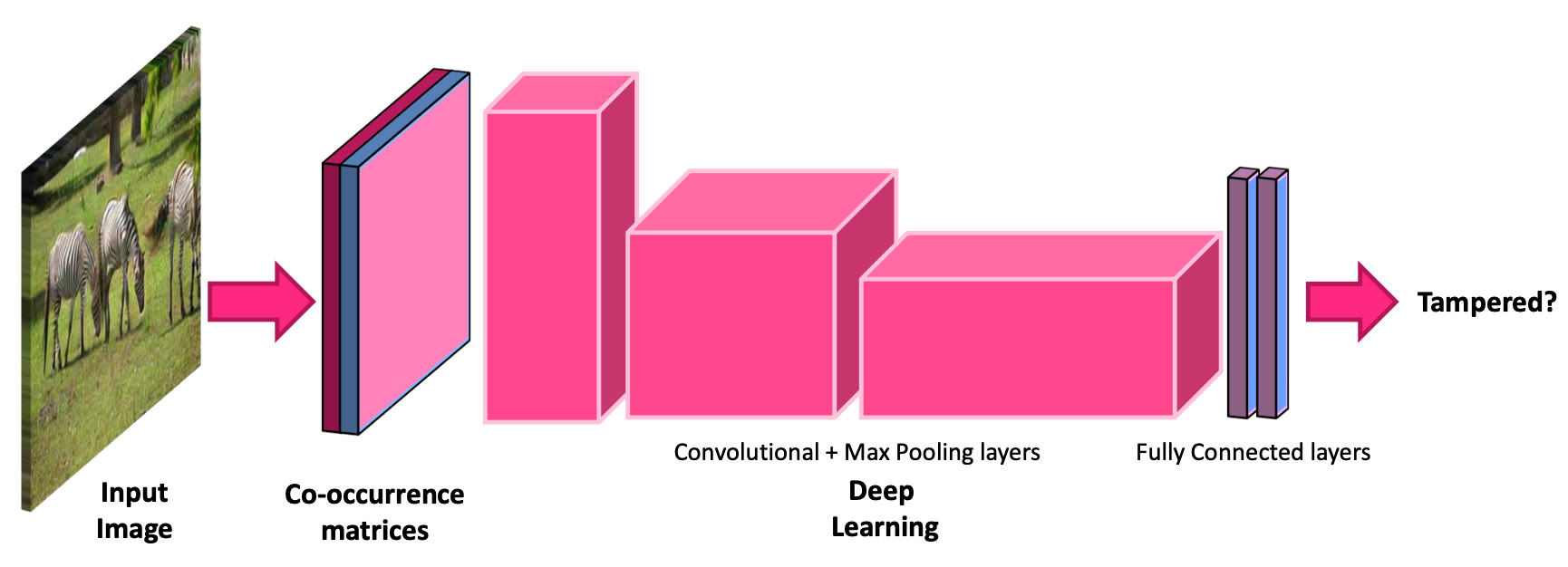}
  \caption{End-to-end pipeline for training a model for detecting GAN generated images. Image source:~\cite{nataraj2019detecting}}
  \label{fig:gan-detect}
\end{figure}

Rössler et~al.~\cite{rossler2019faceforensics++} introduced FaceForensic++, a standardized dataset for detecting fake images. With this, they provide a large-scale standardized benchmark with multiple manipulation methods and realistic compression conditions. The dataset consists of 1.8M manipulated images derived from 1000 real videos using the most advanced methods of the time: Face2Face~\cite{thies2016face2face}, FaceSwap~\cite{kowalski_faceswap_2017}, DeepFakes~\cite{faceswap_deepfakes_2025}, and NeuralTextures~\cite{thies2019deferred}. Later Dolhansky et~al.~\cite{dolhansky2019deepfake} introduced the Deep Fake Detection Challenge Dataset (DFDC). DFDC consists of 5,214 videos , tampered and original with the ratio of 1:0.28. Realizing how fragile the accuracy metrics of DeepFake detection models could be under different scenarios (skin tone, orientation, background color, lighting, etc.), they created this dataset with this in mind, ensuring that it covers all of these scenarios as comprehensively as possible. The community continued its campaign for creating better and more comprehensive datasets, and later Jiang et~al.~\cite{jiang2020deeperforensics} introduced DeeperForensics-1.0 that further increased the number of videos to 60,000.

Nataraj et al.~\cite{nataraj2019detecting} shifted towards an end-to-end framework to detect GAN-generated images, as shown in Fig.~\ref{fig:gan-detect}. They no longer rely on handcrafted features and instead let the CNN learn discriminative patterns from co-occurrence matrices computed separately on the R, G, and B channels. They evaluated the performance of the model on two different datasets: CycleGAN~\cite{zhu2017unpaired} and StarGAN~\cite{choi2018stargan}. CycleGAN contains image-to-image style transfer samples, and StarGAN consists of images of real celebrities, as well as GAN-generated celebrity images. In order to evaluate how well the model generalizes, they train on one dataset and use the other dataset for testing. Table~\ref{tab:nataraj_cross_dataset} shows the performance of the model on the CycleGAN and StarGAN datasets.

\begin{table}[t]
\centering
\begin{tabular}{l l c}
\hline
\textbf{Training dataset} & \textbf{Testing dataset} & \textbf{Accuracy} \\
\hline
CycleGAN & StarGAN & 99.49 \\
StarGAN  & cycleGAN & 93.42 \\
\hline
\end{tabular}
\caption{Cross-dataset detection accuracy when training on CycleGAN vs StarGAN.}
\label{tab:nataraj_cross_dataset}
\end{table}

\begin{figure}[t]
  \centering
  \includegraphics[width=\linewidth]{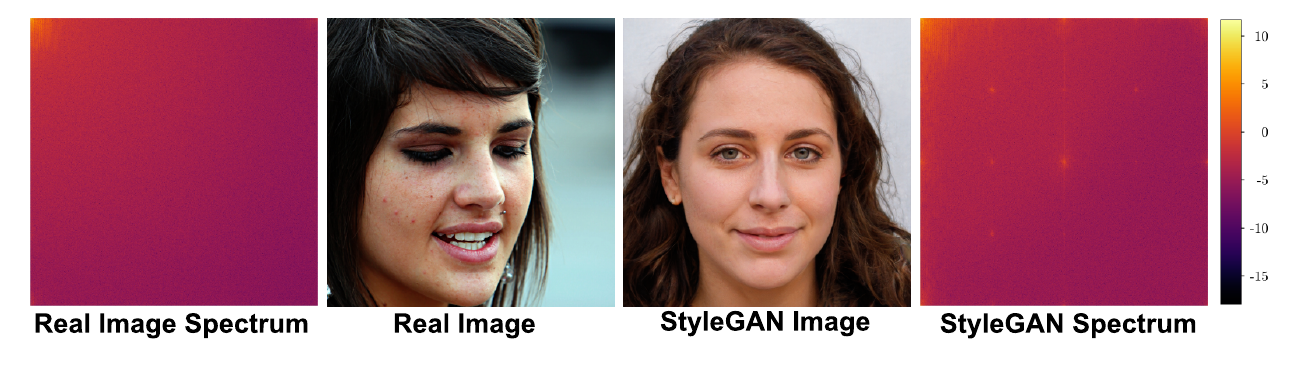}
  \caption{Comparison between the spectrum of a real image and a generated image. Image source:~\cite{frank2020leveraging}}
  \label{fig:spectral-2}
\end{figure}

\begin{figure}[h]
  \centering
  \includegraphics[width=\linewidth]{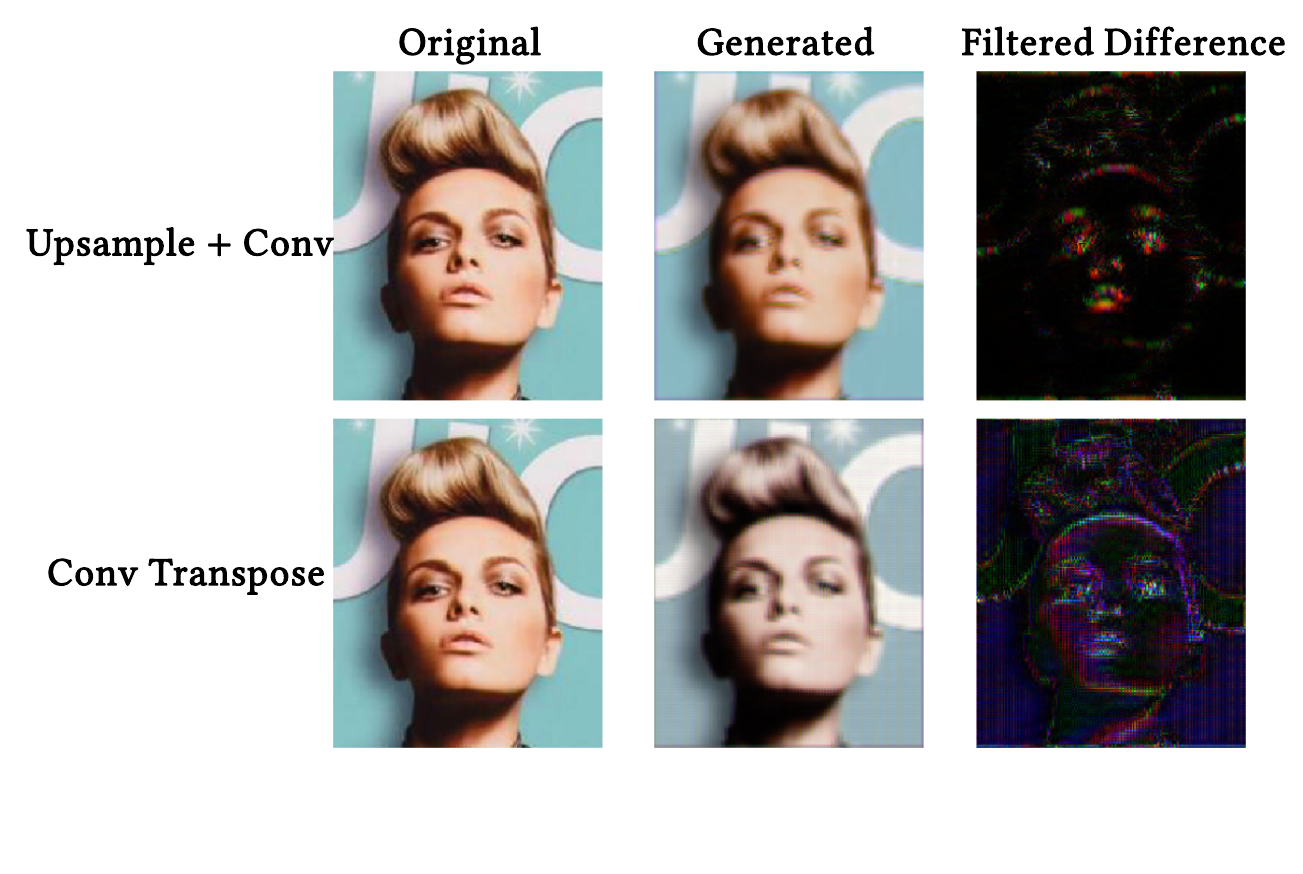}
  \caption{Impact of up convolutional layers in the spectral response of the GAN/VAE generated images. Image source:~\cite{durall2020watch}}
  \label{fig:spectral-gan}
\end{figure}

\subsection{Generated Images Leave Behind Artifacts}

Durall et al.~\cite{durall2020watch} find that GAN-based image generation models leave behind artifacts in the frequency domain that can be used to detect GAN-generated images. Fig.~\ref{fig:spectral-2} compares the spectrum of a real image and a fake image. They argue that convolutional upsampling blocks introduce systematic spectral distortions in the generated image. There are two commonly used methods for up-convolution: upsampling + convolution and transposed convolution. As shown in Fig.~\ref{fig:spectral-gan}, both can cause artifacts. In order to detect these artifacts in the generated images, they first take the Discrete Fourier transform (DFT) of the input image and then convert it to a 1D spectrum. The 1D spectrum is then passed to a classifier to detect fake and real images.

Conversely, to cope with this spectral artifact in generated images, they propose adding a spectral regularization term during training to encourage the generator to match real-image spectral statistics.
\begin{equation}
\mathcal{L}_{\text{final}} = \mathcal{L}_{\text{Generator}} + \lambda \, \mathcal{L}_{\text{spectral}}
\end{equation}
where $\lambda$ is a hyperparameter that weighs the importance of the spectral loss. Later, Frank et~al.~\cite{frank2020leveraging} used a similar frequency based approach to detect AI generated images.

\subsection{Watermarking Generated Images}

As image generation models improved with the introduction of diffusion-based models, it became increasingly difficult to detect images generated by diffusion models. Wang et al.~\cite{wang2023dire} ask what happens if they take the image through the inversion and reconstruction path. For diffusion-generated images, there should be a noise path that leads to a better reconstruction. In order to showcase this, they first take the image through the DDIM inversion process:
\begin{equation}
\begin{aligned}
\frac{x_{t+1}}{\sqrt{\alpha_{t+1}}}
  &= \frac{x_t}{\sqrt{\alpha_t}}
   + \left(
        \sqrt{\frac{1-\alpha_{t+1}}{\alpha_{t+1}}}
        - \sqrt{\frac{1-\alpha_t}{\alpha_t}}
     \right)\epsilon_\theta(x_t, t)
\end{aligned}
\end{equation}
where $\epsilon_\theta$ is a pretrained diffusion model. They continually add noise to the input image and reach $x_T$ after $S$ steps. Next, they run a DDIM deterministic denoising process that takes $x_T$ back to the input image, as shown in Fig.~\ref{fig:dire}:

\begin{figure}[t]
  \centering
  \includegraphics[width=\linewidth]{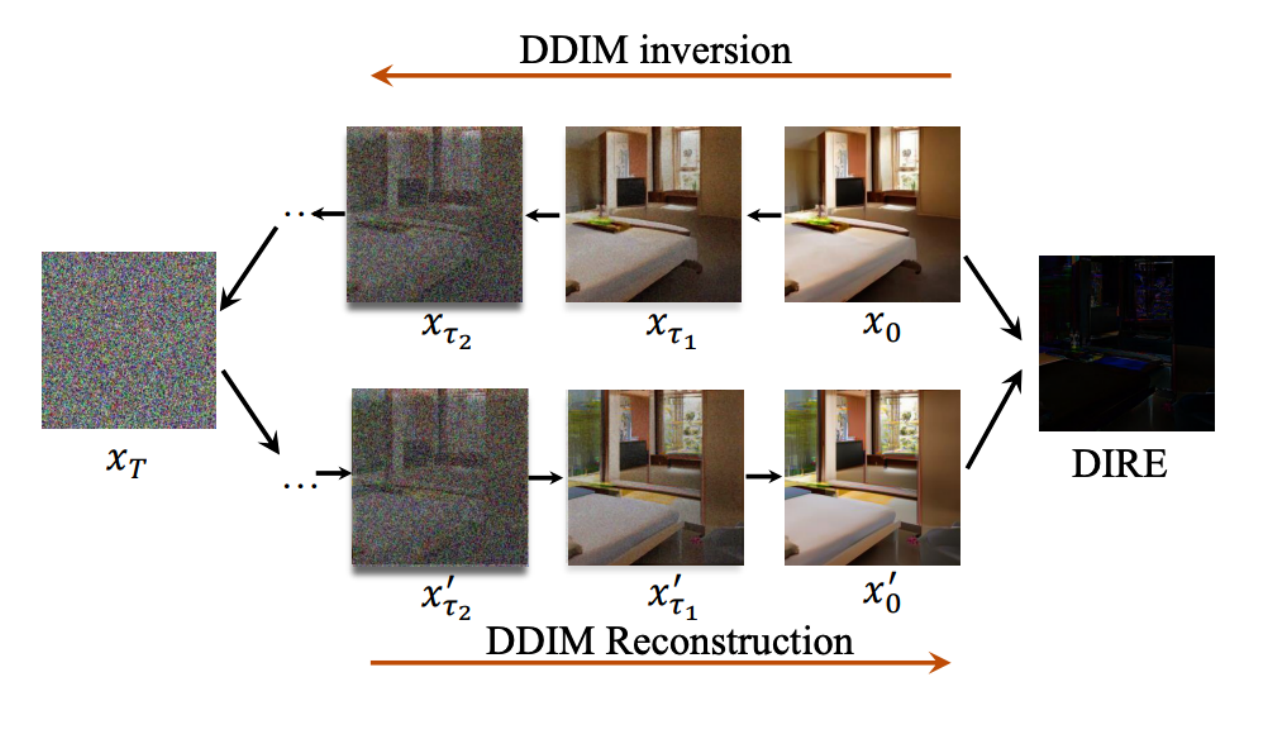}
  \caption{The overall pipeline for Diffusion Reconstruction Error (DIRE). The input image ($x_0$) goes through DDIM inversion, DDIM reconstruction, and DIRE process respectively. Image source:~\cite{wang2023dire}}
  \label{fig:dire}
\end{figure}

\begin{equation}
\begin{aligned}
x_{t-1}
  &= \sqrt{\alpha_{t-1}}
     \left(
       \frac{x_t - \sqrt{1-\alpha_t}\,\epsilon_\theta(x_t, t)}
            {\sqrt{\alpha_t}}
     \right) \\
  &\quad
     + \sqrt{1-\alpha_{t-1}-\sigma_t^{2}}\,\epsilon_\theta(x_t, t)
     + \sigma_t \epsilon_t
\end{aligned}
\end{equation}

\begin{figure}[t]
  \centering
  \includegraphics[width=\linewidth]{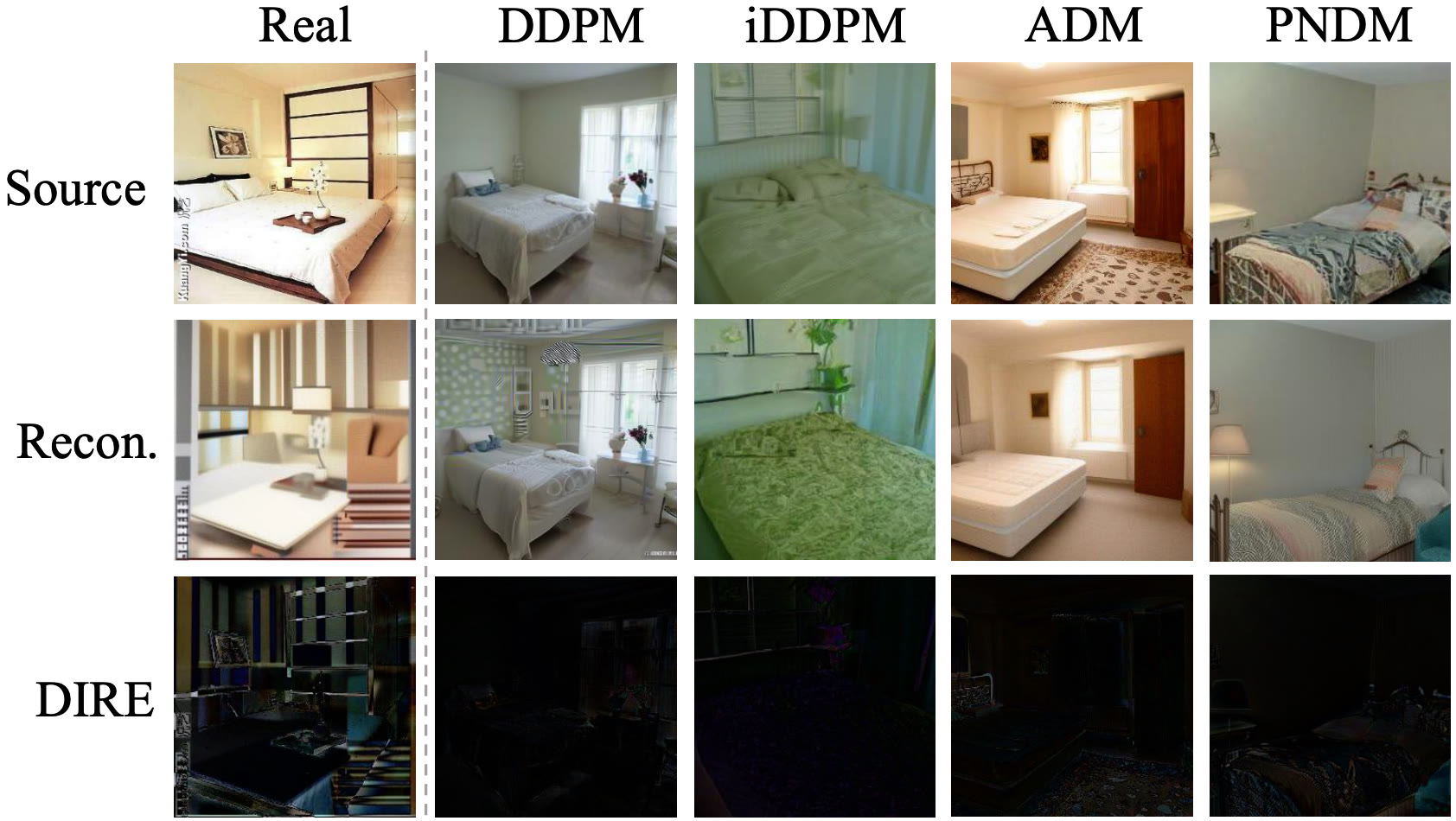}
  \caption{Real images tend to have larger DIRE values compared to diffusion generated images. Image source:~\cite{wang2023dire}}
  \label{fig:dire_vis}
\end{figure}

\noindent Near the end, they build the diffusion reconstruction error (DIRE)
\begin{equation}
\mathrm{DIRE}(x_0) = \lVert x_0 - x_0' \rVert
\end{equation}
where $x_0$ is the original image and $x_0'$ is the reconstructed image. Fig.~\ref{fig:dire_vis} shows the DIRE representation of real and diffusion-generated images. In experiments, real images show larger DIRE values compared to diffusion-generated ones. Lastly, they train a binary (ResNet-based) classifier on DIRE to detect diffusion-generated images. Using this approach, they are able to achieve better accuracy and average precision compared to other state-of-the-art detector models.

\begin{figure*}[t]
  \centering
  \includegraphics[width=\linewidth]{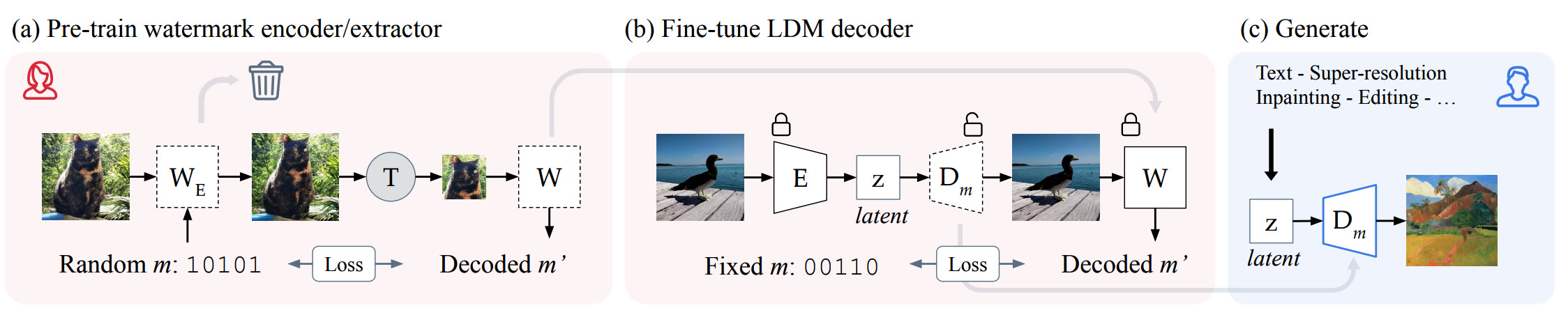}
  \caption{Stable signature steps. (a) Pre-training the watermark encoder and decoder. (b) Finetuning the diffusion decoder. (c) Generating an image with invisible watermarks. Image source:~\cite{fernandez2023stable}}
  \label{fig:fernandez_pipeline}
\end{figure*}

Fernandez et al.~\cite{fernandez2023stable} take a different approach and try to add invisible watermarks to generated images so that any usage of a diffusion model carries that watermark. As shown in Fig.~\ref{fig:fernandez_pipeline}, they first train a watermark encoder and extractor. Given an input image $x_0$ and a random binary message $m \in \{0,1\}^K$, the watermark encoder $W_E$ outputs a residual $\delta$ with the same size as the image $x_0$. The watermarked image is then formed as
\begin{equation}
x_w = x_0 + \alpha \delta
\end{equation}
where $\alpha$ controls the strength of the watermark in the generated image. They next apply random distortions $T$ (cropping, JPEG compression, brightness changes, etc.) to simulate real-world edits,
\begin{equation}
\tilde{x}_w = T(x_w)
\end{equation}
The extractor $W$ is then used to recover the message, and they train it with a binary cross-entropy loss between the true bits $m$ and the predicted bits. They discard the encoder $W_E$ and keep $W$ as a fixed extractor. In the next phase, they start from a pretrained diffusion model with encoder $E$ and decoder $D$, and use the fixed extractor $W$ to teach the diffusion decoder to embed a fixed signature $m$ into all outputs. This results in a signed decoder $D_m$ for each signature $m$. Concretely,
\begin{equation}
\begin{aligned}
z      &= E(x), \\
x_{\text{orig}} &= D(z), \\
x_{\text{sig}}  &= D_m(z), \\
y      &= W(x_{\text{sig}}) \in \mathbb{R}^K
\end{aligned}
\end{equation}
They use a message loss $L_m$ that compares the output $y$ to the original signature $m$ and an image similarity loss $L_i$ that compares $x_{\text{sig}}$ to $x_{\text{orig}}$, and the overall loss becomes
\begin{equation}
L = L_m + \lambda_i L_i ,
\end{equation}
where $\lambda_i$ is a weight on the image similarity loss. Note that only the signed decoder parameters of $D_m$ are updated but the encoder $E$ and extractor $W$ are kept frozen. Using a 48-bit signature, they are able to detect $99\%$ of newly generated images on the MS-COCO dataset while falsely flagging real images as fake only about one in $10^9$. Compared to Corvi et al.~\cite{corvi2023detection}, they achieve an order of magnitude lower false positive rate for a similar true positive rate.

\subsection{Conclusion}

Image generation models have made significant leaps over the past decade. We have moved from models that produced low-quality, smudged outputs to models that generate images and coherent videos that are nearly indistinguishable from reality. Recent empirical scaling trends suggest continued improvements in the capabilities of these models. A central challenge is establishing effective safeguards against potential misuses. Do we have adequate safety measures in place? Are we prepared for a future in which the harmful uses of image generation systems, described above, are realized at scale? Given the risks associated with these powerful image and video generation models, we need both technical and societal solutions to mitigate these risks and to ensure preparedness for worst case scenarios.

\section{Final Conclusion}

Image generation models have gone through a revolution in the past decade. We started from models that were able to generate low quality images in a restricted setting to today's models that are able to generate images with exceptional quality and user control. Users can control different aspects of the generated images including quality, fine details, style, background, coloring, etc.

VAEs provided a probabilistic framework for generative image modeling. They are still valuable when interpretability and controllability of the generated output matter. GANs had probably the most impact by creating the first photorealistic images; however, mode collapse and instability of training were recurring issues with them.

Normalizing flows offered a simple but powerful mathematical framework for image generation. Their invertible formulation, exact likelihood, and one-step sampling made them an excellent candidate for image generation; however, they struggle to model complex image generation tasks as well as diffusion and autoregressive models. Autoregressive models are among the most stable models when it comes to training. They are also a great fit for conditional generation because of their autoregressive nature, but this autoregressive nature comes with a price, since generation can be slow and expensive.

Diffusion models started from iterative pixel domain denoising and evolved to faster, few iteration latent space denoising. The overall complexity of diffusion model architectures increased as their capabilities grew. Architectural advances enabled conditioning on other modalities such as text and image embeddings.

Rectified Flow and Flow Matching provide new objectives, as well as a new framework for designing the next generation of image generation systems. These include defining a transport path, a learned vector field, and a new sampling pipeline with as few function evaluations as possible.

Finally, as image generation models become more capable, the potential misuses (deepfakes, fraud, manipulation, privacy harm) of these models become easier to scale. As a result, safety measures are critical when deploying these models. Moving forward, there are still areas that require further progress. These include efficient generation in fewer diffusion steps, better temporal and 3D consistency, strong conditioning to users' input, watermarking, and responsible usage.

\section*{Acknowledgement}
We would like to thank Sean McGregor for his help with the publication.

\bibliographystyle{IEEEtran}
\bibliography{references}

\begin{thebibliography}{100}
\providecommand{\url}[1]{#1}
\csname url@samestyle\endcsname
\providecommand{\newblock}{\relax}
\providecommand{\bibinfo}[2]{#2}
\providecommand{\BIBentrySTDinterwordspacing}{\spaceskip=0pt\relax}
\providecommand{\BIBentryALTinterwordstretchfactor}{4}
\providecommand{\BIBentryALTinterwordspacing}{\spaceskip=\fontdimen2\font plus
\BIBentryALTinterwordstretchfactor\fontdimen3\font minus \fontdimen4\font\relax}
\providecommand{\BIBforeignlanguage}[2]{{%
\expandafter\ifx\csname l@#1\endcsname\relax
\typeout{** WARNING: IEEEtran.bst: No hyphenation pattern has been}%
\typeout{** loaded for the language `#1'. Using the pattern for}%
\typeout{** the default language instead.}%
\else
\language=\csname l@#1\endcsname
\fi
#2}}
\providecommand{\BIBdecl}{\relax}
\BIBdecl

\bibitem{van2017neural}
A.~Van Den~Oord, O.~Vinyals \emph{et~al.}, ``Neural discrete representation learning,'' \emph{Advances in neural information processing systems}, vol.~30, 2017.

\bibitem{Rumelhart1986LearningRB}
\BIBentryALTinterwordspacing
D.~E. Rumelhart, G.~E. Hinton, and R.~J. Williams, ``Learning representations by back-propagating errors,'' \emph{Nature}, vol. 323, pp. 533--536, 1986. [Online]. Available: \url{https://api.semanticscholar.org/CorpusID:205001834}
\BIBentrySTDinterwordspacing

\bibitem{Vincent2008DAE}
P.~Vincent, H.~Larochelle, Y.~Bengio, and P.-A. Manzagol, ``Extracting and composing robust features with denoising autoencoders,'' in \emph{Proceedings of the 25th International Conference on Machine Learning}.\hskip 1em plus 0.5em minus 0.4em\relax ACM, 2008, pp. 1096--1103.

\bibitem{bengio2013generalized}
Y.~Bengio, L.~Yao, G.~Alain, and P.~Vincent, ``Generalized denoising auto-encoders as generative models,'' \emph{Advances in neural information processing systems}, vol.~26, 2013.

\bibitem{kingma2013auto}
D.~P. Kingma and M.~Welling, ``Auto-encoding variational bayes,'' \emph{arXiv preprint arXiv:1312.6114}, 2013.

\bibitem{rezende2014stochastic}
D.~J. Rezende, S.~Mohamed, and D.~Wierstra, ``Stochastic backpropagation and approximate inference in deep generative models,'' in \emph{International conference on machine learning}.\hskip 1em plus 0.5em minus 0.4em\relax PMLR, 2014, pp. 1278--1286.

\bibitem{higgins2017beta}
I.~Higgins, L.~Matthey, A.~Pal, C.~Burgess, X.~Glorot, M.~Botvinick, S.~Mohamed, and A.~Lerchner, ``beta-vae: Learning basic visual concepts with a constrained variational framework,'' in \emph{International conference on learning representations}, 2017.

\bibitem{burgess2018understanding}
C.~P. Burgess, I.~Higgins, A.~Pal, L.~Matthey, N.~Watters, G.~Desjardins, and A.~Lerchner, ``Understanding disentangling in beta-vae,'' \emph{arXiv preprint arXiv:1804.03599}, 2018.

\bibitem{van2016conditional}
A.~Van~den Oord, N.~Kalchbrenner, L.~Espeholt, O.~Vinyals, A.~Graves \emph{et~al.}, ``Conditional image generation with pixelcnn decoders,'' \emph{Advances in neural information processing systems}, vol.~29, 2016.

\bibitem{daCosta_autoencoder2021}
P.~F. da~Costa, ``Autoregressive models — pixelcnn,'' \url{https://pedroferreiradacosta.github.io/post/auto_encoder/}, 2021, accessed: 2025-09-19.

\bibitem{gulrajani2016pixelvae}
I.~Gulrajani, K.~Kumar, F.~Ahmed, A.~A. Taiga, F.~Visin, D.~Vazquez, and A.~Courville, ``Pixelvae: A latent variable model for natural images,'' \emph{arXiv preprint arXiv:1611.05013}, 2016.

\bibitem{sohn2015learning}
K.~Sohn, H.~Lee, and X.~Yan, ``Learning structured output representation using deep conditional generative models,'' \emph{Advances in neural information processing systems}, vol.~28, 2015.

\bibitem{kingma2014semi}
D.~P. Kingma, D.~J. Rezende, S.~Mohamed, and M.~Welling, ``Semi-supervised learning with deep generative models,'' \emph{Advances in neural information processing systems}, vol.~27, 2014.

\bibitem{burda2015importance}
Y.~Burda, R.~Grosse, and R.~Salakhutdinov, ``Importance weighted autoencoders,'' \emph{arXiv preprint arXiv:1509.00519}, 2015.

\bibitem{gregor2015draw}
K.~Gregor, I.~Danihelka, A.~Graves, D.~Rezende, and D.~Wierstra, ``Draw: A recurrent neural network for image generation,'' in \emph{International conference on machine learning}.\hskip 1em plus 0.5em minus 0.4em\relax PMLR, 2015, pp. 1462--1471.

\bibitem{metzger2021deep}
A.~Metzger, M.~Toscani, A.~Akbarinia, M.~Valsecchi, and K.~Drewing, ``Deep neural network model of haptic saliency,'' \emph{Scientific reports}, vol.~11, no.~1, p. 1395, 2021.

\bibitem{ranganath2016hierarchical}
R.~Ranganath, D.~Tran, and D.~Blei, ``Hierarchical variational models,'' in \emph{International conference on machine learning}.\hskip 1em plus 0.5em minus 0.4em\relax PMLR, 2016, pp. 324--333.

\bibitem{kingma2016improved}
D.~P. Kingma, T.~Salimans, R.~Jozefowicz, X.~Chen, I.~Sutskever, and M.~Welling, ``Improved variational inference with inverse autoregressive flow,'' \emph{Advances in neural information processing systems}, vol.~29, 2016.

\bibitem{sonderby2016ladder}
C.~K. S{\o}nderby, T.~Raiko, L.~Maal{\o}e, S.~K. S{\o}nderby, and O.~Winther, ``Ladder variational autoencoders,'' \emph{Advances in neural information processing systems}, vol.~29, 2016.

\bibitem{klushyn2019learning}
A.~Klushyn, N.~Chen, R.~Kurle, B.~Cseke, and P.~van~der Smagt, ``Learning hierarchical priors in vaes,'' \emph{Advances in neural information processing systems}, vol.~32, 2019.

\bibitem{vahdat2020nvae}
A.~Vahdat and J.~Kautz, ``Nvae: A deep hierarchical variational autoencoder,'' \emph{Advances in neural information processing systems}, vol.~33, pp. 19\,667--19\,679, 2020.

\bibitem{liu2018large}
Z.~Liu, P.~Luo, X.~Wang, and X.~Tang, ``Large-scale celebfaces attributes (celeba) dataset,'' \emph{Retrieved August}, vol.~15, no. 2018, p.~11, 2018.

\bibitem{child2011very}
R.~Child, ``Very deep vaes generalize autoregressive models and can outperform them on images (2020),'' \emph{arXiv preprint arXiv:2011.10650}, 2011.

\bibitem{goodfellow2014generative}
I.~J. Goodfellow, J.~Pouget-Abadie, M.~Mirza, B.~Xu, D.~Warde-Farley, S.~Ozair, A.~Courville, and Y.~Bengio, ``Generative adversarial nets,'' \emph{Advances in neural information processing systems}, vol.~27, 2014.

\bibitem{radford2015unsupervised}
A.~Radford, L.~Metz, and S.~Chintala, ``Unsupervised representation learning with deep convolutional generative adversarial networks,'' \emph{arXiv preprint arXiv:1511.06434}, 2015.

\bibitem{lsundataset}
\BIBentryALTinterwordspacing
F.~Yu, Y.~Zhang, S.~Song, A.~Seff, and J.~Xiao, ``Lsun: Construction of a large-scale image dataset using deep learning with humans in the loop.'' \emph{CoRR}, vol. abs/1506.03365, 2015. [Online]. Available: \url{http://dblp.uni-trier.de/db/journals/corr/corr1506.html#YuZSSX15}
\BIBentrySTDinterwordspacing

\bibitem{salimans2016improved}
T.~Salimans, I.~Goodfellow, W.~Zaremba, V.~Cheung, A.~Radford, and X.~Chen, ``Improved techniques for training gans,'' \emph{Advances in neural information processing systems}, vol.~29, 2016.

\bibitem{szegedy2015going}
C.~Szegedy, W.~Liu, Y.~Jia, P.~Sermanet, S.~Reed, D.~Anguelov, D.~Erhan, V.~Vanhoucke, and A.~Rabinovich, ``Going deeper with convolutions,'' in \emph{Proceedings of the IEEE conference on computer vision and pattern recognition}, 2015, pp. 1--9.

\bibitem{arjovsky2017towards}
M.~Arjovsky and L.~Bottou, ``Towards principled methods for training generative adversarial networks,'' \emph{arXiv preprint arXiv:1701.04862}, 2017.

\bibitem{arjovsky2017wassersteingan}
\BIBentryALTinterwordspacing
M.~Arjovsky, S.~Chintala, and L.~Bottou, ``Wasserstein gan,'' 2017. [Online]. Available: \url{https://arxiv.org/abs/1701.07875}
\BIBentrySTDinterwordspacing

\bibitem{gulrajani2017improved}
I.~Gulrajani, F.~Ahmed, M.~Arjovsky, V.~Dumoulin, and A.~C. Courville, ``Improved training of wasserstein gans,'' \emph{Advances in neural information processing systems}, vol.~30, 2017.

\bibitem{mescheder2018training}
L.~Mescheder, A.~Geiger, and S.~Nowozin, ``Which training methods for gans do actually converge?'' in \emph{International conference on machine learning}.\hskip 1em plus 0.5em minus 0.4em\relax PMLR, 2018, pp. 3481--3490.

\bibitem{sonderby2016amortised}
C.~K. S{\o}nderby, J.~Caballero, L.~Theis, W.~Shi, and F.~Husz{\'a}r, ``Amortised map inference for image super-resolution,'' \emph{arXiv preprint arXiv:1610.04490}, 2016.

\bibitem{roth2017stabilizing}
K.~Roth, A.~Lucchi, S.~Nowozin, and T.~Hofmann, ``Stabilizing training of generative adversarial networks through regularization,'' \emph{Advances in neural information processing systems}, vol.~30, 2017.

\bibitem{mirza2014conditional}
M.~Mirza and S.~Osindero, ``Conditional generative adversarial nets,'' \emph{arXiv preprint arXiv:1411.1784}, 2014.

\bibitem{odena2017conditional}
A.~Odena, C.~Olah, and J.~Shlens, ``Conditional image synthesis with auxiliary classifier gans,'' in \emph{International conference on machine learning}.\hskip 1em plus 0.5em minus 0.4em\relax PMLR, 2017, pp. 2642--2651.

\bibitem{miyato2018cgans}
T.~Miyato and M.~Koyama, ``cgans with projection discriminator,'' \emph{arXiv preprint arXiv:1802.05637}, 2018.

\bibitem{karras2017progressive}
T.~Karras, T.~Aila, S.~Laine, and J.~Lehtinen, ``Progressive growing of gans for improved quality, stability, and variation,'' \emph{arXiv preprint arXiv:1710.10196}, 2017.

\bibitem{karras2019style}
T.~Karras, S.~Laine, and T.~Aila, ``A style-based generator architecture for generative adversarial networks,'' in \emph{Proceedings of the IEEE/CVF conference on computer vision and pattern recognition}, 2019, pp. 4401--4410.

\bibitem{karras2020analyzing}
T.~Karras, S.~Laine, M.~Aittala, J.~Hellsten, J.~Lehtinen, and T.~Aila, ``Analyzing and improving the image quality of stylegan,'' in \emph{Proceedings of the IEEE/CVF conference on computer vision and pattern recognition}, 2020, pp. 8110--8119.

\bibitem{karras2021alias}
T.~Karras, M.~Aittala, S.~Laine, E.~Harkonen, J.~Hellsten, J.~Lehtinen, and T.~Aila, ``Alias free generative adversarial networks,'' \emph{Advances in neural information processing systems}, vol.~34, pp. 852--863, 2021.

\bibitem{heusel2017gans}
M.~Heusel, H.~Ramsauer, T.~Unterthiner, B.~Nessler, and S.~Hochreiter, ``Gans trained by a two time-scale update rule converge to a local nash equilibrium,'' \emph{Advances in neural information processing systems}, vol.~30, 2017.

\bibitem{zhang2019making}
R.~Zhang, ``Making convolutional networks shift-invariant again,'' in \emph{International conference on machine learning}.\hskip 1em plus 0.5em minus 0.4em\relax PMLR, 2019, pp. 7324--7334.

\bibitem{zhang2017stackgan}
H.~Zhang, T.~Xu, H.~Li, S.~Zhang, X.~Wang, X.~Huang, and D.~N. Metaxas, ``Stackgan: Text to photo-realistic image synthesis with stacked generative adversarial networks,'' in \emph{Proceedings of the IEEE international conference on computer vision}, 2017, pp. 5907--5915.

\bibitem{reed2016learning}
S.~E. Reed, Z.~Akata, S.~Mohan, S.~Tenka, B.~Schiele, and H.~Lee, ``Learning what and where to draw,'' \emph{Advances in neural information processing systems}, vol.~29, 2016.

\bibitem{reed2016generative}
S.~Reed, Z.~Akata, X.~Yan, L.~Logeswaran, B.~Schiele, and H.~Lee, ``Generative adversarial text to image synthesis,'' in \emph{International conference on machine learning}.\hskip 1em plus 0.5em minus 0.4em\relax Pmlr, 2016, pp. 1060--1069.

\bibitem{xu2017attnganfinegrainedtextimage}
\BIBentryALTinterwordspacing
T.~Xu, P.~Zhang, Q.~Huang, H.~Zhang, Z.~Gan, X.~Huang, and X.~He, ``Attngan: Fine-grained text to image generation with attentional generative adversarial networks,'' 2017. [Online]. Available: \url{https://arxiv.org/abs/1711.10485}
\BIBentrySTDinterwordspacing

\bibitem{ledig2017photo}
C.~Ledig, L.~Theis, F.~Husz{\'a}r, J.~Caballero, A.~Cunningham, A.~Acosta, A.~Aitken, A.~Tejani, J.~Totz, Z.~Wang \emph{et~al.}, ``Photo-realistic single image super-resolution using a generative adversarial network,'' in \emph{Proceedings of the IEEE conference on computer vision and pattern recognition}, 2017, pp. 4681--4690.

\bibitem{simonyan2014very}
K.~Simonyan and A.~Zisserman, ``Very deep convolutional networks for large-scale image recognition,'' \emph{arXiv preprint arXiv:1409.1556}, 2014.

\bibitem{agnelli2010clustering}
J.~P. Agnelli, M.~Cadeiras, E.~G. Tabak, C.~V. Turner, and E.~Vanden-Eijnden, ``Clustering and classification through normalizing flows in feature space,'' \emph{Multiscale Modeling \& Simulation}, vol.~8, no.~5, pp. 1784--1802, 2010.

\bibitem{papamakarios2021normalizing}
G.~Papamakarios, E.~Nalisnick, D.~J. Rezende, S.~Mohamed, and B.~Lakshminarayanan, ``Normalizing flows for probabilistic modeling and inference,'' \emph{Journal of Machine Learning Research}, vol.~22, no.~57, pp. 1--64, 2021.

\bibitem{dinh2014nice}
L.~Dinh, D.~Krueger, and Y.~Bengio, ``Nice: Non-linear independent components estimation,'' \emph{arXiv preprint arXiv:1410.8516}, 2014.

\bibitem{dinh2016density}
L.~Dinh, J.~Sohl-Dickstein, and S.~Bengio, ``Density estimation using real nvp,'' \emph{arXiv preprint arXiv:1605.08803}, 2016.

\bibitem{weng2018flow}
\BIBentryALTinterwordspacing
L.~Weng, ``Flow-based deep generative models,'' \emph{lilianweng.github.io}, 2018. [Online]. Available: \url{https://lilianweng.github.io/posts/2018-10-13-flow-models/}
\BIBentrySTDinterwordspacing

\bibitem{rezende2015variational}
D.~Rezende and S.~Mohamed, ``Variational inference with normalizing flows,'' in \emph{International conference on machine learning}.\hskip 1em plus 0.5em minus 0.4em\relax PMLR, 2015, pp. 1530--1538.

\bibitem{papamakarios2017masked}
G.~Papamakarios, T.~Pavlakou, and I.~Murray, ``Masked autoregressive flow for density estimation,'' \emph{Advances in neural information processing systems}, vol.~30, 2017.

\bibitem{germain2015made}
M.~Germain, K.~Gregor, I.~Murray, and H.~Larochelle, ``Made: Masked autoencoder for distribution estimation,'' in \emph{International conference on machine learning}.\hskip 1em plus 0.5em minus 0.4em\relax PMLR, 2015, pp. 881--889.

\bibitem{huang2018neural}
C.-W. Huang, D.~Krueger, A.~Lacoste, and A.~Courville, ``Neural autoregressive flows,'' in \emph{International conference on machine learning}.\hskip 1em plus 0.5em minus 0.4em\relax PMLR, 2018, pp. 2078--2087.

\bibitem{kingma2018glow}
D.~P. Kingma and P.~Dhariwal, ``Glow: Generative flow with invertible 1x1 convolutions,'' \emph{Advances in neural information processing systems}, vol.~31, 2018.

\bibitem{ho2019flow++}
J.~Ho, X.~Chen, A.~Srinivas, Y.~Duan, and P.~Abbeel, ``Flow++: Improving flow-based generative models with variational dequantization and architecture design,'' in \emph{International conference on machine learning}.\hskip 1em plus 0.5em minus 0.4em\relax PMLR, 2019, pp. 2722--2730.

\bibitem{durkan2019neural}
C.~Durkan, A.~Bekasov, I.~Murray, and G.~Papamakarios, ``Neural spline flows,'' \emph{Advances in neural information processing systems}, vol.~32, 2019.

\bibitem{grathwohl2018ffjord}
W.~Grathwohl, R.~T. Chen, J.~Bettencourt, I.~Sutskever, and D.~Duvenaud, ``Ffjord: Free-form continuous dynamics for scalable reversible generative models,'' \emph{arXiv preprint arXiv:1810.01367}, 2018.

\bibitem{lugmayr2020srflow}
A.~Lugmayr, M.~Danelljan, L.~Van~Gool, and R.~Timofte, ``Srflow: Learning the super-resolution space with normalizing flow,'' in \emph{European conference on computer vision}.\hskip 1em plus 0.5em minus 0.4em\relax Springer, 2020, pp. 715--732.

\bibitem{zhai2024normalizing}
S.~Zhai, R.~Zhang, P.~Nakkiran, D.~Berthelot, J.~Gu, H.~Zheng, T.~Chen, M.~A. Bautista, N.~Jaitly, and J.~Susskind, ``Normalizing flows are capable generative models,'' \emph{arXiv preprint arXiv:2412.06329}, 2024.

\bibitem{gu2025starflow}
J.~Gu, T.~Chen, D.~Berthelot, H.~Zheng, Y.~Wang, R.~Zhang, L.~Dinh, M.~A. Bautista, J.~Susskind, and S.~Zhai, ``Starflow: Scaling latent normalizing flows for high-resolution image synthesis,'' \emph{arXiv preprint arXiv:2506.06276}, 2025.

\bibitem{vaswani2017attention}
A.~Vaswani, N.~Shazeer, N.~Parmar, J.~Uszkoreit, L.~Jones, A.~N. Gomez, {\L}.~Kaiser, and I.~Polosukhin, ``Attention is all you need,'' \emph{Advances in neural information processing systems}, vol.~30, 2017.

\bibitem{van2016pixel}
A.~Van Den~Oord, N.~Kalchbrenner, and K.~Kavukcuoglu, ``Pixel recurrent neural networks,'' in \emph{International conference on machine learning}.\hskip 1em plus 0.5em minus 0.4em\relax PMLR, 2016, pp. 1747--1756.

\bibitem{salimans2017pixelcnn++}
T.~Salimans, A.~Karpathy, X.~Chen, and D.~P. Kingma, ``Pixelcnn++: Improving the pixelcnn with discretized logistic mixture likelihood and other modifications,'' \emph{arXiv preprint arXiv:1701.05517}, 2017.

\bibitem{chen2018pixelsnail}
X.~Chen, N.~Mishra, M.~Rohaninejad, and P.~Abbeel, ``Pixelsnail: An improved autoregressive generative model,'' in \emph{International conference on machine learning}.\hskip 1em plus 0.5em minus 0.4em\relax PMLR, 2018, pp. 864--872.

\bibitem{parmar2018image}
N.~Parmar, A.~Vaswani, J.~Uszkoreit, L.~Kaiser, N.~Shazeer, A.~Ku, and D.~Tran, ``Image transformer,'' in \emph{International conference on machine learning}.\hskip 1em plus 0.5em minus 0.4em\relax PMLR, 2018, pp. 4055--4064.

\bibitem{chen2020generative}
M.~Chen, A.~Radford, R.~Child, J.~Wu, H.~Jun, D.~Luan, and I.~Sutskever, ``Generative pretraining from pixels,'' in \emph{International conference on machine learning}.\hskip 1em plus 0.5em minus 0.4em\relax PMLR, 2020, pp. 1691--1703.

\bibitem{ramesh2021zero}
A.~Ramesh, M.~Pavlov, G.~Goh, S.~Gray, C.~Voss, A.~Radford, M.~Chen, and I.~Sutskever, ``Zero-shot text-to-image generation,'' in \emph{International conference on machine learning}.\hskip 1em plus 0.5em minus 0.4em\relax Pmlr, 2021, pp. 8821--8831.

\bibitem{WahCUB_200_2011}
C.~Wah, S.~Branson, P.~Welinder, P.~Perona, and S.~Belongie, ``Caltech-ucsd birds-200-2011,'' California Institute of Technology, Tech. Rep. CNS-TR-2011-001, 2011.

\bibitem{esser2021taming}
P.~Esser, R.~Rombach, and B.~Ommer, ``Taming transformers for high-resolution image synthesis,'' in \emph{Proceedings of the IEEE/CVF conference on computer vision and pattern recognition}, 2021, pp. 12\,873--12\,883.

\bibitem{liu2016deepfashion}
Z.~Liu, P.~Luo, S.~Qiu, X.~Wang, and X.~Tang, ``Deepfashion: Powering robust clothes recognition and retrieval with rich annotations,'' in \emph{Proceedings of the IEEE conference on computer vision and pattern recognition}, 2016, pp. 1096--1104.

\bibitem{ding2021cogview}
M.~Ding, Z.~Yang, W.~Hong, W.~Zheng, C.~Zhou, D.~Yin, J.~Lin, X.~Zou, Z.~Shao, H.~Yang \emph{et~al.}, ``Cogview: Mastering text-to-image generation via transformers,'' \emph{Advances in neural information processing systems}, vol.~34, pp. 19\,822--19\,835, 2021.

\bibitem{chang2022maskgit}
H.~Chang, H.~Zhang, L.~Jiang, C.~Liu, and W.~T. Freeman, ``Maskgit: Masked generative image transformer,'' in \emph{Proceedings of the IEEE/CVF conference on computer vision and pattern recognition}, 2022, pp. 11\,315--11\,325.

\bibitem{chang2023muse}
H.~Chang, H.~Zhang, J.~Barber, A.~Maschinot, J.~Lezama, L.~Jiang, M.-H. Yang, K.~Murphy, W.~T. Freeman, M.~Rubinstein \emph{et~al.}, ``Muse: Text-to-image generation via masked generative transformers,'' \emph{arXiv preprint arXiv:2301.00704}, 2023.

\bibitem{raffel2020exploring}
C.~Raffel, N.~Shazeer, A.~Roberts, K.~Lee, S.~Narang, M.~Matena, Y.~Zhou, W.~Li, and P.~J. Liu, ``Exploring the limits of transfer learning with a unified text-to-text transformer,'' \emph{Journal of machine learning research}, vol.~21, no. 140, pp. 1--67, 2020.

\bibitem{yu2022scaling}
J.~Yu, Y.~Xu, J.~Y. Koh, T.~Luong, G.~Baid, Z.~Wang, V.~Vasudevan, A.~Ku, Y.~Yang, B.~K. Ayan \emph{et~al.}, ``Scaling autoregressive models for content-rich text-to-image generation,'' \emph{arXiv preprint arXiv:2206.10789}, vol.~2, no.~3, p.~5, 2022.

\bibitem{yu2021vector}
J.~Yu, X.~Li, J.~Y. Koh, H.~Zhang, R.~Pang, J.~Qin, A.~Ku, Y.~Xu, J.~Baldridge, and Y.~Wu, ``Vector-quantized image modeling with improved vqgan,'' \emph{arXiv preprint arXiv:2110.04627}, 2021.

\bibitem{sohl2015deep}
J.~Sohl-Dickstein, E.~Weiss, N.~Maheswaranathan, and S.~Ganguli, ``Deep unsupervised learning using nonequilibrium thermodynamics,'' in \emph{International conference on machine learning}.\hskip 1em plus 0.5em minus 0.4em\relax pmlr, 2015, pp. 2256--2265.

\bibitem{ho2020denoising}
J.~Ho, A.~Jain, and P.~Abbeel, ``Denoising diffusion probabilistic models,'' \emph{Advances in neural information processing systems}, vol.~33, pp. 6840--6851, 2020.

\bibitem{weng2021diffusion}
\BIBentryALTinterwordspacing
L.~Weng, ``What are diffusion models?'' \emph{lilianweng.github.io}, Jul 2021. [Online]. Available: \url{https://lilianweng.github.io/posts/2021-07-11-diffusion-models/}
\BIBentrySTDinterwordspacing

\bibitem{ronneberger2015u}
O.~Ronneberger, P.~Fischer, and T.~Brox, ``U-net: Convolutional networks for biomedical image segmentation,'' in \emph{International Conference on Medical image computing and computer-assisted intervention}.\hskip 1em plus 0.5em minus 0.4em\relax Springer, 2015, pp. 234--241.

\bibitem{song2020denoising}
J.~Song, C.~Meng, and S.~Ermon, ``Denoising diffusion implicit models,'' \emph{arXiv preprint arXiv:2010.02502}, 2020.

\bibitem{nichol2021improved}
A.~Q. Nichol and P.~Dhariwal, ``Improved denoising diffusion probabilistic models,'' in \emph{International conference on machine learning}.\hskip 1em plus 0.5em minus 0.4em\relax PMLR, 2021, pp. 8162--8171.

\bibitem{song2020score}
Y.~Song, J.~Sohl-Dickstein, D.~P. Kingma, A.~Kumar, S.~Ermon, and B.~Poole, ``Score-based generative modeling through stochastic differential equations,'' \emph{arXiv preprint arXiv:2011.13456}, 2020.

\bibitem{child2019generating}
R.~Child, S.~Gray, A.~Radford, and I.~Sutskever, ``Generating long sequences with sparse transformers,'' \emph{arXiv preprint arXiv:1904.10509}, 2019.

\bibitem{roy2021efficient}
A.~Roy, M.~Saffar, A.~Vaswani, and D.~Grangier, ``Efficient content-based sparse attention with routing transformers,'' \emph{Transactions of the Association for Computational Linguistics}, vol.~9, pp. 53--68, 2021.

\bibitem{salimans2022progressive}
T.~Salimans and J.~Ho, ``Progressive distillation for fast sampling of diffusion models,'' \emph{arXiv preprint arXiv:2202.00512}, 2022.

\bibitem{yin2024one}
T.~Yin, M.~Gharbi, R.~Zhang, E.~Shechtman, F.~Durand, W.~T. Freeman, and T.~Park, ``One-step diffusion with distribution matching distillation,'' in \emph{Proceedings of the IEEE/CVF conference on computer vision and pattern recognition}, 2024, pp. 6613--6623.

\bibitem{song2023consistency}
Y.~Song, P.~Dhariwal, M.~Chen, and I.~Sutskever, ``Consistency models,'' 2023.

\bibitem{zheng2023fast}
H.~Zheng, W.~Nie, A.~Vahdat, K.~Azizzadenesheli, and A.~Anandkumar, ``Fast sampling of diffusion models via operator learning,'' in \emph{International conference on machine learning}.\hskip 1em plus 0.5em minus 0.4em\relax PMLR, 2023, pp. 42\,390--42\,402.

\bibitem{dhariwal2021diffusion}
P.~Dhariwal and A.~Nichol, ``Diffusion models beat gans on image synthesis,'' \emph{Advances in neural information processing systems}, vol.~34, pp. 8780--8794, 2021.

\bibitem{brock2018large}
A.~Brock, J.~Donahue, and K.~Simonyan, ``Large scale gan training for high fidelity natural image synthesis,'' \emph{arXiv preprint arXiv:1809.11096}, 2018.

\bibitem{ho2022classifier}
J.~Ho and T.~Salimans, ``Classifier-free diffusion guidance,'' \emph{arXiv preprint arXiv:2207.12598}, 2022.

\bibitem{nichol2021glide}
A.~Nichol, P.~Dhariwal, A.~Ramesh, P.~Shyam, P.~Mishkin, B.~McGrew, I.~Sutskever, and M.~Chen, ``Glide: Towards photorealistic image generation and editing with text-guided diffusion models,'' \emph{arXiv preprint arXiv:2112.10741}, 2021.

\bibitem{radford2021learning}
A.~Radford, J.~W. Kim, C.~Hallacy, A.~Ramesh, G.~Goh, S.~Agarwal, G.~Sastry, A.~Askell, P.~Mishkin, J.~Clark \emph{et~al.}, ``Learning transferable visual models from natural language supervision,'' in \emph{International conference on machine learning}.\hskip 1em plus 0.5em minus 0.4em\relax PmLR, 2021, pp. 8748--8763.

\bibitem{karras2022elucidating}
T.~Karras, M.~Aittala, T.~Aila, and S.~Laine, ``Elucidating the design space of diffusion-based generative models,'' \emph{Advances in neural information processing systems}, vol.~35, pp. 26\,565--26\,577, 2022.

\bibitem{rombach2022high}
R.~Rombach, A.~Blattmann, D.~Lorenz, P.~Esser, and B.~Ommer, ``High-resolution image synthesis with latent diffusion models,'' in \emph{Proceedings of the IEEE/CVF conference on computer vision and pattern recognition}, 2022, pp. 10\,684--10\,695.

\bibitem{peebles2023scalable}
W.~Peebles and S.~Xie, ``Scalable diffusion models with transformers,'' in \emph{Proceedings of the IEEE/CVF international conference on computer vision}, 2023, pp. 4195--4205.

\bibitem{saharia2022palette}
C.~Saharia, W.~Chan, H.~Chang, C.~Lee, J.~Ho, T.~Salimans, D.~Fleet, and M.~Norouzi, ``Palette: Image-to-image diffusion models,'' in \emph{ACM SIGGRAPH 2022 conference proceedings}, 2022, pp. 1--10.

\bibitem{teterwak2019boundless}
P.~Teterwak, A.~Sarna, D.~Krishnan, A.~Maschinot, D.~Belanger, C.~Liu, and W.~T. Freeman, ``Boundless: Generative adversarial networks for image extension,'' in \emph{Proceedings of the IEEE/CVF international conference on computer vision}, 2019, pp. 10\,521--10\,530.

\bibitem{lin2021infinitygan}
C.~H. Lin, H.-Y. Lee, Y.-C. Cheng, S.~Tulyakov, and M.-H. Yang, ``Infinitygan: Towards infinite-pixel image synthesis,'' \emph{arXiv preprint arXiv:2104.03963}, 2021.

\bibitem{zhou2017places}
B.~Zhou, A.~Lapedriza, A.~Khosla, A.~Oliva, and A.~Torralba, ``Places: A 10 million image database for scene recognition,'' \emph{IEEE Transactions on Pattern Analysis and Machine Intelligence}, 2017.

\bibitem{ramesh2022hierarchical}
A.~Ramesh, P.~Dhariwal, A.~Nichol, C.~Chu, and M.~Chen, ``Hierarchical text-conditional image generation with clip latents,'' \emph{arXiv preprint arXiv:2204.06125}, vol.~1, no.~2, p.~3, 2022.

\bibitem{saharia2022photorealistic}
C.~Saharia, W.~Chan, S.~Saxena, L.~Li, J.~Whang, E.~L. Denton, K.~Ghasemipour, R.~Gontijo~Lopes, B.~Karagol~Ayan, T.~Salimans \emph{et~al.}, ``Photorealistic text-to-image diffusion models with deep language understanding,'' \emph{Advances in neural information processing systems}, vol.~35, pp. 36\,479--36\,494, 2022.

\bibitem{betker2023improving}
J.~Betker, G.~Goh, L.~Jing, T.~Brooks, J.~Wang, L.~Li, L.~Ouyang, J.~Zhuang, J.~Lee, Y.~Guo \emph{et~al.}, ``Improving image generation with better captions,'' \emph{Computer Science. https://cdn. openai. com/papers/dall-e-3. pdf}, vol.~2, no.~3, p.~8, 2023.

\bibitem{podell2023sdxl}
D.~Podell, Z.~English, K.~Lacey, A.~Blattmann, T.~Dockhorn, J.~M{\"u}ller, J.~Penna, and R.~Rombach, ``Sdxl: Improving latent diffusion models for high-resolution image synthesis,'' \emph{arXiv preprint arXiv:2307.01952}, 2023.

\bibitem{chen2018neural}
R.~T. Chen, Y.~Rubanova, J.~Bettencourt, and D.~K. Duvenaud, ``Neural ordinary differential equations,'' \emph{Advances in neural information processing systems}, vol.~31, 2018.

\bibitem{liu2022flow}
X.~Liu, C.~Gong, and Q.~Liu, ``Flow straight and fast: Learning to generate and transfer data with rectified flow,'' \emph{arXiv preprint arXiv:2209.03003}, 2022.

\bibitem{esser2024scaling}
P.~Esser, S.~Kulal, A.~Blattmann, R.~Entezari, J.~M{\"u}ller, H.~Saini, Y.~Levi, D.~Lorenz, A.~Sauer, F.~Boesel \emph{et~al.}, ``Scaling rectified flow transformers for high-resolution image synthesis,'' in \emph{Forty-first international conference on machine learning}, 2024.

\bibitem{ghosh2023geneval}
D.~Ghosh, H.~Hajishirzi, and L.~Schmidt, ``Geneval: An object-focused framework for evaluating text-to-image alignment,'' \emph{Advances in Neural Information Processing Systems}, vol.~36, pp. 52\,132--52\,152, 2023.

\bibitem{lee2024improving}
S.~Lee, Z.~Lin, and G.~Fanti, ``Improving the training of rectified flows,'' \emph{Advances in neural information processing systems}, vol.~37, pp. 63\,082--63\,109, 2024.

\bibitem{hu2025improving}
X.~Hu, R.~Liao, K.~Xu, B.~Liu, Y.~Li, E.~Ie, H.~Fei, and Q.~Liu, ``Improving rectified flow with boundary conditions,'' \emph{arXiv preprint arXiv:2506.15864}, 2025.

\bibitem{lipman2022flow}
Y.~Lipman, R.~T. Chen, H.~Ben-Hamu, M.~Nickel, and M.~Le, ``Flow matching for generative modeling,'' \emph{arXiv preprint arXiv:2210.02747}, 2022.

\bibitem{kornilov2024optimal}
N.~Kornilov, P.~Mokrov, A.~Gasnikov, and A.~Korotin, ``Optimal flow matching: Learning straight trajectories in just one step,'' \emph{Advances in Neural Information Processing Systems}, vol.~37, pp. 104\,180--104\,204, 2024.

\bibitem{vondrick2016generating}
C.~Vondrick, H.~Pirsiavash, and A.~Torralba, ``Generating videos with scene dynamics,'' \emph{Advances in neural information processing systems}, vol.~29, 2016.

\bibitem{saito2017temporal}
M.~Saito, E.~Matsumoto, and S.~Saito, ``Temporal generative adversarial nets with singular value clipping,'' in \emph{Proceedings of the IEEE international conference on computer vision}, 2017, pp. 2830--2839.

\bibitem{tulyakov2018mocogan}
S.~Tulyakov, M.-Y. Liu, X.~Yang, and J.~Kautz, ``Mocogan: Decomposing motion and content for video generation,'' in \emph{Proceedings of the IEEE conference on computer vision and pattern recognition}, 2018, pp. 1526--1535.

\bibitem{aifanti2010mug}
N.~Aifanti, C.~Papachristou, and A.~Delopoulos, ``The mug facial expression database,'' in \emph{11th International Workshop on Image Analysis for Multimedia Interactive Services WIAMIS 10}.\hskip 1em plus 0.5em minus 0.4em\relax IEEE, 2010, pp. 1--4.

\bibitem{cao2017realtime}
Z.~Cao, T.~Simon, S.-E. Wei, and Y.~Sheikh, ``Realtime multi-person 2d pose estimation using part affinity fields,'' in \emph{Proceedings of the IEEE conference on computer vision and pattern recognition}, 2017, pp. 7291--7299.

\bibitem{yan2021videogpt}
W.~Yan, Y.~Zhang, P.~Abbeel, and A.~Srinivas, ``Videogpt: Video generation using vq-vae and transformers,'' \emph{arXiv preprint arXiv:2104.10157}, 2021.

\bibitem{soomro2012ucf101}
K.~Soomro, A.~R. Zamir, and M.~Shah, ``Ucf101: A dataset of 101 human actions classes from videos in the wild,'' \emph{arXiv preprint arXiv:1212.0402}, 2012.

\bibitem{ho2022video}
J.~Ho, T.~Salimans, A.~Gritsenko, W.~Chan, M.~Norouzi, and D.~J. Fleet, ``Video diffusion models,'' \emph{Advances in neural information processing systems}, vol.~35, pp. 8633--8646, 2022.

\bibitem{kahembwe2020lower}
E.~Kahembwe and S.~Ramamoorthy, ``Lower dimensional kernels for video discriminators,'' \emph{Neural Networks}, vol. 132, pp. 506--520, 2020.

\bibitem{gordon2021latent}
C.~Gordon and N.~Parde, ``Latent neural differential equations for video generation,'' in \emph{NeurIPS 2020 Workshop on Pre-registration in Machine Learning}.\hskip 1em plus 0.5em minus 0.4em\relax PMLR, 2021, pp. 73--86.

\bibitem{saito2020train}
M.~Saito, S.~Saito, M.~Koyama, and S.~Kobayashi, ``Train sparsely, generate densely: Memory-efficient unsupervised training of high-resolution temporal gan,'' \emph{International Journal of Computer Vision}, vol. 128, no.~10, pp. 2586--2606, 2020.

\bibitem{clark2019adversarial}
A.~Clark, J.~Donahue, and K.~Simonyan, ``Adversarial video generation on complex datasets,'' \emph{arXiv preprint arXiv:1907.06571}, 2019.

\bibitem{blattmann2023stable}
A.~Blattmann, T.~Dockhorn, S.~Kulal, D.~Mendelevitch, M.~Kilian, D.~Lorenz, Y.~Levi, Z.~English, V.~Voleti, A.~Letts \emph{et~al.}, ``Stable video diffusion: Scaling latent video diffusion models to large datasets,'' \emph{arXiv preprint arXiv:2311.15127}, 2023.

\bibitem{singer2022make}
U.~Singer, A.~Polyak, T.~Hayes, X.~Yin, J.~An, S.~Zhang, Q.~Hu, H.~Yang, O.~Ashual, O.~Gafni \emph{et~al.}, ``Make-a-video: Text-to-video generation without text-video data,'' \emph{arXiv preprint arXiv:2209.14792}, 2022.

\bibitem{ho2022imagen}
J.~Ho, W.~Chan, C.~Saharia, J.~Whang, R.~Gao, A.~Gritsenko, D.~P. Kingma, B.~Poole, M.~Norouzi, D.~J. Fleet \emph{et~al.}, ``Imagen video: High definition video generation with diffusion models,'' \emph{arXiv preprint arXiv:2210.02303}, 2022.

\bibitem{villegas2022phenaki}
R.~Villegas, M.~Babaeizadeh, P.-J. Kindermans, H.~Moraldo, H.~Zhang, M.~T. Saffar, S.~Castro, J.~Kunze, and D.~Erhan, ``Phenaki: Variable length video generation from open domain textual description,'' \emph{arXiv preprint arXiv:2210.02399}, 2022.

\bibitem{bar2024lumiere}
O.~Bar-Tal, H.~Chefer, O.~Tov, C.~Herrmann, R.~Paiss, S.~Zada, A.~Ephrat, J.~Hur, G.~Liu, A.~Raj \emph{et~al.}, ``Lumiere: A space-time diffusion model for video generation,'' in \emph{SIGGRAPH Asia 2024 Conference Papers}, 2024, pp. 1--11.

\bibitem{Goss2024DeepfakesSocietalImpacts}
T.~Goss, ``Deepfakes and societal impacts,'' February 2024, capstone paper, Bellevue University, MBPC680-T301 Business and Professional Communication.

\bibitem{Priti2025DeepfakesDevelopingSocieties}
R.~N. Priti, M.~A. Khan, A.~Rahman, and A.~T. Wasi, ``Deepfakes in developing societies: Handling the societal impacts and cross-disciplinary vulnerabilities in tech-limited environments,'' in \emph{Proceedings of the AAAI Conference on Artificial Intelligence}.\hskip 1em plus 0.5em minus 0.4em\relax Association for the Advancement of Artificial Intelligence, 2025.

\bibitem{Hameleers2024CheapDeep}
M.~Hameleers, ``Cheap versus deep manipulation: The effects of cheapfakes versus deepfakes in a political setting,'' \emph{International Journal of Public Opinion Research}, vol.~36, pp. 1--9, 2024.

\bibitem{brundage2018malicious}
M.~Brundage, S.~Avin, J.~Clark, H.~Toner, P.~Eckersley, B.~Garfinkel, A.~Dafoe, P.~Scharre, T.~Zeitzoff, B.~Filar \emph{et~al.}, ``The malicious use of artificial intelligence: Forecasting, prevention, and mitigation,'' \emph{arXiv preprint arXiv:1802.07228}, 2018.

\bibitem{Chesney2019DeepFakes}
R.~Chesney and D.~Citron, ``Deep fakes: A looming challenge for privacy, democracy, and national security,'' \emph{California Law Review}, vol. 107, pp. 1753--1820, 2019.

\bibitem{Paris2019DeepfakesCheapFakes}
B.~Paris and J.~Donovan, ``Deepfakes and cheap fakes: The manipulation of audio and visual evidence,'' Data \& Society Research Institute, New York, NY, Tech. Rep., 2019.

\bibitem{Bender2021StochasticParrots}
E.~M. Bender, T.~Gebru, A.~McMillan-Major, and M.~Mitchell, ``On the dangers of stochastic parrots: Can language models be too big?'' in \emph{Proceedings of the 2021 ACM Conference on Fairness, Accountability, and Transparency}, ser. FAccT '21.\hskip 1em plus 0.5em minus 0.4em\relax New York, NY, USA: Association for Computing Machinery, 2021, pp. 610--623.

\bibitem{popescu2004exposing}
A.~C. Popescu and H.~Farid, ``Exposing digital forgeries by detecting duplicated image regions,'' 2004.

\bibitem{Farid2009ImageForgeryDetection}
H.~Farid, ``Image forgery detection: A survey,'' \emph{IEEE Signal Processing Magazine}, pp. 16--25, Mar. 2009.

\bibitem{lukas2006digital}
J.~Lukas, J.~Fridrich, and M.~Goljan, ``Digital camera identification from sensor pattern noise,'' \emph{IEEE Transactions on Information Forensics and Security}, vol.~1, no.~2, pp. 205--214, 2006.

\bibitem{thies2016face2face}
J.~Thies, M.~Zollhofer, M.~Stamminger, C.~Theobalt, and M.~Nie{\ss}ner, ``Face2face: Real-time face capture and reenactment of rgb videos,'' in \emph{Proceedings of the IEEE conference on computer vision and pattern recognition}, 2016, pp. 2387--2395.

\bibitem{kim2018deep}
H.~Kim, P.~Garrido, A.~Tewari, W.~Xu, J.~Thies, M.~Niessner, P.~P{\'e}rez, C.~Richardt, M.~Zollh{\"o}fer, and C.~Theobalt, ``Deep video portraits,'' \emph{ACM transactions on graphics (TOG)}, vol.~37, no.~4, pp. 1--14, 2018.

\bibitem{Suwajanakorn2017Obama}
S.~Suwajanakorn, S.~M. Seitz, and I.~Kemelmacher-Shlizerman, ``Synthesizing obama: Learning lip sync from audio,'' \emph{ACM Transactions on Graphics}, vol.~36, no.~4, Jul. 2017.

\bibitem{li2018ictu}
Y.~Li, M.-C. Chang, and S.~Lyu, ``In ictu oculi: Exposing ai generated fake face videos by detecting eye blinking,'' \emph{arXiv preprint arXiv:1806.02877}, 2018.

\bibitem{li2018exposing}
Y.~Li and S.~Lyu, ``Exposing deepfake videos by detecting face warping artifacts,'' \emph{arXiv preprint arXiv:1811.00656}, 2018.

\bibitem{yang2019exposing}
X.~Yang, Y.~Li, and S.~Lyu, ``Exposing deep fakes using inconsistent head poses,'' in \emph{ICASSP 2019-2019 IEEE international conference on acoustics, speech and signal processing (ICASSP)}.\hskip 1em plus 0.5em minus 0.4em\relax IEEE, 2019, pp. 8261--8265.

\bibitem{korshunov2018deepfakes}
P.~Korshunov and S.~Marcel, ``Deepfakes: a new threat to face recognition? assessment and detection,'' \emph{arXiv preprint arXiv:1812.08685}, 2018.

\bibitem{zhou2017two}
P.~Zhou, X.~Han, V.~I. Morariu, and L.~S. Davis, ``Two-stream neural networks for tampered face detection,'' in \emph{2017 IEEE conference on computer vision and pattern recognition workshops (CVPRW)}.\hskip 1em plus 0.5em minus 0.4em\relax IEEE, 2017, pp. 1831--1839.

\bibitem{afchar2018mesonet}
D.~Afchar, V.~Nozick, J.~Yamagishi, and I.~Echizen, ``Mesonet: a compact facial video forgery detection network,'' in \emph{2018 IEEE international workshop on information forensics and security (WIFS)}.\hskip 1em plus 0.5em minus 0.4em\relax IEEE, 2018, pp. 1--7.

\bibitem{nataraj2019detecting}
L.~Nataraj, T.~M. Mohammed, S.~Chandrasekaran, A.~Flenner, J.~H. Bappy, A.~K. Roy-Chowdhury, and B.~Manjunath, ``Detecting gan generated fake images using co-occurrence matrices,'' \emph{arXiv preprint arXiv:1903.06836}, 2019.

\bibitem{rossler2019faceforensics++}
A.~Rossler, D.~Cozzolino, L.~Verdoliva, C.~Riess, J.~Thies, and M.~Nie{\ss}ner, ``Faceforensics++: Learning to detect manipulated facial images,'' in \emph{Proceedings of the IEEE/CVF international conference on computer vision}, 2019, pp. 1--11.

\bibitem{kowalski_faceswap_2017}
\BIBentryALTinterwordspacing
M.~Kowalski, ``{FaceSwap},'' 2017, 3D face swapping implemented in Python. GitHub repository. [Online]. Available: \url{https://github.com/MarekKowalski/FaceSwap}
\BIBentrySTDinterwordspacing

\bibitem{faceswap_deepfakes_2025}
\BIBentryALTinterwordspacing
deepfakes, ``{FaceSwap}: Deepfakes software for all,'' 2025, gitHub repository, last updated Nov 24, 2025. [Online]. Available: \url{https://github.com/deepfakes/faceswap}
\BIBentrySTDinterwordspacing

\bibitem{thies2019deferred}
J.~Thies, M.~Zollh{\"o}fer, and M.~Nie{\ss}ner, ``Deferred neural rendering: Image synthesis using neural textures,'' \emph{Acm Transactions on Graphics (TOG)}, vol.~38, no.~4, pp. 1--12, 2019.

\bibitem{dolhansky2019deepfake}
B.~Dolhansky, R.~Howes, B.~Pflaum, N.~Baram, and C.~C. Ferrer, ``The deepfake detection challenge (dfdc) preview dataset,'' \emph{arXiv preprint arXiv:1910.08854}, 2019.

\bibitem{jiang2020deeperforensics}
L.~Jiang, R.~Li, W.~Wu, C.~Qian, and C.~C. Loy, ``Deeperforensics-1.0: A large-scale dataset for real-world face forgery detection,'' in \emph{Proceedings of the IEEE/CVF conference on computer vision and pattern recognition}, 2020, pp. 2889--2898.

\bibitem{zhu2017unpaired}
J.-Y. Zhu, T.~Park, P.~Isola, and A.~A. Efros, ``Unpaired image-to-image translation using cycle-consistent adversarial networks,'' in \emph{Proceedings of the IEEE international conference on computer vision}, 2017, pp. 2223--2232.

\bibitem{choi2018stargan}
Y.~Choi, M.~Choi, M.~Kim, J.-W. Ha, S.~Kim, and J.~Choo, ``Stargan: Unified generative adversarial networks for multi-domain image-to-image translation,'' in \emph{Proceedings of the IEEE conference on computer vision and pattern recognition}, 2018, pp. 8789--8797.

\bibitem{frank2020leveraging}
J.~Frank, T.~Eisenhofer, L.~Sch{\"o}nherr, A.~Fischer, D.~Kolossa, and T.~Holz, ``Leveraging frequency analysis for deep fake image recognition,'' in \emph{International conference on machine learning}.\hskip 1em plus 0.5em minus 0.4em\relax PMLR, 2020, pp. 3247--3258.

\bibitem{durall2020watch}
R.~Durall, M.~Keuper, and J.~Keuper, ``Watch your up-convolution: Cnn based generative deep neural networks are failing to reproduce spectral distributions,'' in \emph{Proceedings of the IEEE/CVF conference on computer vision and pattern recognition}, 2020, pp. 7890--7899.

\bibitem{wang2023dire}
Z.~Wang, J.~Bao, W.~Zhou, W.~Wang, H.~Hu, H.~Chen, and H.~Li, ``Dire for diffusion-generated image detection,'' in \emph{Proceedings of the IEEE/CVF International Conference on Computer Vision}, 2023, pp. 22\,445--22\,455.

\bibitem{fernandez2023stable}
P.~Fernandez, G.~Couairon, H.~J{\'e}gou, M.~Douze, and T.~Furon, ``The stable signature: Rooting watermarks in latent diffusion models,'' in \emph{Proceedings of the IEEE/CVF International Conference on Computer Vision}, 2023, pp. 22\,466--22\,477.

\bibitem{corvi2023detection}
R.~Corvi, D.~Cozzolino, G.~Zingarini, G.~Poggi, K.~Nagano, and L.~Verdoliva, ``On the detection of synthetic images generated by diffusion models,'' in \emph{ICASSP 2023-2023 IEEE International Conference on Acoustics, Speech and Signal Processing (ICASSP)}.\hskip 1em plus 0.5em minus 0.4em\relax IEEE, 2023, pp. 1--5.

\end{thebibliography}
\end{document}